\RequirePackage{fix-cm} 
\documentclass[ twoside,openright,titlepage,numbers=noenddot,headinclude,
                footinclude=true,cleardoublepage=empty,abstractoff, 
                BCOR=5mm,paper=a4,fontsize=11pt,
                ngerman,american,%
                ]{scrreprt}

\PassOptionsToPackage{utf8}{inputenc}
	\usepackage{inputenc}

\PassOptionsToPackage{eulerchapternumbers,listings,
					 pdfspacing,
					 subfig,beramono,eulermath,parts,dottedtoc}{classicthesis}                                        

\newcommand{\myTitle}{Contextualization and Generalization\\ in Entity and Relation Extraction \xspace}

\newcommand{\myName}{Bruno Taillé \xspace}
\newcommand{\myProf}{Vincent Guigue \xspace}
\newcommand{\myOtherProf}{Patrick Gallinari \xspace}

\newcommand{\myUni}{Sorbonne Université\xspace}

\newcommand{\myTime}{March 2022\xspace}
\newcommand{\myVersion}{version 0.0\xspace}

\newcounter{dummy} 
\providecommand{\mLyX}{L\kern-.1667em\lower.25em\hbox{Y}\kern-.125emX\@}

	\usepackage{babel}                  

\usepackage{marginnote}
\usepackage{sidenotes}

\usepackage{csquotes}
\PassOptionsToPackage{%
	backend=bibtex8,bibencoding=ascii,%
	language=auto,%
    style=authoryear, 
    citestyle=authoryear-comp,
    dashed=false,
    sorting=nyt, 
    maxbibnames=10, 
    maxcitenames=2,
    natbib=true 
}{biblatex}
    \usepackage{biblatex}

\PassOptionsToPackage{fleqn}{amsmath}       
    \usepackage{amsmath}

\PassOptionsToPackage{T1}{fontenc} 
    \usepackage{fontenc}     
\usepackage{textcomp} 
\usepackage{scrhack} 
\usepackage{xspace} 
\usepackage{mparhack} 
\usepackage{fixltx2e} 
\PassOptionsToPackage{printonlyused,smaller}{acronym} 
    \usepackage{acronym} 
    
\usepackage{tabularx} 
\usepackage{caption}
\usepackage{subfig}  

\usepackage{listings} 
\PassOptionsToPackage{pdftex,hyperfootnotes=false,pdfpagelabels}{hyperref}
    \usepackage{hyperref}  

\PassOptionsToPackage{pdftex}{graphicx}
    \usepackage{graphicx}

\hypersetup{%
    colorlinks=false, linktocpage=false, pdfstartpage=3, pdfstartview=FitV, pdfborder={0 0 0},%
    breaklinks=true, pdfpagemode=UseNone, pageanchor=true, pdfpagemode=UseOutlines,%
    plainpages=false, bookmarksnumbered, bookmarksopen=true, bookmarksopenlevel=1,%
    hypertexnames=true, pdfhighlight=/O,
    urlcolor=webbrown, linkcolor=RoyalBlue, citecolor=webgreen, 
    pdftitle={\myTitle},%
    pdfauthor={\textcopyright\ \myName, \myUni},%
    pdfsubject={},%
    pdfkeywords={},%
    pdfcreator={pdfLaTeX},%
    pdfproducer={LaTeX with hyperref and classicthesis}%
}  

\makeatletter
\@ifpackageloaded{babel}%
    {%
       \addto\extrasamerican{%
                }%
    }{\relax}
\makeatother

\usepackage{amsmath,amsfonts,amsthm,bm}
\usepackage{pifont}
\newcommand{\cmark}{\ding{51}}%
\newcommand{\xmark}{\ding{55}}%
\usepackage{amssymb}
\usepackage{booktabs}
\renewcommand{\arraystretch}{1.2}
\usepackage{fixltx2e}
\usepackage{arydshln}
\usepackage{booktabs}
\usepackage{tabularx}
\usepackage{multirow}
\usepackage{siunitx}

\usepackage{lscape}
\usepackage{etoc}

\DeclareFieldFormat{citehyperref}{%
  \DeclareFieldAlias{bibhyperref}{noformat}
  \bibhyperref{#1}}

\DeclareFieldFormat{textcitehyperref}{%
  \DeclareFieldAlias{bibhyperref}{noformat}
  \bibhyperref{%
    #1%
    \ifbool{cbx:parens}
      {\bibcloseparen\global\boolfalse{cbx:parens}}
      {}}}

\savebibmacro{cite}
\savebibmacro{textcite}

\renewbibmacro*{cite}{%
  \printtext[citehyperref]{%
    \restorebibmacro{cite}%
    \usebibmacro{cite}}}

\renewbibmacro*{textcite}{%
  \ifboolexpr{
    ( not test {\iffieldundef{prenote}} and
      test {\ifnumequal{\value{citecount}}{1}} )
    or
    ( not test {\iffieldundef{postnote}} and
      test {\ifnumequal{\value{citecount}}{\value{citetotal}}} )
  }
    {\DeclareFieldAlias{textcitehyperref}{noformat}}
    {}%
  \printtext[textcitehyperref]{%
    \restorebibmacro{textcite}%
    \usebibmacro{textcite}}}

\usepackage{classicthesis} 

\newcolumntype{R}{>{\raggedleft\arraybackslash}X}

\DeclareMathOperator*{\argmax}{argmax} 

\begin{document}
\frenchspacing
\raggedbottom
\selectlanguage{american} 
\pagenumbering{roman}
\pagestyle{plain}
\begin{titlepage}
    \begin{addmargin}[-1cm]{-4cm}
    
    \includegraphics[width=5cm]{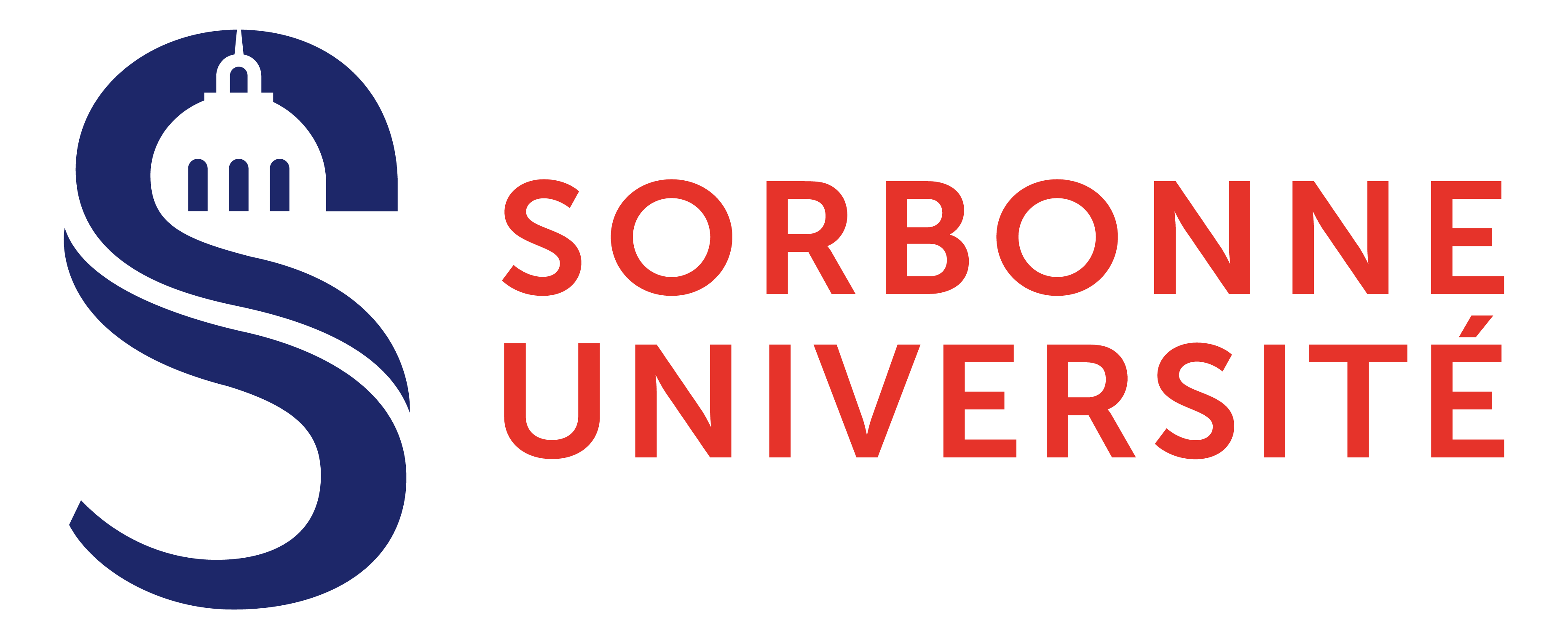}
    
        
        
        
    

	
        
    \medskip
    \begin{center}
        \large  

        \hfill
        
        \spacedallcaps{\textbf{Thèse de Doctorat de Sorbonne Université}}\\
        \textbf{Spécialité Informatique}\\
        \small{École Doctorale Informatique, Télécommunication et Électronique (Paris)}
        
        \vfill

        \begingroup
            \Large
            \color{Maroon}\spacedallcaps{\myTitle} \\ \bigskip
        \endgroup

        {\large \spacedlowsmallcaps{\myName}}
        
        \vfill
        
        \large
        Dirigée par\\
        \spacedlowsmallcaps{\myProf} et \spacedlowsmallcaps{\myOtherProf}
        
    \vfill
    
    Soutenue publiquement pour obtenir le grade de \\
    \textbf{Docteur en Informatique de Sorbonne Université}\\
    le 11 mars 2022 devant un jury composé de :
    
    \end{center}
    
    \vfill

    \medskip
    \begin{center}
    \begin{tabular}{lr}
        \spacedlowsmallcaps{Patrice Bellot} & Rapporteur\\
        \addlinespace[-1.3ex]\textit{Professeur, Aix-Marseille Université} \\
        
        \addlinespace[-0.7ex]\spacedlowsmallcaps{Antoine Doucet} & Rapporteur\\
        \addlinespace[-1.3ex]\textit{Professeur, La Rochelle Université} \\
        
        \addlinespace[-0.7ex]\spacedlowsmallcaps{Maud Ehrmann} & Examinatrice\\
        \addlinespace[-1.3ex]\textit{Research Scientist, École Polytechnique Fédérale de Lausanne}\\
        
        \addlinespace[-0.7ex]\spacedlowsmallcaps{Xavier Tannier} & Examinateur \\
        \addlinespace[-1.3ex]\textit{Professeur, Sorbonne Université} \\
        
        \addlinespace[-0.7ex]\spacedlowsmallcaps{Vincent Guigue} & Directeur de thèse\\
        \addlinespace[-1.3ex]\textit{Maître de Conférences, Sorbonne Université} \\
        
        \addlinespace[-0.7ex]\spacedlowsmallcaps{Patrick Gallinari} & Directeur de thèse\\ 
        \addlinespace[-1.3ex]\textit{Professeur, Sorbonne Université} \\
    \end{tabular}
    \end{center}
    
    
  \end{addmargin}       
\end{titlepage}   

\cleardoublepage\thispagestyle{empty}

\hfill

\vfill

\noindent\myName: \textit{\myTitle,} 
\textcopyright\ \myTime\\


%
%
%
%
%

\cleardoublepage
\thispagestyle{empty}
\refstepcounter{dummy}
\pdfbookmark[1]{Dedication}{Dedication}

\vspace*{3cm}

\begin{center}
    À mes grands-parents, mes parents et mon frère. 
\end{center}

\cleardoublepage
\pdfbookmark[1]{Abstract}{Abstract}
\begingroup
\let\clearpage\relax
\let\cleardoublepage\relax
\let\cleardoublepage\relax

\chapter*{Abstract}
During the past decade, neural networks have become prominent in Natural Language Processing (NLP), notably for their capacity to learn relevant word representations from large unlabeled corpora.
These word embeddings can then be transferred and finetuned for diverse end applications during a supervised training phase. 
More recently, in 2018, the transfer of entire pretrained Language Models and the preservation of their contextualization capacities enabled to reach unprecedented performance on virtually every NLP 
benchmark, sometimes even outperforming human baselines.
However, as models reach such impressive scores, their comprehension abilities still appear as shallow, which reveal limitations of benchmarks to provide useful insights on their factors of performance and to accurately measure understanding capabilities. 

In this thesis, we study the behaviour of state-of-the-art models regarding generalization to facts unseen during training in two important Information Extraction tasks: Named Entity Recognition (NER) and Relation Extraction (RE).
Indeed, traditional benchmarks present important lexical overlap between mentions and relations used for training and evaluating models, whereas the main interest of Information Extraction is to extract previously unknown information.
We propose empirical studies to separate performance based on mention and relation overlap with the training set and find that pretrained Language Models are mainly beneficial to detect unseen mentions, in particular out-of-domain.
While this makes them suited for real use cases, there is still a gap in performance between seen and unseen mentions that hurts generalization to new facts.
In particular, even state-of-the-art ERE models rely on a shallow retention heuristic, basing their prediction more on arguments surface forms than context.

In this work, we also consolidate the foundations of evaluation in End-to-end Relation Extraction that were undermined by previous incorrect comparisons and propose a basis for a finer-grained evaluation and comprehension of End-to-end Relation Extraction models regarding generalization to new relations.
We finally suggest ideas to improve context incorporation in the creation of both future models and datasets.

\endgroup			

\vfill
\cleardoublepage

\pdfbookmark[1]{Résumé}{Résumé}
\chapter*{Résumé}
\begingroup
\let\clearpage\relax
\let\cleardoublepage\relax
\let\cleardoublepage\relax
Au cours de la dernière décennie, les réseaux de neurones sont devenus incontournables
dans le Traitement Automatique du Langage (TAL), notamment pour leur capacité à
apprendre des représentations de mots à partir de grands corpus non étiquetés.
Ces plongements de mots peuvent ensuite être transférés et raffinés pour des applications diverses au cours d'une phase d'entraînement supervisé.
Plus récemment, en 2018, le transfert de modèles de langue pré-entraînés et la préservation de leurs capacités de contextualisation ont permis 
d'atteindre des performances sans précédent sur pratiquement tous les benchmarks de TAL,
surpassant parfois même des performances humaines de référence.
Cependant, alors que ces modèles atteignent des scores impressionnants, leurs capacités de compréhension apparaissent toujours assez peu développées, révélant les limites des jeux de données de référence pour identifier leurs facteurs de performance et pour mesurer précisément leur capacité de compréhension.

Dans cette thèse, nous étudions la généralisation à des faits inconnus par des modèles état de l'art dans deux tâches importantes en Extraction d'Information : la Reconnaissance d'Entités Nommées et l'Extraction de Relations.
En effet, les benchmarks traditionnels présentent un recoupement lexical important entre les mentions et les relations utilisées pour
l'entraînement et l'évaluation des modèles.
Au contraire, l'intérêt principal de l'Extraction d'Information est d'extraire des informations inconnues jusqu'alors.
Nous proposons plusieurs études empiriques pour séparer les performances selon le recoupement des mentions et des relations avec le jeu d'entraînement.
Nous constatons que les modèles de langage pré-entraînés sont principalement bénéfiques pour détecter les mentions non connues, en particulier dans des genres de textes nouveaux.
Bien que cela les rende adaptés à des cas d'utilisation concrets, il existe toujours un écart de performance important entre les mentions connues et inconnues, ce qui nuit à la généralisation à de nouveaux faits.
En particulier, même les modèles d'Extraction d'Entités et de Relations les plus récents reposent sur une heuristique de rétention superficielle, basant plus leur
prédiction sur les arguments des relations que sur leur contexte.

Nous consolidons également les bases de l'évaluation de l'Extraction d'Entités et de Relations qui ont été sapées par des comparaisons incorrectes et nous proposons une base pour une évaluation et une compréhension plus fines des modèles concernant leur généralisation à de nouvelles relations.
Enfin, nous suggérons des pistes pour améliorer l'incorporation du contexte dans la création de futurs modèles et jeux de données.
\endgroup			
\vfill
\cleardoublepage\pdfbookmark[1]{Remerciements}{Remerciements}

\chapter*{Remerciements}

Arrivant au terme de ce long voyage qu'est le doctorat, il me reste à remercier les nombreuses personnes que j'ai rencontrées et qui m'ont accompagné le long du chemin.
Bien qu'il me soit impossible de tous vous nommer ici, je tiens à vous exprimer toute ma gratitude.

Je tiens d'abord à remercier mes directeurs de thèse Vincent et Patrick auprès desquels j'ai appris les rudiments de la recherche académique ainsi que Geoffrey qui m'a accompagné du côté de BNP Paribas.
J'ai toujours trouvé auprès d'eux une oreille attentive et des suggestions pertinentes, le tout dans un cadre agréable et bienveillant m'offrant une grande liberté dans mes directions de recherche.

J'ai ensuite une pensée singulière pour Clément avec qui j'ai partagé le même double statut durant presque quatre ans et avec qui l'union a fait la force tout au long du périple.

J'aimerais également remercier les membres de l'équipe MLIA, que ce soit les permanents pour les différents conseils que j'ai pu recevoir, Nadine et Christophe pour leurs rôles de support indispensables et l'intégralité des doctorants que j'ai côtoyés, dans un même bureau, au détour d'un couloir ou autour d'un repas ou d'un verre.
Votre cohésion sans faille a été d'autant plus importante en ces temps de restrictions sanitaires.
Sans pouvoir tous vous citer, j'ai une pensée particulière pour Éloi, Arthur, Edouard, Mickaël, Jérémie, Jean-Yves, les deux Thomas, Étienne, Clara, Yuan, Adrien, Valentin, Agnès et Matthieu. 

J'ai aussi côtoyé de merveilleux collègues à la BNP, là encore dans un cadre de travail très agréable.
Je pense d'abord à Edouard qui m'a fait confiance et a rendu possible cette collaboration et à tous ceux avec qui j'ai partagé des moments formidables que ce soit au quotidien, au Portugal ou à la Taverne.
Encore une fois au risque d'en oublier certains, je citerais Baoyang, Thomas, Bruce, Frank, Anne, Ludan, Alexis, Pirashanth, Tom, Mhamed, Mathis, Charline et Pierre.

J'ai également aujourd'hui une pensée pour tous les professeurs qui ont pu m'inculquer un goût pour les sciences et leur rigueur assez tôt. Que ce soit Alain Lique ou mon cher oncle au collège, l'iconique Jean-Pierre Sanchez ou le regretté Prebagarane Mouttou au lycée, ou encore Emmanuelle Tosel ou Stéphane Olivier en prépa.

Pour finir, j'exprime ma profonde gratitude à toute ma famille.
Merci à mes grands-parents pour leurs encouragements.
Merci à mes parents et à mon frère pour leur indéfectible soutien. 

\cleardoublepage
\pdfbookmark[1]{Publications}{publications}
\chapter*{Publications}
Some ideas and figures have appeared previously in the following publications:


\begin{refsection}[ownpubs]
    \small
    \nocite{cap2019, ecir2020, emnlp2020, emnlp2021} 
    \printbibliography[heading=none]
\end{refsection}


\bigskip
\noindent
The code accompanying these publications is publicly released at \href{https://www.github.com/btaille}{github.com/btaille}

\nocite{Taille2019UneDEntites, taille-etal-2021-separating}
\pagestyle{scrheadings}
\cleardoublepage
\refstepcounter{dummy}
\pdfbookmark[1]{\contentsname}{tableofcontents}
\setcounter{tocdepth}{2} 
\setcounter{secnumdepth}{3} 
\manualmark
\markboth{\spacedlowsmallcaps{\contentsname}}{\spacedlowsmallcaps{\contentsname}}
\tableofcontents 
\automark[section]{chapter}
\renewcommand{\chaptermark}[1]{\markboth{\spacedlowsmallcaps{#1}}{\spacedlowsmallcaps{#1}}}
\renewcommand{\sectionmark}[1]{\markright{\thesection\enspace\spacedlowsmallcaps{#1}}}
\clearpage

\begingroup 
    \let\clearpage\relax
    \let\cleardoublepage\relax
    \let\cleardoublepage\relax
    \refstepcounter{dummy}
    \pdfbookmark[1]{\listfigurename}{lof}
    \listoffigures

\endgroup

\begingroup 
    \let\cleardoublepage\relax
    \clearpage
    \refstepcounter{dummy}
    \pdfbookmark[1]{\listtablename}{lot}
    \listoftables
\endgroup
    

       

\cleardoublepage\pagenumbering{arabic}
\cleardoublepage

\etocsettocstyle{\vskip0.3\baselineskip}{\noindent\rule{\linewidth}{0.5pt}\vskip0.5\baselineskip} 

\chapter{Introduction}

Language, whether signed, spoken or written, is at the heart of human societies and cultures as the principal means of inter-human communication.
Its ability to convey ideas, knowledge or emotions makes it a defining human trait, often considered as the hallmark of human intelligence.
For this reason, understanding and producing coherent language has long been viewed as a milestone goal in the development of computers, ever since \citet{Turing1950ComputingIntelligence}'s famous ``imitation game''.

This motivated the development of two complementary fields: \textbf{Computational Linguistics} (CL) which aims at studying languages using the ability of computers to process large corpora and \textbf{Natural Language processing} (NLP) which uses the same means to build systems with useful applications.
These include now commonly used tools such as Machine Translation and Speech Recognition systems or conversational assistants such as Alexa or Siri that are able to detect a user's intents or answer some of their questions.

Within language, text in particular has played a key role in the way humans have stored and broadcasted information for centuries.
Whether in laws, administrative records, news, novels, scientific articles, letters, emails, SMS or internet forums and comments; text has imposed itself as an efficient means of inter-human communication that displays a diversity of languages, usages, domains and forms.
The role of text has only been increased by the development of Information Technologies such as Internet and social media, leading to an ever growing quantity of text produced and stored daily.
Given the scales at hand, automatic processing of text seems necessary to detect hateful or harmful contents, spams, duplicate questions or increase accessibility of content with e.g. automatic translation.

Because of the complexity and the vastness of language, NLP is often decomposed into tasks that are designed to reflect one or several aspects of language and can be divided into two broad categories: \textbf{Natural Language Understanding} (NLU) and \textbf{Natural Language Generation} (NLG).
NLU aims at capturing elements of meaning in a text or speech such as its genre, its polarity, the spatial or temporal markers or real-life entities it mentions and the relations that are expressed between them.
NLG comprises all tasks where the system produces text or speech conditionally on an input that can itself be text or speech such as in Machine Translation or Speech Recognition or not, for example to generate weather broadcast or financial reports from tabular data.

\section{Natural Language Understanding and Information Extraction}

Hence, Natural Language Understanding can be used to process the information contained in large corpora of texts to automate or assist decisions with applications as diverse as using clinical reports for diagnostics or news broadcasts for stock trading.
Going even further, the ability to grasp the meaning of a text could be a first step towards building systems able to reason logically over the numerous facts stored in a textual form to answer complex questions or even automate scientific discovery.

Converting the information expressed in textual documents into a machine-readable structured format is thus an important issue and the goal of the \textbf{Information Extraction} (IE) field.
Such information often involves real-world beings, objects or concepts that are connected with one another to a certain extent.
Consequently, one proposal in this direction, is to build a database in which facts expressed in a document are stored in the form of a \textbf{Knowledge Graph} (KG) whose nodes typically represent real-life \textbf{entities} such as people, organization, location, dates and the edges represent \textbf{relations} between them.
Facts are thus stored as \textbf{triples} such as (Apple Inc., founded by, Steve Jobs) or (Victor Hugo, birth date, 26-02-1802).
Entity and Relations types are predefined, following a format of knowledge representation called an \textbf{Ontology}.

\begin{figure}[!h]
\begin{adjustwidth*}{}{-.3\textwidth}
\caption{Illustration of the structure of a Knowledge Graph}
\label{fig:01:kb}
\centering\includegraphics[width=1.3\textwidth]{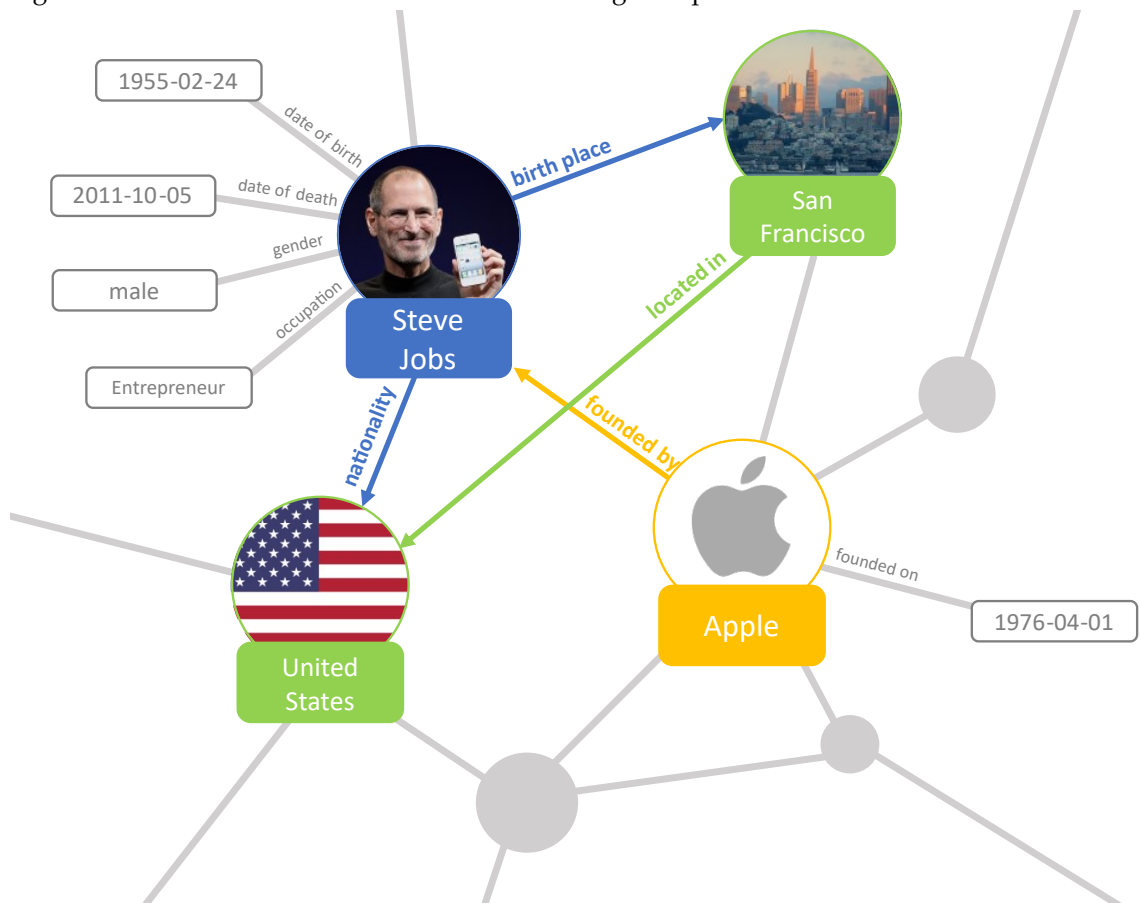}
\end{adjustwidth*}
\end{figure}

The graph structure of such databases enables to perform semantic queries more conveniently than in a classical Relational Database, typically when retrieving all entities linked to a given one at a certain level (e.g. neighbours of neighbours) or all entities with a given value of an attribute.
An application example of such a structure is Google's Knowledge Graph that has been introduced in 2012 and is used in Search to store and display information regarding entities in infoboxes or in Google assistant to answer questions regarding entities.
The structure of Wikipedia where each entity is represented by a page and where pages are connected by hyperlinks also follows a graph and there are initiatives to build a corresponding knowledge graph such as \textbf{Wikidata} through crowd-sourced annotation.

While these graphs can be constructed by hand, one of the goals of IE is \textbf{Automatic Knowledge Graph Construction} (AKBC) from textual documents alone, task that can be decomposed in several subtasks such as Named Entity Recognition, Coreference Resolution, Entity Linking and Relation Extraction.

Although these four tasks are necessary for KBC, \textbf{Named Entity Recognition} (NER) and \textbf{Relation Extraction} (RE), are more specifically at the heart of the process, making them important NLU tasks.
NER aims at detecting textual mention of entities and classifying them by type (e.g. person or location).
RE proposes to further extract the relations that are expressed between these entities.
Because of the apparent interdependency between these tasks, they can be tackled as a single joint Entity and Relation Extraction task, referred to as \textbf{End-to-end Relation Extraction} (ERE).

\section{Recent Advances in Natural Language Processing}
While early NLP approaches proposed to leverage linguistics to design rules to grasp the meaning of a text \citep{Weizenbaum1966ELIZA:Machine, Winograd1972UnderstandingLanguage}, they appeared restricted to the situations envisioned at their creation, unable to adapt to unseen inputs or domains.

These early rule-based models left the floor to a new paradigm: using statistical models to learn these rules automatically from data.
This was envisioned as a method easier to adapt to new domains as it relied less on human expertise.
Hence, human effort moved from designing rules to data annotation and feature engineering: designing relevant data representations that can be leveraged by these Machine Learning models.
Nevertheless, these models still show a gap between the performance on the specific data used to train the algorithm and on new unseen data, which is characteristic of a lack of \textbf{generalization}. 

Following the same idea to reduce dependency on human expertise, Deep Learning proposes to in turn learn these data representations automatically, using a hierarchical multi-layer structure to learn features.
In the past decade, these neural networks were successfully applied to a wide variety of fields including Computer Vision and Natural Language Processing.
For text processing in particular, neural networks have been used to automatically learn word representations called \textbf{word embeddings} using a simple self-supervised Language Model pretraining objective \citep{Bengio2003AModel, Mikolov2013EfficientSpace}.
Because this pretraining can be used to encode semantic information into word representations without annotated data, word embeddings have been used as an effective \textbf{Transfer Learning} method in NLP to improve generalization \citep{Collobert2008ALearning}.
In fact, the recent advances in NLP models performance, as measured by common benchmarks, mainly stem from using more and more data to pretrain and transfer deeper and deeper pretrained neural networks \citep{devlin-etal-2019-bert}.
This advances have been favored by the development of mature software frameworks such as Tensorflow \citep{Abadi2016TensorFlow:Learning} or Pytorch \citep{Paszke2019PyTorch:Library} that integrate GPU optimization, as well as initiatives to simplify the transfer of entire pretrained models, such as Huggingface Transformers library \citep{wolf-etal-2020-transformers}. 

Whereas Deep Neural Networks showed impressive successes, outperforming previous models in e.g. Image Classification \citep{Krizhevsky2012ImageNetNetworks} or Machine Translation \citep{Wu2016GooglesTranslation}, they have known shortcomings.
First, their training often requires to be supervised with very large datasets of data samples labeled according to their final objective.
Second, their prediction is hard to explain or interpret and they are often viewed as blackboxes.
Third, while very effective in tackling data similar to their training data, recent studies show that \textbf{these models can adopt shallow heuristics that hurt generalization} to examples too different from their training data \citep{jia-liang-2017-adversarial, mccoy-etal-2019-right}.

Despite the recent language model pretraining strategy \citep{devlin-etal-2019-bert} to leverage the vast amount of unlabeled data and reduce human annotation cost to obtain better, sometimes even superhuman, performance on numerous benchmarks, this last drawback ultimately boils down to a lack of generalization, similar to early rule-based approaches.

\section{Context and Contributions}
\subsection{Industrial Perspectives}
This work was initiated with and partly financed by BNP Paribas CIB's AI Lab which identified Information Extraction as a central part of numerous applications across all the departments of the group.
In particular, identifying datapoints such as people, organizations, locations, dates but also fund names or transaction ids with Named Entity Recognition can be used for automatic processing of orders or as an automatic contract screening preprocessing step to identify any mention of an entity under embargo.

In a more ambitious long term perspective, one banking application of end-to-end Relation Extraction is to automatically build a Knowledge Graph of facts between people, organizations and locations from public text sources such as newsfeeds in order to construct a Know Your Customer (KYC) system. Such system can be used to fight money laundering and prevent the financing of terrorist organizations.

In this industrial context, models performance on specific internal data is critical but the budget allocated for data annotation is also limited. 
Hence, generalization beyond the training data is an imperative requirement to obtain useful and cost effective solutions.

\subsection{Natural Language Processing Context}
The work presented in this dissertation was performed between mid 2018 and mid 2021, in a context of a quick evolution in the NLP field both in terms of technology and community.
Indeed, the introduction of contextual embeddings and the major shifts in performance induced by these representations led BERT-based models \citep{devlin-etal-2019-bert} to replace the previously ubiquitous recurrent neural networks in every NLP task in less than a year.
In parallel, the NLP community has grown massively with unprecedented numbers of contribution proposals for new models, data and evaluation resources or empirical studies.

However, it also appears that the recent progress in NLP mainly stems from the introduction of BERT and its variations that largely make use of more unlabeled data with more parameters and computation power only accessible to a few key actors such as Google, Facebook, Nvidia or Microsoft.
Given the practical impossibility to compete in the field of Language Model pretraining, we can see a uniformization of NLP models that rely on large publicly released pretrained Language Models, finetuned along with a few simple additional layers.

While these models have achieved superhuman performance on common benchmarks such as SQUAD in Question Answering or GLUE in Natural Language Understanding, these impressive results encouraged the development of works in two directions in particular: understanding the reasons of the effectiveness of BERT-like approaches with works coined as BERTology \citep{rogers-etal-2020-primer} and designing new evaluation settings and benchmarks that put in perspective the limitations of these approaches whose comprehension capabilities are nowhere near human-level \citep{mccoy-etal-2019-right, ribeiro-etal-2020-beyond}. 

\subsection{Contributions}

In the previously described context, we study the fundamental Knowledge Base Construction tasks of Named Entity Recognition and Relation Extraction.
Following the more abundant related works and resources for this language, we focus on \textbf{English} corpora with the belief that our findings also apply to at least numerous other languages.

Language Model Pretraining was introduced with models such as ELMo \citep{peters-etal-2018-deep} or BERT \citep{devlin-etal-2019-bert} that were originally tested on multiple NLP tasks and outperformed previous state-of-the-art-models even when used with simple baselines.
This impressive leap in performance led to their quick adoption by the community.
However, because they were originally tested on several tasks at once, their evaluation was limited to single scores on benchmarks that did not reflect the origin of this performance.

In particular, this original evaluation includes the Named Entity Recognition task, but limits to a single F1 score on a single benchmark.
We propose to more precisely analyze their performance in End-to-end Relation Extraction with a focus on their \textbf{generalization capability} beyond the mere memorization of training examples.
This capacity is both a key issue in real-life applications and a key aspect of comprehension that we believe is overlooked by standard benchmarks and metrics.

Indeed, a specificity of text is to rely on a finite number of words that are used in sequence to express a large variety of concepts.
This leads a given entity or concept to be designated by a limited set of rigid designators which can simply be memorized by models.
Introducing contextual information thus seems useful to reduce the dependency on exact surface forms, and contextual embeddings obtained from Language Model Pretraining precisely incorporate contextual information in word representations.
In particular, they are intuitively particularly useful in an entity-centric task such as Named Entity Recognition which is an integral part of Knowledge Base Construction.
Our first contribution is thus an empirical study for a finegrained quantification of their impact on Named Entity Recognition, in particular on generalization to unseen mentions and new domains (\autoref{chapter:03:ner}).

Then, we tackle the global End-to-end Relation Extraction setting for which numerous settings and models have been introduced.
First, we put this abundant literature into order with a proposal for a taxonomy (\autoref{chapter:04:re-taxonomy}) and the identification of several previous incorrect evaluations in the literature (\autoref{chapter:05:rethinking}).
Second, we extend our previous study on generalization to End-to-end Relation Extraction showing that a simple retention heuristics can partly explain the performance of state-of-the-art-models on standard benchmarks.

Finally we propose our perspectives on methods to use self-attention patterns from BERT-like Language Model to better incorporate contextual information for Relation Extraction (\autoref{chapter:06:attention}). 

\section{Outline}

In \autoref{chapter:02:bert}, we provide an overview of the evolution of word representations used in Machine Learning, in particular in Deep Learning models.
We review the ideas and techniques that led to the evolution from handcrafted features and one-hot vector representations to distributed word embeddings learned using Language Models. Furthermore, we present the Transformer architecture that together with Language Model pretraining led to recent breakthroughs in virtually every NLP task with the BERT model \citep{devlin-etal-2019-bert}. 

In \autoref{chapter:03:ner}, we focus on the Named Entity Recognition task and present the lexical overlap issue that questions the ability of standard benchmarks to accurately measure generalization to unseen mentions.
Then, we propose an empirical study that both confirms that lexical overlap plays a key role in the performance of state-of-the-art models and shows that recent pretrained Language Models are a helpful way to incorporate context and generalize to unseen mentions and domains.

The following chapters tackle the broader and even more challenging End-to-end Relation Extraction scenario.
In \autoref{chapter:04:re-taxonomy}, we review previously explored Entity and Relation Extraction approaches.
We briefly introduce the pipeline approach before proposing a taxonomy of numerous End-to-end Relation Extraction models.
We argue that we can observe a triple evolution of word representations, joint learning strategy and NER strategy that prevents drawing useful conclusions from the literature alone.

In \autoref{chapter:05:rethinking}, we first identify several incorrect comparisons in the End-to-end Relation Extraction literature that only makes comparison between models worse.
We obtain a leaderboard of published results on five main benchmarks corrected from identified mistakes and call for an unified evaluation setting.
Moreover, we perform the ablations of two recent developments that we believe were missing: pretrained Language Models and Span-level NER.
We confirm that improvements on classical benchmarks are mainly due to the former.
Second, we extend our previous study of lexical overlap in NER to end-to-end RE and show that memorization of training triples can explain a part of performance on standard benchmarks.

\autoref{chapter:06:attention} presents explored lines of research towards proposing models better able to generalize beyond memorisation.
Following some previous BERTology works that evidence that BERT's attention patterns capture syntactic properties \citep{clark-etal-2019-bert}, we propose an approach called Second Order Attention.
It uses attention heads to model syntactic structures useful to detect which words in the context of argument candidates are reflective of a relation.

\autoref{chapter:07:conclusion} finally summarizes our findings and proposes our perspectives on the future of End-to-end Relation Extraction evaluation and models.  

\chapter{From handcrafted features to BERT}
\label{chapter:02:bert}

In the last three years, Language Model pretraining has become the new de facto standard to obtain state-of-the-art Natural Language Processing models.
While this was reflected by the sudden adoption of BERT-like models for virtually every NLP task in less than a year, the ideas used to obtain such universal word representations date back to the early 2000's with the introduction of word embeddings.

Indeed, early NLP algorithms were based on set of \textbf{handcrafted rules} that, for example, enabled to fake comprehension by detecting keywords and rephrasing user inputs in a conversational system such as \textbf{ELIZA} \citep{Weizenbaum1966ELIZA:Machine}. 
\textbf{Regular expressions} were used to recognize predefined text patterns and have also been used e.g. in early Named Entity Recognition programs \citep{Rau1991ExtractingText}.
Such handcrafted patterns are still used in current conversational assistants, most notably by Apple's Siri.
Nevertheless, maintaining and expanding such sets of rules for more and more applications and domains requires expensive human expertise and it seemed useful to use \textbf{Machine Learning} algorithms to learn these rules automatically from data.

However, a natural problematic in using ML algorithms is the \textbf{data representation} step that in turn first relied on feature engineering from domain experts. 
The same reasoning led to the development of \textbf{Deep Neural Networks}, whose Representation Learning capabilities were proved to be effective to tackle varied tasks in many application domains, including Natural Language Processing as soon as the early 2000's.

After an introduction to  Natural Language Processing tasks (\autoref{sec:02:nlp}) and some important Deep Learning architectures (\autoref{sec:02:deep learning}), this chapter reviews two key recent evolutions of Neural Networks for Natural Language Processing that led to successive leaps in benchmark performance over the past decade: 1) the use of neural language models to learn distributed word representations from unlabeled text (\autoref{sec:02:word representations}) and 2) the introduction of self-attention in the efficient Transformer architecture (\autoref{sec:02:transformer}). These two advances were recently combined in BERT \citep{devlin-etal-2019-bert}, which is now the standard approach to obtain state-of-the-art results for every NLP problem. This model is further described in \autoref{sec:02:bert} as well as several works trying to better understand the underlying reasons for the provided performance gain and framed as Bertology in \autoref{sec:02:bertology}.

\section{Classification in Natural Language Processing}
\label{sec:02:nlp}

Either for Computer Vision or Natural Language Processing applications, deep neural networks are mainly used for \textbf{classification}.
Their output must then be a \textbf{label} representative of a class, even though very diverse settings are used depending on the final application.

In NLP for example, \textbf{Text Classification} tasks aim at assigning a label to an entire sentence or document that can be indicative of its domain, its sentiment polarity or even if its a spam or not.
Other tasks propose to take as input pairs of sentences and output a single label predicting if sentences share a same meaning in \textbf{Paraphrase Detection} or if the first one entails or contradicts the second in \textbf{Natural Language Inference}.

Classification can also be made at a word or span level to predict grammatical properties in \textbf{Part-of-Speech Tagging} or the type of real-world entities they refer to in \textbf{Named Entity Recognition}.
It is also useful for \textbf{Extractive Question Answering} where given a question, we can tag every word in a context document as being part of the answer or not. 
Like for sentences, classification can be made for pairs of spans for example to predict if two mentions refer to a same entity in \textbf{Coreference Resolution} or the relation that holds between them in \textbf{Relation Extraction}.

Even Natural Language Generation tasks are viewed as classification tasks where the classes correspond to the different words in a vocabulary.
At each step, given a query and the sequence of previous outputs the model must find the most likely next word in the entire vocabulary.
This applies to \textbf{Language Modeling}, \textbf{Speech Recognition}, \textbf{Neural Machine Translation} or \textbf{Abstractive Question Answering} for example.

\section{An Introduction to Deep Neural Networks}
\label{sec:02:deep learning}

\subsection{Deep Learning}
Artificial Intelligence is currently one of the most thriving research fields, with a renewed interest from both academia and industry, partly due to recent breakthroughs in Deep Learning.
The first approach in AI was to hard-code logic rules and knowledge, which proved successful on well formalized tasks not requiring real-world knowledge, such as chess with the iconic win of IBM's Deep Blue over Garry Kasparov in 1996.
The machine was quickly able to tackle formal logic tasks impossible for human beings but failed to replicate basic human abilities such as text or speech understanding or visual recognition. 
To face these tasks, it took a paradigm shift from the hard-coded rules approach to \textbf{Machine Learning} which gives the machine the ability to extract its own knowledge from raw data.

The first algorithm showing this ability was developed as early as 1952 by Arthur Samuel who set the base of Machine Learning with a checkers program based on search trees which improved with the number of games it played \citep{Samuel1959SomeCheckers}.
Another breakthrough came in 1957 when Rosenblatt introduced the perceptron unit inspired by the structure of a neuron, upon which today's Deep Learning state-of-the-art algorithms are based \citep{Rosenblatt1958TheBrain.}. 
The perceptron, alongside later linear classifier algorithms could not deal with nonlinear problems. 
This led to the development of feature engineering, which relies on human knowledge to select and transform raw data into representations that are relevant for the task at hand and easy for the computer to classify. This approach however tends to move away from the spirit of Machine Learning, strongly relying on human knowledge, and is less effective for tasks such as computer vision where for example it is hard to handcraft features based on pixels to detect objects.
Hence the introduction of \textbf{Representation Learning}, a subfield of Machine Learning in which algorithms not only learns mappings from representations to outputs but also learn representations from raw inputs. 

\textbf{Deep Learning} \citep{LeCun2015DeepLearning, Goodfellow2016DeepLearning} is a particular area of Representation Learning which proposes to learn representations with artificial neural networks.
Neurons are arranged in a \textbf{hierarchy of layers}, each of which takes as input the output of the previous one.
This enables the network to combine the linear separation ability of each neuron to learn more and more abstract features by stacking layers.
These architectures are originally inspired by the research of \citet{Hubel1959ReceptiveCortex.} who studied the primary visual cortex of cats and proposed a hierarchical biological model.
They noticed that some parts of the brain were responding to very low-level stimuli such as edges instead of full object images. 
Quite naturally, Computer Vision became a major application in the development of neural networks, Convolutional Neural Networks in particular.

In the 1990s, \citet{LeCun1989BackpropagationRecognition} applied \textbf{backpropagation} to these architectures to classify hand-written digits and read ZIP Codes or paychecks. 
In the early 2010s, the development of GPU-accelerated implementations and the introduction of large annotated datasets such as ImageNet \citep{Russakovsky2015ImageNetChallenge} made of more than a million images of objects labeled in a thousand categories enabled to efficiently train deep architectures which rely on large amounts of training data. 
In 2012, AlexNet \citep{Krizhevsky2012ImageNetNetworks}, a deep convolutional neural network outperformed other traditional computer vision algorithms on the ImageNet classification challenge and every year since then new architectures with more and more layers perform better results, diverting the interest from feature engineering based methods.

\subsection{Multilayer Perceptron}

The smallest unit in deep neural networks is an \textbf{artificial neuron}, inspired by the activity of biological neurons \citep{Mcculloch1943AACTIVITY}.
This mathematical model takes a multidimensional input vector \textbf{$x$} and computes an output \textbf{$y$} with a weighted sum of its components followed by a non-linear \textbf{activation function $\sigma$}.

This can be divided in two steps:

\begin{align*}
  &- \text{Linear pre-activation:} &\textbf{l}(\textbf{x}) &= \textbf{w}.\textbf{x}+\textbf{b}\\
  &- \text{Activation:} &\textbf{y} &= \sigma(\textbf{l}(\textbf{x}))
\end{align*}

Where $\textbf{w}$ and $\textbf{b}$ are the weights and bias \textbf{parameters} of the neuron that are modified during training and $\sigma$ is a non-linear activation function such as sigmoid, tanh or ReLU~=~$\text{max}({0,.})$.

\begin{figure}[!h]
\caption{Schema of an artificial neuron.}
\label{fig:02:neuron}
\centering\includegraphics[width=\textwidth]{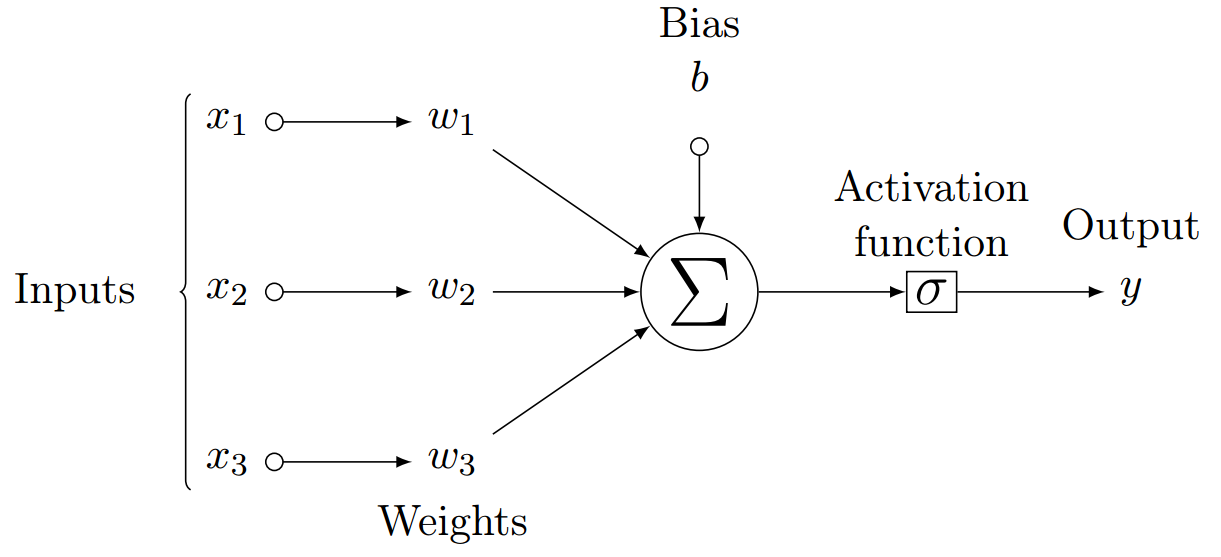}
\end{figure}

The simplest deep architecture is the\textbf{ Multilayer Perceptron (MLP)}, a multi-layered, fully-connected, feed forward neural network.
It means that all the neurons of a current layer are taken into account for the calculation of the next layer’s ones. 
The introduction of non-linear activation functions between layers enables to capture non-linear patterns by stacking several layers.

\begin{figure}[!h]
\caption{Structure of a 1-hidden layer MLP.}
\label{fig:02:neuron}
\centering\includegraphics[width=0.5\textwidth]{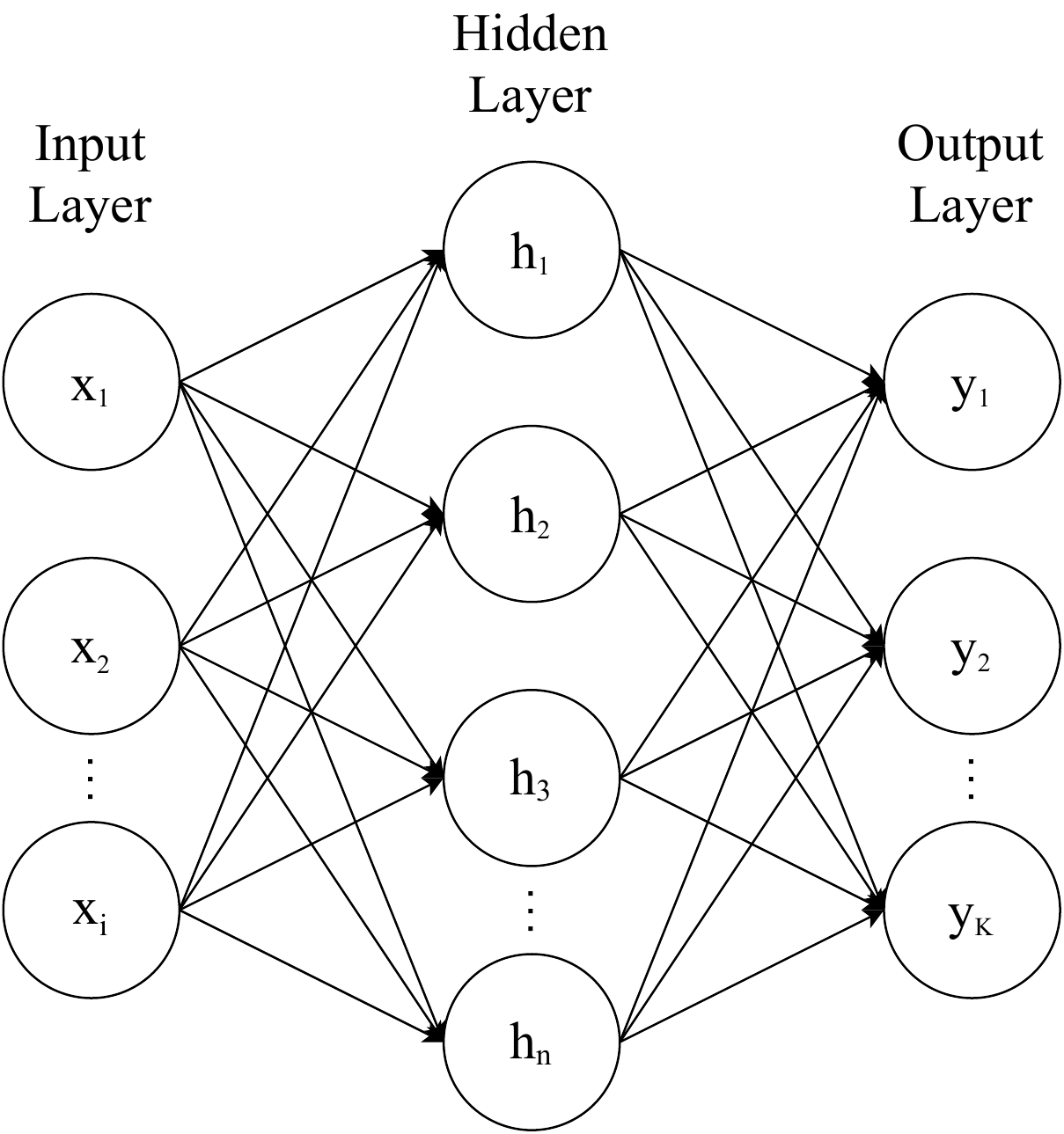}
\end{figure}

For classification, the size of the output layer is often taken as the number of different classes $K$.
It enables to represent the ground truth label $\textbf{y}$ as a one-hot vector which elements are zeros except at the index corresponding to the correct class where it is one. To model the prediction, the last layer often uses the \textbf{softmax} function to normalize the scores corresponding to each class so that they have the properties of a probability distribution.
Assuming that before softmax, the output scores are stored in a vector $\textbf{s}$ of size $K$, the final predicted output \textbf{$\hat{y}$} is computed as 

\begin{equation*}
\forall k \leq K, \quad \hat{y}_k = \text{softmax}(\textbf{s})_k = \frac{exp(s_k)}{\sum_{k'=1}^K{exp(s_{k'})}}.
\end{equation*}

\subsection{Gradient Descent Optimization}

The parameters of a neural network are usually randomly initialized and learned during a training phase.
Given pairs of inputs and associated ground truth labels (\textbf{$x^{i}$},~\textbf{$y^{i}$}), they must be adjusted so that the outputs $\hat{y}^{i}$ are close to the ground truth $y^{i}$ and the \textbf{loss function $\mathcal{L}(\hat{y}^{i}, y^{i})$}, that must be \textbf{differentiable} and reflect the difference between the prediction and ground truth, is minimized.

The classical loss function used for classification is \textbf{Cross-Entropy}, which can be defined as 
\begin{equation*}
 \mathcal{L}(y,\hat{y}) = H(y,\hat{y}) = -\sum_{k=1}^K{y_k log(\hat{y}_k)} \quad \text{where $K$ is the number of classes}
\end{equation*}

Parameters tuning is performed with \textbf{Gradient Descent Optimization} algorithms: the gradient of the loss function with respect to the parameters \textbf{$\theta$} is estimated with \textbf{backpropagation} and the parameters are updated in the opposite direction to decrease the loss function 
\begin{equation*}
\theta_{t+1} = \theta_{t} - \eta \nabla_\theta \mathcal{L}(\theta) \quad \text{where $\eta$ is the \textbf{learning rate}}
\end{equation*}

In classical Gradient Descent, the gradient  $\nabla_{\theta} \mathcal{L}(\theta)$ is computed on the entire training set before every update which can be very slow and cannot be used for online training.

An alternative is \textbf{Stochastic Gradient Descent (SGD)} where an update is performed for each training example which is must faster and enables to learn online. 
The gradient is estimated for individual training examples as $\nabla_{\theta} \mathcal{L}(\theta; \hat{y}^i, y^i)$.
However this leads to high variance in the estimation of gradient which can be beneficial to jump out of local minima but can also complicate convergence to the global minimum.

Hence, \textbf{Mini-batch Gradient Descent} has been proposed of as an efficient trade-off to compute the gradient on a small subset of the training data $\nabla_{\theta} \mathcal{L}(\theta; \hat{y}^{(i:i+bs)}, y^{(i:i+bs)})$ to reduce the variance of parameter updates.
Furthermore, recent deep learning libraries use efficient GPU-based implementation that enable computation parallelization tailored for mini-batch processing.

Numerous variants of these algorithms have been proposed to improve optimization.
We can cite for example \textbf{Momentum} \citep{Qian1999OnAlgorithms} that proposes to keep a fraction of the previous update direction to reduce variations in gradient computed from one time step to another.
\begin{align*}
v_t &= \gamma v_{t-1} + \eta \nabla_\theta \mathcal{L}(\theta)\\
\theta_{t+1} &= \theta_{t} - v_t
\end{align*}

Another popular algorithm is \textbf{Adam} \citep{Kingma2015Adam:Optimization} that computes adaptative learning rates for each parameter.
It implements a mechanism similar to momentum with bias-corrected exponential moving averages of past first and second moments $\hat{m}$ and $\hat{v}$.
\begin{align*}
m_t &= \beta_1 m_{t-1} + (1 - \beta_1) \nabla_\theta \mathcal{L}(\theta) \quad &\hat{m}_t = \frac{m_t}{1-\beta_1^t}\\
v_t &= \beta_2 v_{t-1} + (1 - \beta_2) \nabla_\theta \mathcal{L}(\theta)^2 \quad &\hat{v}_t = \frac{v_t}{1-\beta_2^t}\\
\theta_{t+1} &= \theta_{t} - \frac{\eta}{\sqrt{\hat{v}} + \epsilon} \hat{m}_t 
\end{align*}

These gradient descent algorithms are used to train deep neural networks, from simple Multi Layer Perceptron to more complex architectures further introduced in this section.

\subsection{Convolutional Neural Networks}
\textbf{Convolutional Neural Networks (CNN)} \citep{Lecun1995ConvolutionalTime-Series} have the same feed-forward structure than MLPs. 
However, they differ because of their \textbf{locally-connected} architecture, taking advantage of the structure of data such as images or texts. 
Such data can be analyzed at different scales, each structure being built with components of a smaller scale that follows properties such as translation invariance.
In an image, objects are made of smaller parts, in turn made of textures, in turn made of edges and colors.
A textual document is made of sentences, in turn made of chunks, in turn made of tokens.
CNNs are built following this observation and the deeper the layer the larger the \textbf{receptive field}, part of the original data that influence the neuron.

In a convolutional layer, a neuron is only connected to neurons of the previous layer corresponding to the same neighbouring area in the original input. 
Weights of these local connections form small \textbf{filters}, also called kernels, shared throughout the whole input via the convolution operation which ensures translation invariance. 
Finally, additional pooling layers are used to reduce the spatial dimension of the data as it goes deeper in the network, typically dividing the resolution by 4 at each layer for images, to significantly increase the size of receptive field of higher layers. 
The last layers are often fully connected which enables to use information from the entire input in the final softmax classification layer.

\begin{figure}[!h]
\caption{Architecture of AlexNet. Figure from \citep{Krizhevsky2012ImageNetNetworks}.}
\label{fig:02:alexnet}
\centering\includegraphics[width=\textwidth]{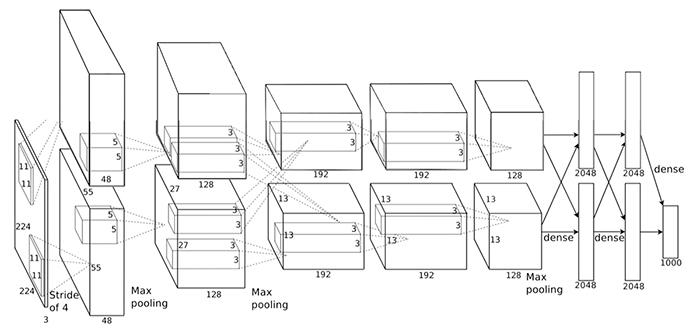}
\end{figure}

\begin{figure}[!h]
\centering
\caption[Visualization of activation of neurons at different layers of a derivative of AlexNet.Figure adapted from \citep{Zeiler2014VisualizingNetworks}.]{Visualization of activation of neurons at different layers of a derivative of AlexNet. For each neuron it shows a representation of its triggers in the 9 images of the dataset that activate it the most. We can distinguish that low-level features combine in higher layers to create neurons activated by car wheels or dog faces. Figure adapted from \citep{Zeiler2014VisualizingNetworks}.}
\label{fig:02:zeiler}
\includegraphics[width=\textwidth]{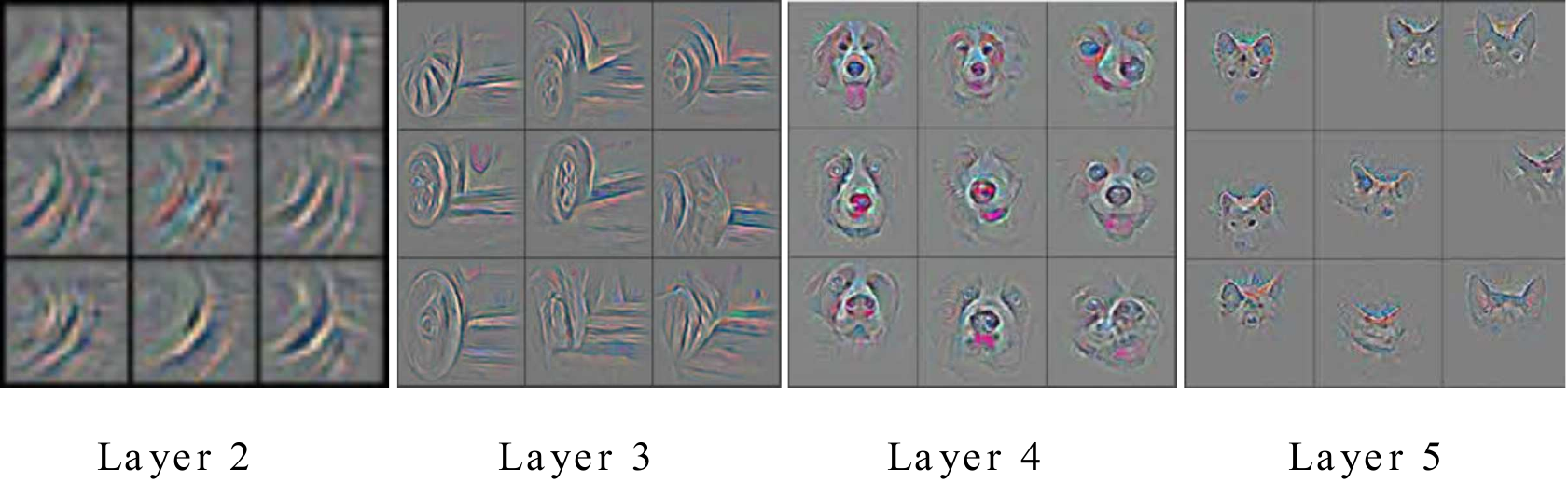}
\end{figure}

The most spectacular example of the hierarchical feature extraction capacity of CNN can be seen in Computer Vision for which they were originally conceived.
\citet{Zeiler2014VisualizingNetworks} developed a way to visualize the patterns learned by high-level layers  as shown in \autoref{fig:02:zeiler}. 
While first layers show low-level filters responding to edge orientation and color opponency, high-levels can respond to very specific objects like a particular race of dog.

Convolutional Neural Networks were originally designed to process images, with square filters and 2D convolutions reflecting horizontal and vertical translation invariance.
However, 1D CNNs can be used on sequences, in particular on textual inputs which are sequences of words \citep{Collobert2008ALearning, kim-2014-convolutional}.
Because unlike datasets like ImageNet \citep{Russakovsky2015ImageNetChallenge} with images of same dimensions, sentences are sequences of words of different sizes some null tokens are added at the beginning of shorter sentences to obtain input of same size in an additional \textbf{padding} step.

\begin{figure}[!h]
\centering
\caption[Illustration of a 1D CNN for sentence classification. Figure from \citep{kim-2014-convolutional}.]{Illustration of a 1D CNN for sentence classification.
A representation for each token is obtained with filters taking into account a local neighborhood of two (red) or three (yellow) tokens.
They are aggregated with a pooling function into a global sentence representation.
Figure from \citep{kim-2014-convolutional}.}
\label{fig:02:kim}
\includegraphics[width=\textwidth]{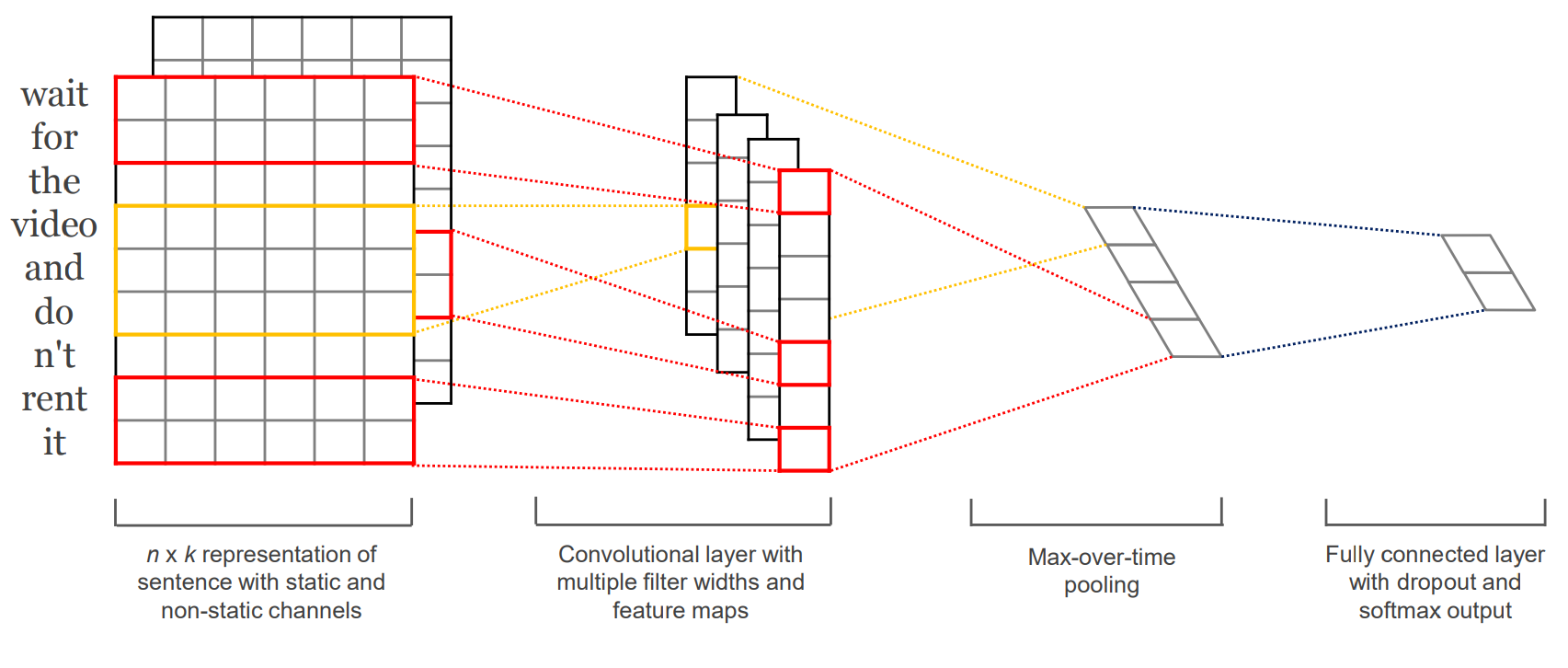}
\end{figure}

Nevertheless, another type of architecture has been proposed to process sequential inputs with varying lengths like videos or texts: Recurrent Neural Networks.

\subsection{Recurrent Neural Networks and LSTM} 
\label{sec:02:lstm}

The main feature of \textbf{Recurrent Neural Networks (RNN)} \citep{Rumelhart1986LearningErrors} is their loop structure which enables them to process sequences of different sizes by adapting the number of iterations of the loop.
For each element of the input sequence $x_t$, an output $h_t$ can be computed based on $x_t$ as well as the previous output $h_{t-1}$.
\begin{align*}
h_t &= tanh(W.[h_{t-1},x_t] + b)
\end{align*}

They are still similar to a classical neural network, as we can see by unrolling the network where the loop can be interpreted as a succession of the same layer passing its output to its successor. 
Unrolling enables to perform the backpropagation algorithm in what is called Back-Propagation Through Time (BPTT). 
However, as introduced by \citet{Hochreiter1998TheSolutions}, \textbf{vanishing or exploding gradient} issues prevent the network from learning long-term dependencies between distant inputs in the sequence.
Indeed, backpropagation is done through time and the gradient at a step is essentially obtained by multiplying the gradient at the previous step by the weights of the recurrent unit $W$.
Hence, the gradient at time step $t+l$ is roughly obtained by multiplying the gradient at time $t$ by $W^{l}$ and its norm exponentially decreases (resp. increases) if $\lVert W \rVert < 1$ (resp. $> 1$).

\begin{figure}[!h]
\caption{An unrolled Recurrent Neural Network, $x_t$ is the input at step $t$ and $h_t$ the corresponding output. Figure from \citep{Olah2015UnderstandingNetworks}.}
\label{unrolled}
\centering\includegraphics[width=0.8\textwidth]{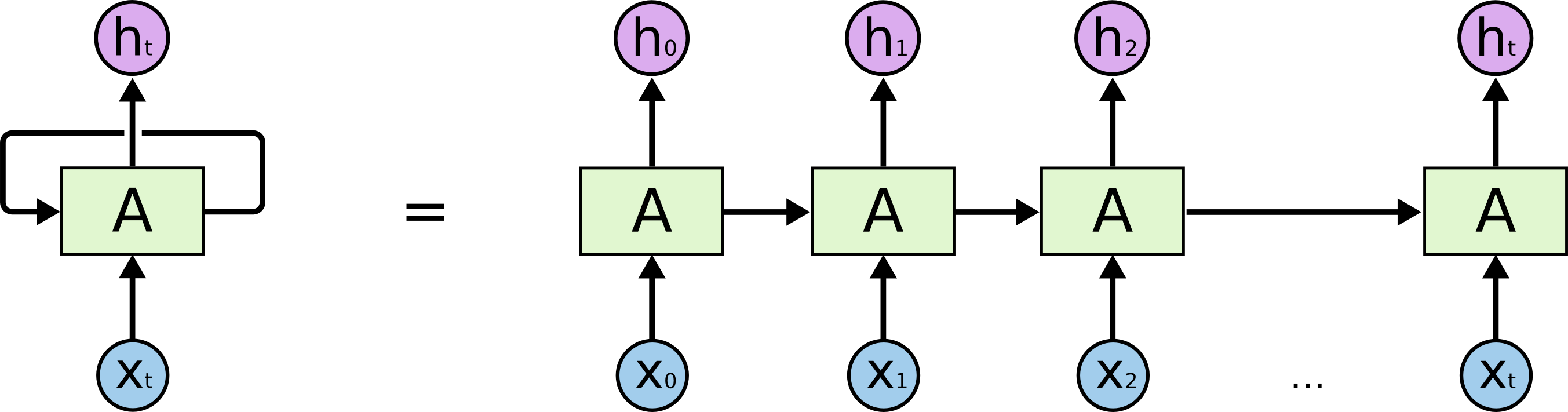}
\end{figure}

\begin{figure}[!h]
\caption{A LSTM. The three $\sigma$ represent sigmoid activation gates and correspond to the forget, input and output gates.
Figure from \citep{Olah2015UnderstandingNetworks}.}
 \label{fig:02:lstm}
\centering\includegraphics[width=0.8\textwidth]{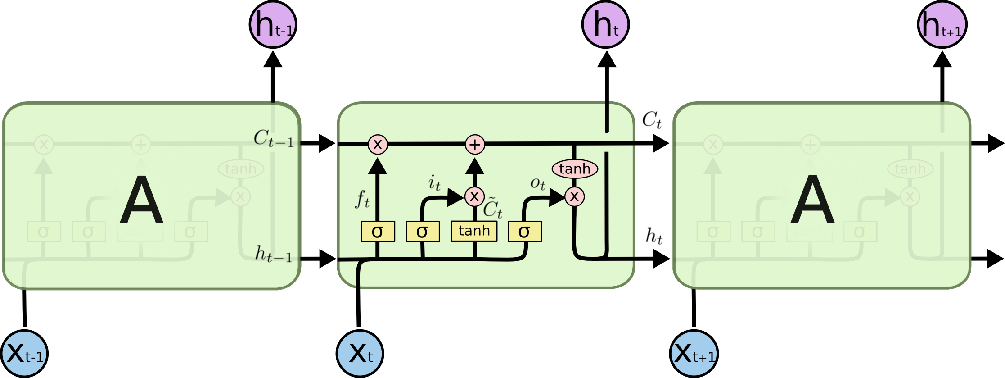}
\end{figure}

To tackle this issue, \citet{Hochreiter1997LongMemory} introduced \textbf{Long Short-Term Memory networks (LSTM)}. 
Additionally to the classical non-linear activation function contained in the loop, the LSTM contains three gates enabling it to keep or remove information stored in a cell state $C_t$ designed to retain long-term information. Hence the weights of the gates are tuned to learn which part of the input $x_t$ (input gate) and the previously stored cell state $C_{t-1}$ (forget gate) to keep in the current cell state $C_t$. Finally the output gate learns which part of the input $x_t$ to use in combination with $C_t$ to compute the output $h_t$ (see \autoref{fig:02:lstm}). 

\begin{figure}[!h]
\caption{Vanilla Recurrent Neural Network. Figure from \citep{Olah2015UnderstandingNetworks}.}
\label{fig:02:rnn}
\centering\includegraphics[width=0.8\textwidth]{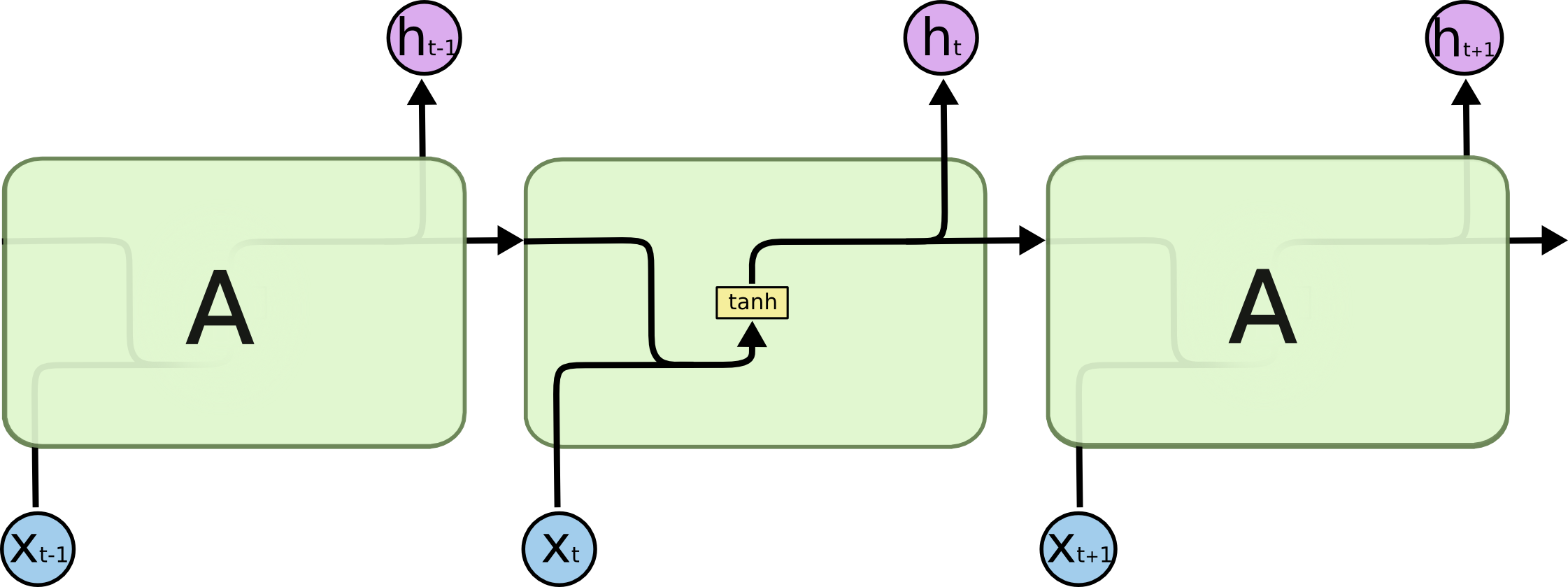}
\end{figure}

The behaviour of a LSTM can be summed up with the following equations:
\begin{align*}
f_t &= \sigma(W_f.[h_{t-1},x_t] + b_f)\\
i_t &= \sigma(W_i.[h_{t-1},x_t] + b_i)\\
\tilde{C}_t &= \tanh(W_C.[h_{t-1},x_t] + b_C)\\
C_t &= f_t*C_{t-1}+i_t*\tilde{C}_t\\
o_t &= \sigma(W_o.[h_{t-1},x_t] + b_o)\\
h_t &= o_t*tanh(C_t)
\end{align*}

Regardless of their architecture, deep neural networks have been used as an effective way to learn models from data while minimizing human efforts to design effective representations.
In the case of Natural Language Processing, it enabled to move away from handcrafted word representations to learned ones.

\section{The Evolution of Word Representations}
\label{sec:02:word representations}

The first step to build a mathematical model of a phenomenon is to determine which type of information is available and relevant in the description of such phenomenon. 
Likewise, the first step in Machine Learning is to define the \textbf{representation} of the data that an algorithm can access.
While this step was primarily undertaken by human experts that built handcrafted features, this \textbf{feature engineering} step has been replaced by \textbf{representation learning}. 
This consists in minimising the use of prior human knowledge and letting a Machine Learning algorithm discover the optimal representations from raw data.
Along with the hierarchical multi-layered structure of neural networks, representation learning is a  key factor of the recent successes of \textbf{Deep Learning} \citep{LeCun2015DeepLearning} in numerous applications from Computer Vision to Speech and Language Processing. 

In Natural Language Processing, in particular, text representation also evolved from handcrafted features to more and more complex learned representations.
Although depending on the final application, we would have to represent a simple word or an entire document, most systems use word-level representations. 
This implies a \textbf{word tokenization} step that splits a text into tokens: signs that include words or punctuation symbols.

\subsection{Bag of Words}

A classical method is to consider words as atomic units and to compute representations based on word counts.
Given a fixed vocabulary of possible words $V$, a word $w_i \in V$ can thus be represented with a \textbf{one-hot representation}, a vector of dimension $|V|$ where the i$^{th}$ component is 1 and every other component is 0.
A document can then be modeled without taking word order into account with a representation called \textbf{Bag of Words} and aggregating one-hot representations of the words it contains, for example with a one-hot representation with a 1 at the component corresponding to each word.

These representations can be useful to model similarity between documents based on the assumption that similar documents contain the same subsets of words.
They can be extended to word frequencies, possibly normalized to give more importance to words occurring more rarely like in the TF-IDF model \citep{SparckJones1972ARetrieval}.

However, although such representations have been shown useful associated to Naive Bayes or Logistic Regression classifiers in early NLP models, they suffer from several shortcomings.
First, one-hot representations are \textbf{sparse high-dimensional vectors}, with vocabularies often in the order of magnitude of tens of thousands to millions of different words. 
Second, using words as atomic units prevents this discrete representation from modeling semantic similarity between them since they are \textbf{all orthogonal pairwise}.

\subsection{Lexicons}
A naive approach to represent word similarity is to use \textbf{lexicons} to compute handcrafted features.
For example in \textbf{polarity classification}, a framing of \textbf{sentiment analysis} where a document, typically a review, must be classified as positive or negative, we can maintain lists of words annotated with positive (e.g. good, love, beautiful) or negative sentiment.  
We can then use occurrences of each class as an additional feature \citep{Hu2004MiningReviews}.
Such lexicons have also been used in Named-Entity Recognition: \textbf{gazetteers} are lists regrouping names of geographic locations or people for example \citep{florian-etal-2003-named}.
However, such an approach is limited since it requires to maintain a list for each class of words we want to model. 

\subsection{Distributional Semantics}
Another approach to bridge the semantic gap between words as symbols and their meaning is to hypothesize that some information on the meaning can be accessed through statistics.
Following the \textbf{distributional hypothesis} that ``a word is characterized by the company it keeps'' \citep{Firth1957A1930-1955}, a representation of a word can be obtained by counting its occurrences in documents among a corpus.
Word representations are obtained with the reverse of bag of words hypothesis, i.e. the hypothesis that similar words appear in the same subsets of documents. 
Such a term-document co-occurrence matrix can then be factorized to obtain a lower dimension vector representation of words using a method called \textbf{Latent Semantic Analysis} \citep{Deerwester1990IndexingAnalysis}.

This distributional hypothesis is still used to compute current word representations, but instead of designing these representations based on statistics, they are learned using predictive Language Models trained to capture this distribution. 

\subsection{n-gram Language Models}
Historically used for Natural Language Generation tasks such as sentence completion, spelling correction, translation, Optical Character Recognition or Speech Recognition, \textbf{Language Models} are now a key representation learning component of every modern NLP system.

They are probabilistic models designed to predict the words most likely to appear next, given the beginning of a sequence.
It must hence estimate the probability that each word in a vocabulary at position $k$ given the previous sequence of words $w_{1:k-1}$: $P(w_{k} | w_{1}, ..., w_{k-1})$.

Early Language Models make the simplifying \textbf{Markov chain approximation} \citep{Markov1913EssaiChane} that the appearance of a word is uniquely conditioned on the n-1 previous words and are called \textbf{n-gram Language Models} \citep{Jurafsky2020SpeechModels}.

Thus,
\begin{equation}
    P(w_{k} | w_{1}, ... w_{k-1}) \approx P(w_{k} | w_{k-n}, ... w_{k-1})
\end{equation}

Where this can be statistically approximated in a sufficiently large training corpus by counting the occurrences of n-grams, typically limited to n~$\leq$~5, since the average number of occurrences of an n-gram statistically drops when n increases.

\begin{equation}
    P(w_{k} | w_{k-n}, ... w_{k-1}) \approx \frac{count(w_{k-n}, ..., w_{k})}{count(w_{k-n}, ..., w_{k-1})}
\end{equation}

Again, a limitation of count based methods is to take words as atomic units and being unable for example to model the semantic or syntactic similarity of words.

\subsection{Neural Language Models and Pretrained Word Embeddings}
Instead of n-gram counts statistics, \citet{Bengio2003AModel} propose to use a neural network as a Language Model (LM).
The neural network is designed to learn both ``distributed word feature vectors'' jointly with ``the probability function for word sequence''.
Each word is associated with a continuous real-vector of dimension ($\sim$ 10-100) orders of magnitude smaller than the size of the vocabulary ($\sim$ 10 000).
These vectors are fed to a Multi Layer Perceptron (MLP) neural network that models a n-gram Language Model, outperforming previous statistical models on the Brown corpus \citep{Francis1979BrownManual}.
Such representations, referred to as \textbf{word embeddings}, enable to fight the \textbf{curse of dimensionality} and to model word similarity in a dense vector space.

\citet{collobert-weston-2007-fast, Collobert2008ALearning, Collobert2011NaturalScratch} then pretrain a neural Language Model to obtain word embeddings and use them in several NLP tasks, in a Multi-Task learning setting, providing the first demonstration of the \textbf{Transfer Learning} capabilities of Language Model pretraining.
Furthermore, they introduce two key differences with \citet{Bengio2003AModel}'s neural Language Model.
First, because the objective is to learn word representations, the LM can use both the \textbf{context before and after the predicted word}.
Second, instead of casting LM as a classification task over every word in the vocabulary, they model it as a binary classification task thanks to \textbf{Negative Sampling}.
Given the context, the actual corresponding word and a different random word, the network is trained to rank the positive and negative examples.
The embeddings obtained in their method are referred to as \textbf{SENNA}.

What later popularized word embeddings pretraining is the \textbf{word2vec} framework that further reduces the computational cost of pretraining \citep{Mikolov2013EfficientSpace, Mikolov2013DistributedCompositionality}. The \textbf{Skip-Gram with Negative Sampling} (SGNS) model proposes two variations for improved efficiency. First, as suggested by the name, the Language Model is simplified to estimate the probability that a word occurs in the context of a center word (typically a window of $\leq~10$ words). This similarity between two words is simply estimated as the \textbf{dot product} of their representations. Second, \textbf{Negative Sampling} consists in selecting k (typically~$\leq~20$) random negative words for each positive (center word, context word) pair so that the computational cost is largely reduced compared to computing a cost for each word in the vocabulary.
The availability of efficient implementations as well as simple geometric interpretations of semantic and syntactic similarities in the embedding space, such as the notorious "king - man + woman = queen" helped popularize embedding pretraining \footnote{Several later studies \citep{levy-goldberg-2014-linguistic, nissim-etal-2020-fair, fournier-etal-2020-analogies} question the validity of these experiments since the original word ("king" in our example) is manually removed from the target space.}.

Another popular choice of pretrained embebddings is \textbf{GloVe (Global Vectors)} \citep{pennington-etal-2014-glove}, which later unifies Global Matrix Factorizaton methods such as Latent Semantic Analysis (LSA) and local context window models such as Skip-Gram. It learns an embedding matrix in a log-bilinear model that approximates global co-occurence statistics of words inside a fixed-size window.

\subsection{Character-level and subword representations}
Although pretrained word embeddings led to breakthroughs in Deep Learning applied to NLP, word-level representations are \textbf{lexical representations} : each word in a vocabulary is mapped to a dense vector.
This implies two shortcomings: 1) a word not present in the training corpus, referred to as \textbf{out-of-vocabulary (OOV)}, has no learned representation 2) these representations cannot capture \textbf{morphological information}, for example contained in affixes.

Thus, character-based representations have been introduced to model this morphological information for tasks where it seems particularly important such as \textbf{Part-of-Speech Tagging} \citep{ling-etal-2015-finding}, \textbf{Dependency Parsing} \citep{ballesteros-etal-2015-improved} or \textbf{Named-Entity Recognition} \citep{lample-etal-2016-neural}.
In these models, for each word, a Convolutional Neural Network or a LSTM is fed with embeddings of its characters to learn a character-based representation in addition with the traditional word-level representations. They are often denoted \textbf{charCNN or charLSTM} representations.
The obtained vocabulary of characters is then limited to the alphabet, numbers and punctuation symbols, which reduces the chance of encountering OOV symbols. 

\textbf{FastText} embeddings \citep{bojanowski-etal-2017-enriching} are another example of how morphology can be taken into account in word representations. FastText is an extension of the word2vec skip-gram model to \textbf{character n-grams} that represents a word as the sum of the representations of its character n-grams.

In between word and character representations, \textbf{subword tokenization} was introduced to limit the size of the vocabulary that is tied to the number of parameters of neural models in Natural Language Generation tasks.
Different variants have been proposed, inspired by \textbf{Byte Pair Encoding (BPE)} \citep{Gage1994ACompression}.
The vocabulary is built recursively from the initial set of characters.
The most frequent bigram of vocabulary units is merged and added to the vocabulary and the process is repeated until reaching a predefined size (typically 8000 tokens).
\textbf{WordPiece} \citep{Wu2016GooglesTranslation} follows the same idea but the bigram is not chosen as the most frequent but as maximizing the likelihood of a language model on the training data when added to the vocabulary.

\subsection{Pretrained Language Models and Contextual Word Embeddings}
An additional issue of previous lexical or morphological representations is that they do not incorporate contextual information, for example useful when encountering \textbf{polysemous words}.
Indeed, once pretrained, word or character embeddings are \textbf{static} and do not depend on the context of the considered word.
For example, in the sentence ``Georges Washington lived in Washington D.C.'', the two occurrences of ``Washington'' would share the same representation whereas they refer to different real-world entities. 
If the word Washington refers most of the time to the US president in the training corpus, its representation should be closer to representations of other people or even US presidents in the embedding space, otherwise it should be closer to other capital cities.

Based on this observation, it is interesting to have a representation that depends on the context and is able to disambiguate when a word has several senses.
\citet{peters-etal-2017-semi} propose to use a ``Language Model embedding'' in addition to the classical SENNA word embedding in the \textbf{TagLM} model for Named-Entity Recognition.
The idea is simple: instead of only using the first embedding layer of a pretrained Language Model that maps each word in a vocabulary to a vector, we can use the full prediction capability of the LM.
Hence, in our previous example we can expect that the LM might learn that ``Georges'' and ``lived'' appear in the context of people while ``in'' is followed by locations.
This idea is the basis for obtaining word representations that depend on the context that are called \textbf{Contextual Word Embeddings}.

Concurrently, other tasks have been explored for pretraining inspired by the simple and efficient Transfer Learning paradigm in Computer Vision that consists in training
a Convolutional Neural Network for Image Classification on ImageNet \citep{Russakovsky2015ImageNetChallenge} and
finetuning it with only the last few layers modified \citep{Oquab2014LearningNetworks}.
In contrast, only the first word embeddings layer was traditionally transfered in NLP.
Hence, \citet{Mccann2017LearnedVectors} propose to pretrain an attentional BiLSTM Seq2Seq model on Neural Machine Translation from English to German.
The encoder part that treats English input text is then used to obtain contextual representations named \textbf{Context Vectors (CoVe)} that are combined
with GloVe embeddings to improve performance on several tasks such as Natural Language Inference (NLI), Semantic Similarity and Question Answering.
\citet{conneau-etal-2017-supervised} use NLI for pretraining a BiLSTM network to obtain sentence representations that can be transfered
in other sentence-level NLP tasks such as Semantic Similarity or Polarity Classification but also in a multimodal Image or Caption Retrieval setting.

Language Models have also been examined for transfering representations to other tasks.
\citet{Radford2017LearningSentiment} train a character-based LSTM Language Model on Amazon reviews and show that the activation
of a single unit, the "\textbf{sentiment neuron}", can be used to predict the polarity of a review on another domain such as IMDB more effectively than a
supervised Naive Bayes model.
\citet{howard-ruder-2018-universal} propose to pretrain a multi-layer LSTM model for Language Modeling and
then finetune it for several text classification tasks in a paradigm they coin \textbf{Universal Language Model Finetuning (ULMFiT)}.
They obtain new state-of-the-art results on six datasets, outperforming the previous CoVe representations and show an impressive
sample efficiency of such finetuning compared to training the whole network from scratch with randomly initialized weights.

\citet{peters-etal-2018-deep} then refine their TagLM model in \textbf{ELMo} (Embeddings from Language Models).
They replace the rather old word-level SENNA representations with a charCNN embedding layer and the bidirectional
Language Model is implemented with a forward and a backward 2-layer LSTM.
This papers also shows the effectiveness of such representations by improving the state-of-the-art on benchmarks
representative of a larger range of NLP tasks: Question Answering, Natural Language Inference, Semantic Role labeling,
Coreference Resolution, Named-Entity Recognition and Sentiment Analysis.

Shortly after ELMo, two influential works propose to use the recent \textbf{Transformer} architecture \citep{Vaswani2017AttentionNeed} which had been shown effective in Neural Machine Translation. 
\textbf{GPT} (Generative Pretrained Transformer) \citep{Radford2018ImprovingPre-Training} use the decoder part of the Transformer to pretrain a traditional autoregressive Language Model and then fine-tune this network with an additional classification layer for QA, NLI, Semantic Similarity and Text Classification.
Then, \citet{devlin-etal-2019-bert} propose \textbf{BERT} (Bidirectional Encoder Representations from Transformers) that uses the encoder part of the Transformer for \textbf{Masked Language Modeling} at the \textbf{subword-level}.
Contrary to GPT, BERT has access to both the left and right contexts of the predicted subword for Language Modeling.
Thanks to the computational efficiency of the Transformer architecture, GPT and BERT can be pretrained on larger corpora and show additional quantitative improvements over ELMo on a large number of tasks.
This increased effectiveness, as well as Huggingface's initiative to implement BERT and its variations in an accessible transformers library \citep{wolf-etal-2020-transformers}, led BERT to be the new de facto state-of-the-art baseline in virtually every NLP tasks, including Named-Entity Recognition and Relation Extraction.

This is why we dedicate the rest of this chapter to the introduction of the underlying Transformer architecture (\autoref{sec:02:transformer}), a more detailed description of BERT and some of its subsequent variations (\autoref{sec:02:bert}), as well as some highlights in works regarding Bertology  (\autoref{sec:02:bertology}).

\section{The Transformer}
\label{sec:02:transformer}

The \textbf{Transformer} is a neural network architecture originally introduced for \textbf{Neural Machine Translation} (NMT) and only relying on attention and fully-connected layers \citep{Vaswani2017AttentionNeed}.
It was proposed to reduce the computational complexity of the widely used \textbf{recurrent or convolutional mechanisms}, which enables to train on larger sets of data in a realistic time.
This enabled to improve state-of-the-art results mainly in Neural Machine Translation \citep{Vaswani2017AttentionNeed} and Language Modeling \citep{Radford2018ImprovingPre-Training, Radford2020LanguageLearners},  which are tasks where training resources are important relatively to other supervised ones.

The Transformer follows the classical encoder-decoder structure with classical \textbf{attention} (\autoref{sec:02:transformer:attention}).
Its main specificity is in the architecture of both the encoder and the decoder, where the recurrent mechanism is replaced by \textbf{Multi-Head Self-Attention} (MHSA) (\autoref{sec:02:transformer:mhsa}).

\subsection{Attention}
\label{sec:02:transformer:attention}
\textbf{Attention} is a Deep Learning mechanism inspired by human visual attention that enables us to focus our gaze only on the parts of our environment that seem the most relevant and filters the amount of information our brain needs to process. 
It has been notably introduced in image captioning \citep{Xu2015ShowAttention} to select relevant parts of an image and Neural Machine Translation (NMT) \citep{Bahdanau2015NeuralTranslate} to select relevant parts of an input sentence. 
For NMT, it has been used as an improvement over the \textbf{sequence to sequence model (Seq2Seq)} \citep{Sutskever2014SequenceNetworks}, which is based on Recurrent Neural Networks (RNN).
In this \textbf{encoder-decoder model}, a first RNN encodes a sentence regarded as a sequence of word embeddings into a single vector, the last hidden state of the RNN.
A second RNN, the decoder, is then conditioned on this encoded representation to retrieve a sequence of vectors that are interpreted as word embeddings.

\begin{figure}[!h]
\begin{adjustwidth*}{}{-.3\textwidth}
\caption[Schema of the Seq2seq architecture for NMT.]{Schema of the Seq2seq architecture for NMT. The input sentence is encoded into a single vector given as input to an autoregressive decoder network.}
\label{fig:02:seq2seq}
\centering\includegraphics[width=1.3\textwidth]{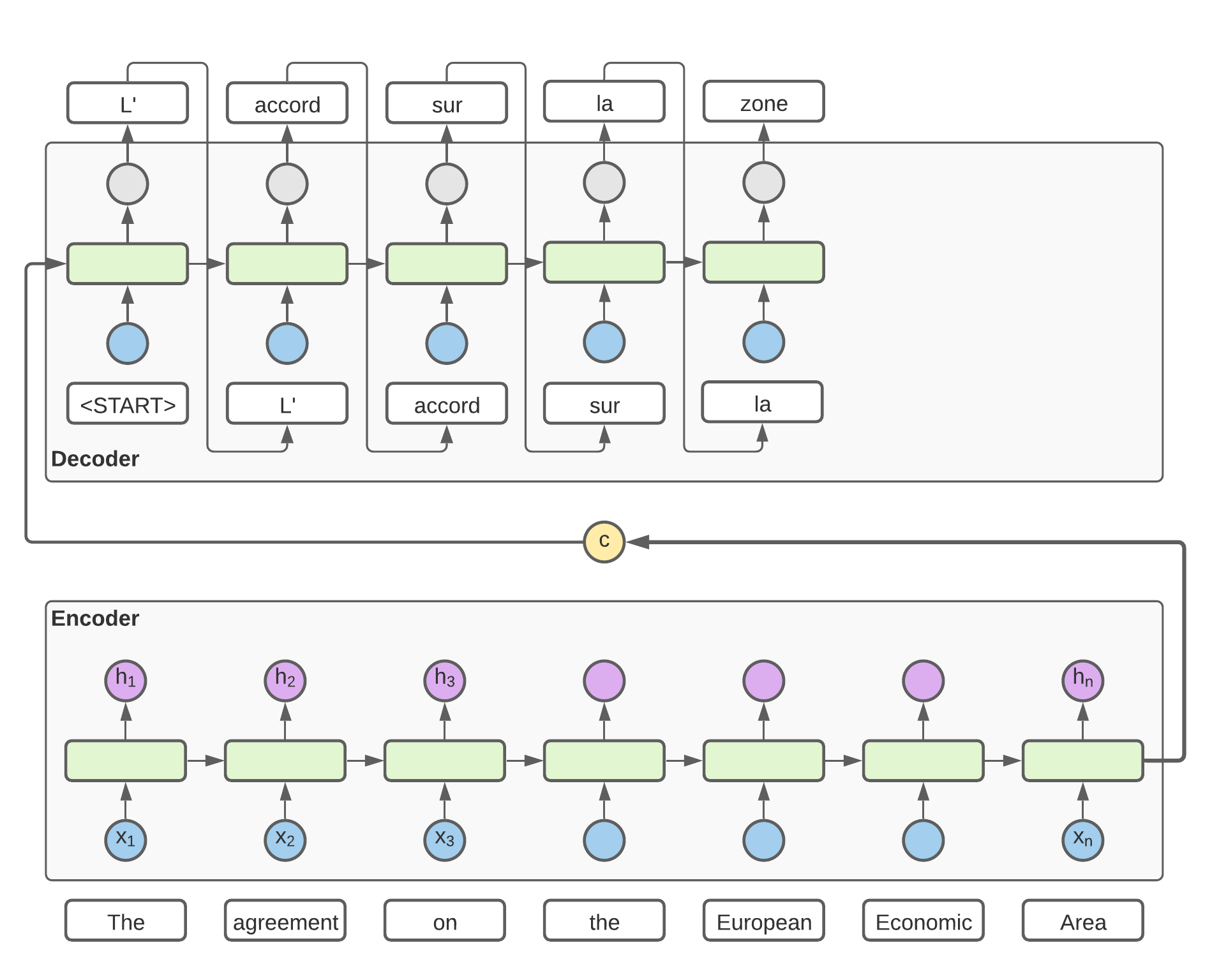}
\end{adjustwidth*}
\end{figure}

\begin{figure}[!h]
\caption[Attention weights in a EN-FR NMT model. Figure from \citep{Bahdanau2015NeuralTranslate}.]{Attention weights in a EN-FR NMT model. Each row represents the weights assigned to the words of the source sentence used to generate the corresponding output word.
We can observe that the order of words is similar except for the French "zone économique européenne" for which the order is reversed with "European Economic Area".
Figure from \citep{Bahdanau2015NeuralTranslate}.}
\label{fig:02:alignment}
\centering\includegraphics[width=0.8\textwidth]{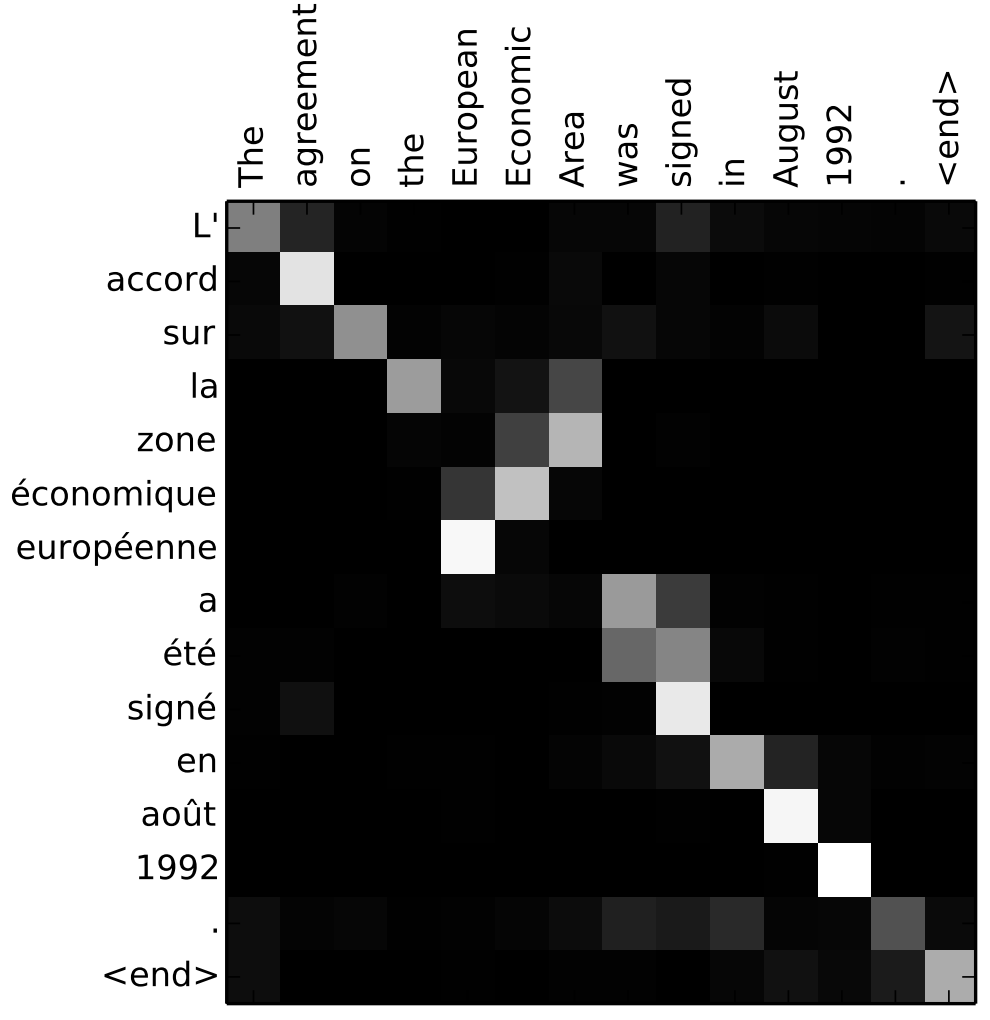}
\end{figure}

The problem is the difficulty to encode and retrieve the whole meaning of the input sentence in a single vector, especially for complex and long sentences. 
The idea behind attention in NMT is to sequentially select the part of encoded representations relevant to predict the next word.
Hence, an alignment function computes weights between each encoded representation and the current decoder state, reflecting this selection.
In practice each output word often corresponds to a single input word which shows that the network learns to align sentences, as illustrated in \autoref{fig:02:alignment}.

\begin{figure}[!h]
\begin{adjustwidth*}{}{-.3\textwidth}
\caption[Schema of the Seq2seq architecture with attention for NMT.]{Schema of the Seq2seq architecture with attention for NMT. In this model, at each decoding step the input context vector is the sum of every encoded representations weighted by an alignment score with a current query.}
\label{fig:02:bahdanau}
\centering\includegraphics[width=1.3\textwidth]{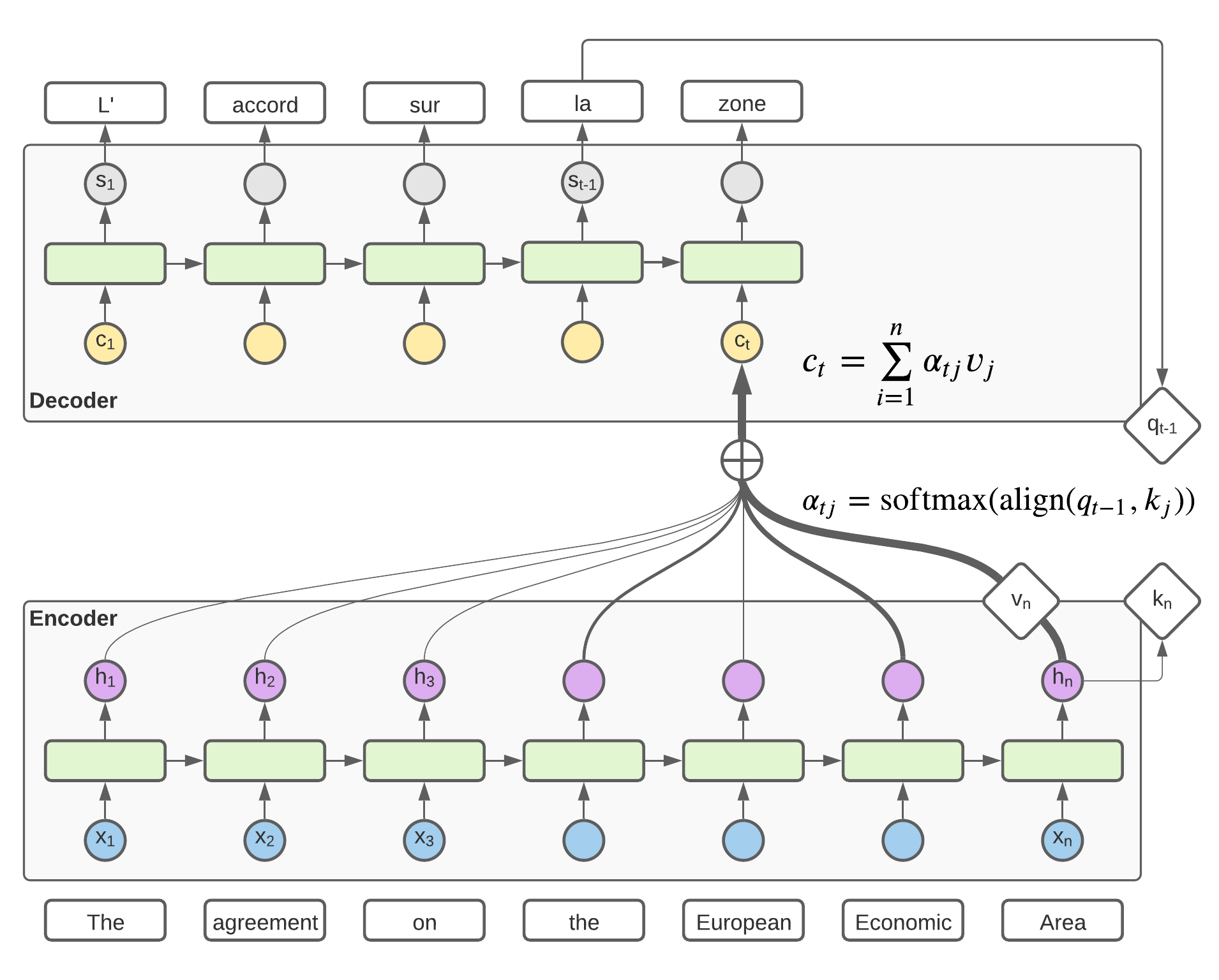}
\end{adjustwidth*}
\end{figure}

Although several implementations of attention have been proposed \citep{Bahdanau2015NeuralTranslate, luong-etal-2015-effective, Vaswani2017AttentionNeed}, they can be framed in a single setting with three types of \textbf{vectors}: \textbf{query}, \textbf{key} and \textbf{value}.

\begin{enumerate}
    \item Each token $i$ in the source input has a key $\boldsymbol{k_i}$ and a corresponding value $\boldsymbol{v_i}$.
    \item Given a query $\boldsymbol{q}$, an alignment score is computed with each input key $\boldsymbol{k_i}$:
$\alpha_{i} = align(\boldsymbol{q}, \boldsymbol{k_i})$, often normalized with a softmax function to obtain a probability distribution on the input tokens.
    \item The values $\boldsymbol{v_i}$ of the source inputs are then summed, weighted with the alignment scores into a single context vector\\
    ${\boldsymbol{c}~=~\sum_{i=1}^{n}~\alpha_{i}~\boldsymbol{v_i}}$.
\end{enumerate}

In the case of NMT, there is a query at each decoding step $t$, corresponding to the $(t-1)^{th}$ hidden state of the decoder which depends on the previous output words $y_1, ..., y_{t-1}$.

\begin{figure}[!h]
\begin{adjustwidth*}{}{-.3\textwidth}
\caption[Schema of Multihead Self Attention.]{Schema of Multihead Self Attention. For each token embedding in the input and in each attention head, a corresponding query, key and value are obtained by linear projection. Each token $i$ is represented by the sum of all input values $v_j$ weighted by the alignment of its query $q_i$ with every key $k_j$.}
\label{fig:02:mhsa}
\centering\includegraphics[width=1.3\textwidth]{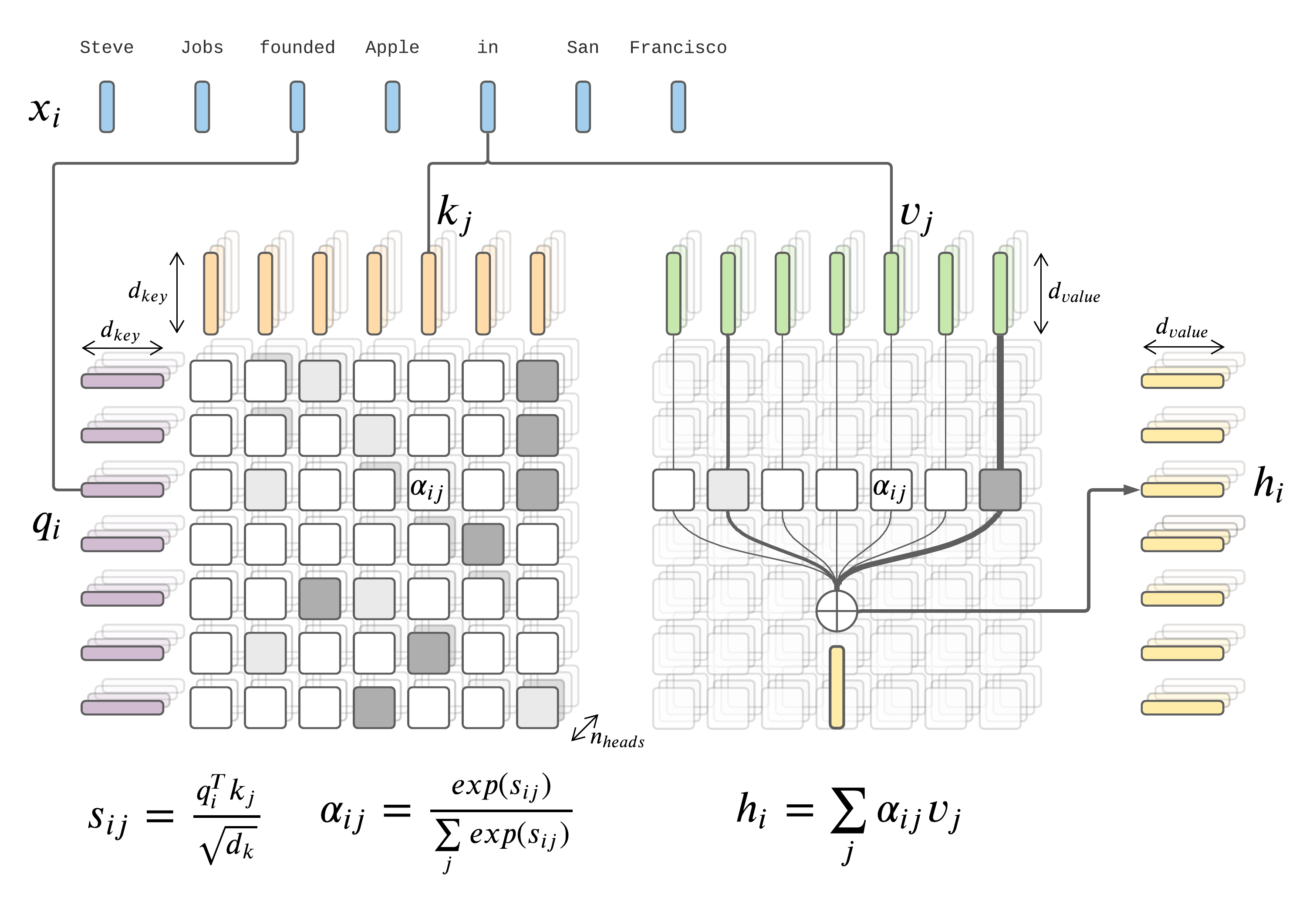}
\end{adjustwidth*}
\end{figure}

\subsection{Multi-Head Self-Attention}
\label{sec:02:transformer:mhsa}
\paragraph{Self-Attention} \textbf{Self-Attention} is simply an attention mechanism where the previously described alignment is computed between the input and itself.
Hence a set of $q_i$, $k_i$, $v_i$ corresponds to each token $i$ in the input, and the alignment is computed for each pair $(i,j)$ between $q_i$ and $k_j$. 
This enables to obtain a representation $h_i$ of each token as an aggregation of all the values $v_j$ and thus incorporate contextual information in a non autoregressive manner, without any recurrent or convolutional mechanism.

Specifically, \citet{Vaswani2017AttentionNeed} use Scaled-Dot Product attention:

\begin{enumerate}
    \item Given a sequence of input vecotrs $x_{i}$, the three vectors $q_i$, $k_i$, $v_i$ of same dimension $d_k$ are obtained with a linear projection
    \item For all $(i,j)$, the alignment is given by $\alpha_{i,j} = \frac{exp(q_i^T k_j / \sqrt{d_k})}{\sum_{j} exp(q_i^T k_j / \sqrt{d_k})}$
    \item The resulting token representation is $h_i=\sum_{j} \alpha_{i,j} v_{j}$
\end{enumerate}

This is often summed up in a single equation : 
\begin{equation}
    \text { Attention }(Q, K, V)=\operatorname{softmax}\left(\frac{Q K^{T}}{\sqrt{d_{k}}}\right) V
\end{equation}

\paragraph{Multihead Attention} \citet{Vaswani2017AttentionNeed} propose to compute $h$ (query, key, value) sets for each input words in order to ``allow the model to jointly attend to information from different representation subspaces at different positions", where $h$ stands for the number of \textbf{attention heads}. 
Each head can thus detect a different type of relevant pattern and the outputs of every heads are simply concatenated.

\paragraph{Computational Complexity}
The main advantage of such Multihead Self Attention (MHSA) architecture over Convolutional Neural Networks (CNN) or Recurrent Neural Networks (RNN) is the fact that \textbf{long range dependencies are modeled immediately at every layer} of the network because every element in the sequence interacts with all the other ones.
Furthermore, in a given layer the output for a given token can be computed independently of the other ones which enables parallelization similar to CNNs but more efficient that RNNs that are autoregressive and require $O(n)$ sequential operations where $n$ is the number of tokens.

However, this efficient time complexity is at the cost of a more \textbf{expensive memory consumption in $O(n^2)$} since an attention score must be computed and stored for every pair of words.
This leads to limiting the input length of Transformers to fixed numbers of tokens such as 128 or 512, making them suited to process sentences or paragraphs but restricting their direct use on entire documents.

\subsection{Additional Implementation Details}
\paragraph{Positional Encoding}
Because contrary to a recurrent or convolutional mechanism, there is no notion of positional structure in the attention mechanism, a \textbf{positional encoding} (a fixed function of the position in the sequence) is added to a more classical token embedding to incorporate such a positional feature.

While in the original version of the Transformer, this positional encoding resorts to trigonometric functions sin and cos, they can also be replaced by a learned \textbf{positional embedding}.

\begin{figure}[!h]
\caption{The Transformer architecture. Figure from \citep{Vaswani2017AttentionNeed}.} 
\label{fig:transformer}
\centering\includegraphics[width=.7\textwidth]{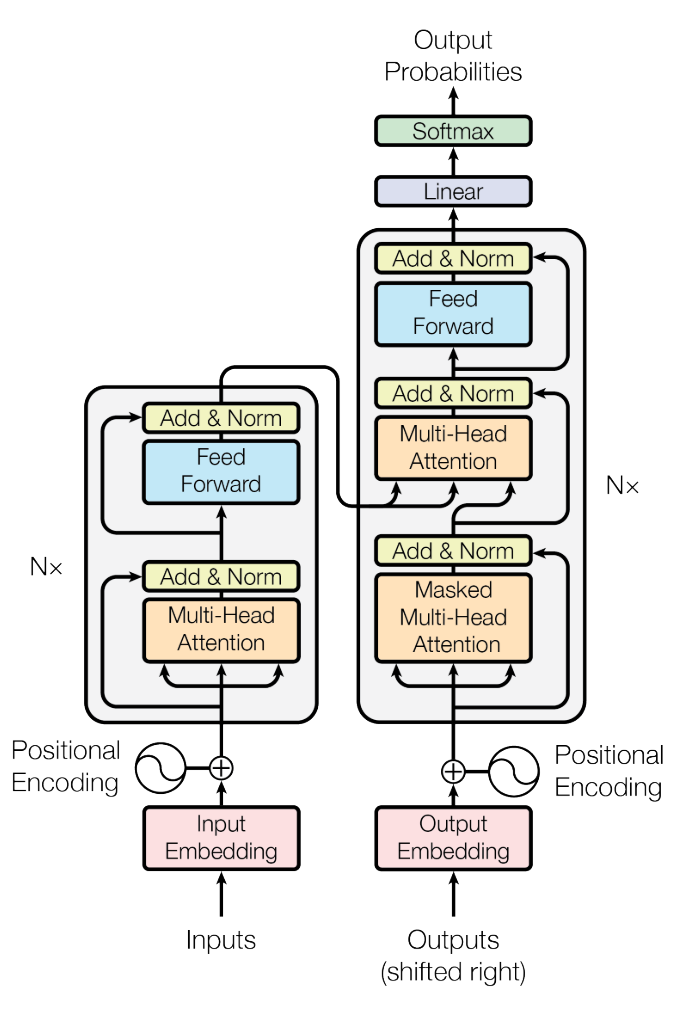}
\end{figure}

\paragraph{Network Architecture} The complete Transformer architecture follows the classical Encoder-Decoder with Attention \citep{Bahdanau2015NeuralTranslate} but without recurrent networks. Each of the Encoder and Decoder are a succession of layers consisting in Multi-Head Self-Attention followed by a \textbf{Positionwise Feed-Forward Neural Network} (the representation of each token is encoded separately). In the original paper, there are 6 layers, 8 heads and $d_k = 64$. The overall network is illustrated in \autoref{fig:transformer}.

\section{BERT and variations}
\label{sec:02:bert}

\subsection{Additional BERT implementation details}
As described in \autoref{sec:02:word representations}, \textbf{BERT} \citep{devlin-etal-2019-bert} is a model designed to obtain contextual representations from Language Model Pretraining.
More precisely, the model is the Encoder part of a Transformer and is introduced in two versions:
BERT$_{BASE}$ with 12 layers and 12 heads per layer and BERT$_{LARGE}$ with 24 layers and 16 attention heads.
While using BERT$_{LARGE}$ often enables to outperform the base version, the difference in performance is marginal and the base version is more often used in later works because it is easier to finetune.

BERT is pretrained in an self-supervised manner on two tasks: \textbf{Masked Language Modeling (MLM)} and \textbf{Next Sentence Prediction (NSP)}.
Sentences are tokenized into subwords using WordPiece \citep{Wu2016GooglesTranslation}.

For MLM, the network is trained to recover an original input sentence given a corrupted version where every token is replaced by the special token \textbf{[MASK]} with a given probability. The final hidden representations of masked tokens are fed to a softmax over the vocabulary for prediction.

\begin{figure}[!h]
\caption{BERT's input representation. Figure from \citep{devlin-etal-2019-bert}.} 
\label{fig:bert_input}
\centering\includegraphics[width=\textwidth]{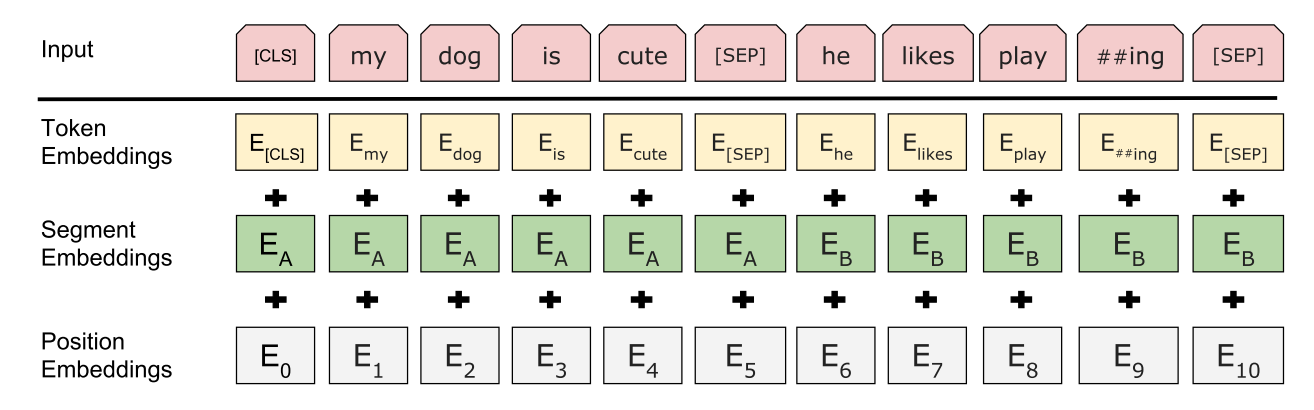}
\end{figure}

For NSP, two sentences are given to the model and it is trained to predict whether they were originally consecutive in the source corpus (binary classification).
In practice, it uses subword token embeddings to represent a sentence and sentences are separated by special tokens \textbf{[CLS]} (classification) and \textbf{[SEP]} (separation), as shown in \autoref{fig:bert_input}.
The [CLS] token's final hidden representation is the one used for binary classification, and intuitively encodes sentence-level information.

Using these representations with a simple additional linear layer and finetuning the whole model leads to new state-of-the-art results in virtually every NLP tasks.

\subsection{Highlights on some variations of BERT}
The success of BERT led to the exploration of several variations upon its architecture in different directions. We can thus cite works that tried to reduce the important pretraining cost of BERT (estimated at 280 days on a single Nvidia V100) such as \textbf{DistilBERT} \citep{Sanh2019DistilBERTLighter} that uses knowledge distillation or \textbf{ALBERT} \citep{Lan2020ALBERT:Representations} that shares parameters of Transformer layers.

On the contrary, others tried to further scale both the amount of training data and time.
\textbf{RoBERTa} \citep{Liu2019RoBERTa:Approach} is a version with a few minor tweaks on the BERT architecture and without the Next Sentence Prediction pretraining objective but mostly pretrained on ten times more data. 
\textbf{XLNet} \citep{Yang2019XLNet:Understanding} aims at representing larger contexts by using the Transformer-XL architecture and predicting a sequence in a random order, again on an order of magnitude more data.
In the same line of scaled up models, we can cite the subsequent versions of GPT: GPT-2 \citep{Radford2020LanguageLearners} and GPT-3 \citep{Brown2020LanguageLearners} that share a similar architecture but successively inflate the number of parameters from 117 M in GPT to 1.5 B in GPT-2 to 175B in GPT-3.
These new papers demonstrate the zero-shot to few-shot learning capabilities of such massive neural language models.

As soon as the introduction of BERT, Google released a multilingual version of BERT trained on texts in around 100 different languages.
While not directly discussed in the original paper, later studies reveal that it enables to perform transfer from one language to another, for example in Named Entity Recognition \citep{pires-etal-2019-multilingual}. 
Another multilingual language model study is performed with a variant called XLM \citep{Conneau2019Cross-lingualPretraining} and evaluated on Neural Machine Translation and Cross-lingual NLI.
Since the initial success of BERT, which main versions are trained on English, numerous variations have been trained secifically on other languages such as CamemBERT in French \citep{martin-etal-2020-camembert} or German BERT in German \citep{chan-etal-2020-germans}.

\section{Bertology}
\label{sec:02:bertology}
Since the introduction of BERT, several works have tried to understand or interpret the effectiveness of BERT and Transformer LM pretraining in general, a field of research referred to as Bertology \citep{rogers-etal-2020-primer}. 
We can distinguish several types of studies including \textbf{probing}, \textbf{pruning} of attention heads or layers and analysis of \textbf{self-attention patterns}.
Although we refer to \citep{rogers-etal-2020-primer} as a more exhaustive summary of previous works, we highlight some key findings in this section. 

\subsection{Probing}

\textbf{Probing} studies consist in freezing BERT representations and feeding its hidden representations at each layer into simple probe networks (often a Multilayer Perceptron) supervised for different NLP tasks.
Several studies \citep{tenney-etal-2019-bert, jawahar-etal-2019-bert} suggest that representations at lower layers of BERT are more useful for lower-level tasks that rely less on context such as Part of Speech tagging, while higher layers are necessary for task that require a deeper understanding of context such as Semantic Role Labeling or Coreference Resolution.
This hierarchy of tasks is similar to the NLP pipeline used more classically. 
\citet{lin-etal-2019-open} argue that BERT representations are hierarchical rather than linear which enables to encode linguistic information in each token representation such as if it is the main auxiliary in a sentence.

One difficulty of probing studies is to make sure that the linguistic performance measured are inherent to the language model and not learned by the probing networks, as simple as they might be.
To this extent, \citet{hewitt-liang-2019-designing} propose to randomize the mapping between inputs and outputs of linguistic tasks as control tasks.
Good performance on these tasks cannot be explained by learned linguistic knowledge from BERT and are mostly indicative of the learning capacity of the probe.

Whereas these probing studies tackle \textbf{syntactic} tasks such as Part-of-Speech tagging or \textbf{semantic} tasks such as Named Entity Recognition, another type of probing has been proposed to examine world knowledge encoded in BERT representations by filling the blank in a \textbf{cloze style} task.
\citet{petroni-etal-2019-language} show that for some relation types, triples of facts can be retrieved from BERT with probing sentences such as "Dante was born in [MASK]" and that BERT is competitive with other Open Relation Extraction methods relying on knowledge bases.

\subsection{Pruning}
Another way to analyze the behavior of BERT is to ablate elements of its architecture and see the resulting effects which is the principle of \textbf{pruning} studies.
This can also be used as an alternative to \textbf{knowledge distillation} \citep{Hinton2015DistillingNetwork} to compress large models.

\citet{voita-etal-2019-analyzing} propose to use layer-wise relevance propagation \citep{ding-etal-2017-visualizing} to identify the most important attention heads in a Transformer trained for English-Russian Neural Machine Translation (NMT) and prune a significant proportion of heads without significant effect on performance.
They conclude that only a small subset of heads are important for translation.

\citet{Michel2019AreOne} argue that BERT is overparametrized and show that a large proportion of attention heads can be pruned without significant performance loss, even when reducing whole layers to a single head.
They even show that disabling some heads could result in performance gain in machine translation.

\subsection{Analyzing Self-Attention Patterns}
What is more related to the work we present in \autoref{chapter:06:attention} is the analysis of self-attention patterns.
Indeed, as soon as the introduction of the Transformer trained for NMT \citep{Vaswani2017AttentionNeed}, some visualization of Multi-Head Self-Attention showed interesting patterns, with qualitative examples where words are aligned to their coreferent pronouns for example.
Since then, several papers tried to more precisely analyze the learned self-attention patterns, whether in a full Transformer trained for NMT \citep{raganato-tiedemann-2018-analysis, voita-etal-2019-analyzing} or in BERT \citep{clark-etal-2019-bert, kovaleva-etal-2019-revealing}.

\begin{figure}[!h]
\begin{adjustwidth*}{}{-.3\textwidth}
\caption{The several patterns identified in BERT's attention heads. Figure from \citep{kovaleva-etal-2019-revealing}.} 
\label{fig:02:bert_patterns}
\centering\includegraphics[width=1.3\textwidth]{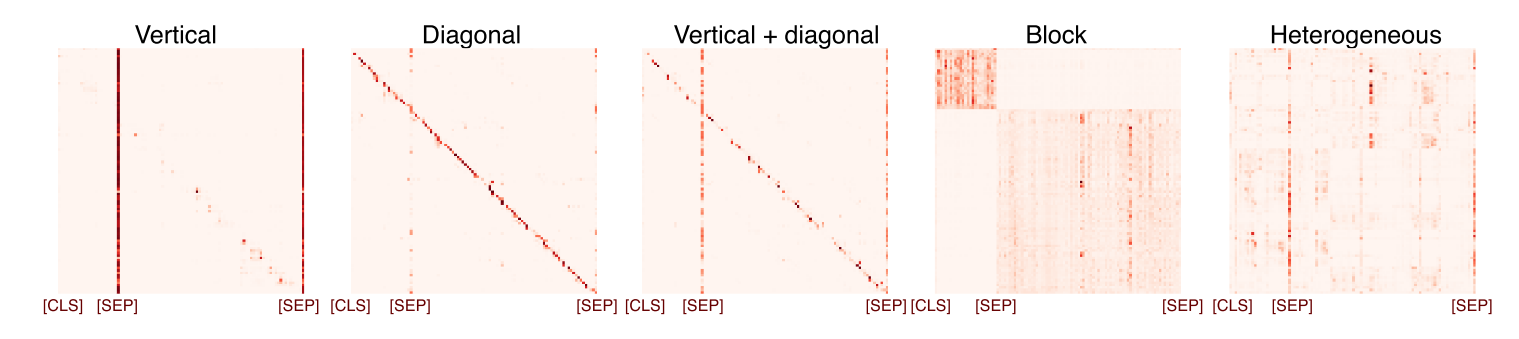}
\end{adjustwidth*}
\end{figure}

In the case of BERT, \citep{clark-etal-2019-bert} and \citet{kovaleva-etal-2019-revealing} identify similar recurrent patterns in attention heads, reported in \autoref{fig:02:bert_patterns}.
\textit{Vertical}: all tokens are aligned to the same token in the sentence (often special tokens [CLS] or [SEP], sometimes a rare word).
\textit{Diagonal}: tokens are aligned with themselves, or the previous or next tokens.
\textit{Block}: blocks of successive tokens are aligned with each other.

More interestingly, several papers raise a correlation between some attention heads in BERT and syntactic trees.
\citet{jawahar-etal-2019-bert} and \citet{kovaleva-etal-2019-revealing} propose qualitative visualizations of such structures. 
However, two works propose a quantitative evaluation of how well attention scores reflect some \textbf{syntactic relations}.

First, in their previously described pruning study of Transformer trained for NMT, \citet{voita-etal-2019-analyzing} notice that the heads they identify as the most important have roles such as attending to rare words or to the previous or next word, but this role can also be syntactic.
They study two English corpora (WMT and OpenSubtitles) and consider the following important syntactic relations: nominal subject (between a verb and its subject), direct object (between a verb and its object), adjectival modifier (between a noun and its adjective) and adverbial modifier (between a verb and its adverb).
They show that for some heads, the token attended with maximum weight (excluding the end of sentence token EOS) often corresponds to one of the aforementioned relations.
They show that for EN-RU, EN-DE and EN-FR models, the attention head that best corresponds to each syntactic relation has a better accuracy than the naive baseline of predicting the most frequent relative position for this relation.

\begin{figure}[!h]
\begin{adjustwidth*}{}{-.3\textwidth}
\caption{Examples of attention weights reflecting syntactic relations in specific BERT heads. Figure adapted from \citep{clark-etal-2019-bert}.} 
\label{fig:02:bert_weights}
\centering\includegraphics[width=1.3\textwidth]{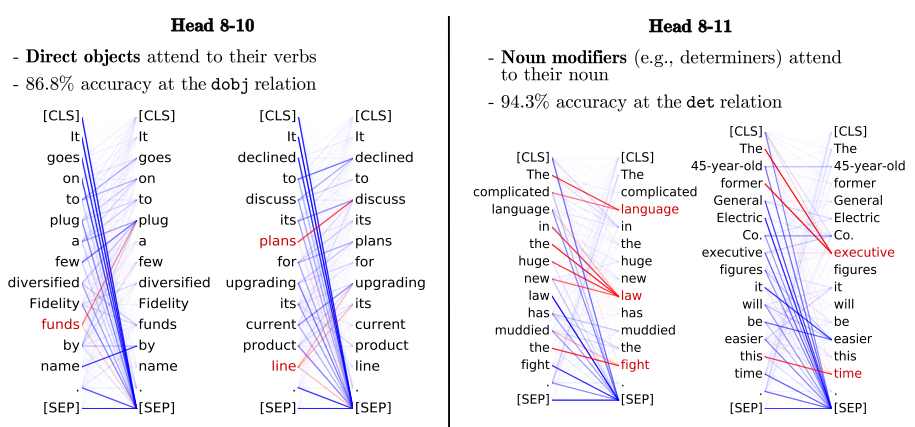}
\end{adjustwidth*}
\end{figure}

Second, \citet{clark-etal-2019-bert} analyze BERT's attention heads on Wall Street Journal articles from the Penn Treebank annotated with Stanford Dependencies. And perform a study similar to \citet{voita-etal-2019-analyzing} with a similar fixed-offset baseline but extended to more syntactic relations.
They obtain similar results with the best corresponding head in BERT at least on par with the baseline and sometimes significantly better.
They also extend this study to \textbf{Coreference Resolution} on CoNLL2012 \citep{pradhan-etal-2012-conll} and show that the best head can detect coreference links with results almost comparable to a rule-based system \citep{lee-etal-2011-stanfords} while much further from a more recent supervised neural coreference model \citep{wiseman-etal-2015-learning}. 
They conclude that this ability of BERT's heads to capture syntactic relations might be one explanation for its performance.

\section{Conclusion}
The introduction of Deep Neural Networks as a means to learn word representations has been pivotal in Natural Language Processing, enabling recent breakthroughs and diverting interest from previous rule-based or feature engineering solutions.
In particular, Neural Language Modeling has been used as an effective self-supervised pretraining objective to learn word embeddings that can be transferred to other NLP tasks.
Its success can be explained by the distributional hypothesis that distribution of words can be indicative at least to a certain extent of their semantic similarity but also by the capacity of neural networks to be trained online on unprecedented amounts of data.

Latest reported improvements of NLP models on standard benchmarks mainly stem from the transfer of richer contextual word representations.
Indeed, transferring word embeddings that can be obtained as the first layer of a LM gave way to transferring entire LM networks able to take into account each specific context of a word.
This was allowed by the development of mature software frameworks and the increase in available computational power and it was only accelerated by the introduction of the efficient Transformer neural architecture, suited for GPU parallelization.

Combining Language Modeling with Transformer networks in BERT-based architectures led to new state-of-the-art scores in benchmarks for virtually every task, even reaching superhuman scores on recent Question Answering and Natural Language Understanding benchmarks.
The analysis of the self-attention patterns suggest that the performance of LM pretraining comes from its ability to capture and transfer syntactic knowledge in addition to the semantic information contained in non contextual word embeddings.
However, when challenging these NLP models with handcrafted examples different from the data they were trained on, it quickly appears that they are nowhere near human-level of understanding. This raises question regarding how they can obtain superhuman performance on some benchmarks and how to design evaluation settings better measuring their ability to generalize to unseen, surprising or complex scenarios which is both a key aspect of comprehension and a major industrial stake.

In particular, detecting previously unknown facts expressed in a text is the main interest of Information Extraction.
More specifically, identifying and classifying unseen named entities and extracting relations in which they are involved is a key aspect of End-to-end Relation Extraction.
However, standard evaluation benchmarks in Named Entity Recognition or End-to-end Relation Extraction often limit to a single F1 score that do not take into account such consideration.
And although when they were introduced, LM pretraining models such as ELMo or BERT were shown to improve previous state-of-the-art on such benchmarks, their behavior regarding this type of generalization was ignored.

In the next chapters, we study the behavior of state-of-the-art models based on LM pretraining with respect to the extraction of unknown facts in End-to-end Relation Extraction.
We first focus on the Named Entity Recognition subtask in \autoref{chapter:03:ner}.
We exhibit an important lexical overlap between mentions in the test and train sets in standard benchmarks and perform an empirical study suggesting that contextual embeddings obtained by LM pretraining are mainly beneficial for generalization to unseen mentions and to new domains.

Second, we tackle the broader End-to-end Relation Extraction task for which we first propose a taxonomy of the various proposed approaches in \autoref{chapter:04:re-taxonomy}.
In \autoref{chapter:05:rethinking}, we then identify several evaluation inconsistencies in previously published articles which led to incorrect comparisons and call for using a unified evaluation setting and reduce confusion in this field.
We finally extend the previous empirical study on NER to End-to-end Relation Extraction where we show that recent models can be subject to a simple retention heuristics that is encouraged by standard datasets with relatively high lexical overlap such as CoNLL04 and ACE05.

\cleardoublepage

\chapter[Contextual Embeddings in NER]{Contextual Embeddings in Named Entity Recognition}
\label{chapter:03:ner}

\newcolumntype{Y}{>{\centering\arraybackslash}X}

The first key step to extract relational facts between entities is to identify
how and where the said entities are referred to in the text.
While Relation Extraction can be treated as a separate task given ground truth entity mentions,
this assumes an unrealistic perfection in the extraction of entity mentions and might lead to ignore
how errors in this step propagate to the final Relation Extraction task.
It thus seems essential to first closely examine Named Entity Recognition, as an inevitable step in Relation Extraction.

As introduced in \autoref{chapter:02:bert}, Language Model pretraining can be used to compute representations of words depending on their context called \textbf{contextual embeddings}.
This is intuitively useful for generalization, especially in Named-Entity Recognition for entity type disambiguation or detection of mentions never seen during training.
In this chapter, we propose to quantify the generalization capability of contextual embeddings when evaluated on unseen mentions as well as new domains.

In \autoref{sec:03:ner models}, we introduce the NER task and present the evolution of NER models. \autoref{sec:03:ner eval} presents the main datasets and related work regarding the evaluation of generalization of NER algorithms.
Finally, we present our empirical study of generalization capabilities of contextual embeddings in \autoref{sec:03:contener}.

\section{History of Named Entity Recognition Models}
\label{sec:03:ner models}

A natural idea to automatically process language is to use concepts human have built for thousands of years in the vast field of \textbf{Linguistics}, the study of language.
It is often divided in seven branches, Phonetics, Phonology, Morphology, Syntax, Semantics and Pragmatics \citep{}.
\textbf{Syntax} studies the way words can combine into sentences following a structure that makes them grammatical.
Words can be categorized depending on their function in such structures into \textbf{Parts of Speech (POS)}: noun, verb, pronoun, preposition, adverb, conjunction, participle and article.
Identifying the syntactic structure of a sentence is a key step to understand its meaning since it can help to disambiguate word senses and identify agents, actions and objects.

Among these parts of speech, \textbf{nouns} are words that refer to people, places, things, ideas or concepts and can take many forms. \textbf{Proper nouns} or \textbf{names} are sets of words, often multi-words, used to designate a particular person, place or thing that always begin with a capital letter in languages such as English.
They play an important role in the communication of information because they refer to specific real-world \textbf{entities} such as people, organization or locations.

Hence, when designing a framework for studying \textbf{Information Extraction}, \textbf{Named Entity Recognition (NER)} was defined as one key subtask.
It consists in detecting spans of words (the \textbf{entity mentions}) that refer to real-world entities (the \textbf{named entities}) and classifying them into predefined \textbf{entity types}.
These types typically include person, organization, and location but are also often extended to numeric expressions such as time, date or amount of money.
It can also be extended to the detection of pronouns that refer to entities in a task called \textbf{Entity Mention Detection (EMD)}.
However, because the framing is very similar, "Named Entity Recognition" can also refer to this task.

\subsection{Rule-based pattern matching}

According to a survey by \citet{Nadeau2007AClassification}, one of the first research work on NER is presented in \citep{Rau1991ExtractingText} where a rule-based algorithm is designed to extract names of companies in English financial news stories.
\citet{Rau1991ExtractingText} identifies several main difficulties in this task.
First, companies are created, closed or renamed at a relatively frequent pace which make it difficult to maintain lists of known companies.
Second, several variations of a name can refer to a company such as acronyms.
She proposes heuristics combining rules such as the presence of a company name indicator (such as Inc., Ltd., Corp. ...), creation of variations of company names (such as acronyms), presence of capitalization and neighbouring conjunctions.

What accelerated the development of NER algorithms is the creation of \textbf{shared tasks} and \textbf{public datasets} that help benchmarking different techniques.
The first shared task dedicated to the task, and in fact coining the terms "Named Entity Recognition", is \textbf{MUC-6} (the 6$^{th}$ Message Understanding Conference) \citep{grishman-sundheim-1996-message}, where four Information Extraction subtasks are evaluated on Wall Street Journal articles:  NER, Coreference Resolution, Word Sense Disambiguation and Predicate-argument Structure Detection.

This led to a first round of papers using \textbf{rule-based finite state automata} to match predefined patterns \citep{appelt-etal-1995-sri}.
These rules can be applied differently to various words in the vocabulary to optimize performance on a validation set in an automatic manner \citep{aberdeen-etal-1995-mitre} as previously used for Part-of-Speech tagging \citep{brill-1992-simple}.
However crafting these rules is time consuming all the more as they are specific to language, domain and entity types.

\subsection{Feature-based Supervised Machine Learning}
Hence, inspired by the previous successes of Machine Learning models such as \textbf{Hidden Markov Models (HMM)} in POS tagging \citep{church-1988-stochastic} \textbf{feature-based supervised learning} is explored for NER with different algorithms.
\citet{bikel-etal-1997-nymble} use a model inspired by HMMs on English and Spanish news.
\citet{sekine-etal-1998-decision} train a decision tree on Japanese texts.
\citet{borthwick-etal-1998-exploiting} apply Maximum Entropy on English texts.
\citet{asahara-matsumoto-2003-japanese} use a Support Vector Machine (SVM) based algorithm for Japanese texts.
Finally, \citet{mccallum-li-2003-early} use a \textbf{linear-chain Conditional Random Field (CRF)} on CoNLL03 in English and German, a probabilistic graphical model that would play a more important role in the history of NER models.

In this setting, similarly to POS tagging, the task is cast as \textbf{sequence labelling}: the model must associate a tag to each word in the sequence.
However, because named entities mentions can be multi-word expressions, assigning an entity type to each word is not sufficient: words must also be grouped into \textbf{chunks}. 
Hence, early works design tags to contain two types of information: the named entity type (including "not an entity") as well as the position of the word inside the expression.
The first occurrence of such tagging scheme for text chunking appears in \citet{ramshaw-marcus-1995-text} that introduces the \textbf{IOB tagging scheme}.
In NER, \citet{sekine-etal-1998-decision} and \citet{borthwick-etal-1998-exploiting} use settings similar to the extended \textbf{IOBES} setting illustrated in \autoref{fig:03:iobes}.

In both cases, words that do not belong to named entities are tagged with "O", standing for outside.
Words belonging to entities are assigned tags of the form "P-TYPE" where the prefix P can be B or I and indicates the position of the word inside the name chunk. B stands for "\textbf{beginning}", the first word of the entity; I for "\textbf{inside}", any following word. In the IOBES setting, P can also be E for "\textbf{end}", the last word of the chunk or S for "\textbf{singleton}" when a name is composed of a unique word.
The IOBES setting can also be referred to as BILOU where E is replaced by L for "\textbf{last}" and S by U for "\textbf{unit}". 

\begin{figure}[!h]
\caption{Example of different NER tagging schemes.}
\label{fig:03:iobes}
\centering\includegraphics[width=\textwidth]{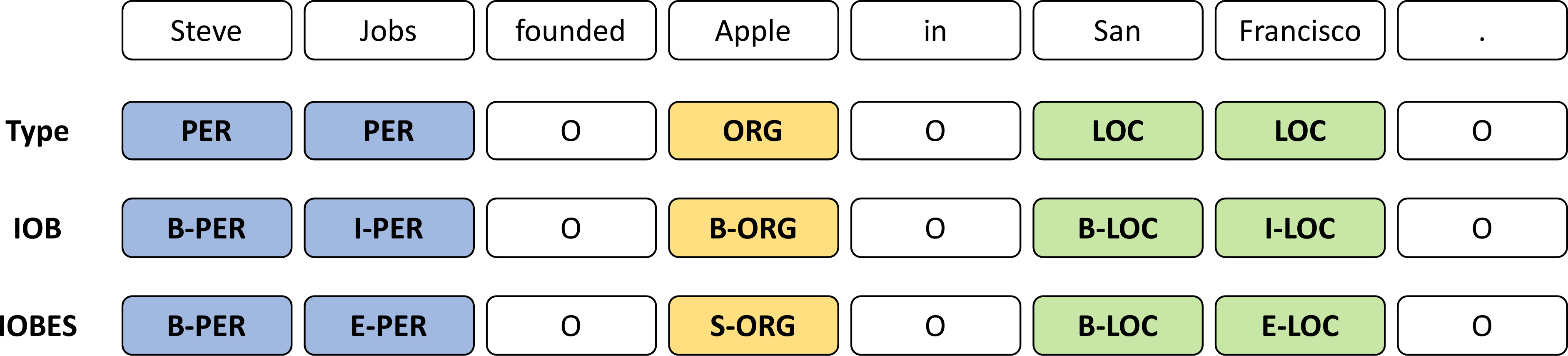}
\end{figure}

The task is modeled as predicting the sequence of tags $Y$ corresponding to the sequence of input words $X$ estimated by $\hat{Y} = \argmax_{Y \in \mathcal{Y}} P(Y|X)$.
In a linear-chain CRF, there is a conditional independence assumption that the current tag $y_t$ only depends on the previous one $y_{t-1}$ as well as the entire input sequence $X$ and the goal is to find: 
$\hat{Y} = \argmax_{Y \in \mathcal{Y}} P(Y|X)$. This Markov chain assumption intends to model the fact that some bigrams of tags are more likely to appear, such as ``B-PER E-PER'' that corresponds to the frequent name surname pattern designating people, whereas other transitions are impossible.

In a linear-chain CRF, the conditional probability of a label sequence can be written as:
\begin{align*}
  P(Y|X) &= \frac{1}{Z(X)} exp(\sum_{t=1}^{T}\sum_{k=1}^{K} w_k f_k(y_t, y_{t-1}, X, t)\\
  Z(X) &= \sum_Y exp(\sum_{t=1}^{T}\sum_{k=1}^{K} w_k f_k(y_t, y_{t-1}, X, t)
\end{align*}
where $f_{k}$ are local handcrafted feature functions only depending on the whole input sentence and current and previous output tokens $y_t$ and $y_{t-1}$, and $w_k$ the corresponding learned weights.

\textbf{Feature Engineering} is hence a key part of such algorithms, to design the most relevant features useful to predict named entity tags. Additionally to a one-hot word feature that corresponds to the training vocabulary, several additional rules have been used to create such features.
These rules can be based on \textbf{word shapes}, for example whether the first or all letters of a word are capitalized, which is a key indicator of a proper name when its not the first word of a sentence.
Word shape can also refer to the fact that a word includes digits as well as their number and separation signs that can be indicative of time, dates or amounts.

Furthermore, list of known entities known as \textbf{gazetteers}, often derived from encyclopedias can be used to compute binary features indicating the presence or absence of a word in a list of locations or people for example.
Such lists can include whole words or limit to common affixes frequent in some word types.

Finally, grammatical information such as \textbf{Part-Of-Speech tags}, either from ground truth or predictions of a preliminary model, are also useful in the prediction since Named Entities should mostly correspond to the proper noun POS class.

\subsection{The BiLSTM-CRF architecture}

Following the general trend in Natural Language Processing, Deep neural networks and word embeddings have been used to reduce the remaining dependency on handcrafted rules in the design and computation of the previously described features.
As described in \autoref{sec:02:word representations}, \citet{Collobert2008ALearning, Collobert2011NaturalScratch} introduce the foundations of modern neural networks for Natural Language Processing with word embeddings learned from Language Model Pretraining and Multitask Learning.
Among many other tasks, they tackle Named-Entity Recognition with a 1D Convolutional Neural Network (CNN) \citep{Waibel1989PhonemeNetworks}.
In \citep{Collobert2011NaturalScratch}, they use a CRF-like loss function by introducing a learned transition matrix to model the different likelihood of tags bigrams in tasks like chunking, NER or SRL.
And thus this model has been referred to as \textbf{Conv-CRF}.
In NER, they improve performance over previous feature-based baselines on the standard CoNLL03 benchmark.

As an alternative to CNNs which are not designed to process sequences with different lengths such as sentences, Recurrent Neural Networks \citep{Rumelhart1986LearningErrors} have largely been used in Natural Language Processing.
As described in \autoref{sec:02:lstm}, the necessity to backpropagate through the sequence leads to vanishing or exploding gradient issues and \citet{Hochreiter1997LongMemory} proposed \textbf{Long Short-Term Memory networks (LSTM)} to address them.

LSTMs, along with Gated Recurrent Units (GRU) \citep{cho-etal-2014-learning} which follow the same idea in a simpler implementation with only one gate, have been popular in Natural Language Processing since their first impressive application on handwriting recognition \citep{Graves2008OfflineNetworks, Graves2009ARecognition}.
It has been used in voice recognition, Neural Machine Translation in the Seq2seq architecture and became the de facto standard for NLP models with attention between 2015 and 2018.

The LSTM enables to process a sequence one item at a time and compute an hidden state representation depending on the current item as well as every previously seen ones, i.e. the beginning of the sequence.
However, in sequence labelling, we have access to the entire text sequence and can use the right context in addition to the left context.
To combine both context, two opposite directions recurrent neural networks can be combined: one in the forward direction computes a state depending on the left context and one in the backward direction computes a state depending on the right context.
This architecture is called Bidirectional RNN (BiRNN) \citep{Schuster1997BidirectionalNetworks}.
\citet{Graves2013SPEECHNETWORKS} use a Bidirectional LSTM (BiLSTM) for Speech Processing.

\begin{figure}[!h]
\begin{adjustwidth*}{}{-.3\textwidth}
\caption[Schema of the BiLSTM-CRF architecture]{Schema of the BiLSTM-CRF architecture. To compute the hidden representation of ``Apple'', a forward LSTM takes into account the left context to compute the $\textbf{l}_{Apple}$ representation that is concatenated to the $\textbf{r}_{Apple}$ representation computed by a backward LSTM.}
\label{fig:03:iobes}
\centering\includegraphics[width=1.3\textwidth]{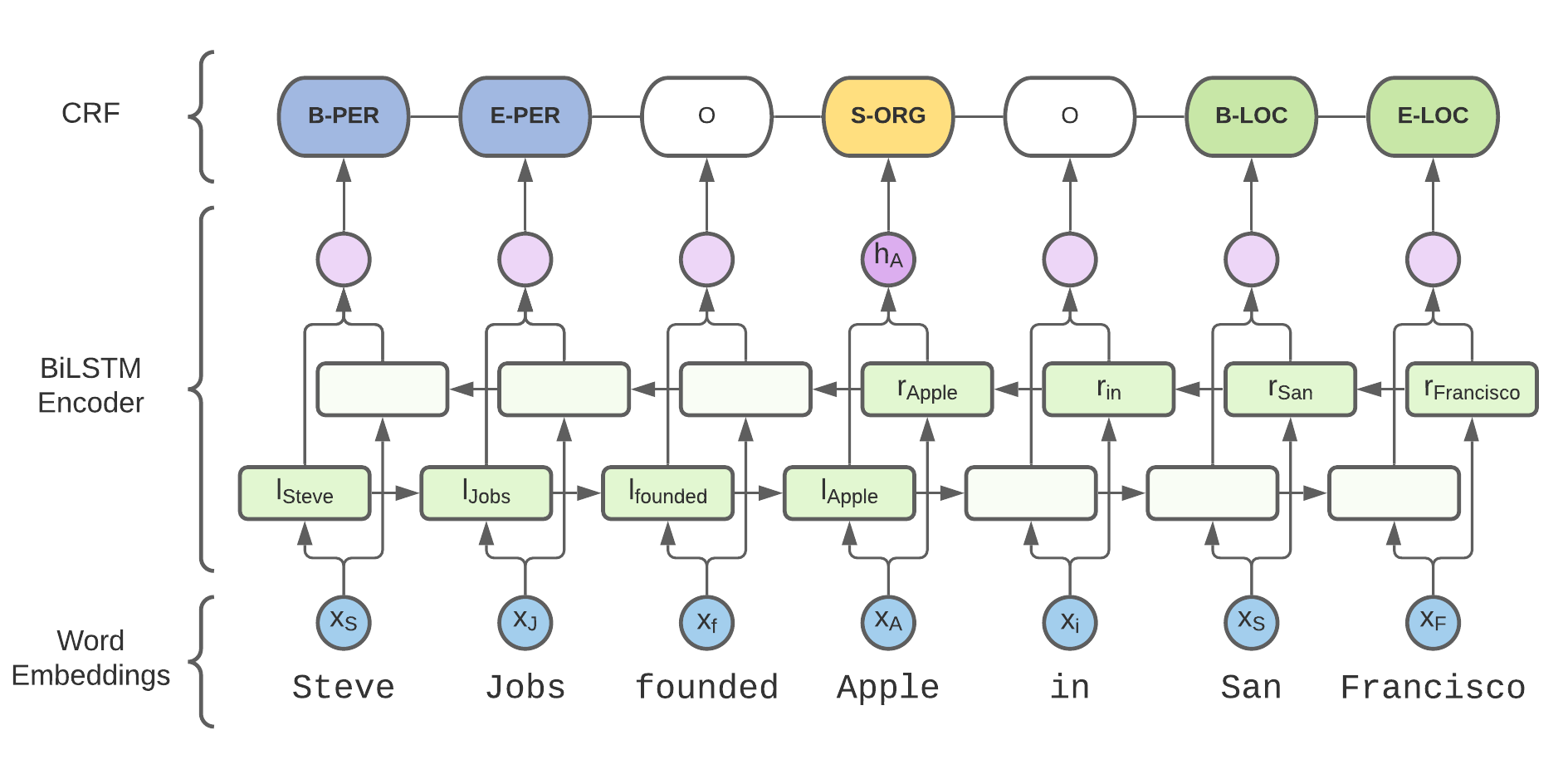}
\end{adjustwidth*}
\end{figure}

For NER but also POS and chunking, \citet{Huang2015BidirectionalTagging} take inspiration from \citep{Collobert2011NaturalScratch} to introduce the BiLSTM-CRF architecture.
In this \textbf{BiLSTM-CRF} model, the BiLSTM is used as an encoder to obtain contextual representations of words depending both on the left and right context.
They show very marginal improvements on standard English benchmarks when using the same pretrained word embeddings (SENNA) but a larger improvement with random embeddings.
This architecture has since become the most popular in state-of-the-art NER models, although alternatives have been explored
to reduce the time complexity of these models that is linear with respect to the size of the sequence.
For example we can mention the Iterated Dilated Convolutional Neural Network architecture \citep{strubell-etal-2017-fast}
that proposes to stack Dilated Convolution layers to be competitive on CoNLL03 and OntoNotes while leading to a 8x speed up enabling to consider larger contexts than sentences.
Further gains on benchmarks mainly stem from using richer word representations: learned \textbf{character-level word embeddings} and \textbf{contextual embeddings} derived from language models.

\subsection{Character-level and contextual word representations}
\label{sub:03:representations}
Character-level word embeddings are learned by a word-level neural network from character  embeddings  to  incorporate  orthographic  and  morphological  features. \citet{lample-etal-2016-neural}, use a BiLSTM-CRF model similar to \citet{Huang2015BidirectionalTagging}'s and add a character-level BiLSTM (\textbf{charBiLSTM}) that learns a representation concatenated to traditional word embeddings pretrained with skip-n-gram \citep{ling-etal-2015-contexts}, a variation of word2vec \citep{Mikolov2013EfficientSpace}.  \citet{ma-hovy-2016-end} propose a character-level Convolutional Neural Network (\textbf{charCNN}) to learn such representations.
Although they report new state-of-the-art results on the respective CoNLL standard benchmarks for POS and NER, the architecture of the network is very similar to the previous charBiLSTM representations that remained more popular in related work.

The second evolution in word representations used in NER is a major shift introduced in \textbf{TagLM}, a model designed for sequence labelling \citep{peters-etal-2017-semi}.
Some word representations are derived from the prediction of a forward and a backward language models that uniquely depend on the left or right contexts, thus contextualizing word representations.  
They study several architectures for LMs inspired by \citet{Jozefowicz2016ExploringModeling} who previously studied the impact of using word embeddings or charCNN representations in Language Modeling.
The language models are two-layers LSTMs that either take word embeddings or charCNN representations as input and are frozen when training the sequence tagging model.
Concatenating these "LM embeddings" with more traditional SENNA embeddings in a BiLSTM-CRF architecture enables to outperform previous models, even when they used external resources such as gazetteers in both POS and NER.

\citet{peters-etal-2018-deep} then propose \textbf{ELMo}, an evolution of TagLM, and extend its study to additional NLP tasks.
The Language Model in ELMo uses charCNN embeddings as input to its first layer, again following \citet{Jozefowicz2016ExploringModeling}.
This non contextual representation is then used in replacement of SENNA embeddings.
Additionally, the final ELMo representation combines hidden states of the two layers of the LSTM LMs, with the intuition that each layer can capture different types of information that are more or less useful depending on the final task.
Hence, they obtain an intermediate representation at each layer of the LMs: a non contextual representation learned by a \textbf{charCNN} in the first layer and the hidden state of each LSTM layer.
To obtain a task specific representation when training a model with ELMo, they propose to \textbf{freeze the weights of the LM} and learn \textbf{task-specific weighted sums} of the three layers of representation.
The authors demonstrate significant improvements over previous baselines by simply changing word representations to ELMo in standard benchmarks for Question Answering, Textual Entailment, Semantic Role Labeling, Coreference Resolution, Named Entity Recognition and Sentiment Analysis.

\citet{akbik-etal-2018-contextual} propose to use a similar idea specifically for sequence tagging with a character-level Language Model, trained to predict a string character by character.
This model can be referred to as \textbf{Flair}, the name of the software framework released along with the paper.
Like in ELMo, they use a forward and a backward LSTM Language Models that are frozen but are limited to one layer.
A word is then represented as the concatenation of the hidden states of its last character in the forward model (thus depending on the left context and the word itself) and of its first character in the backward model (thus depending on itself and its right context).
They report results competitive with ELMo for NER, Chunking and POS when Flair representations are combined with GloVe embeddings and charBiLSTM representations. 

Finally, \textbf{BERT} \citep{devlin-etal-2019-bert} takes inspiration from ELMo to pretrain a large Language Model and transfer the learned knowledge to a multitude tasks.
As described in more details in  \autoref{chapter:02:bert}, the \textbf{subword level Language Model} is based on the architecture of a \textbf{Transformer Encoder} \citep{Vaswani2017AttentionNeed}.
Because of the computing efficiency of the Transformer architecture over RNNs, BERT has more layers than ELMo and is released in two versions: BERT$_{BASE}$ with 12 layers and BERT$_{LARGE}$ with 24.
Contrary to ELMo or Flair, BERT weights are classically \textbf{finetuned} along with the task specific model and the final text representation is simply the hidden state of the last layer. 
However, the original paper also introduced a feature-based model in which the LM is frozen and the final representation is the concatenation of the hidden states of the last four layers.

Because the whole Transformer architecture is finetuned and already capable of contextualizing, the standard practice to use BERT for sequence tagging is not to add an additional BiLSTM-CRF model but simply a randomly initialized Linear Layer or Multilayer Perceptron with one hidden layer.

\subsection{Span-level NER}
Although the traditional view is to model Named Entity Recognition as a sequence labelling problem, this approach has a major drawback when it comes to detecting overlapping or nested mentions.
Indeed, for example "Bank of China" refers to an organisation but the word "China" that is nested in this expression refers to a Geopolitical Entity that might be interesting to detect.

\citet{sohrab-miwa-2018-deep} propose to tackle Named Entity Recognition as a span classification task by enumerating all subsequences of words (up to a realistic maximum length) and classify each one with an entity type.
They use a BiLSTM fed with pretrained word embeddings and charBiLSTM representations to encode the entire sequence of words.
Then, each span of successive words is represented with the concatenation of the hidden states corresponding to its first and last words as well as the mean of all its corresponding hidden states.
They focus on the biomedical domain more subject to nested entities on the GENIA \citep{Kim2003GENIABio-textmining} and JNLPBA \citep{collier-kim-2004-introduction} corpora with specific entity types such as "protein", "DNA" or "cell".
They show significant quantitative gains compared to previous nested NER algorithms.

This span based models are also used in multitask learning settings such as end-to-end Relation Extraction which is tackled in \autoref{chapter:04:re-taxonomy} and \autoref{chapter:05:rethinking}.

\section{Evaluating Performance of NER models}
\label{sec:03:ner eval}

As it is common in the Machine Learning field, the development of Named Entity Recognition models is linked to the creation of public datasets and shared evaluation settings.

\subsection{Metrics}
The traditional metric used to assess the performance of NER models is the \textbf{F1 score} that balances the contribution of \textbf{precision} and \textbf{recall}.
Precision measures the exactitude of retrieved samples while recall assesses their exhaustivity and the F1 score is the harmonic mean of the two measures.

\begin{align*}
\mathrm{P} &= \frac{TP}{TP + FP} & \mathrm{R} &= \frac{TP}{TP + FN} & \mathrm{F_1} &= \frac{2 \mathrm{P} \mathrm{R}}{\mathrm{P} + \mathrm{R}}\\
\end{align*}

where $TP$ is the number of true positives, $FN$ of false negatives and $FP$ of false positives.

Each metric is separated by entity type, since some types are easier to detect than others, and the global score is traditionally obtained with a \textbf{micro-average} of the scores.

Although in the first MUC conferences several metrics are used to distinguish \textbf{boundaries detection} errors from \textbf{typing} errors, nowadays the standard evaluation is an \textbf{exact match} one.
This is strictest setting where an entity is correctly detected if and only if both its boundaries and type are correctly detected. 

\subsection{Datasets}

While numerous datasets have been proposed in several application domains and languages, we only review the ones we identified as the most popular and thus linked to the current assessment of NER algorithm performance.

The first shared task introducing Named Entity Recognition is \textbf{MUC-6} (the 6$^{th}$ Message Understanding Conference) \citep{grishman-sundheim-1996-message} that proposes to extract the names of people, organisations and places as well some numerical and temporal expressions in Wall Street Journal articles. It was followed by MUC-7 in 1997 that processes New York Times news.
Although such conferences were focused on English documents, generalization to others languages was concurrently considered in the \textbf{Multilingual Entity Task (MET)} \citep{merchant-etal-1996-multilingual} that proposed to measure NER performance in Spanish, Japanese and Chinese documents.

CoNLL (Conference on Computational Natural Language Learning) is another conference that proposed successive shared tasks that became current standard benchmarks for several NLP tasks.
For Named Entity Recognition, the 2002 and 2003 instances of CoNLL introduced tagged news article data in Spanish and Dutch for \textbf{CoNLL02} \citep{tjong-kim-sang-2002-introduction} and English and German for \textbf{CoNLL03} \citep{tjong-kim-sang-de-meulder-2003-introduction}.
In these two datasets, entities are assigned one of the four types: person, organisation, location or miscellaneous (often corresponding to languages or dates).
Because evaluation is often centered on English, the English part of CoNLL03 has become the de facto standard benchmark for reporting NER performance, still used for example to demonstrate the capabilities of BERT in \citep{devlin-etal-2019-bert}.

A second dataset that is more and more used in addition to CoNLL03 to evaluate performance in English is \textbf{OntoNotes 5} \citep{Weischedel2013OntoNotesLDC2013T19}. It is a larger dataset comprised of various genres (news, talk shows, telephone conversation, blogs and forums) and annotated with 18 entity types, also including temporal and numerical information.

Numerous other datasets have been proposed to tackle different domains, problematics or languages.
We can thus mention datasets designed for \textbf{fine-grained NER} with order of magnitudes more entity types than in the previously described ones and with a hierarchy between types (for example a person can be an artist or political leader).
FG-NER \citep{mai-etal-2018-empirical} or HYENA \citep{yosef-etal-2012-hyena} are examples of datasets for Fine-grained NER with 200 and 505 entity types respectively.
Regarding specialized domains, a particularly active field of application is the \textbf{biomedical and clinical text} domain.
Such applications on the biomedical litterature include the detection of proteins or cells names in GENIA \citep{Kim2003GENIABio-textmining} or of diseases and drugs in BC5CDR \citep{Li2016BioCreativeExtraction}.

Another domain specific dataset that is of particular interest in our study on generalization is \textbf{WNUT 2017} that was introduced as a shared task at the Workshop on Noisy User Generated Text \citep{derczynski-etal-2017-results}. It is composed of \textbf{user generated texts} such as tweets or YouTube comments. To model the ever evolving events or celebrities that are referred to in tweets, the test sets are designed so that no entity mentions are present in the training set thus resulting in no \textbf{lexical overlap}.

\begin{table}[h]

\begin{adjustwidth*}{}{-.3\textwidth}
\caption[Statistics of CoNLL03, OntoNotes and WNUT 2017 datasets.]{Statistics of CoNLL03, OntoNotes and WNUT 2017 datasets. We report both the number of mention occurrences and unique mentions. We take type into account to compute the latter.}
\label{table:supplementary-datasets}
\small
\begin{center}
\begin{tabularx}{1.3\textwidth}{@{}l*{3}{R}r*{3}{R}r*{3}{R}@{}}
\toprule
                    &   \multicolumn{3}{c}{CoNLL03}     &   &  \multicolumn{3}{c}{OntoNotes} &   & \multicolumn{3}{c}{WNUT} \\
\cmidrule{2-4} \cmidrule{6-8} \cmidrule{10-12}
                    & Train     & Dev       & Test      &   & Train     & Dev       & Test &        & Train     & Dev       & Test\\
\midrule
Sentences           & 14,041    & 3,250     & 3,453     &   & 59,924    & 8,528     & 8,262 &       & 3,394     & 1,009     & 1,287\\
Tokens              & 203,621   & 51,362    & 46,435    &   & {\scriptsize 1,088,503} & 147,724   & 152,728  &    & 62,730    & 15,733    & 23,394 \\
Mentions            & 23,499    & 5,942     & 5,648     &   & 81,828    & 11,066     & 11,257  &    & 1,975     & 836       & 1,079 \\
Unique              & 8,220     & 2,854     & 2,701     &   & 25,707     & 4,935     & 4,907  &     & 1,604    & 747        & 955 \\
\bottomrule
\end{tabularx}
\end{center}
\end{adjustwidth*}
\end{table}

\subsection{Related Work on Generalization of NER models}

Since the very work of \citet{Rau1991ExtractingText} on the extraction of company names, the diversity and variability of mentions appeared as major difficulties of Named Entity Recognition.
This variability was first handled by designing handcrafted rules specific to languages, text domains and entity types, like in \citep{appelt-etal-1995-sri} or \citep{aberdeen-etal-1995-mitre}.

Supervised learning methods were introduced to automatically learn patterns from labeled data and reduce human engineering.
This quickly raised several questions regarding the ability of such methods to be easily transferable between languages, domains or new entity types.
Hence, the Multilingual Entity Task (MET) was proposed soon after the first Named Entity Recognition shared task.

Another important feature of NER algorithms is their capability to detect unseen mentions.
\citet{palmer-day-1997-statistical} propose a simple memorization baseline as a lower bound for supervised models by building entity lists from the training set and searching for matches in the test set.
They introduce the \textbf{vocabulary transfer rate} as the proportion of words in the test vocabulary (i.e. without repetition) that appear in the training vocabulary and show that it is correlated to the performance of this memorization baseline on corpora for different languages.

This idea that the \textbf{lexical diversity} of a corpus makes it harder to detect named entities is further explored by \citet{Augenstein2017GeneralisationAnalysis}. They propose to measure the lexical diversity of a corpus as the ratio of mentions in the test set that are not present in the training set, referred to as \textbf{unseen mentions}.
They propose to study the performance of three NER models on seven different corpora and measure the impact of unseen entities on performance as well as measure out-of-domain performance. The three off-the-shelf models are \textbf{Stanford NER} \citep{finkel-etal-2005-incorporating}, \textbf{CRFsuite} \citep{Okazaki2007CRFsuite:CRFs} and \textbf{SENNA} \citep{Collobert2011NaturalScratch}.

Stanford NER and CRFsuite both use a feature-based linear chain CRF with the difference that CRFsuite does not use external knowledge such as a gazetteer or unsupervised representations whereas Stanford NER features have been tuned on CoNLL03.
SENNA is a Conv-CRF neural network that use word embeddings as well as gazetteer features.
They conclude that the use of word embeddings in SENNA enables to achieve the best generalization from training to test data and that NER performance can be predicted with a simple memorization baseline that predicts the most frequent label for each token, confirming a correlation between performance and \textbf{lexical overlap} between train and test entity mentions, i.e. the ratio of seen test entities.

The notion of lexical overlap is not specific to NER but is also applicable to NLP tasks involving spans of words.
For example, \citet{moosavi-strube-2017-lexical} study its impact on Coreference Resolution on CoNLL2012 \citep{pradhan-etal-2012-conll} with state-of-the-art neural models.
They notice that non-pronominal coreferent mentions largely cooccur in both the train and test sets of CoNLL 2012 with overlap ratios from 37 \% to 76 \% on the different genres of the dataset. 
They show that for out-of-domain evaluation where these ratios are lowered, the performance gap between Deep Learning models and a rule-based system fades away. 
And they add linguistic features (such as gender, NER, POS...) to improve out-of-domain generalization in a subsequent study \citep{Moosavi2018UsingResolvers}. 
Nevertheless, as evidenced by \citet{Augenstein2017GeneralisationAnalysis}, such features are obtained using models in turn based on lexical features and at least for NER the same lexical overlap issue arises.

\section[Generalization Capabilities of Contextual Embeddings]{An Empirical Study on Generalization of Contextual Embeddings in Named Entity Recognition}
\label{sec:03:contener}

The previous work by \citet{Augenstein2017GeneralisationAnalysis} is a consequent empirical study noticing that standard NER benchmarks such as CoNLL03 or OntoNotes present an important
lexical overlap between mentions in the train set and dev / test sets which leads to a poor evaluation of generalization to unseen mentions.
However, it is limited to models dating back from 2005 to 2011 that are not representative of the latest improvements in terms of NER models and word representations.
In particular, the introduction of contextual embeddings from Language Model pretraining is intuitively useful to better incorporate \textbf{syntactic features} and lower the dependency on purely \textbf{lexical features}.

Indeed, LM pretraining enables to obtain contextual word representations and reduce the dependency of neural networks on hand-labeled data specific to tasks or domains \citep{howard-ruder-2018-universal, Radford2018ImprovingPre-Training}. 
This contextualization ability can particularly benefit to NER domain adaptation which is often limited to training a network on source data and either feeding its predictions to a new classifier or finetuning it on target data \citep{Lee2018TransferNetworks, Rodriguez2018TransferClasses}.
Yet, because the successive contextual embeddings models such as ELMo or BERT are tested on a variety of tasks, their original evaluation on NER is often shallow and limits to a single score on CoNLL03, disregarding the impact of linguistic phenomena such as lexical overlap.
Hence, we propose to better quantify the contribution of Language Model pretraining on generalization in Named Entity Recognition by using an evaluation setting closely inspired by \citet{Augenstein2017GeneralisationAnalysis}.

We choose to use the BiLSTM-CRF architecture as a backbone of our study since it became the de facto standard in sequence tagging since its introduction by \citet{Huang2015BidirectionalTagging} and until the development of BERT-like architectures.
Indeed, it is used to obtain a representation of each words in a sequence depending on the context, ultimately close to the role of contextual embeddings.
This is why we propose to distinguish two contextualization effects of using the BiLSTM-CRF architecture with contextual embeddings: one \textbf{unsupervised Language Model contextualization} that we denote $\textbf{C}_{\textbf{LM}}$ and one \textbf{task supervised contextualization}, $\textbf{C}_{\textbf{NER}}$.
We show that the former mainly benefits unseen mention detection, all the more out-of-domain where it is even more beneficial than the latter.

\subsection{Lexical Overlap}
While \citet{Augenstein2017GeneralisationAnalysis} separate test mentions into seen and unseen mentions to  measure the impact of lexical overlap on NER performance, we introduce a slightly finer-grained partition by further separating unseen mentions into \textit{partial match} and \textit{new} categories.
We obtain a partition with three categories: \textbf{exact match} (\textbf{EM}), \textbf{partial match} (\textbf{PM}) and \textbf{new}.

 A mention is an exact match if it appears in the exact same case-sensitive form in the train set, tagged with the same type.
It is a partial match if at least one of its non stop words appears in a mention of same type. 
Every other mentions are new: none of their non stop words have been seen in a mention of same type.

We study lexical overlap in \textbf{CoNLL03} \citep{tjong-kim-sang-de-meulder-2003-introduction} and \textbf{OntoNotes 5} \citep{Weischedel2013OntoNotesLDC2013T19} , the two main English NER datasets, as well as \textbf{WNUT17} \citep{derczynski-etal-2017-results} which is smaller, specific to user generated content (tweets, comments) and was designed without exact overlap.
This measure of lexical overlap can be used both in the classical in-domain setting but also for \textbf{out-of-domain evaluation} where we train on one dataset and test on another.

To study out-of-domain generalization, we propose to train on CoNLL03, composed of news articles and test on the larger and more diverse OntoNotes (see \autoref{table:03:genres} for genres) as well as on the very specific WNUT.
We take the four entity types of CoNLL03 (Person, Location, Organization and Miscellaneous) as standard types and we remap OntoNotes and WNUT entity types to match these standard types and denote the obtained datasets OntoNotes$^{\ast}$ and WNUT$^{\ast}$.

\begin{table}[h]
\begin{adjustwidth*}{}{-.3\textwidth}
\caption{Per type lexical overlap of test mention occurrences with respective train set in-domain and with CoNLL03 train set in the out-of-domain scenario. (EM / PM = \textit{exact / partial match})
}
\label{table:03:overlap}
\scriptsize
\begin{tabularx}{1.3\textwidth}{@{}ll*{5}{Y}rYr*{5}{Y}rYr*{4}{Y}r@{}}
\toprule
& &\multicolumn{5}{c}{CoNLL03} & & ON &  & \multicolumn{5}{c}{OntoNotes$^{\ast}$} & & WNUT & & \multicolumn{4}{c}{WNUT$^{\ast}$}\\
\cmidrule{3-7} \cmidrule{9-9} \cmidrule{11-15} \cmidrule{17-17} \cmidrule{19-22}
 & & LOC & MISC & ORG & PER & ALL & & ALL& & LOC & MISC & ORG & PER & ALL & &  ALL &  & LOC & ORG & PER & ALL\\ 
\midrule
\parbox[t]{3mm}{\multirow{3}{*}{\rotatebox[origin=c]{90}{\textbf{Self}}}} 
& EM & 82\% & 67\% & 54\% & 14\% & 52\% &
        & 67\% & 
       
        & 87\% & 93\% & 54\% & 49\% & 69\% &
         & - & 
         & - & - &- &- \\
        
& PM & 4\% & 11\% & 17\% & 43\% & 20\% & 
        & 24\% &
        
        & 6\% & 2\% & 32\% & 36\% & 20\% &
        & 12\% & 
        & 11\% & 5\% & 13\% & 12\% \\
        
& New  & 14\% & 22\% & 29\% & 43\% & 28\% &
        & 9\% & 
        
        & 7\% & 5\% & 14\% & 15\% & 11\% &
        & 88\% &
        & 89\% & 95\% & 87\% & 88\% \\
        
\midrule
\parbox[t]{3mm}{\multirow{3}{*}{\rotatebox[origin=c]{90}{\textbf{CoNLL03}}}}  
        &  EM   &-  & - &  -& - &-  &
        &- & 
        
        & 70\% & 78\% & 18\% & 16\% & 42\%  &
        & - & 
        & 26\% & 8\% & 1\% & 7\%\\
        
& PM &-  & - &-  &  -& - & 
        & - &
        & 7\% & 10\% & 45\% & 46\% & 28\% &
        & - & 
        & 9\% & 15\% & 16\% & 14\% \\
        
& New  & -  & - & -  & - & - &
        & - & 
        & 23\% & 12\% & 38\% & 38\% & 30\%&
        & - &
        & 65\% & 77\% & 83\% & 78\%\\
        
\bottomrule
\end{tabularx}
\end{adjustwidth*}

\end{table}

As reported in \autoref{table:03:overlap}, the two main benchmarks for English NER mainly evaluate performance on occurrences of mentions already seen during training, although they appear in different sentences. \textbf{Such lexical overlap proportions are unrealistic in real-life} where the model must process orders of magnitude more documents in the inference phase than it has been trained on, to amortize the annotation cost.
Hence the importance of specifically improving performance on unseen mentions.
On the contrary, WNUT proposes a particularly challenging low-resource setting with no exact overlap.

Furthermore, the overlap depends on the entity types: Location and Miscellaneous are the most overlapping types, even out-of-domain, whereas Person and Organization present a higher lexical diversity.
This observation fits an application setting where locations mentions have a vocabulary limited to countries and their main regions or cities while the name of organizations or people mentions in news articles are more subject to \textbf{evolve with time} and \textbf{domain}.

\subsection{Evaluated Word Representations}
In this work, we mainly evaluate the  state-of-the-art BiLSTM-CRF architecture \citep{Huang2015BidirectionalTagging} using different types of representations detailed in \autoref{sub:03:representations}.

\paragraph{}
\textbf{Word Embeddings}
map each word to a single vector which results in a lexical representation. 
We take \textbf{GloVe 840B} embeddings \citep{pennington-etal-2014-glove} trained on Common Crawl as the pretrained word embeddings baseline and fine-tune them as traditionally done in related work.

\paragraph{} \textbf{Character-level word embeddings} are learned by a word-level neural network from character embeddings to incorporate orthographic and morphological features.
We reproduce the \textbf{Char-BiLSTM} from \citep{lample-etal-2016-neural}. 
It is trained jointly with the NER model and its outputs are concatenated to GloVe embeddings.
We also separate the non contextual Char-CNN layer in ELMo to isolate the effect of LM contextualization and denote it \textbf{ELMo[0]}.

\paragraph{} \textbf{Contextualized word embeddings} take into account the context of a word in its representation, contrary to previous representations.
A LM is pretrained and used to predict the representation of a word given its context. 
\textbf{ELMo} \citep{peters-etal-2018-deep} uses a Char-CNN to obtain a context-independent word embedding and the concatenation of a forward and backward two-layer LSTM LM for contextualization.
These representations are summed with weights learned for each task as the LM is frozen after pretraining.
\textbf{BERT} \citep{devlin-etal-2019-bert} uses WordPiece subword embeddings \citep{Wu2016GooglesTranslation} and learns a representation 
modeling both left and right contexts by training a Transformer encoder \citep{Vaswani2017AttentionNeed} for Masked LM and next sentence prediction.
For a fairer comparison, we use the BERT\textsubscript{LARGE} feature-based approach where the LM is not fine-tuned and its last four hidden layers are concatenated.
\textbf{Flair} \citep{akbik-etal-2018-contextual} uses a character-level LM for contextualization. 
As in ELMo, they train two opposite LSTM LMs, freeze them and concatenate the predicted states of the first and last characters of each word.
Flair and ELMo are pretrained on the 1 Billion Word Benchmark \citep{Chelba2013OneModeling} while BERT uses Book Corpus \citep{Zhu2015AligningBooks} and English Wikipedia.

\subsection{Experiments}

In order to compare the different embeddings, we feed them as input to a classifier.

We first use the state-of-the-art \textbf{BiLSTM-CRF} \citep{Huang2015BidirectionalTagging} with hidden size 100 in each direction and present in-domain results on all datasets in \autoref{table:03:in-domain}.

\def\s{0.55}
\newcommand{\stddev}{\scalebox{\s}}{}
\def\dimspace{0.5em}

\begin{table}[h]
\begin{adjustwidth*}{}{-.3\textwidth}
\caption{In-domain micro-F1 scores of the BiLSTM-CRF. We split mentions by novelty: \textit{exact match} (EM), \textit{partial match} (PM) and \textit{new}. Average of 5 runs, subscript denotes standard deviation. 
}

\label{table:03:in-domain}
\centering
 \begin{tabularx}{1.3\textwidth}{@{}lr@{\hspace{\dimspace}}*{4}{Y}r*{4}{Y}r*{3}{Y}r@{}}

 \toprule
  &
 & \multicolumn{4}{c}{CoNLL03} & \hspace{\dimspace} & \multicolumn{4}{c}{OntoNotes$^{\ast}$} & \hspace{\dimspace} & \multicolumn{3}{c}{WNUT$^{\ast}$}\\
\cmidrule{3-6} \cmidrule{8-11} \cmidrule{13-15} 
Embedding  & Dim\hspace{\dimspace} & EM & PM & New & All & & EM & PM & New & All & & PM & New & All\\
\midrule

BERT                & 4096\hspace{\dimspace} & 95.7$_{\stddev{.1}}$ & 88.8$_{\stddev{.3}}$ & 82.2$_{\stddev{.3}}$ & 90.5$_{\stddev{.1}}$ &
                    & 96.9$_{\stddev{.2}}$ & 88.6$_{\stddev{.3}}$ & 81.1$_{\stddev{.5}}$ & \textbf{93.5}$_{\stddev{.2}}$ &
                    & 77.0$_{\stddev{4.6}}$ & 53.9$_{\stddev{.9}}$ & \textbf{57.0}$_{\stddev{1.0}}$\\

ELMo                & 1024\hspace{\dimspace} & 95.9$_{\stddev{.1}}$ & 89.2$_{\stddev{.5}}$ & 85.8$_{\stddev{.7}}$ & \textbf{91.8}$_{\stddev{.3  }}$ &
                    & 97.1$_{\stddev{.2}}$ & 88.0$_{\stddev{.2}}$ & 79.9$_{\stddev{.7}}$ & \textbf{93.4}$_{\stddev{.2}}$ &
                    & 67.7$_{\stddev{3.2}}$ & 49.5$_{\stddev{.9}}$ & 52.1$_{\stddev{1.0}}$\\

Flair               & 4096\hspace{\dimspace} & 95.4$_{\stddev{.1}}$ & 88.1$_{\stddev{.6}}$ & 83.5$_{\stddev{.5}}$ & 90.6$_{\stddev{.2}}$ &
                    & 96.7$_{\stddev{.1}}$ & 85.8$_{\stddev{.5}}$ & 75.0$_{\stddev{.6}}$ & 92.1$_{\stddev{.2}}$ &
                    & 64.9$_{\stddev{.7}}$ & 48.2$_{\stddev{2.0}}$ & 50.4$_{\stddev{1.8}}$\\

\cdashline{1-15}
ELMo[0]             & 1024\hspace{\dimspace}
                    & 95.8$_{\stddev{.1}}$ & 87.2$_{\stddev{.2}}$ & 83.5$_{\stddev{.4}}$ & 90.7$_{\stddev{.1}}$ &
                    & 96.9$_{\stddev{.1}}$ & 85.9$_{\stddev{.3}}$ & 75.5$_{\stddev{.6}}$ & 92.4$_{\stddev{.1}}$ &
                    & 72.8$_{\stddev{1.3}}$ & 45.4$_{\stddev{2.8}}$ & 49.1$_{\stddev{2.3}}$ \\
                    
GloVe + char & 350\hspace{\dimspace} 
                    & 95.3$_{\stddev{.3}}$ & 85.5$_{\stddev{.7}}$ & 83.1$_{\stddev{.7}}$ & 89.9$_{\stddev{.5}}$ &
                    & 96.3$_{\stddev{.1}}$ & 83.3$_{\stddev{.2}}$ & 69.9$_{\stddev{.6}}$ & 91.0$_{\stddev{.1}}$ &
                    & 63.2$_{\stddev{4.6}}$ & 33.4$_{\stddev{1.5}}$ & 38.0$_{\stddev{1.7}}$\\
                    
GloVe               & 300\hspace{\dimspace} 
                    & 95.1$_{\stddev{.4}}$ & 85.3$_{\stddev{.5}}$ & 81.1$_{\stddev{.5}}$ & 89.3$_{\stddev{.4}}$ &
                    & 96.2$_{\stddev{.2}}$ & 82.9$_{\stddev{.2}}$ & 63.8$_{\stddev{.5}}$ & 90.4$_{\stddev{.2}}$ &
                    & 59.1$_{\stddev{2.9}}$ & 28.1$_{\stddev{1.5}}$ & 32.9$_{\stddev{1.2}}$ \\
 \bottomrule
 \end{tabularx}
\end{adjustwidth*}
\end{table}

We then report out-of-domain performance in \autoref{table:03:ood}. In order to better capture the intrinsic effect of LM contextualization, we introduce the\textbf{ Map-CRF} baseline from \citep{akbik-etal-2018-contextual} where the BiLSTM is replaced by a simple linear projection of each word representation. We only consider domain adaptation from CoNLL03 to OntoNotes$^\ast$
and WNUT$^\ast$
assuming that labeled data is scarcer, less varied and more generic than target data in real use cases.

We use the IOBES tagging scheme for NER and no preprocessing. We fix a batch size of 64, a learning rate of 0.001 and a 0.5 dropout rate at the embedding layer and after the BiLSTM or linear projection.
The maximum number of epochs is set to 100 and we use early stopping with patience 5 on validation global micro-F1.
For each configuration, we use the best performing optimization method between SGD and Adam with $\beta_1=0.9$ and $\beta_2=0.999$. We report the mean and standard deviation of five runs.

 \begin{table}[h]
 
 \begin{adjustwidth*}{}{-.3\textwidth}
\caption{Micro-F1 scores of models trained on CoNLL03 and tested in-domain and out-of-domain on OntoNotes$^\ast$ and WNUT$^\ast$. 
Average of 5 runs, subscript denotes standard deviation.}
\label{table:03:ood}
\centering

 \begin{tabularx}{1.3\textwidth}{@{}l@{\hspace{0em}}lr*{4}{Y}r*{4}{Y}r*{4}{Y}r@{}}
 \toprule


 &  & & \multicolumn{4}{c}{CoNLL03} & \hspace{.5em} & \multicolumn{4}{c}{OntoNotes$^\ast$} & \hspace{.5em} & \multicolumn{4}{c}{WNUT$^\ast$}\\
  \cmidrule{4-7} \cmidrule{9-12} \cmidrule{14-17} 
 & Emb. & & EM & PM & New & All & & EM & PM & New & All & & EM & PM & New & All \\
\midrule

 \parbox[t]{4mm}{\multirow{6}{*}{\rotatebox[origin=c]{90}{\textbf{BiLSTM-CRF}}}} 
                      & BERT                   &    & 95.7$_{\stddev{.1}}$ & 88.8$_{\stddev{.3}}$ & 82.2$_{\stddev{.3}}$ & 90.5$_{\stddev{.1}}$ &
                                                    & 95.1$_{\stddev{.1}}$ & 82.9$_{\stddev{.5}}$ & 73.5$_{\stddev{.4}}$ & \textbf{85.0}$_{\stddev{.3}}$ & 
                                                    & 57.4$_{\stddev{1.0}}$ & 56.3$_{\stddev{1.2}}$ & 32.4$_{\stddev{.8}}$ & 37.6$_{\stddev{.8}}$\\
                                                    
                      & ELMo                   &    & 95.9$_{\stddev{.1}}$ & 89.2$_{\stddev{.5}}$ & 85.8$_{\stddev{.7}}$ & \textbf{91.8}$_{\stddev{.3}}$ &
                                                    & 94.3$_{\stddev{.1}}$ & 79.2$_{\stddev{.2}}$ & 72.4$_{\stddev{.4}}$ & 83.4$_{\stddev{.2}}$ &
                                                    & 55.8$_{\stddev{1.2}}$ & 52.7$_{\stddev{1.1}}$ & 36.5$_{\stddev{1.5}}$ & \textbf{41.0}$_{\stddev{1.2}}$\\
                                                    
                      & Flair                  &    & 95.4$_{\stddev{.1}}$ & 88.1$_{\stddev{.6}}$ & 83.5$_{\stddev{.5}}$ & 90.6$_{\stddev{.2}}$ &
                                                    & 94.0$_{\stddev{.3}}$ & 76.1$_{\stddev{1.1}}$ & 62.1$_{\stddev{.5}}$ & 79.0$_{\stddev{.5}}$ &
                                                    & 56.2$_{\stddev{2.2}}$ & 49.4$_{\stddev{3.4}}$ & 29.1$_{\stddev{3.3}}$ & 34.9$_{\stddev{2.9}}$\\
\cdashline{2-17}
                      & ELMo[0] &    & 95.8$_{\stddev{.1}}$ & 87.2$_{\stddev{.2}}$ & 83.5$_{\stddev{.4}}$ & 90.7$_{\stddev{.1}}$ &
                                                    & 93.6$_{\stddev{.1}}$ & 76.8$_{\stddev{.6}}$ & 66.1$_{\stddev{.3}}$ & 80.5$_{\stddev{.2}}$ &
                                                    & 52.3$_{\stddev{1.2}}$ & 50.8$_{\stddev{1.5}}$ & 32.6$_{\stddev{2.2}}$ & 37.6$_{\stddev{1.8}}$\\
                                                    
                      & G + char              &     & 95.3$_{\stddev{.3}}$ & 85.5$_{\stddev{.7}}$ & 83.1$_{\stddev{.7}}$ & 89.9$_{\stddev{.5}}$ &
                                                    & 93.9$_{\stddev{.2}}$ & 73.9$_{\stddev{1.1}}$ & 60.4$_{\stddev{.7}}$ & 77.9$_{\stddev{.5}}$ &
                                                    & 55.9$_{\stddev{.8}}$ & 46.8$_{\stddev{1.8}}$ & 19.6$_{\stddev{1.6}}$ & 27.2$_{\stddev{1.3}}$\\
                                                    
                      & GloVe                  &    & 95.1$_{\stddev{.4}}$ & 85.3$_{\stddev{.5}}$ & 81.1$_{\stddev{.5}}$ & 89.3$_{\stddev{.4}}$ &
                                                    & 93.7$_{\stddev{.2}}$ & 73.0$_{\stddev{1.2}}$ & 57.4$_{\stddev{1.8}}$ & 76.9$_{\stddev{.9}}$ &
                                                    & 53.9$_{\stddev{1.2}}$ & 46.3$_{\stddev{1.5}}$ & 13.3$_{\stddev{1.4}}$ & 27.1$_{\stddev{1.0}}$\\
\cmidrule{1-17}
\parbox[t]{4mm}{\multirow{6}{*}{\rotatebox[origin=c]{90}{\textbf{Map-CRF}}}}
                      & BERT                   &    & 93.2$_{\stddev{.3}}$ & 85.8$_{\stddev{.4}}$ & 73.7$_{\stddev{.8}}$ & 86.2$_{\stddev{.4}}$ &
                                                    & 93.5$_{\stddev{.2}}$ & 77.8$_{\stddev{.5}}$ & 67.8$_{\stddev{.9}}$ & 80.9$_{\stddev{.4}}$ &
                                                    & 57.4$_{\stddev{.3}}$ & 53.5$_{\stddev{2.6}}$ & 33.9$_{\stddev{.6}}$ & 38.4$_{\stddev{.4}}$\\
                                                    
                      & ELMo                   &    & 93.7$_{\stddev{.2}}$ & 87.2$_{\stddev{.6}}$ & 80.1$_{\stddev{.3}}$ & \textbf{88.7}$_{\stddev{.2}}$ &
                                                    & 93.6$_{\stddev{.1}}$ & 79.1$_{\stddev{.5}}$ & 69.5$_{\stddev{.4}}$ & \textbf{82.2}$_{\stddev{.3}}$ &
                                                    & 61.1$_{\stddev{.7}}$ & 53.0$_{\stddev{.9}}$ & 37.5$_{\stddev{.7}}$ & \textbf{42.4}$_{\stddev{.6}}$\\

                      & Flair                   &   & 94.3$_{\stddev{.1}}$ & 85.1$_{\stddev{.3}}$ & 78.6$_{\stddev{.3}}$ & 88.1$_{\stddev{.03}}$ &
                                                    & 93.2$_{\stddev{.1}}$ & 74.0$_{\stddev{.3}}$ & 59.6$_{\stddev{.2}}$ & 77.5$_{\stddev{.2}}$ &
                                                    & 52.5$_{\stddev{1.2}}$ & 50.6$_{\stddev{.4}}$ & 28.8$_{\stddev{.5}}$ & 33.7$_{\stddev{.5}}$\\
\cdashline{2-17}
                      & ELMo[0]                 &   & 92.2$_{\stddev{.3}}$ & 80.5$_{\stddev{1.0}}$ & 68.6$_{\stddev{.4}}$ & 83.4$_{\stddev{.4}}$ &
                                                    & 91.6$_{\stddev{.4}}$ & 69.6$_{\stddev{1.0}}$ & 56.8$_{\stddev{1.5}}$ & 75.0$_{\stddev{1.0}}$ &
                                                    & 51.9$_{\stddev{1.1}}$ & 42.6$_{\stddev{.9}}$ & 32.4$_{\stddev{.3}}$ & 35.8$_{\stddev{.4}}$\\

                      & G + char                &   & 93.1$_{\stddev{.3}}$ & 80.7$_{\stddev{.9}}$ & 69.8$_{\stddev{.7}}$ & 84.4$_{\stddev{.4}}$ &
                                                    & 91.8$_{\stddev{.3}}$ & 69.3$_{\stddev{.3}}$ & 55.6$_{\stddev{1.1}}$ & 74.8$_{\stddev{.5}}$ & 
                                                    & 50.6$_{\stddev{.9}}$ & 42.5$_{\stddev{1.4}}$ & 20.6$_{\stddev{2.8}}$ & 28.7$_{\stddev{2.5}}$\\
                                                    
                      & GloVe                   &   & 92.2$_{\stddev{.1}}$ & 77.0$_{\stddev{.4}}$ & 61.7$_{\stddev{.3}}$ & 81.5$_{\stddev{.05}}$ &
                                                    & 89.6$_{\stddev{.3}}$ & 62.8$_{\stddev{.6}}$ & 38.5$_{\stddev{.4}}$ & 68.1$_{\stddev{.4}}$ & 
                                                    & 46.8$_{\stddev{.8}}$ & 41.3$_{\stddev{.5}}$ & 3.2$_{\stddev{.2}}$ & 18.9$_{\stddev{.7}}$\\
 \bottomrule
 \end{tabularx}

\end{adjustwidth*}
\end{table}

\subsection{General Observations} \label{general}

\paragraph{Comparing ELMo, BERT and Flair}
Drawing conclusions from the comparison of ELMo, BERT and Flair is difficult because there is no clear hierarchy across datasets and they differ in dimensions, tokenization, contextualization levels and pretraining corpora.
However, although BERT is particularly effective on the WNUT dataset in-domain, probably due to its subword tokenization, ELMo yields the most stable results in and out-of-domain.

Furthermore, Flair globally underperforms ELMo and BERT, particularly for unseen mentions and out-of-domain. 
This suggests that LM pretraining at a lexical level (word or subword) is more robust for generalization than at a character level.
In fact, Flair only beats the non contextual ELMo[0] baseline with Map-CRF which indicates that \textbf{character-level contextualization is less beneficial than word-level contextualization} with character-level representations.
However, \citet{akbik-etal-2018-contextual} show that Flair is at least complementary with traditional word embeddings such as GloVe.

\paragraph{Comparing ELMo[0] and GloVe+char}

It is also interesting to compare the two non contextual character-level representations of our study: ELMo[0] and GloVe+char.

Overall, \textbf{using ELMo[0] enables to outperform the GloVe+char baseline}, particularly
on unseen mentions, out-of-domain and on WNUT$^\ast$.
The main difference is the incorporation of morphological features: in ELMo[0] they are learned jointly with the LM on a huge dataset whereas the char-BiLSTM is only trained on the source NER training set. Yet, morphology is crucial to represent words never encountered during pretraining and in WNUT$^\ast$ around 20\% of words in test mentions are out of GloVe's pretrained vocabulary against 5\% in CoNLL03 and 3\% in OntoNotes$^\ast$. This explains the poor performance of GloVe baselines on WNUT$^\ast$, all the more out-of-domain, and why a model trained on CoNLL03 with ELMo outperforms one trained on WNUT$^\ast$ with GloVe+char.
Thus, ELMo's improvement over previous state-of-the-art does not
only stem from contextualization but also from \textbf{more effective non-contextual word representations}.

\sloppy \paragraph{Lexical Overlap Bias}
In every configuration and on every dataset, $F1_{EM}~>~F1_{PM}~>~F1_{new}$
with more than 10 points difference between Exact Match and New mentions.
This gap is wider out-of-domain where the context differs more from training data than in-domain.
NER models thus poorly generalize to unseen mentions, and datasets with high lexical overlap only encourage this behavior.
However, \textbf{this generalization gap is reduced by two types of contextualization} described hereafter.

\def\s{0.7}
\newcommand{\stddevtwo}{\scalebox{\s}}{}

\begin{table}[h!]
  \begin{adjustwidth*}{}{-.3\textwidth}
\centering
\caption[Per-genre micro-F1 scores of the BiLSTM-CRF trained on CoNLL03 and tested on OntoNotes$^\ast$.]{Per-genre micro-F1 scores of the BiLSTM-CRF model trained on CoNLL03 and tested on OntoNotes$^\ast$
 (broadcast conversation, broadcast news, news wire, magazine, telephone conversation and web text).
 $C_{LM}$ mostly benefits genres furthest from the news source domain.}
\label{table:03:genres}

 \begin{tabularx}{1.3\textwidth}{@{}l*{7}{Y}@{}}
 \toprule
  & bc & bn & nw & mz & tc & wb & All\\
  \midrule
 BERT           & 87.2$_{\stddevtwo{.5}}$ & 88.4$_{\stddevtwo{.4}}$ & 84.7$_{\stddevtwo{.2}}$ & 82.4$_{\stddevtwo{1.2}}$ & 84.5$_{\stddevtwo{1.1}}$ & 79.5$_{\stddevtwo{1.0}}$ & \textbf{85.0}$_{\stddevtwo{.3}}$ \\
 ELMo           & 85.0$_{\stddevtwo{.6}}$ & 88.6$_{\stddevtwo{.3}}$ & 82.9$_{\stddevtwo{.3}}$ & 78.1$_{\stddevtwo{.7}}$ & 84.0$_{\stddevtwo{.8}}$ & 79.9$_{\stddevtwo{.5}}$ & 83.4$_{\stddevtwo{.2}}$ \\
 Flair          & 78.0$_{\stddevtwo{1.1}}$ & 86.5$_{\stddevtwo{.4}}$ & 80.4$_{\stddevtwo{.6}}$ & 71.1$_{\stddevtwo{.4}}$ & 73.5$_{\stddevtwo{1.8}}$ & 72.1$_{\stddevtwo{.8}}$ & 79.0$_{\stddevtwo{.5}}$ \\
 \cdashline{1-8}
 ELMo[0]        & 82.6$_{\stddevtwo{.5}}$ & 88.0$_{\stddevtwo{.3}}$ & 79.6$_{\stddevtwo{.5}}$ & 73.4$_{\stddevtwo{.6}}$ & 79.2$_{\stddevtwo{1.2}}$ & 75.1$_{\stddevtwo{.3}}$ & 80.5$_{\stddevtwo{.2}}$ \\
 GloVe + char \hspace{2em}& 80.4$_{\stddevtwo{.8}}$ & 86.3$_{\stddevtwo{.4}}$ & 77.0$_{\stddevtwo{1.0}}$ & 70.7$_{\stddevtwo{.4}}$ & 79.7$_{\stddevtwo{1.8}}$ & 69.2$_{\stddevtwo{.8}}$ & 77.9$_{\stddevtwo{.5}}$ \\
 \bottomrule
 \end{tabularx}

  \end{adjustwidth*}
\end{table}

\subsection{LM and NER Contextualizations} \label{analysis}

\begin{figure}[!h]
\begin{adjustwidth*}{}{-.3\textwidth}
\caption[Schema of the two types of contextualizations: $\mathcal{C}_{LM}$ and $\mathcal{C}_{NER}$.]{Schema of the two types of contextualizations: $\mathcal{C}_{LM}$ and $\mathcal{C}_{NER}$.}
\label{fig:03:iobes}
\centering\includegraphics[width=1.3\textwidth]{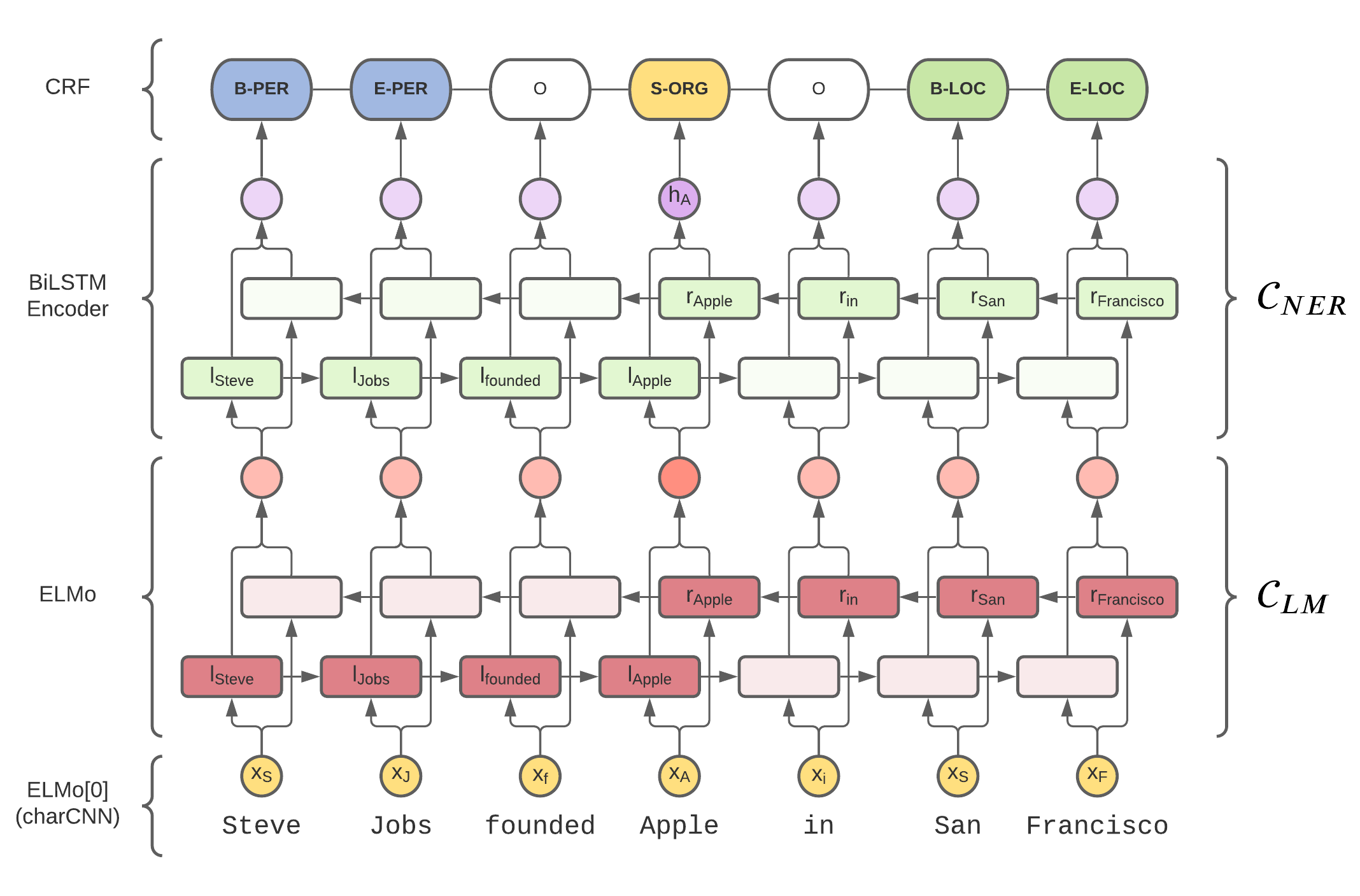}
\end{adjustwidth*}
\end{figure}

The ELMo[0] and Map-CRF baselines enable to distinguish contextualization due to LM pretraining
(\textbf{$C_{LM}$: ELMo[0] to ELMo}) from task supervised contextualization induced by the BiLSTM network (\textbf{$C_{NER}$: Map to BiLSTM}).
Indeed, in both cases a BiLSTM incorporates syntactic information which improves generalization to unseen mentions for which
context is decisive, as shown in \autoref{table:03:ood}.
The model cannot learn to associate their lexical features to the corresponding entity type during training
and must rely on contextual cues.
The main distinction lies in their training: ELMo is trained as a Language Model on a large corpus of text (0.8 Billion tokens) while the NER
model BiLSTM is trained on an orders of magnitude smaller dataset (0.2 Million tokens) with entity type annotations.

\paragraph{Comparison}

We can first compare the separate contribution of both contextualizations.
Our experiments indicate that ${C_{NER}}$ is more valuable than $C_{LM}$ in-domain, which can be explained because 
the supervised BiLSTM is specifically trained on the source dataset, hence more adapted to the test domain.

On the contrary, we can observe that $C_{LM}$ is particularly helpful out-of-domain.
In the latter setting, the benefits from $C_{LM}$ even surpass those from ${C_{NER}}$, specifically on domains further
from source data such as web text in OntoNotes$^\ast$ (see \autoref{table:03:genres}) or WNUT$^\ast$.
This is again explained by the difference in quantity and quality of the corpora on which these contextualizations are learned.
The much larger and more generic unlabeled corpora on which LM are pretrained lead to contextual representations more
robust to domain adaptation than ${C_{NER}}$ learned on a small and specific NER corpus.

Similar behaviors can be observed when comparing BERT and Flair to the GloVe baselines, although we cannot separate the effects of representation and contextualization.

\paragraph{Complementarity}

We can then observe the complementarity of both effects comparing their combination in the BiLSTM-CRF + ELMo baseline to
their individual applications in either Map-CRF + ELMo or BiLSTM-CRF + ELMo[0].
Both in-domain and in out-of-domain evaluation from CoNLL03 to OntoNotes$^\ast$, the two types of contextualization transfer complementary syntactic
features leading to the best configuration.
However,  in the most difficult case of zero-shot domain adaptation from CoNLL03 to WNUT$^\ast$,
${C_{NER}}$ is detrimental with ELMo and BERT. This is probably due to the specificity of the user generated target domain,
excessively different from news articles in the source data.

\subsection{Qualitative Analysis}

\begin{figure}[!h]
\begin{adjustwidth*}{}{-.3\textwidth}
\caption{In-domain qualitative examples from CoNLL03 test set. Bold words are not present in the training vocabulary.}
\label{fig:03:qualitative conll}

\centering\includegraphics[width=1.3\textwidth]{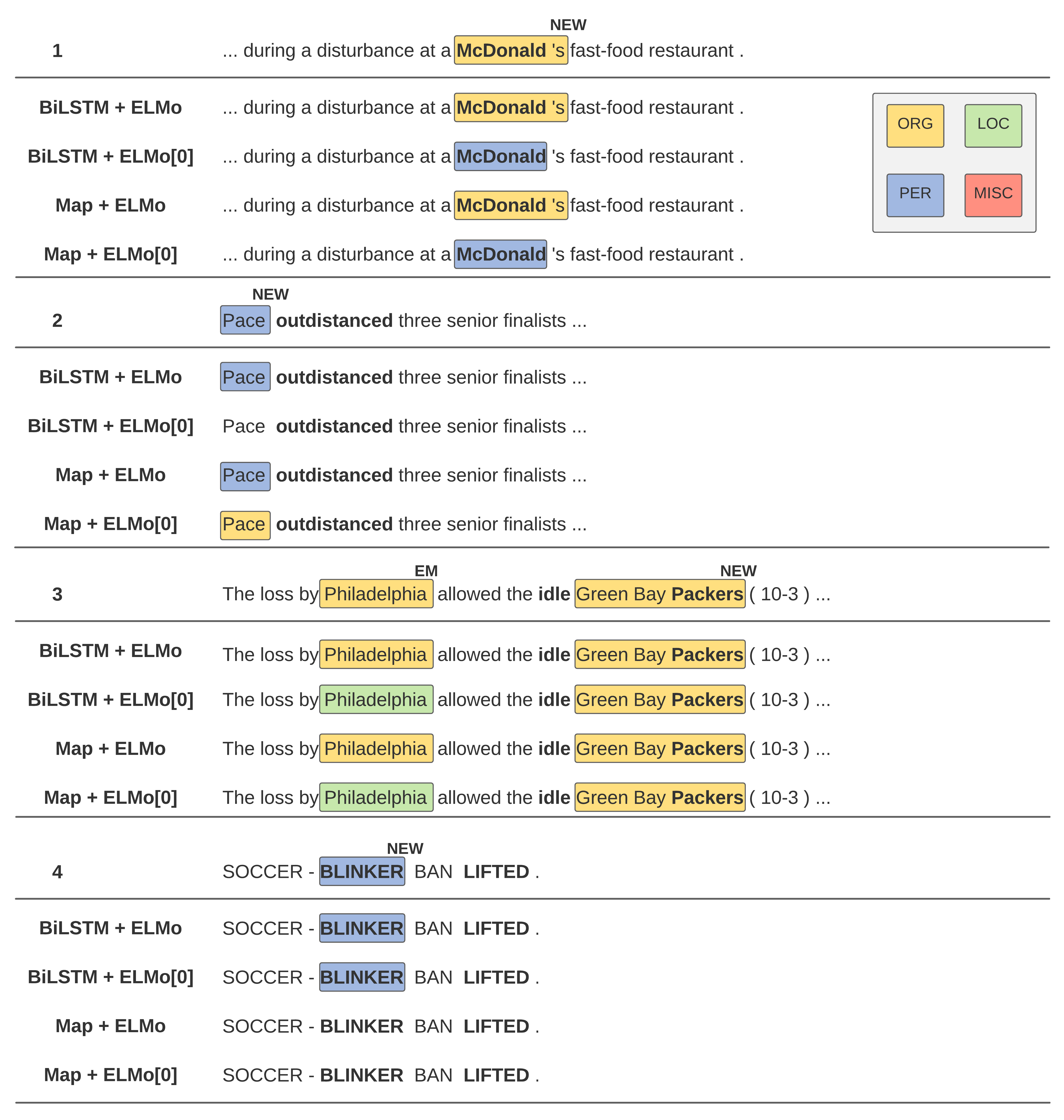}
\end{adjustwidth*}
\end{figure}

\begin{figure}[!h]
\begin{adjustwidth*}{}{-.3\textwidth}
\caption{Out-of-domain qualitative examples from OntoNotes$^\ast$ test set. Bold words are not present in the training vocabulary.}
\label{fig:03:qualitative ood}

\centering\includegraphics[width=1.3\textwidth]{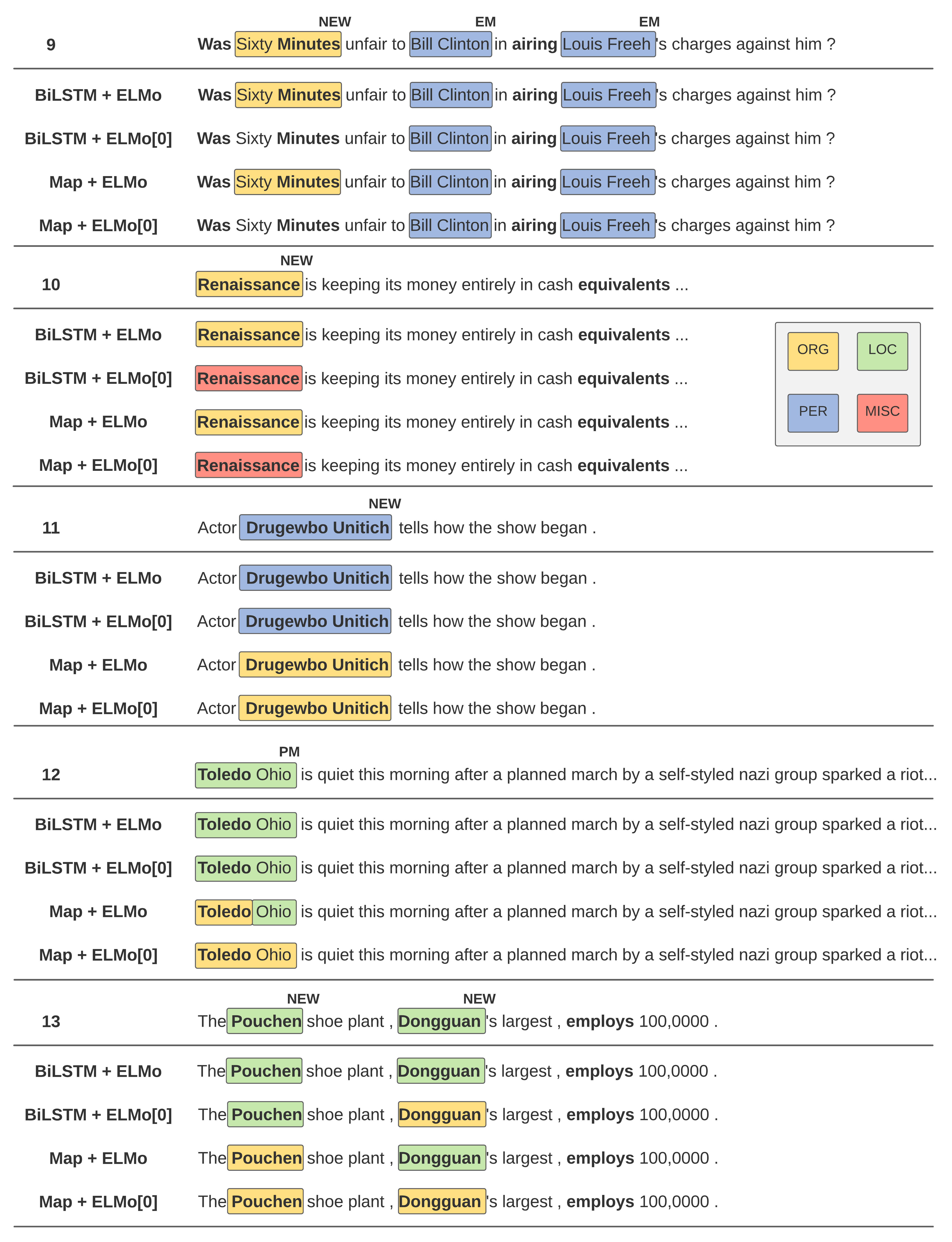}
\end{adjustwidth*}
\end{figure}

 In order to perform qualitative comparison of the effects of LM contextualization and task contextualization, we report qualitative results obtained with Map-CRF and BiLSTM-CRF with both ELMo and ELMo[0].
 For each configuration, we select the model yielding the in-domain test micro-F1 score the closest to the reported average over our five runs.
 In-domain examples are reported in \autoref{fig:03:qualitative conll} whereas some out-of-domain examples on OntoNotes are presented in \autoref{fig:03:qualitative ood}.

 We observe that contextualization is mainly useful for disambiguation as mention detection is mostly correct and classification is the true difficulty.
 This can be explained by the semantic features learned by ELMo[0] during pretraining (as likely in examples 3, 10, 17) or with more orthographic and morphological features such as first letter capitalization (as likely in examples 11 and 14 where the non contextual baseline predicts an incoherent type).

 We can first see that ELMo seems to favor detection of frequent n-grams entities such as \textit{McDonald's} in example 1 or \textit{Sixty Minutes} in example 9.
 In both examples, supervised contextualization is ineffective because the entities are not present in the training data and for example the ELMo[0] embedding for the unigram McDonald is close to a person.
 This shows that immediate context influences the representation as in the often cited\textit{ Washington DC.} / \textit{Georges Washington} example.
 However examples 2, 3 and 10 seem to be more based on lexical field disambiguation, e.g. in 10 whereas the ELMo[0] embedding for \textit{Renaissance} is closer to MISC, the money lexical field is a clue for an organization.

 On the contrary examples where $C_{NER}$ outperforms $C_{LM}$ on CoNLL03 are very specific, often corresponding to sentences in all caps and in the sports domain.
 In this case, as in example 5, city names often refer to clubs and are thus tagged as ORG.
 The majority of the LM pretraining corpus is not capitalized and names of cities most often denote locations, hence its inability to process such data.
 Such examples are scarcer out-of-domain and it is more difficult to explain them, as in examples 11 or 12.
 Similarly it is difficult to propose a convincing interpretation of examples where both contextualizations are required to obtain the correct tag.

\begin{figure}[ht]
\begin{adjustwidth*}{}{-.3\textwidth}
\centering\includegraphics[width=1.3\textwidth]{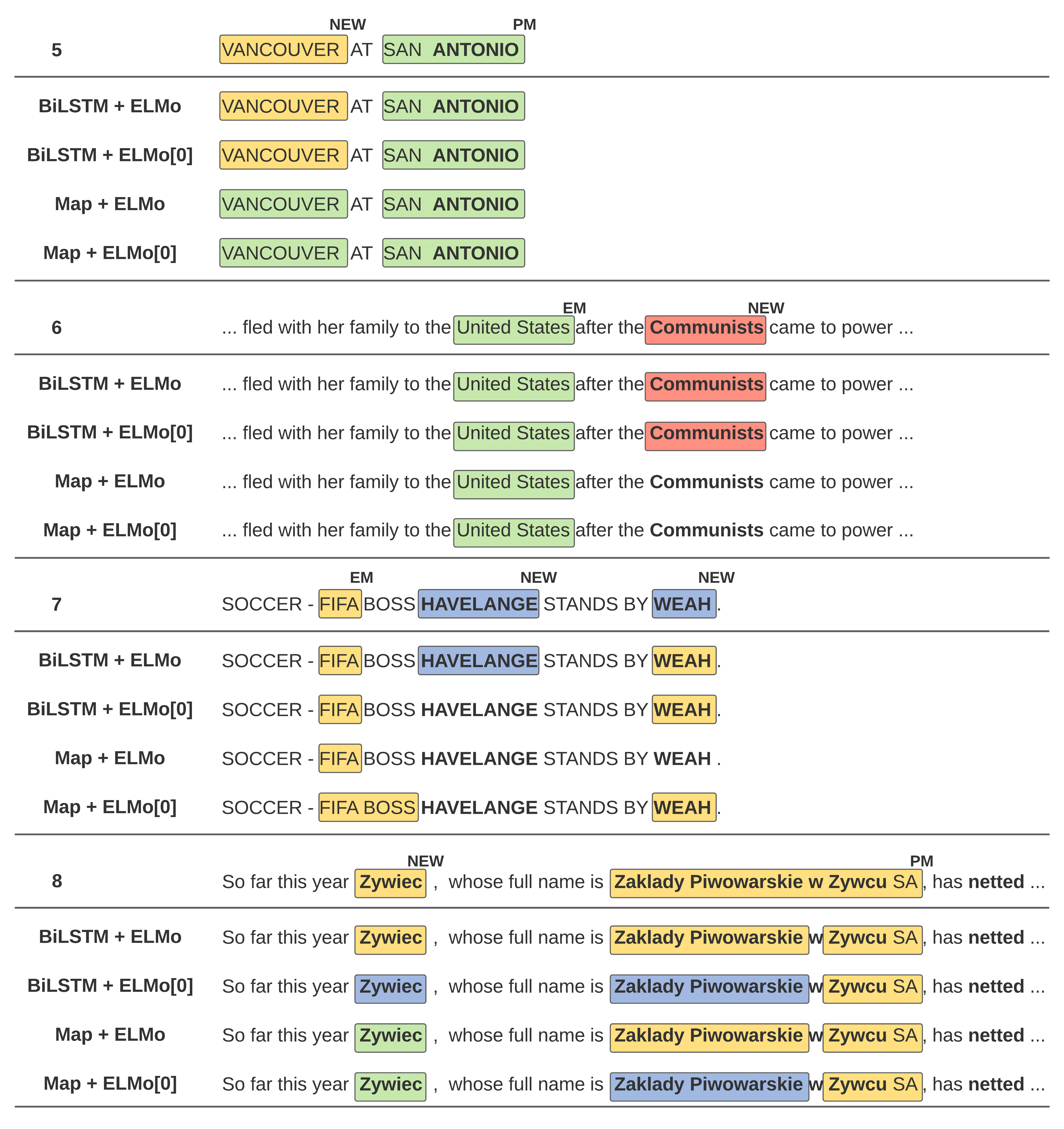}
\end{adjustwidth*}
\end{figure}

\begin{figure}[ht]
\begin{adjustwidth*}{}{-.3\textwidth}
\centering\includegraphics[width=1.3\textwidth]{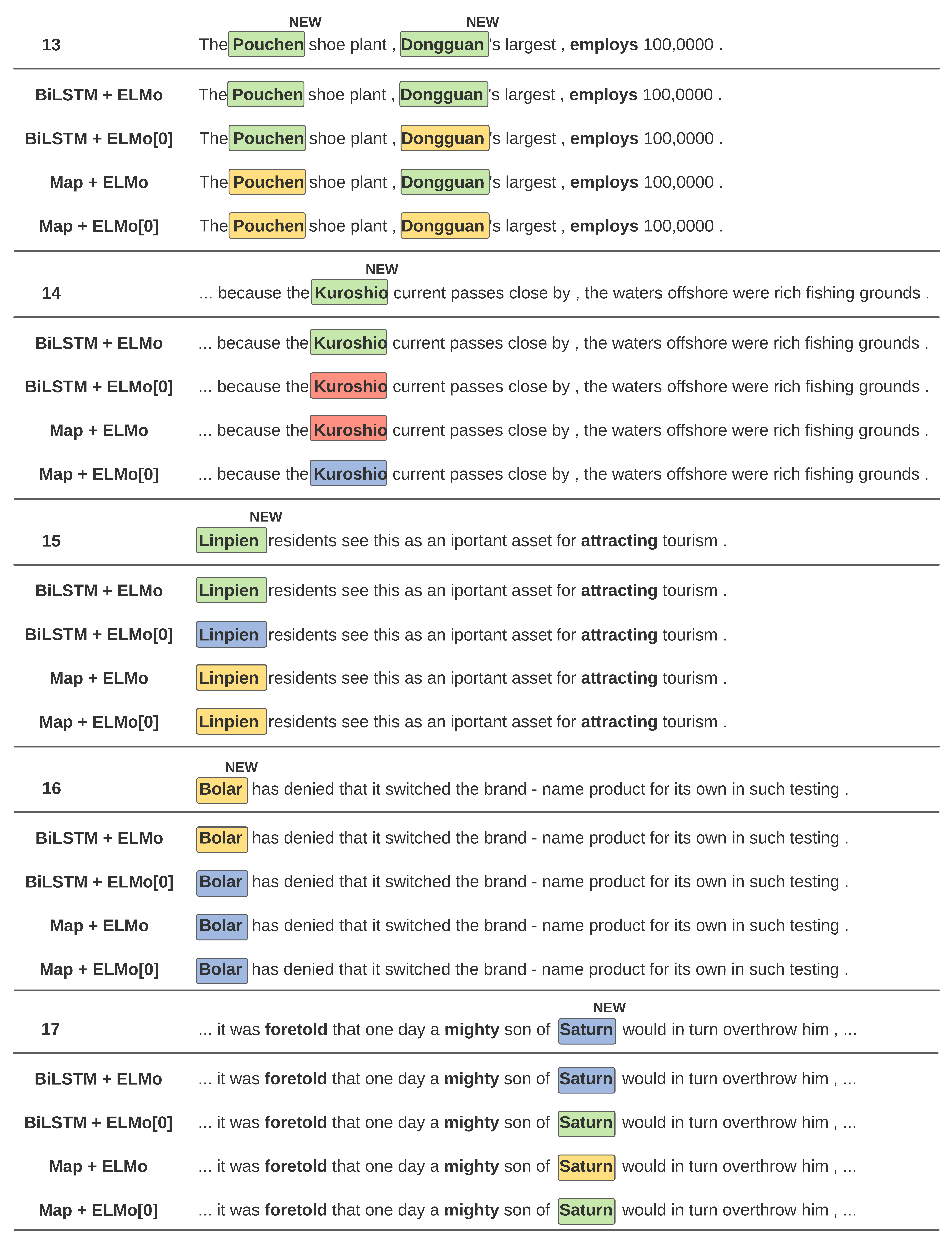}
\end{adjustwidth*}
\end{figure}

\clearpage
\section{Conclusion and Perspectives}
Standard English NER benchmarks are biased towards seen mentions whereas Named Entity Recognition is mostly useful to detect unknown mentions in real-life applications.
It is hence necessary to disentangle performance on seen and unseen mentions and test out-of-domain to better measure the generalization capabilities of NER models.
In such setting, we show that contextualization from LM pretraining is particularly beneficial for generalization to unseen mentions, all the more out-of-domain where it surpasses supervised contextualization.

Other complementary works have been proposed concurrently or after ours.
\citet{pires-etal-2019-multilingual} explore the zero-shot cross-lingual capabilites of Multilingual BERT on a Google internal corpus for NER in 16 languages. They study generalization in all pairs of these languages and propose to measure the vocabulary overlap between BERT subtokens in the entities in the train and evaluations sets. They show that Multilingual BERT's zero-shot performance does not depend on this overlap, indicating that it learns multilingual representations.

\citet{arora-etal-2020-contextual} study the influence of lexical overlap, length of entities and ambiguity in NER with Transformer based LM pretraining as well as traditional GloVe embeddings.
Alternatively to our separation, they classify the test samples according to these properties and separate them in two halves.
They then measure the relative improvement of using BERT on the lower and higher half and confirm that it is more useful on less seen mentions and more ambiguous or long ones.
Alternatively, \citet{Fu2020RethinkingStudy, fu-etal-2020-interpretable} propose to measure the impact of similar properties in a similar manner with more than two buckets. They introduce the label consistency of a mention in the test set as the number of occurrences it is tagged with the same label in the training set divided by the total number of occurrences if it appears, and zero otherwise.
It hence regroups two phenomena: lexical overlap (zero label consistency) and ambiguity. 
They conclude that entity length is the factor that is the more negatively correlated with NER performance.
This work is also the first introduction of \href{http://explainaboard.nlpedia.ai/}{Explainaboard} \citep{liu-etal-2021-explainaboard}, an interactive website which goal is to compare performance of state-of-the-art NLP models depending on test sample properties to identify their individual strengths and weaknesses.

In the following chapters, we extend this study of lexical overlap to the broader end-to-end Relation Extraction task that aims at identifying both entity mentions and the relations expressed between them.

\chapter[A Taxonomy of Entity and Relation Extraction]{A Taxonomy of Entity and Relation Extraction}
\label{chapter:04:re-taxonomy}

Although entity mentions play an important role as explicit descriptors of the Who, What, Where, and When elements in a sentence, understanding a statement also requires to identify how these elements are linked with one another.
This structure between elements in a sentence plays an important role in Information Extraction which role can be defined as \textbf{converting the information expressed in free text into a predefined structured format of knowledge}.

Hence, Information Extraction was quickly modeled as a \textbf{template filling} task by the second Message Understanding Conference (MUC-2) in 1989 after a first exploratory meeting in 1987 initiated by the US National Institute of Standards and Technology (NIST). 
Given a description of different classes of \textbf{events}, one as to fill a template corresponding to each event by filling \textbf{slots} for information regarding for example the type of event, its agents, time and place \citep{ Hirschman1998TheConferences}.
This requires to extract \textbf{relational information} to correctly associate an event with its characteristics.
MUC-6 \citep{grishman-sundheim-1996-message} further distinguishes events and entities by introducing Named Entity Recognition, Coreference Resolution and the "Template Element" task.
While NER and Coreference Resolution aim at detecting textual mentions of entities, the Template Element task serves as a bridge between them by extracting \textbf{entity templates} with different slots such as NAME (all aliases), TYPE (entity types) or CATEGORY (entity subtypes).

The notion of \textbf{relation} only explicitly appears in MUC-7 \citep{chinchor-1998-overview} which introduces the additional ``Template Relation'' task that consists in linking template elements with three types of relations to organizations: ``employee of'', ``product of'' and ``location of''.

Relation Extraction is further explored in a punctual SemEval shared task \citep{hendrickx-etal-2010-semeval} and the successor of the MUCs: the Automatic Content Extraction (ACE) program launched in 1999 \citep{doddington-etal-2004-automatic} on English, Arabic and Chinese texts.
Initially dedicated to ``Entity Detection and Tracking'', relations between entities were introduced in 2002 with five coarse grained types of semantic relations (such as ``role'' between a person and an organization) further divided into 24 subtypes (such as ``founder'', ``member'' or ``client'') in ACE-2 and ACE2003. 
The ACE2004 and ACE2005 datasets are still standard benchmarks for testing Information Extraction algorithms.
ACE progressively evolved into the latest Information Extraction program supported by NIST is the Text Analysis Conference (TAC) that started in 2008 and is still active nowadays.
In particular, the TAC-KBP (Knowledge Base Population) shared tasks since 2009 \citep{ji-grishman-2011-knowledge} provide useful resources for studying Relation Extraction more recently compiled into the popular TACRED Relation Extraction dataset \citep{zhang-etal-2017-position}.

These initiatives enabled to build standard datasets and benchmarks to develop models designed to specifically tackle the Relation Extraction subtask, that is a key element in a more general Information Extraction goal.
However, identifying relations between entities supposes to detect these entities and RE is hence tightly linked to Named Entity Recognition.
Thus, even though the traditional approach is to consider NER and RE as separate tasks and apply a NER and a RE model sequentially in a \textbf{pipeline structure}, an \textbf{end-to-end approach} has been explored to better model the interdependency of both tasks.

\begin{figure}[h!]
    \centering
    \caption{Example of Gold standard annotation for end-to-end Relation Extraction.}
    \label{fig:04:ere_gt}
    \includegraphics[width=.9\textwidth]{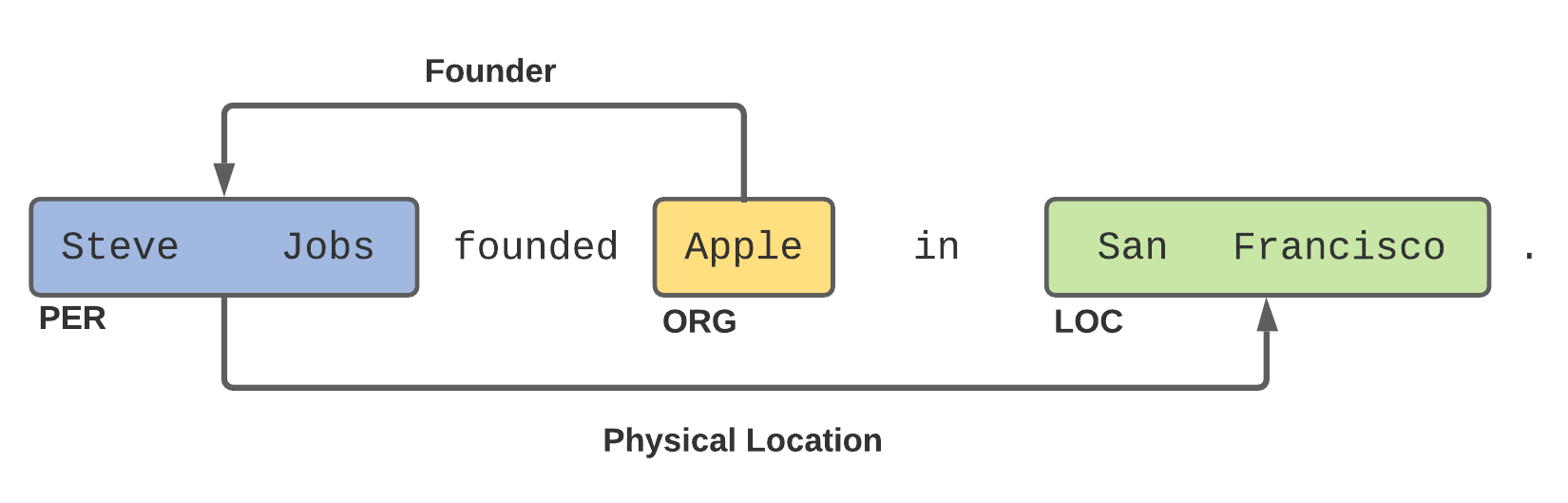}
\end{figure}

In this chapter, we propose a general taxonomy of Relation Extraction models focused on end-to-end Relation Extraction.

\section{Context}
\label{sec:04:taxonomy}

Relation Extraction (RE) aims at detecting and classifying \textbf{semantic relations} between entities mentioned in a text.
However the notion of relation can enclose different settings for this task which it is necessary to properly distinguish.

First, it is common to limit to \textbf{binary relations} that connect two entities such as ``child of'', ``employed by'' or ``physical location'' and that can be extended to n-ary relations.
Directed binary relations can thus be represented as \textbf{(subject, predicate, object) triples} such as (Apple, founder, Steve Jobs).

Second, depending on the final application, RE can be performed at the \textbf{textual mention level} or at the \textbf{entity level}, requiring the additional Coreference Resolution and Entity Linking steps that \textbf{link mentions to entities in an ontology}.
This entails different granularity levels of extraction since RE between textual mentions are extracted at a \textbf{sentence or document level} whereas relations between entities can also be extracted at a \textbf{corpus level} by aggregating all the instances expressing the same relations, for example with \textbf{Multi-Instance Learning}.

Another distinction is in the degree of supervision of RE. \textbf{Open Information Extraction (Open IE)} proposes to extract such relations without labels, mainly relying on the detection of verbs and their respective subjects and objects. One difficulty is to then map the extracted relation phrases to the corresponding predicate in an ontology.
More classically, RE is \textbf{supervised} with relation instances manually annotated for a predefined set of relation types.
Because this manual annotation is tedious and expensive, \textbf{distant supervision} \citep{mintz-etal-2009-distant} proposes to automatically annotate a text corpus with facts present in an existing \textbf{Knowledge Base}. 

In this work, we focus on the supervised extraction of binary relations between textual mentions at the sentence-level and on English documents.

\section{An Introduction to the Pipeline Approach}

The pipeline approach views NER and RE as two separate tasks performed with different models \textbf{designed separately and applied sequentially}.
In this case, Relation Extraction is performed with the assumption that entities have been previously detected and RE is thus trained with ground truth information about relation arguments.
It can thus be formulated as a \textbf{Relation Classification} task given a sentence and two entity mentions it contains 

The first set of methods explored were \textbf{rule-based}, using handcrafted patterns corresponding to specific relations. For example for the relation ``hyponym'', lexico-syntactic patterns appearing between noun phrases such as ``including'', ``such as'' or ``especially'' can be used as cues for RE. For this specific relation, other patterns can be added to obtain around 60\% of recall on a subset of an encyclopedia \citep{hearst-1992-automatic}. The main drawback of this approach is the fact that every pattern needs to be carefully designed and they are totally different from one relation to the other.

Consequently, and following the same logic as in other NLP tasks, RE models followed the classical evolution from feature-based ML models to deep neural networks with learned representations.
Traditional handcrafted features can include \textbf{word features} such as the \textbf{headword} of both mentions or bags of their word or character n-grams, as well as \textbf{words in particular positions}, specifically words immediately preceding or following a mention, or all words between the mentions.
Models can also make use of the \textbf{part-of-speech} of the words in the mentions or their corresponding \textbf{entity types}, but also a more complete description of the syntactic structure of the sentence such as a \textbf{Dependency Tree} that represents grammatical relations between words.
These features were notably used with Support Vector Machines (SVMs) \citep{Cortes1995Support-VectorNetworks} with different type of kernels: from string kernels representative of word features \citep{Lodhi2002TextKernels}, to bag of features kernels that incorporated NE or POS informations
\citep{Bunescu2005SubsequenceExtraction, zhou-etal-2005-exploring}, to parse tree \citep{Zelenko2003KernelExtraction} or dependency tree kernels \citep{culotta-sorensen-2004-dependency}.

The cost of annotation in supervised setting quickly appeared as a major hurdle to develop supervised Relation Extraction algorithms, leading to the development of semi-supervised techniques to make use of the abundant unlabeled text data.
The classical approach follows \textbf{bootstrapping}: iteratively augmenting a small set of labeled data using the most confident predictions of a weak learner trained on this data.
Several models follow this idea, in fact already proposed in \citep{hearst-1992-automatic}. 
For example, the DIPRE \citep{Brin1998ExtractingWeb} and Snowball \citep{Agichtein2000ExtractingCollections} systems iteratively find patterns and argument pairs expressing a given relation in a large unlabeled corpus starting from small initial seed of known pairs (e.g. only five author-book pairs in DIPRE). 
Given a known argument pair, every sentence containing this pair is split into three contexts: prefix (subsentence before the first argument), middle (between the arguments) and suffix after the second argument).
In DIPRE for example, relation patterns are defined as frequently occurring (order, middle context) groups with optional additional common substrings in the prefix and suffix. 
This constraint comes from the observation that relations are most often expressed between the arguments in English sentences.

A similar idea is to directly use a complete preexisting Knowledge Base as a weak supervision signal instead of iteratively augmenting a small seed of handcrafted patterns.
This idea coined as \textbf{distant supervision} by \citet{mintz-etal-2009-distant} consists in using a knowledge base to automatically label every sentence mentioning two related entities as expressing this relation.
This method enables to obtain very large corpora of weakly labeled text and is for example used in the TAC-KBP 2013 leading system in Slot Filling using SVM classifiers \citep{Roth2014EffectiveMethods}. The candidate validation component is based on per-relation SVM classifiers which are trained using distant supervision.
However, it still presents apparent shortcomings since, for example, a sentence can mention two entities without expressing all known relations between them.

Nevertheless, distantly annotated datasets such as the \textbf{NYT-Freebase dataset} \citep{Riedel2010ModelingText} played a key role in the development of Relation Classification models and in particular neural networks that could use this large amount of silver standard data.
In particular, the \textbf{Piecewise Convolutional Neural Network architecture (PCNN)} \citep{zeng-etal-2015-distant} has long been a standard for state-of-the-art models.
It proposes to classically process a sentence with two candidate arguments with a 1D CNN to obtain contextual word representations.
But the sentence-level representation used for classification is not obtained with a classical global max pooling but with \textbf{piecewise max pooling}.
Following the same intuition as \citet{Brin1998ExtractingWeb}, the sentence is split in three parts at the entities and the representations of each of them are pooled separately. 
The final softmax classifier takes the concatenation of these three representations as input.

\begin{figure}[h!]
    \centering
    \caption[Illustration of the Piecewise Architecture.]{Illustration of the Piecewise Architecture. Encoded representations, typically with a CNN or a RNN, are split in five parts corresponding to the arguments and their surrounding contexts. The representation of each part is obtained with a pooling function.}
    \label{04:fig:pcnn}
    
    \includegraphics[width=\textwidth]{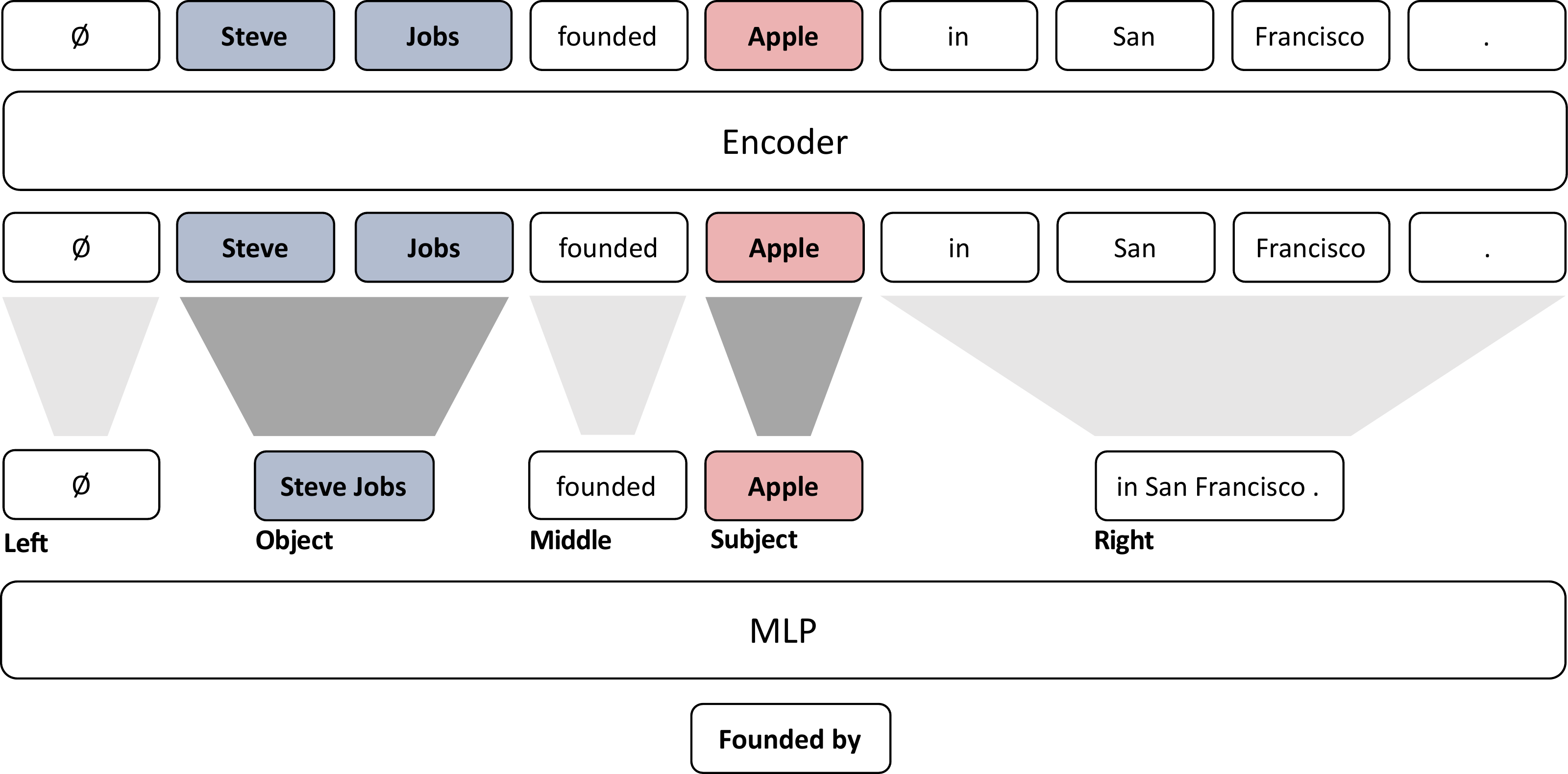}
\end{figure}

Following the recent BERT breakthrough, more recent works use pretrained Transformer language models and finetune them to classify relations given the whole sentence as well as the two entities \citep{Alt2019ImprovingRepresentations, baldini-soares-etal-2019-matching}.

Several ways to use BERT have been tested, especially by \citet{baldini-soares-etal-2019-matching}, either using the [CLS] token as input for classification, or the concatenation of arguments representations.
Arguments can in turn be represented by pooling the words they contain or with additional special tokens surrounding the arguments and possibly representative of their types. 

The current state-of-the-art system is thus "Matching the Blanks" (MTB) \citep{baldini-soares-etal-2019-matching} where a BERT model is first finetuned with entity linked text from Wikipedia.
Given two relation statements (i.e. sentences with at least two entity mentions), it must predict if the pairs of entities are identical whereas entities are masked with a special token with a probability $\alpha = 0.7$.
This large scale distantly supervised finetuning enables to obtain general relation representations and further tuning on a supervised dataset leads to new state-of-the-art results, particularly in the few-shot learning setting.

\begin{figure}[h!]
    \begin{adjustwidth*}{}{-.3\textwidth}
    \caption{Illustration of the Transformer for Relation Extraction (TRE). 
    Figure from \citep{Alt2019ImprovingRepresentations}}
    \label{fig:alt2019}
    \centering
    \includegraphics[width=1.3\textwidth]{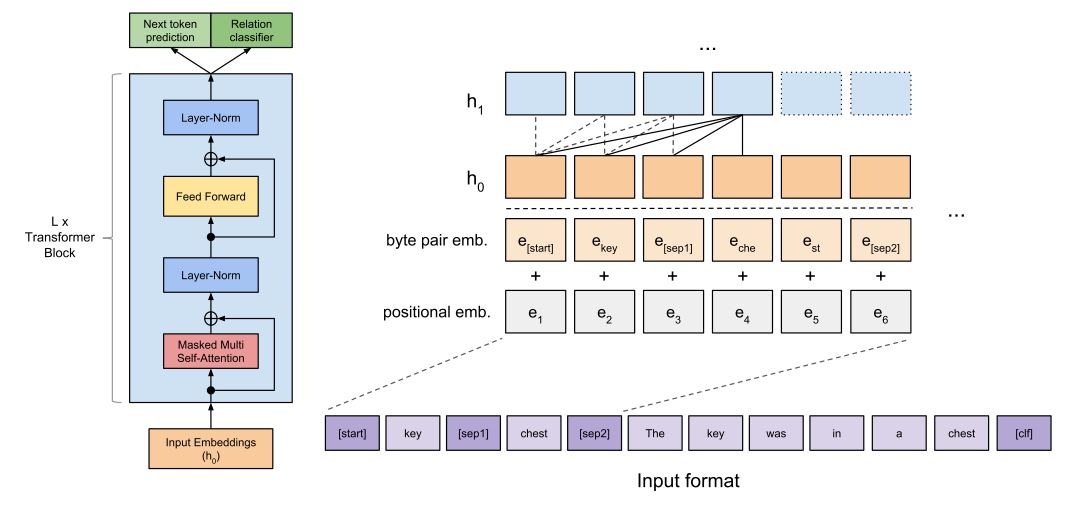}
    \end{adjustwidth*}
\end{figure}

\section[A Taxonomy of End-to-end RE Models]{A Taxonomy of End-to-end Relation Extraction Models}

The pipeline setting views Entity and Relation Extraction as separate tasks treated by two systems designed independently and applied sequentially.
While this setting is highly modular and reflects the intuitive dependency of Relation Extraction on Entity Extraction, it also fails to model the opposite dependency of Entity Extraction on Relation Extraction and can favor \textbf{error propagation}.
Indeed, while Relation Classification models are trained given ground truth candidate arguments, the Named Entity Recognition system inevitably makes mistakes at inference time that cannot be corrected when extracting the relation.
However, features extracted for RE can help NER since the expression of a relation in a sentence necessarily implies the presence of two argument mentions and can also give indication regarding their types.
Indeed, for example, the relation ``founded by'' only holds between a person and an organization.
To better take this reverse dependency into account and evaluate Relation Extraction in a more realistic end-to-end setting has been explored, evaluated the performance of the whole Named Entity Recognition and Relation Extraction process.

In this setting, various methods have been proposed to tackle both tasks and we propose a taxonomy of these models.

\subsection{Local classifiers}
The first attempts to model the mutual dependency between NER and RE combined the predictions of independent local classifiers according to global constraints 
(e.g. the arguments of the ``Live In" relation must be a Person and a Location).
These constraints restrict the search space of the most probable assignments between the entities and relations, unifying the predictions of the two systems a posteriori.
Several articles propose to explore this approach using different methods such as Probabilistic Graphical Models \citep{roth-yih-2002-probabilistic}, Integer Linear Programming \citep{roth-yih-2004-linear} or Card Pyramid Parsing \citep{kate-mooney-2010-joint}.
These models are built over NER and RE models using traditional word-level features such as neighbouring words, word shape and part-of-speech.

Although their study is focused on the two relations ``born in'' and ``kill'', \citet{roth-yih-2002-probabilistic} are the first to demonstrate that modeling the known interdependency between NER and RE can lead to significant improvement over the pipeline approach in the final RE evaluation.
This initial result served as the main motivation for the development of later end-to-end Relation Extraction systems that often propose to learn a single model that is used to predict both types of information instead of a posteriori reconciling the predictions of two separate classifiers. 

\subsection{Incremental Joint Training}

\begin{figure}[h!]
    \centering
    \caption[Illustration of the Incremental Approach to ERE.]{Illustration of the Incremental Approach to ERE. In this approach a sentence is parsed word by word with two types of prediction: entity tag and relation with previously detected entities if the word is predicted as the last word of an entity.}
    \label{04:fig:incremental}
    
    \includegraphics[width=.9\textwidth]{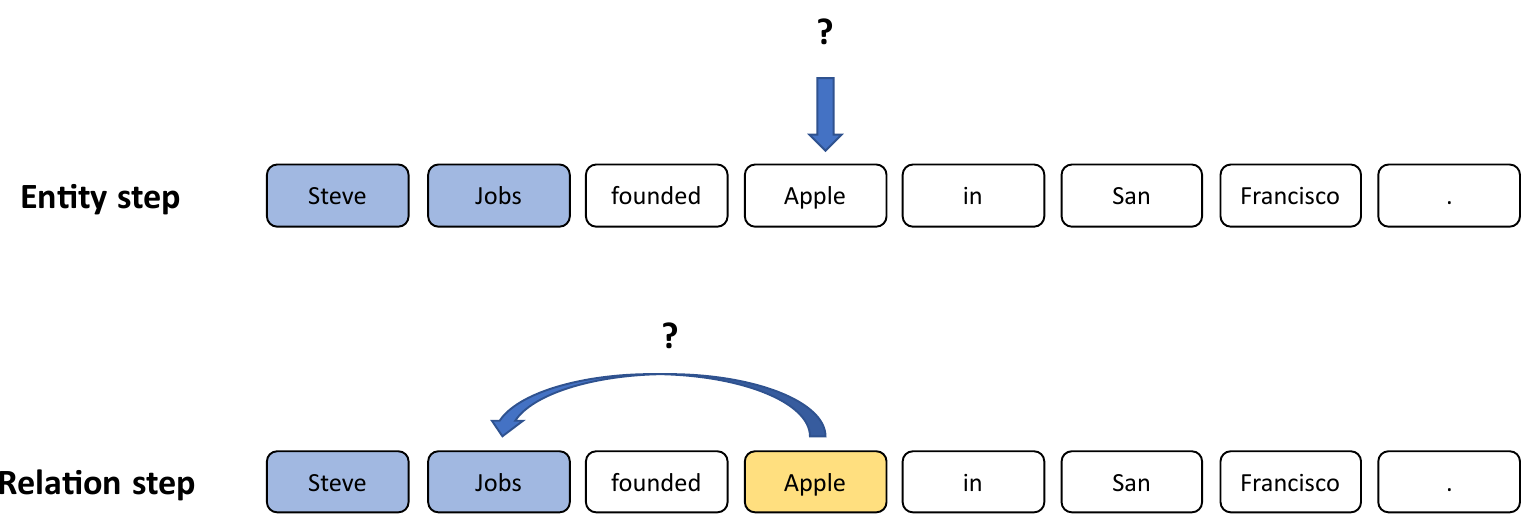}
\end{figure}

The first type of joint NER and RE models are \textbf{incremental models} which parse the sentence once and make a joint prediction for entity mentions and relations. Because here the relations are predicted sequentially, previously predicted entities and relations impact the current entity or relation prediction, allowing to model dependency inside and between tasks.

\citet{li-ji-2014-incremental}
propose the first end-to-end RE model, based on a structured perceptron and beam search.
The model parses a sentence and predicts if the current word is the last word of a mention, if so its type and whether there is a relation between this mention and a previously encountered one.
This model is based on handcrafted features introduced in \citep{florian-etal-2003-named} (such as word case, gazeeter, neighbours unigrams and bigrams), constituency and dependency parses as well as mentions coreference and a large set of handcrafted rules regarding the coherence of predicted entities and relations.

\citet{katiyar-cardie-2017-going}
adopt the same framing but replace handcrafted and external features by learned word embeddings. 
They first use a BiLSTM for entity detection and keep a stack of tokens previously detected as part of entities. 
Then, if the current token is predicted as a part of an entity, its representation is compared to those of every tokens in the stack with a pointer network \citep{Vinyals2015PointerNetworks} to predict the relation.
One issue with such an incremental aproach is that it is unnatural to detect a pattern where an entity is involved in several relations with different other entities, even though \citet{katiyar-cardie-2017-going} propose strategies to deal with this case.

\subsection{Table Filling}

\begin{figure}[h!]
    \centering
    \caption[Illustration of the Table Filling approach.]{Illustration of the Table Filling approach. The whole information useful for ERE can be formatted as a table with NER information on the diagonal and RE information on off-diagonal cells. ERE can then be modeled as a sequence labelling task where the sequence includes every cell of the table.}
    \label{04:fig:table filling}
    
    \includegraphics[width=.7\textwidth]{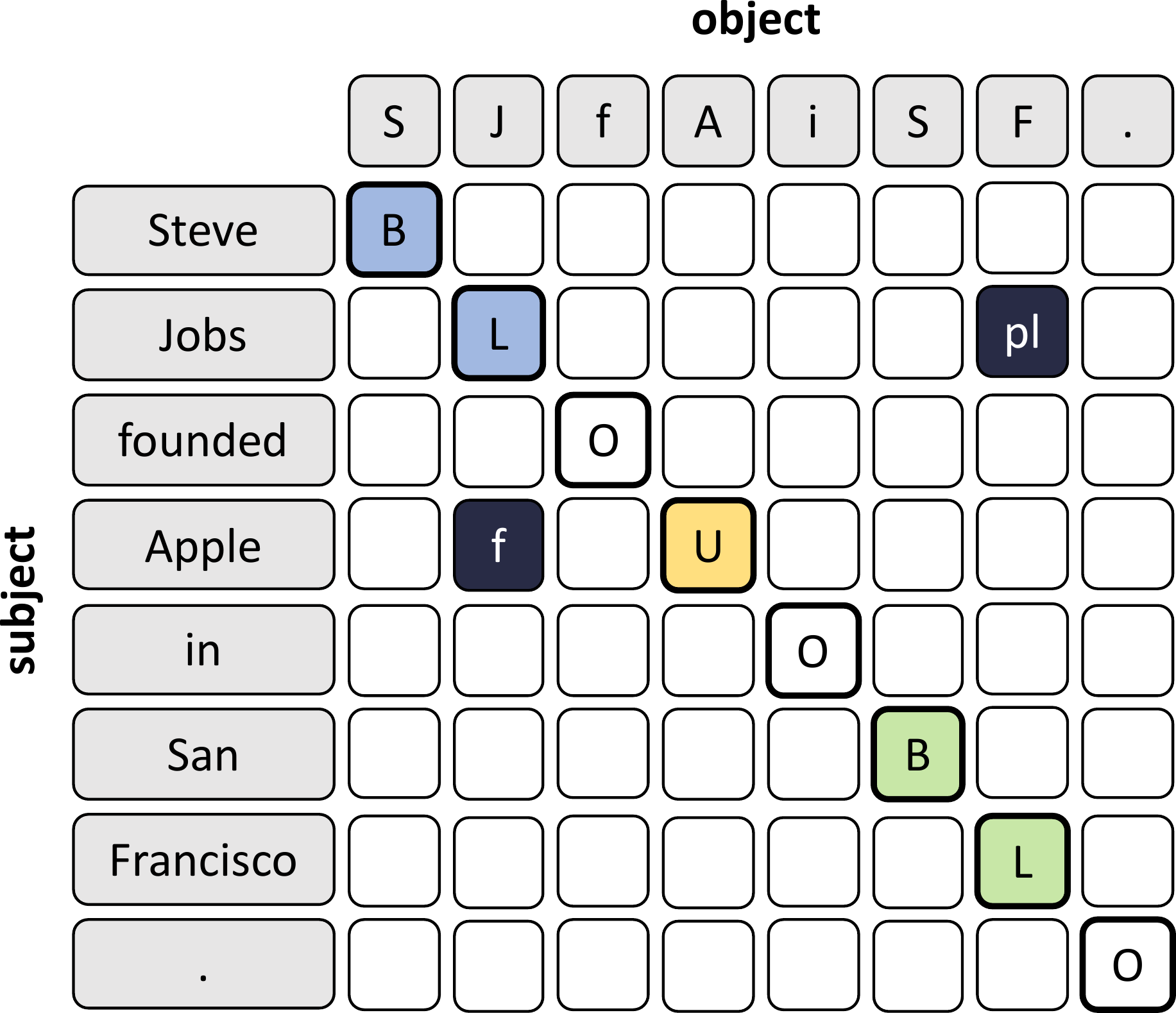}
\end{figure}

\citet{miwa-sasaki-2014-modeling} propose to simplify this incremental setting by sequentially filling a table representation which contains all mentions and relations.
This \textbf{Table Filling} approach enables to view Relation Extraction as a sequence labeling problem and they experiment with different prediction orders, showing that predicting mentions first is more effective.
This setting enables to very simply handle sentences with several relations, even involving a same entity, because each cell of the table corresponds to an entity pair.  
Similarly to \citep{li-ji-2014-incremental}, this model relies on handcrafted features and rules and a dependency parse.
\citet{gupta-etal-2016-table} take up this Table Filling (TF) approach but use an RNN with a multitask approach.
Similarly, \citet{zhang-etal-2017-end} use LSTMs but add syntactic features from \citep{Dozat2017DeepParsing}'s Dependency Parser.

More recently, this approach is also adopted by \citet{wang-lu-2020-two} that use BERT-like representations and intertwine a sequence encoder and a table encoder.
One originality is that the pretrained Language Model is frozen and that the final hidden states are used by both encoders but the attention weights are also used by the table encoder.
The prediction is finally performed by a Multi-Dimensional RNN (MD-RNN).
The idea of the Multi-Dimensional RNN is to extend the classical RNN to the prediction of a structure of higher dimension that a sequence such as a 2D table.
The hidden state of the current cell can for example depend on the hidden states corresponding to the cell at its left (previous horizontally) and to the cell above (previous vertically).
Hence in this case the computation of the hidden state depends on the entire up left rectangle, similarly to how the left context is taken into account in a classical RNN.
Then, like in a BiLSTM, the representations of different the contexts are concatenated.
In 2D, we can concatenate four contexts (up left, up right, bottom left an bottom right) or limit to two (e.g. up left and bottom right).

\subsection{Shared Encoder}

Another type of models use \textbf{parameter sharing}, a common Deep Multi-task Learning method, to tackle joint NER and RE.
The NER and RE modules share an encoder, whose weights are updated to simultaneously optimize the NER and RE objective functions.
The interdependency of both tasks can thus be modeled by this sole shared encoder.

We propose to further subdivide shared encoder models in two additional subclasses: \textbf{Entity Filtering} models and \textbf{Multi Head Selection} models.

\paragraph{Entity Filtering}
A first set of shared encoder models keep the pipeline structure with a preliminary NER module whose predictions are used as input for RE.
As in the pipeline, RE is viewed as classification given a sentence and a pair of arguments.
This requires passing each pair of candidate entities detected by the NER through the RE classifier, hence the Entity Filtering denomination.
The only difference with the pipeline approach is that the NER and RE models share some parameters in end-to-end RE, often in a BiLSTM encoder.
Indeed, as in the previous incremental setting, NER is modeled as sequence labeling using IOBES tags \citep{ratinov-roth-2009-design} and the NER module is often a BiLSTM following \citep{Huang2015BidirectionalTagging}, although it is now replaced by BERT in recent works.
The two modules are jointly trained by optimizing for the (weighted) sum of their losses.

\citet{miwa-bansal-2016-end} use a sequential BiLSTM for NER and a Tree-LSTM over the shortest path between candidate arguments in a Dependency tree predicted by an external parser.
The idea of using dependency trees follows the same idea as previous dependency tree kernels
that the syntactic structure of a sentence is useful to detect a relation.
In particular, we can expect that the predicate is expressed by a verb that is present in the shortest dependency path linking its subject to its object.
\citet{Li2017AText} apply this model to biomedical data.

The Piecewise CNN (PCNN) architecture \citep{zeng-etal-2015-distant} has also been explored by numerous works in the entity filtering approach.

\citet{adel-schutze-2017-global} use the structure of a Piecewise CNN but add a``global normalization layer" composed of a length three linear CRF which linearly takes the predicted sequence (argument 1, predicate, argument 2) to model the interdependency. By analysing the transition matrix they show that the CRF learns strong correlations between relations and corresponding entity types.

\citet{zhang-etal-2017-end} proposes to use a classical BiLSTM EMD and a Piecewise LSTM (PLSTM) RE module.
They are fed with GloVe embeddings and POS tags as well as the output of an external dependency parser encoder.
The novelty is ``global optimization" which consists in not only calculating a loss for the whole sentence but add a loss corresponding to all right-truncated subsentences.
This leads to an improvement mainly visible for longer sentences.

\citet{sun-etal-2018-extracting} introduce Minimum Risk Training to incorporate both RE and EMD sentence-level F1 scores in the loss function. They use a BiLSTM NER module and a PCNN RE module with GloVe embeddings and a word-level charCNN.

Finally, whereas all previously cited works rely on the sequence tagging view of the NER task, \textbf{span-level NER and RE} has also been explored in different settings.
Hence, \citet{luan-etal-2018-multi} propose an end-to-end model to perform NER, RE and Coreference Resolution with span-level representations.
This architecture is inspired by previous end-to-end architectures used in Coreference Resolution (CR) \citep{lee-etal-2017-end} or Semantic Role Labeling (SRL) \citep{he-etal-2018-jointly}.
In this setting, all spans of consecutive words (up to a fixed length) are independently classified as entities, which enables detecting overlapping entities,
and they use an element-wise biaffine RE classifier to classify all pairs of detected spans. 
In a later work \citep{luan-etal-2019-general}, they then propose to iteratively refine predictions with dynamic graph propagation of RE and CR confidence scores. 
This work is adapted with BERT as an encoder in \citep{wadden-etal-2019-entity}.

\citet{dixit-al-onaizan-2019-span} use a model very similar to \citet{luan-etal-2018-multi}'s but restrict to end-to-end RE.
\citet{Eberts2020Span-basedPre-training} take a similar approach with BERT as an encoder.
For RE, in addition to the representations of both argument spans, they use a max pooled representation of the middle context for RE, similarly to piecewise models.

\paragraph{Multi-Head Selection}

To avoid relying explicitly on NER prediction, \citet{Bekoulis2018JointProblem, bekoulis-etal-2018-adversarial}, propose Multi-Head Selection where RE classification is made for every pair of words.
As in the original implementations of Table Filling, relations should only be predicted between the last words of entity mentions to avoid redundancy and inconsistencies.
While this enables to make the complete prediction in a single pass, contextual information must be implicitly encoded in all word representations since the Linear RE classifier is only fed with representations of both arguments and a label embedding of NER predictions.

\renewcommand{\tabcolsep}{3pt}
\renewcommand{\arraystretch}{1}

\begin{table}[h]
\begin{adjustwidth*}{}{-.3\textwidth}

\caption[Proposed classification of end-to-end RE models in antichronological order.]{Proposed classification of end-to-end RE models in antichronological order.\\
Language Model pretraining:  \textbf{E}LMo \citep{peters-etal-2018-deep} / \textbf{B}ERT \citep{devlin-etal-2019-bert} / \textbf{AlB}ERT \citep{Lan2020ALBERT:Representations}.\\
Word embeddings: \textbf{S}ENNA \citep{Collobert2011NaturalScratch} / \textbf{W}ord2Vec \citep{Mikolov2013DistributedCompositionality} / \textbf{G}loVe \citep{pennington-etal-2014-glove} / \textbf{T}urian \citep{turian-etal-2010-word}.\\
Character embeddings pooling: \textbf{C}NN / (Bi)\textbf{L}STM.\\
Hand: handcrafted features. POS/DEP: use of Ground Truth or external Part-of-Speech tagger or Dependency Parser.\\
Encoder: (Bi)\textbf{L}STM. NER Tag: \textbf{B}ILOU / \textbf{S}pan.\\
Decoders: I- = Incremental, TF- = Table Filling, MHS=Multi-Head Selection, SP=Shortest Dependency Path.\\ ns=Not Specified, for words it might be randomly initialized embeddings.
}

\label{table:04:taxonomy}

\centering
\scriptsize
\begin{tabularx}{1.3\textwidth}{@{}ll*{6}{c}lcclll@{}}

\toprule
    &  
    & \multicolumn{6}{c}{Representations} & & Enc. & \multicolumn{2}{c}{NER} & & RE \\
            \cline{3-8} \cline{11-12}

Reference   &  
& LM & Word & Char & Hand & POS & DEP & & &  Tag & Dec. & & Dec.\\
\midrule

\citep{zhong-chen-2021-frustratingly} & 
                                            & Alb & & & & & &
                                            & - 
                                            & B & MLP &
                                            & Biaff. \\

\citep{wang-lu-2020-two} & 
                                            & Alb & & & & & &
                                            & - 
                                            & B & MLP &
                                            & TF-MDRNN \\

\citep{Giorgi2019End-to-endModels} & 
                                            & B & & & & & &
                                            & - 
                                            & B & MLP &
                                            & Biaff. \\
                                            
\citep{Eberts2020Span-basedPre-training} & 
                                            & B & & & & & &
                                            & - 
                                            & S& MLP &
                                            & PMaxPool\\
                                            
\citep{wadden-etal-2019-entity} & 
                                            & B & & & & & &
                                            & - 
                                            & S & MLP &
                                            & Biaff. \\ 
                                            
\citep{li-etal-2019-entity} & 
                                            & B & & & & &  &
                                            & -  
                                            & - & MT QA &  & MT QA \\
                                            
\citep{dixit-al-onaizan-2019-span} & 
                                            & E & S & C & & & &
                                            & L 
                                            & S& MLP &
                                            & Biaff.\\
                                            
\citep{luan-etal-2019-general} & 
                                            & E & G & ns & & & &
                                            & L 
                                            & S & MLP &
                                            & Biaff.\\
                                            
\citep{Nguyen2019End-to-endAttention} & 
                                            & & G & L & & &  &
                                            & L 
                                            & B& MLP &
                                            & MHS-Biaff.\\
                                            
\citep{Sanh2019ATasks} & 
                                            & E & G & C & & &  &
                                            & L 
                                            & B & CRF &
                                            & MHS-Lin.\\
                                            
\citep{luan-etal-2018-multi} &  
                                            & E & G & ns & & &  &
                                            &   
                                            & S & MLP &
                                            & Biaff.\\
                                            
\citep{sun-etal-2018-extracting} & 
                                            & & ns & C & & &  &
                                            & L  
                                            & B & MLP &
                                            & PCNN\\
                                            
\citep{bekoulis-etal-2018-adversarial, Bekoulis2018JointProblem} & 
                                            & & S/W & L & & &  &
                                            & L 
                                            & B & CRF &
                                            & MHS-Lin.\\

\citep{zhang-etal-2017-end} & 
                                            & & G & C & & \cmark & \cmark &
                                            & L 
                                            & B & TF-LSTM & & TF-LSTM \\

\citep{Li2017AText} & 
                                            & & ns & C & & \cmark & \cmark  &
                                            & L 
                                            & B & MLP &
                                            & SP LTSM\\

\citep{katiyar-cardie-2017-going} & 
                                            & & W & & & &  &
                                            & L 
                                            & B & I-MLP & & I-Pointer\\

\citep{Zheng2017JointNetwork} & 
                                            & & ns & & & &  &
                                            & L 
                                            & B & MLP &
                                            & PCNN\\

\citep{adel-schutze-2017-global} & 
                                            & & W & & & & &
                                            & - 
                                            & B & CNN  &
                                            & PCNN+CRF \\

\citep{gupta-etal-2016-table} & 
                                            & & T & & \cmark &  \cmark & &
                                            & - 
                                            & B & TF-RNN & & TF-RNN \\

\citep{miwa-bansal-2016-end} & 
                                            & & ns & & & \cmark & \cmark &
                                            & L 
                                            & B & MLP &
                                            & SP LSTM\\

\citep{miwa-sasaki-2014-modeling} & 
                                            & & & & \cmark & & &
                                            & - 
                                            & B & TF-SVM & & TF-SVM \\

\citep{li-ji-2014-incremental} & 
                                            & & & &  \cmark & & &
                                            & - 
                                            & B & I-Perc.& & I-Perc. \\

\bottomrule

\end{tabularx}

\end{adjustwidth*}
\end{table}

\renewcommand{\arraystretch}{1.2}

\citet{Nguyen2019End-to-endAttention} replace this linear RE classifier by the bilinear scorer from \citet{Dozat2017DeepParsing}'s Dependency Parser.
A similar biaffine architecture is extended with BERT representations by \citet{Giorgi2019End-to-endModels}.

Finally, \citet{Sanh2019ATasks} build on \citep{Bekoulis2018JointProblem} to explore a broader multitask setting incorporating Coreference Resolution (CR) as well as NER learned on the classical OntoNotes corpus. 
They use ELMo contextualized embeddings \citep{peters-etal-2018-deep} with a multi-layer BiLSTM architecture.

\subsection{Question Answering}

\begin{figure}[h!]
    \centering
    \caption[Illustration of the Multiturn Question Answering approach.]{Illustration of the Multiturn Question Answering approach. A BERT-based model is trained for extractive Question Answering to answer successive questions in a template corresponding to each relation type.}
    \label{fig:04:mtqa}
    \includegraphics[width=\textwidth]{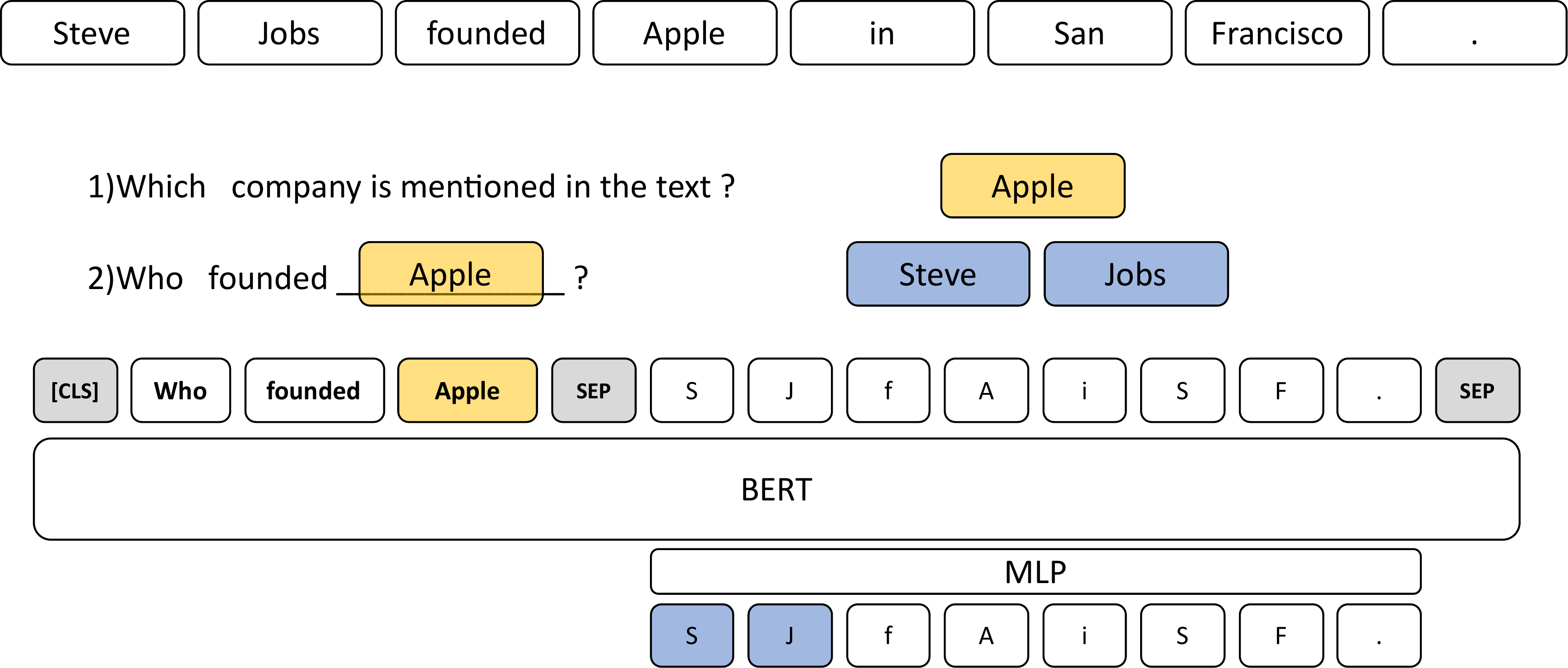}
\end{figure}

Finally, a few papers propose to view Relation Extraction as Question Answering (QA).
\citet{levy-etal-2017-zero} first propose to use Reading Comprehension QA to perform RE, even in the zero-shot setting with relation type never encountered at test time.
They consider the Slot Filling scenario where given a sentence, an entity $e$ and a relation type $r$, they must find the set of entities $A$ (possibly empty) such that for each $a \in A$ the relation $(e, r, a)$ holds.

\citet{li-etal-2019-entity} tackle end-to-end Relation Extraction as Multi-turn QA. For each relation, they design a template of successive questions to iteratively identify mentions and relations in a paragraph, with the help of a state-of-the-art QA model using BERT embeddings.

\subsection{Pipeline Models in End-to-end RE}
While since the assertion by \citet{roth-yih-2002-probabilistic} that joint inference of NER and RE is beneficial over a pipeline baseline and empirical confirmations in the works of \citet{li-ji-2014-incremental} and \citet{miwa-sasaki-2014-modeling}, end-to-end RE has mostly been treated with a joint model approach, taking its benefits for granted.
In parallel, pipeline models were developed on separate benchmarks focused on Relation Classification such as TACRED \citep{zhang-etal-2017-position} without consideration of the end-to-end performance.
However, with the rise of LM pretraining and the performance improvements in both tasks, the error rate of NER models decreased, which naturally addresses the error propagation issue.

Recently, \citet{zhong-chen-2021-frustratingly} propose a pipeline system based on state-of-the-art NER and RE models based on BERT-like pretrained models.
The overall system lead to new state-of-the-art results on ACE04, ACE05 and SciERC at the cost of finetuning two separate BERT-like encoders.
Their experiments support the fact that parameter sharing in a single BERT encoder slightly hurts the final performance.
While this does not prove that joint NER and RE learning is no longer useful with BERT based models, it suggests that these benefits would come from different strategies or perhaps on more difficult scenarios or datasets.  

\section{A lack of clear conclusions}
Considering the multiplication of settings to tackle entity and relation extraction, we might simply wonder what are strenghts and weaknesses of each proposal or if we can draw clear insights from their comparison on what architecture design choices to adopt.
However, \textbf{several aspects prevents us to do draw clear conclusions} from this sole literature review.

In the first place, we can distinguish the concurrent evolutions of three main elements in the proposed models.
And in the absence of appropriate ablation studies, it is impossible to conclude on the true impact of each of them.
First, following a general NLP trends, \textbf{word representations} have evolved from handcrafted features, possibly augmented with external syntactic information such as part of speech or dependency trees, to pretrained word embeddings and to entire pretrained language models.
Second, we can also see that the traditional \textbf{view of NER} as a IOBES sequence tagging task is progressively leaving a place for span-level classification.
Third, there is an evolution of the \textbf{joint Entity and Relation Extraction strategy}, from the reconciliation of local classifiers predictions to incremental settings, to shared encoder strategies, and now even going back to a simple pipeline approach.

In the second place, like in NER and numerous other NLP tasks, this comparison between models often rely on a few standard benchmarks and a single metric that is taken as an indication of ``state-of-the-art results''.
In the specific case of end-to-end Relation Extraction, different views of what should be the end result of the two tasks led to \textbf{several evaluation settings and metrics} to be developed \citep{bekoulis-etal-2018-adversarial}.
While it is interesting to have several measures for a finer-grained evaluation, in this case, it led to \textbf{confusion} and several \textbf{incorrect comparisons} in the literature.

Third, as initially shown for NER by \citet{palmer-day-1997-statistical} or \citet{Augenstein2017GeneralisationAnalysis}, and developed in \autoref{chapter:03:ner}, a single global metric cannot fully reflect the influence of linguistic properties of individual data samples such as lexical overlap.
In fact, these studies are now part of a more general trend to propose fine-grained evaluations of NLP models that can reach ``super-human'' scores on some benchmarks, while showing a sudden drop in performance when the evaluation samples are slightly modified \citep{mccoy-etal-2019-right, ribeiro-etal-2020-beyond}.

We propose to tackle these three aspects in the following chapter.
\chapter[Rethinking Evaluation in end-to-end RE]{Rethinking Evaluation in end-to-end Relation Extraction}
\label{chapter:05:rethinking}

Numerous and diverse approaches have been explored in end-to-end Relation Extraction, departing from handcrafted rules to neural networks and following the general evolution of word representations described in \autoref{chapter:02:bert}.
However, despite the now commonly known fact that using BERT enables to outperform models using classical word embeddings, it is difficult to draw conclusions on the respective strengths and weaknesses of the numerous proposed architecture.
Worse, because of different views of what constitutes a relation, several evaluation settings have been used, leading to confusion and incorrect comparisons.

Furthermore, like in Named Entity Recognition, the evaluation of end-to-end Relation Extraction is often limited to a sole Precision, Recall and F1 score evaluation, overlooking key linguistic specificities of individual data samples such as lexical overlap.

In this chapter, we propose to address the two first issues in \autoref{sec:05:sincere} by identifying the main sources for incorrect comparisons in the literature and performing a double ablation study of two recent evolutions: pretrained Language Models, specifically BERT, and span-level NER from the SpERT model \citep{Eberts2020Span-basedPre-training} that combines both.

We then present a contained study on retention in end-to-end Relation Extraction in \autoref{sec:05:retention} which extends our work on Named Entity Recognition presented in \autoref{chapter:03:ner}.
This study suggests that a simple retention heuristic can explain the relatively better performance of RE models on the standard CoNLL04 and ACE05 datasets compared to the more recent SciERC dataset that presents a reduced proportion of relation overlap between the train and test sets.
This indicates that despite the recent advances brought by BERT pretraining, there is room to improve generalization of ERE models to relations not seen during training. 

\section[Let's Stop Incorrect Comparisons in End-to-end RE]{Let's Stop Incorrect Comparisons in End-to-end Relation Extraction}
\label{sec:05:sincere}

Despite previous efforts to distinguish three different evaluation setups in end-to-end Relation Extraction \citep{bekoulis-etal-2018-adversarial, Bekoulis2018JointProblem}, several end-to-end RE articles present unreliable performance comparison to previous work that we clearly identify in this section.
Our goal is to highlight the confusion that arose from these multiple evaluation settings and in order to prevent future replication and propagation of such erroneous comparisons.

We then perform an empirical study with two goals: 1) quantifying the impact of the most common mistake that can lead to overestimating the final RE performance by around 5\% on ACE05 and 2) perform the unexplored ablations of two recent developments, namely the use of language model pretraining (specifically BERT) and span-level NER.

This meta-analysis emphasizes the need for rigor in the report of both the evaluation setting and the dataset statistics.
We finally call for unifying the evaluation setting in end-to-end RE.

Before diving into the identification of patterns for incorrect comparisons, we introduce common benchmarks and evaluation settings used to evaluate end-to-end RE.

\subsection{Datasets and Metrics}

\paragraph{Datasets}
Following \citet{roth-yih-2002-probabilistic}, end-to-end RE has traditionally been explored on English news articles, which is reflected in the domain of its historical benchmarks, the ACE datasets and CoNLL04.

Although the Automatic Content Extraction program tackled the study of relations as soon as 2002, only the last two datasets are still currently used as benchmarks for end-to-end RE.
The \textbf{ACE04} dataset \citep{doddington-etal-2004-automatic} defines seven coarse entity types and seven relation types.
\textbf{ACE05} \citep{Walker2006ACECorpus} resumes this setting but merges two relation types leading to six of them.
Originally, the ACE program proposed a custom and convoluted scoring setting separated into three separate scores: entity, relation and event \citep{doddington-etal-2004-automatic}.
These scores require to map each system output to a corresponding Ground Truth element in a way that maximize the final score.
It proposed to evaluate extraction performance both at the entity and mention level for entity detection and at the relation and argument level for relations.
Because of its complexity and the difficulty of interpreting these scores, this ACE scoring was mostly used for ACE shared tasks and the traditional Precision, Recall and F1 score setting was preferred in end-to-end RE evaluation.

\textbf{CoNLL04} \citep{roth-yih-2004-linear} is annotated for four entity types and five relation types and specifically only contains sentences with at least one relation.
It is composed of news articles from a TREC (Text REtrieval Conference) dataset which compiles different sources such as the Wall Street Journal or the Associated Press.

More recently, other datasets have been proposed for more specific domains, as original as Dutch real-estate advertisement in the DREC dataset \citep{Bekoulis2018AnAds} or more classically the biomedical or scientific domain. 
Hence, \citet{Gurulingappa2012DevelopmentReports} propose the \textbf{ADE} dataset in the biomedical domain, which focuses on one relation, the \textbf{Adverse Drug Event} between a Drug and one of its Adverse Effects.

In the scientific domain, \citet{luan-etal-2018-multi} introduce \textbf{SciERC} composed of 500 scientific article abstracts annotated with six types of scientific entities, coreference clusters, and seven relations between them.

Although other datasets have been proposed for end-to-end RE, we identify ACE04, ACE05, CoNLL04, ADE and SciERC as the five main current benchmarks and limit our report on these five datasets.

\paragraph{Metrics}
Apart from initial tentatives to propose more detailed metrics to evaluate end-to-end RE performance from MUC \citep{grishman-sundheim-1996-message} or ACE \citep{doddington-etal-2004-automatic} conferences, the traditional metrics for assessing both NER and RE performance are Precision, Recall and F1 scores.
However, there are two points of attention when reporting such global metrics: the use of \textbf{micro or Macro averaged metrics} across types and the \textbf{criterion} used to consider a prediction as \textbf{true positive}.

On this second point, there is no difficulty for NER where a consensus has been reached in both considering mention detection and typing as necessary for a correct prediction.
However, compared to the pipeline Relation Classification, this end-to-end RE setting adds a source of mistake in the identification of arguments that are no longer given as input.
And while there is an agreement that the relation type must be correctly detected with the correct subject-object order, several evaluation settings have been introduced with different requirements regarding the detection of arguments. 

Hence, \citet{bekoulis-etal-2018-adversarial} distinguishes three evaluation settings: 

\textit{\textbf{Strict}}: both the boundaries and the entity type of each argument must be correct.

\textit{\textbf{Boundaries}}: argument type is not considered and boundaries must be correct.

\textit{\textbf{Relaxed}}: NER is reduced to Entity Classification i.e. predicting a type for each token. A multi-token entity is considered correct if at least one token is correctly typed.

\begin{figure}[h!]
    \begin{adjustwidth*}{}{-.3\textwidth}
    \caption{Illustration of the different evaluation settings in End-to-end RE.}
    \label{05:fig:ere-eval}
    \centering
    \includegraphics[width=1.3\textwidth]{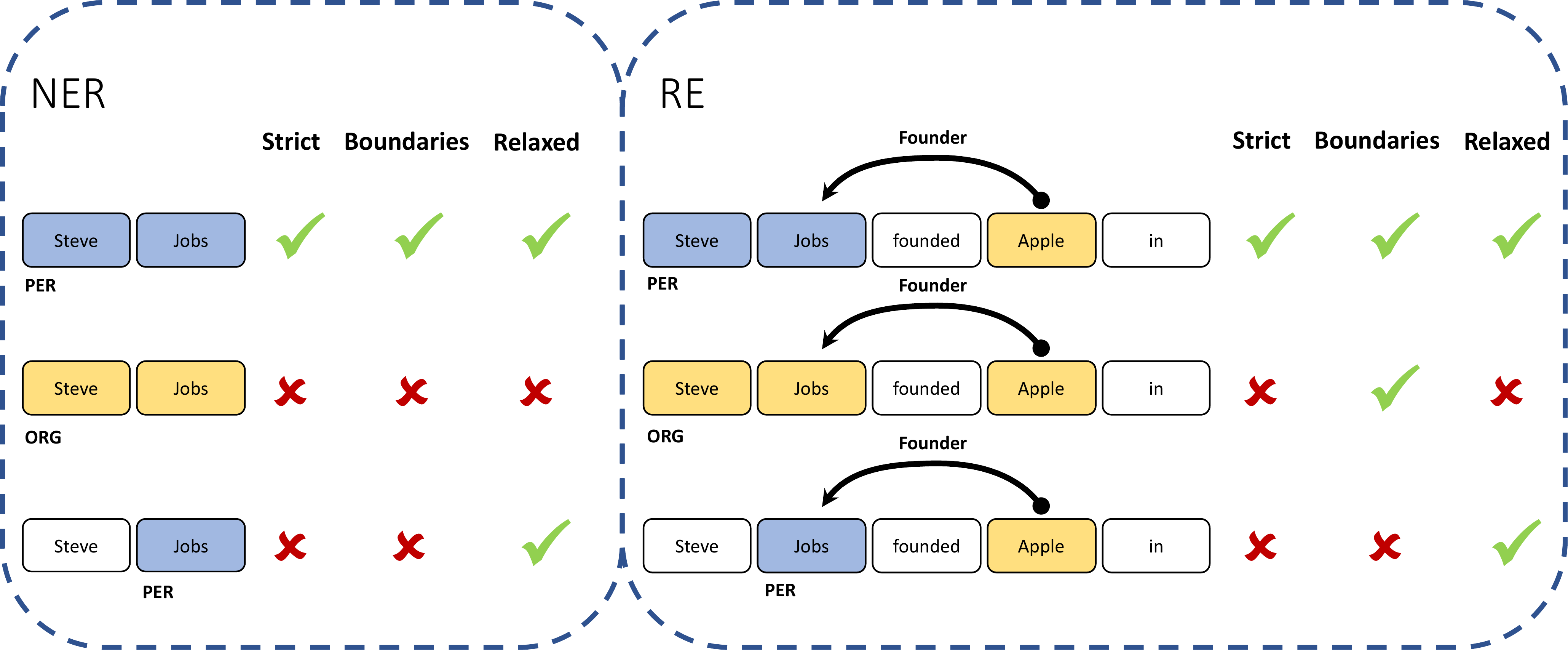}
    \end{adjustwidth*}
\end{figure}

\subsection{Identified Issues in Published Results}
Despite the previous unambiguous distinction by \citet{bekoulis-etal-2018-adversarial}, This variety of evaluation settings, visible in \autoref{table:04:taxonomy}, has unfortunately led to confusion which in turn favored recurring mistakes.
By a careful examination of previous work and often only thanks to released source codes and/or sufficiently detailed descriptions, we identified several patterns responsible for incorrect comparisons in previous literature.
Because these precious sources of information are sometimes missing, we cannot assert that we are exhaustive. 
However, we will now list them to avoid their propagation and present a curated summary of supposedly comparable results in \autoref{table:04:re_sota}.

\paragraph{Comparing Boundaries to Strict results on ACE datasets}
The most common mistake is the comparison of Strict and Boundaries results. 
Indeed, several works  \citep{Zheng2017JointNetwork, luan-etal-2019-general, wadden-etal-2019-entity} use the Boundaries setting to compare to previous Strict results.
This mistake is limited to the ACE datasets because both metrics have been used on them since the work of \citep{li-ji-2014-incremental}.
However, because the Strict setting is more restrictive, this always leads to overestimating the benefit of the proposed model over previous SOTA.
Because it is the most common mistake, we propose to quantify the resulting improper gain in \autoref{subsec:05:empirical_study}.

\renewcommand{\arraystretch}{1}

 \begin{table}[!hbtp]
 \begin{adjustwidth*}{}{-.3\textwidth}
 
    \caption[Summary of recently published results in end-to-end RE on five datasets.]{Summary of recently published results in end-to-end RE on five datasets.\\
    Models over the dashed lines use LM pretraining.\\
    \xmark = some results were incorrectly reported as Strict.
    $^{\dagger}$ = explicit use of train+dev.\\
    $^{\ast}$ = partition from \citep{miwa-sasaki-2014-modeling}.
    \textbf{+} = experiments on additional datasets.}
    \label{table:04:re_sota}
    
    \centering
    \small
    
    \begin{tabularx}{1.3\textwidth}{@{}l*{2}{Y}r*{2}{Y}r*{2}{Y}r*{2}{Y}r*{2}{Y}c@{}}
    
    \toprule
    & \multicolumn{2}{c}{ACE 05} &  & \multicolumn{2}{c}{ACE 04} & & \multicolumn{2}{c}{CoNLL04} & &\multicolumn{2}{c}{ADE} & &\multicolumn{2}{c}{SciERC} &\\
     \cmidrule{2-3} \cmidrule{5-6} \cmidrule{8-9} \cmidrule{11-12} \cmidrule{14-15}
    Reference  & Ent & Rel & & Ent & Rel & & Ent & Rel & & Ent & Rel & & Ent & Rel\\

    \midrule
    \midrule
    
    \textbf{Strict Evaluation}
    & \multicolumn{2}{c}{$\mu$F1} &
    & \multicolumn{2}{c}{$\mu$F1} &
    & \multicolumn{2}{c}{$\mu$F1} &
    & \multicolumn{2}{c}{$\mu$F1} &
    & \multicolumn{2}{c}{$\mu$F1} \\
    
    \citep{zhong-chen-2021-frustratingly}
    & \textbf{89.7} & \textbf{65.6} &   
    & \textbf{88.8} & \textbf{60.2} &
    & - & - &
    & - & - &
    & \textbf{66.6} & \textbf{35.6} & 
    - \\

    \citep{wang-lu-2020-two}
    & \textbf{89.5} & 64.3 &   
    & \textbf{88.6} & 59.6 &
    & \textbf{90.1} & \textbf{73.6} &
    & - & - & 
    & & &
    - \\

    \citep{Giorgi2019End-to-endModels}
    & 87.2$^{\dagger}$ & 58.6$^{\dagger}$ &   
    & 87.6$^{\dagger}$ & 54.0$^{\dagger}$ &
    & 89.5$^{\dagger}$ & 66.8$^{\dagger}$ &
    & \textbf{89.6} & \textbf{85.8} & 
    & & &
    \textbf{+} \\

    \citep{Eberts2020Span-basedPre-training}                 
    & - & - &   
    & - & - &
    & 88.9$^{\dagger}$ & 71.5$^{\dagger}$ &
    & \textbf{89.3}$^{\dagger}$ & 79.2$^{\dagger}$&
    & & &
    -\\
                        
    \citep{dixit-al-onaizan-2019-span}                    
    & 86.0 & \textbf{62.8} &   
    & - & - &
    & - & - &
    & - & - &
    & & &
    -\\
                                                            
    \citep{li-etal-2019-entity}
    & 84.8 & 60.2 &
    & 83.6 & 49.4 &
    & 87.8$^{\ast}$ & 68.9$^{\ast}$ & 
    & - & - &
    &  &  &
    \textbf{+}\\
    
    \cdashline{1-16}
     
    \citep{sun-etal-2018-extracting}                    
    & 83.6 & 59.6 &
    & - & - &
    & - & - &
    & - & - &
    & & &
    \textbf{+}\\

    \citep{bekoulis-etal-2018-adversarial}
    & - & - &
    & 81.6$^{\dagger}$ & 47.5$^{\dagger}$ &
    & 83.6$^{\dagger}$ & 62.0$^{\dagger}$ &
    & 86.7 & 75.5 &
    & & &
    \textbf{+}\\
                            
    \citep{Bekoulis2018JointProblem}
    & - & - &
    & 81.2$^{\dagger}$ & 47.1$^{\dagger}$ & 
    & 83.9$^{\dagger}$ & 62.0$^{\dagger}$ &
    & 86.4 & 74.6 &
    & & &
    \textbf{+}\\
    
    \citep{zhang-etal-2017-end}
    & 83.6 & 57.5 &
    & - & - &
    & 85.6$^{\ast}$ & 67.8$^{\ast}$ & 
    & - & - &
    & & &
    -\\
    
    \citep{Li2017AText}
    & - & - &
    & - & - &
    & - & - &
    & 84.6 & 71.4 &
    & & &
    \textbf{+}\\
                            
    \citep{katiyar-cardie-2017-going}
    & 82.6 & 53.6 &
    & 79.6 & 45.7 &
    & - & - &
    & - & - &
    & & &
    -\\
                            
    \citep{Li2016JointText}
    & - & - &
    & - & - &
    & - & - &
    &  79.5 & 63.4 &
    & & &
    -\\
                            
                            
    \citep{miwa-bansal-2016-end}
    & 83.4 & 55.6 &
    & 81.8 & 48.4 &
    & - & - &
    & - & - &
    & & &
    -\\
                            
    \citep{miwa-sasaki-2014-modeling}
    & - & - &
    & - & - &
    & 80.7$^{\ast}$ & 61.0$^{\ast}$ &
    & - & - &
    & & &
    -\\
    
    \citep{li-ji-2014-incremental}
    & 80.8 & 49.5 &
    & 79.7 & 45.3 &
    & - & - &
    & - & - &
    &  &  &
    -\\
    
    \midrule
    \textbf{Boundaries Evaluation} \\
    
    \citep{zhong-chen-2021-frustratingly}
    & \textbf{89.7} & \textbf{69.0} &   
    & \textbf{88.8} & \textbf{64.7} &
    &  &  &
    &  &  &
    & 66.6 & 48.2 &
    - \\
    
    \citep{wang-lu-2020-two}
    & \textbf{89.5} & 67.6 &   
    & \textbf{88.6} & 63.3 &
    & \textbf{90.1} & \textbf{73.8} &
    &  &  & 
    & - & - &
    - \\
    
    \citep{Eberts2020Span-basedPre-training}                 
    & - & - &   
    & - & - &
    &  &  &
    &  &  &
    & \textbf{70.3}$^{\dagger}$ & \textbf{50.8}$^{\dagger}$&
    -\\
    
    \citep{wadden-etal-2019-entity} \xmark
    & 88.6 & 63.4 &   
    & - & - &
    &  &  &
    &  &  &
    & 67.5 & 48.4 &
    \textbf{+}\\
    
    \citep{luan-etal-2019-general} \xmark
    & 88.4 & 63.2 &   
    & 87.4 & 59.7 &
    &  &  &
    &  &  &
    & 65.2 & 41.6 &
    \textbf{+}\\

    
    \citep{luan-etal-2018-multi}
    & - & - &
    & - & - &
    &  &  &
    &  &  &
    & 64.2 & 39.3 &
    \textbf{+}\\
    
    \cdashline{1-16}
    
    \citep{Zheng2017JointNetwork} \xmark
    & - & 52.1 &
    & - & - &
    &  &  &
    &  &  &
    & - & - &
    -\\
    
    \citep{li-ji-2014-incremental}
    & 80.8 & 52.1 &
    & 79.7 & 48.3 &
    &  &  &
    &  &  &
    & - & - &
    -\\
    
    \midrule
    \midrule
    
    \textbf{Relaxed Evaluation}
    &  &  &
    &  &  &
    & \multicolumn{2}{c}{MF1} \\
    
    \citep{Nguyen2019End-to-endAttention} \xmark
    &  &  &
    &  &  & 
    & \textbf{93.8} & \textbf{69.6} &
    &  & &
    & & &
    -\\
    
    \citep{bekoulis-etal-2018-adversarial}
    &  &  &
    &  &  &
    & 93.0$^{\dagger}$ & 68.0$^{\dagger}$ &
    & & &
    & & &
    \textbf{+}\\
    
    \citep{Bekoulis2018JointProblem}
    &  &  &
    &  &  &
    & 93.3$^{\dagger}$ & 67.0$^{\dagger}$ &
    & & & 
    & & & 
  \textbf{+}\\
                            
    \citep{adel-schutze-2017-global}
    &  &  &
    &  &  &
    & 82.1 & 62.5 &
    &  & &
    & & &
    -\\
                            
    \citep{gupta-etal-2016-table}
    &  &  & 
    &  &  & 
    & 92.4$^{\dagger}$ & \textbf{69.9}$^{\dagger}$ &
    & & &
    & & &
    -\\
    
    \midrule
    \midrule
    \textbf{Not Comparable}\\
    
    \citep{Sanh2019ATasks} \xmark
    & 85.5 & 60.5 &
    &  &  &
    &  &  &
    &  &  &
    &  &  &
    -\\
    \cdashline{1-16}                              
    \bottomrule
    
    \end{tabularx}
    
    \end{adjustwidth*}
    \end{table}
    
\renewcommand{\arraystretch}{1}

\paragraph{Confusing Settings on CoNLL04}
On the CoNLL04 dataset, the two settings that have been used are even more different.
Indeed, while \citet{miwa-sasaki-2014-modeling} use the Strict evaluation, \citet{gupta-etal-2016-table}, who build upon the same Table Filling idea, introduce a different setting.
They 1) use the Relaxed criterion; 2) discard the ``Other" entity type; 3) release another train / test split; 4) use Macro-F1 scores.

This inevitably leads to confusions, first on the train / test splits, e.g. \citet{Giorgi2019End-to-endModels} 
claim to use the splits from \citep{miwa-sasaki-2014-modeling} while they link to \citep{gupta-etal-2016-table}'s. 
Second, \citet{Nguyen2019End-to-endAttention} unconsciously introduce a different \textit{Strict setup} because it ignores the ``Other" entity type and considers Macro-F1 instead of micro-F1 scores. This leads to unfair comparisons.

\paragraph{Altering both Metrics and Data}
\citet{Sanh2019ATasks} propose a multitask Framework for NER, RE and CR and use ACE05 to evaluate end-to-end RE.
However, they combine two mistakes: incorrect metric comparison and dataset alteration.
First, they use the typical formulation to describe a Strict setting but, in fact, use a setting looser than Boundaries.
Indeed, they do not consider the type of arguments and only their last word must be correctly detected.
Second, they truncate the ACE05 dataset to sentences containing at least one relation both in train and test sets, which leads to an even more favorable setting.

What is worrisome is that both these mistakes are almost invisible in their paper and can only be detected in their code.
The only hint for incorrect evaluation is that they report a score for a setting where they only supervise RE, which is impossible in any standard setting. 
For the dataset, the fact that they do not use the standard preprocessing from \citep{miwa-bansal-2016-end}\footnotemark[1] might be a first clue.

\paragraph{Are We Even Using the Same Datasets?}
Without going this far into data alteration, a first source of ambiguity resides in the use or not of the validation set as additional training data.
While on CoNLL04, because there was no initial agreement on a dev set, the final model is often trained on train+dev by default; the situation is less clear on ACE.
And our following experiments show that this point is already critical w.r.t SOTA claims.

Considering data integrity and keeping the ACE datasets example, even when the majority of works refer to the same preprocessing scripts\footnotemark[1] \footnotetext[1]{\href{https://github.com/tticoin/LSTM-ER}{github.com/tticoin/LSTM-ER}} there is no way to check the integrity of the data without a report of complete dataset statistics.
This is especially true for these datasets whose license prevents sharing of preprocessed versions.

Yet, we have to go back to \citep{roth-yih-2004-linear} to find the original CoNLL04 statistics and \citep{li-ji-2014-incremental} for ACE datasets. 
To our knowledge, only a few recent works report in-depth datasets statistics \citep{adel-schutze-2017-global,Sanh2019ATasks,Giorgi2019End-to-endModels}.
We report them for CoNLL04 and ACE05 in \autoref{table:04:data_stats} along with our own.

We observe differences in the number of sentences, entity mentions and relations.
Minor differences in the number of annotated mentions likely come from evolutions in datasets versions.
Their impact on performance comparison should be limited, although problematic.   
But we also observe more impactful differences, e.g. with \citep{Giorgi2019End-to-endModels} for both datasets and despite using the same setup and preprocessing.

Such a difference in statistics reminds us that the dataset is an integral part of the evaluation setting.
And in the absence of sufficiently detailed reports, we cannot track when and where they have been changed since their original introduction.

\begin{table}[!h]
\begin{center}
 \begin{tabular}{@{}ll*{4}{c}@{}}

\toprule
\parbox[t]{2mm}{\multirow{4}{*}{\rotatebox[origin=c]{90}{CoNLL04}}} &  & (R\&Y, 04) & (A\&S, 17) & (G, 19) & Ours\\
\cline{3-6}
& \# sents  &  1,437 &  -       & - & 1,441\\
& \# ents   &  5,336  &   5,302          & 14,193 & 5,349\\
& \# rels  &  2,040 &  2,043             & 2,048 & 2,048\\ 

\midrule
\parbox[t]{2mm}{\multirow{4}{*}{\rotatebox[origin=c]{90}{ACE05}}}&   &   (L\&J, 14) & (S, 19)  &  (G, 19) & Ours \\
\cline{3-6}
& \# sents  &  10,573     &     10, 573       & -         & 14,521 \\
& \# ents   &  38,367      & 34,426 & 38,383    & 38,370 \\
& \# rels   &  7,105       & 7,105 & 6,642     & 7,117 \\

\bottomrule
 \end{tabular}
\end{center}
\caption{Global datasets statistics in CoNLL04 and ACE05 as reported by different sources. More detailed statistics are available in Appendix.}
\label{table:04:data_stats}
\end{table}

\subsection{A Small Empirical Study}
\label{subsec:05:empirical_study}
Given these previous inconsistencies, we can legitimately wonder what is the impact of different evaluation settings on quantitative performance.
However, it is also unrealistic to reimplement and test each and every paper in a same setting to establish a benchmark.
Instead, we propose a small empirical study to quantify the impact of using the Boundaries setting instead of the Strict setting on the two main benchmarks: \textbf{CoNLL04} and \textbf{ACE05}.
We discard the Relaxed setting because it cannot evaluate true end-to-end RE without strictly taking argument detection into account.
It is also limited to CoNLL04 and we have not found any example of misuse.

We will consider a limited set of models representative of the main \textbf{Entity Filtering} approach.
And we seize this opportunity to perform two ablations that correspond to meaningful recent proposals and we believe are missing in related work.

First, when looking at \autoref{table:04:re_sota}, it is difficult to draw general conclusions beyond the now established improvements due to the evolution of word representations to using \textbf{pretrained Language Models}.
And in the absence of ablation studies on the matter\footnotemark[1], it is impossible to compare models using pretrained LM and anterior works.
For example, in the novel work of \citet{li-etal-2019-entity}, we cannot disentangle the quantitative effects of pretrained LM and the proposed MultiTurn Question Answering approach, which is a shame given its originality. 

\footnotetext[1]{Excepting in \citep{Sanh2019ATasks} which ablates ELMo}

Second, to our knowledge, no article compares the recent use of \textbf{span-level NER} instead of classical sequence tagging in end-to-end RE.
And while Span-level NER does seem necessary to detect overlapping or nested mentions, we can wonder if it is already beneficial on datasets without overlapping entities (like CoNLL04 and ACE05), as suggested by \citep{dixit-al-onaizan-2019-span}.

\newcommand{\raisemath}[1]{\mathpalette{\raisemeth{#1}}}
\newcommand{\raisemeth}[3]{\raisebox{#1}{$#2#3$}}

\renewcommand{\stddev}{\raisemath{-3pt}}{}
\renewcommand{\arraystretch}{1.4}

\begin{table}[h]
\begin{adjustwidth*}{}{-.3\textwidth}

\caption[Double ablation study of BERT and Span-level NER.]{Double ablation study of BERT and Span-level NER. We report the average of five runs and their standard deviation in subscript. For RE we consider both the Strict and Boundaries settings, RE Strict score is used as the criterion for early stopping.}
\label{table:experiments}

\centering
\small
\begin{tabularx}{1.3\textwidth}{@{}llr*{2}{Y}r*{2}{Y}r*{2}{Y}r*{2}{Y}r*{2}{Y}r*{2}{Y}r@{}}

\toprule

\multicolumn{3}{c}{\multirow{3}{*}{$\mu$F1}} & \multicolumn{8}{c}{CoNLL04} & & \multicolumn{8}{c}{ACE05}\\
\cline{4-11} \cline{13-20}
 & & & \multicolumn{2}{c}{NER} & & \multicolumn{2}{c}{RE (S)} &  & \multicolumn{2}{c}{RE (B)}&  & \multicolumn{2}{c}{NER} & & \multicolumn{2}{c}{RE (S)} & & \multicolumn{2}{c}{RE (B)}\\

\cline{4-5} \cline{7-8} \cline{10-11} \cline{13-14} \cline{16-17} \cline{19-20}

& & & Dev & Test & & Dev & Test & & Dev & Test & & Dev & Test & & Dev & Test & & Dev & Test\\
\midrule
\parbox[t]{2mm}{\multirow{4}{*}{\rotatebox[origin=c]{90}{BERT}}}
    &\parbox[t]{2mm}{\multirow{2}{*}{\rotatebox[origin=c]{90}{Span}}}
    & train
    & 85.2$_{\stddev{1.9}}$ & 86.5$_{\stddev{1.4}}$ &
    & 69.5$_{\stddev{1.9}}$ & \textbf{67.8}$_{\stddev{.6}}$ &
    & 69.6$_{\stddev{2.0}}$ & \textbf{68.0}$_{\stddev{.5}}$ & 
    
    & 84.6$_{\stddev{.6}}$ & 86.2$_{\stddev{.4}}$ &
    & \textbf{60.1}$_{\stddev{1.0}}$ & \textbf{59.6}$_{\stddev{1.0}}$ &
    &\textbf{63.2}$_{\stddev{.9}}$ &\textbf{62.9}$_{\stddev{1.2}}$ \\
    
    & 
    & +dev 
    & - & 87.5$_{\stddev{.8}}$ &
    & - & \textbf{70.1}$_{\stddev{1.2}}$ & 
    & - & \textbf{70.4}$_{\stddev{1.2}}$ &
    
    & - & 86.5$_{\stddev{.4}}$ &
    & - & \textbf{61.2}$_{\stddev{1.3}}$ &
    & - &\textbf{64.2}$_{\stddev{1.3}}$ \\ 

    & \parbox[t]{2mm}{\multirow{2}{*}{\rotatebox[origin=c]{90}{Seq}}} 
    & train
    & \textbf{86.4}$_{\stddev{1.0}}$ & \textbf{87.4}$_{\stddev{.8}}$ &
    & \textbf{71.0}$_{\stddev{1.8}}$ & \textbf{68.3}$_{\stddev{1.9}}$ &
    & \textbf{71.1}$_{\stddev{1.7}}$ & \textbf{68.5}$_{\stddev{1.8}}$ &
    
    & \textbf{85.7}$_{\stddev{.2}}$ & \textbf{87.0}$_{\stddev{.3}}$ &
    & \textbf{60.1}$_{\stddev{.8}}$ & \textbf{59.7}$_{\stddev{1.1}}$ &
    & 62.6$_{\stddev{1.1}}$ & \textbf{62.9}$_{\stddev{1.2}}$ \\
    
    & 
    & +dev 
    & - & \textbf{88.9}$_{\stddev{0.6}}$ &
    & - & \textbf{70.0}$_{\stddev{1.2}}$ & 
    & - & \textbf{70.2}$_{\stddev{1.2}}$ &
    
    & - & \textbf{87.4}$_{\stddev{.3}}$ &
    & - & \textbf{61.2}$_{\stddev{1.1}}$ &
    & - & \textbf{64.4}$_{\stddev{1.6}}$ \\

\cdashline{1-20}

\parbox[t]{2mm}{\multirow{4}{*}{\rotatebox[origin=c]{90}{BiLSTM}}}
    & \parbox[t]{2mm}{\multirow{2}{*}{\rotatebox[origin=c]{90}{Span}}}
    & train
    & 79.8$_{\stddev{1.6}}$ & 80.3$_{\stddev{1.2}}$ &
    & 61.0$_{\stddev{1.2}}$ & 56.1$_{\stddev{1.4}}$ &
    & 61.2$_{\stddev{1.1}}$ & 56.4$_{\stddev{1.4}}$ &
    
    & 80.0$_{\stddev{.2}}$ & 81.3$_{\stddev{.4}}$ &
    & 46.5$_{\stddev{.8}}$ & 49.4$_{\stddev{1.3}}$ &
    & 49.3$_{\stddev{.9}}$ & 51.9$_{\stddev{1.3}}$ \\
    
    & 
    & +dev 
    & - & 82.7$_{\stddev{1.2}}$ &
    & - & 58.2$_{\stddev{1.5}}$ & 
    & - & 58.5$_{\stddev{1.6}}$ &
    
     & - & 82.2$_{\stddev{.3}}$ &
    & - & 49.3$_{\stddev{.2}}$ &
    & - & 51.9$_{\stddev{.6}}$ \\

    & \parbox[t]{2mm}{\multirow{2}{*}{\rotatebox[origin=c]{90}{Seq}}} 
    & train
    & 80.5$_{\stddev{.7}}$ & 82.0$_{\stddev{.3}}$ &
    & 62.8$_{\stddev{.6}}$ & 60.6$_{\stddev{1.9}}$ &
    & 63.3$_{\stddev{.9}}$ & 60.7$_{\stddev{1.8}}$ &
    
    & 80.8$_{\stddev{.5}}$ & 82.5$_{\stddev{.4}}$ &
    & 47.2$_{\stddev{.5}}$ & 50.3$_{\stddev{1.4}}$ &
    & 49.3$_{\stddev{.5}}$ & 52.8$_{\stddev{1.4}}$ \\
    
    & & +dev 
    & - & 82.6$_{\stddev{.9}}$ &
    & - & 61.6$_{\stddev{1.8}}$ & 
    & - & 61.7$_{\stddev{1.6}}$ & 
    
    & - & 82.8$_{\stddev{.2}}$ &
    & - & 50.1$_{\stddev{1.4}}$ & 
    & - & 52.9$_{\stddev{1.6}}$ \\

\bottomrule

\end{tabularx}

\end{adjustwidth*}
\end{table}

\renewcommand{\arraystretch}{1}

\paragraph{Dataset preprocessing and statistics}
\label{sec:data}
We use the standard preprocessing from \citep{miwa-bansal-2016-end} to preprocess ACE05\footnotemark[1].

For CoNLL04, we take the preprocessed dataset and train / dev / test split from \citep{Eberts2020Span-basedPre-training}\footnotemark[2] and check that it corresponds to the standard train / test split from \citep{gupta-etal-2016-table}\footnotemark[3].
We report global dataset statistics in \autoref{table:04:data_stats}.

\footnotetext[1]{\href{https://github.com/tticoin/LSTM-ER}{github.com/tticoin/LSTM-ER}}
\footnotetext[2]{\href{https://github.com/markus-eberts/spert}{github.com/markus-eberts/spert}}
\footnotetext[3]{\href{https://github.com/pgcool/TF-MTRNN}{github.com/pgcool/TF-MTRNN}}

\paragraph{Models}

We propose to use a model inspired by \citep{Eberts2020Span-basedPre-training} as a baseline for our ablation study since they combine BERT finetuning and Span-level NER.
We then perform two ablations: replacing BERT by a BiLSTM encoder with non-contextual representations and substituting Span-level NER with IOBES sequence tagging.

\paragraph{Encoder : BiLSTM vs BERT}
We use BERT \citep{devlin-etal-2019-bert} as LM pretraining baseline, expecting that the effects of ELMo \citep{peters-etal-2018-deep} would be similar.
As in related work, we use cased BERT$_{\text{BASE}}$ and finetune its weights.
A word is represented by max-pooling of the last hidden layer representations of all its subwords.

For our non-contextual baseline, we take the previously ubiquitous BiLSTM encoder and choose a 384 hidden size in each direction so that the encoded representation matches BERT's dimension.
We feed this encoder with the concatenation of 300d GloVe 840B word embeddings \citep{pennington-etal-2014-glove} and a reproduction of the charBiLSTM from \citep{lample-etal-2016-neural} (100d char embeddings and hidden size 25 in each direction).

\paragraph{NER Decoder : IOBES vs Span}
In the sequence tagging version, we simply feed the previously encoded word representation $\textbf{h}_i$ into a linear layer with a softmax to predict IOBES tags.
\begin{equation}
    \hat{\textbf{y}}^{seq}_i = \text{softmax}(W^{seq} . \textbf{h}_i + \textbf{b}^{seq})
\end{equation}

For span-level NER,
we only consider spans up to maximal length 10, which are represented by the max pooling of the representations of their tokens.
An additional span width embedding \textbf{w} of dimension 25 is concatenated to this representation as in \citep{lee-etal-2017-end}.
The only difference with \citep{Eberts2020Span-basedPre-training} is that they also concatenate the representation of the [CLS] token in all span representations to incorporate sentence-level information.
We discard this specificity of BERT-like models.
All these span-level representations are classified using a linear layer followed by a softmax to predict entity types (including None).
We also use negative sampling by randomly selecting 100 negative spans during training.
\begin{align}
    \textbf{h}(s) & = \text{MaxPool}(\textbf{h}_{i}, ...\textbf{h}_{i+l-1})\\
    \textbf{e}(s) & = \lbrack \textbf{h}(s); \textbf{w} (l) \rbrack\\
    \hat{\textbf{y}}^{span}(s) & = \text{softmax}(W^{span} . \textbf{e}(s) + \textbf{b}^{span})
\end{align}
The NER loss $\mathcal{L}_{NER}$ is the cross-entropy over either IOBES tags or entity classes.

\paragraph{RE Decoder}
For the RE Decoder, we first filter candidate entity pairs i.e. all the ordered pairs of entity mentions detected by the NER decoder.
Then, for every pair, the input of the relation classifier is the concatenation of each span representation \textbf{e}(s$_{i}$) and a context representation \textbf{c}(s$_{1}$, s$_{2}$), the max pooling of all tokens strictly between the two spans\footnotemark[1].
\footnotetext[1]{If there are none, $\textbf{c}(s_{1}, s_{2}) = \bm{0}$}
Once again, this pair representation is fed to a linear classifier but with a sigmoid activation so that multiple relations could be predicted for each pair.
\begin{align}
    \textbf{x}(s_1, s_2) & = \lbrack \textbf{e}(s_1); \textbf{e}(s_2); \textbf{c}(s_{1}, s_{2}) \rbrack \\
    \hat{\textbf{y}}^{rel}(s_1, s_2) & = \sigma (W^{rel} . \textbf{x}(s_1, s_2)  + \textbf{b}^{rel})
\end{align}
$\mathcal{L}_{RE}$ is computed as the binary cross-entropy over relation classes. 
During training, we sample up to 100 random negative pairs of detected or ground truth spans, which is different from \citep{Eberts2020Span-basedPre-training} in which negative samples contain only ground truth spans.

\paragraph{Joint Training}
As in most related work, we simply optimize for $\mathcal{L} = \mathcal{L}_{NER} + \mathcal{L}_{RE}$.

\paragraph{Experimental Setting}
We implement these models with Pytorch \citep{Paszke2019PyTorch:Library} and Huggingface Transformers \citep{wolf-etal-2020-transformers}.
For all settings, we fix a dropout rate of 0.1 across the entire network, a 0.1 word dropout for Glove embeddings and a batch size of 8.
We use Adam optimizer \citep{Kingma2015Adam:Optimization} with $\beta_1=0.9$ and $\beta_2=0.999$.
A preliminary grid search on CoNLL04 led us to select a learning rate of  10$^{-5}$ when using BERT and 5.10$^{-4}$ with the BiLSTM\footnotemark[2].

\footnotetext[2]{Search in $\{10^{-6}, 5.10^{-6}, \bm{10^{-5}}, 5.10^{-5}, 10^{-4}\}$ with BERT and $\{10^{-4}, \bm{5.10^{-4}}, 10^{-3}, 5.10^{-3}, 10^{-2}\}$ otherwise.} 

We perform early stopping with patience 5 on the dev set Strict RE $\mu$ F1 score with a minimum of 10 epochs and a maximum of 100.
To compare to related work on CoNLL04, we retrain on train+dev for the optimal number of epochs as determined by early stopping.\footnotemark[3]
\footnotetext[3]{This is not a reproduction of the experimental setting used in \citep{Eberts2020Span-basedPre-training}.}

We report aggregated results from five runs in \autoref{table:experiments}.

\subsubsection{Quantifying the Impact of Comparing Boundaries and Strict Setups}
\label{sec:boundvsstrict}
This humble study first quantifies the impact of using Boundaries instead of Strict evaluation to an overestimation of 2.5 to 3 F1 points on ACE05 (i.e. a 5\% relative improvement), which is far from negligible.

But it is also interesting to see that such a mistake has almost no impact on CoNLL04, which highlights an overlooked difference between the two datasets.
A simple explanation is the reduced number of entity types (4 against 7) which reduces the chance to wrongly type an entity.
But we can also notice the difference in the variety of argument types in each relation.
Indeed, in CoNLL04 there is a \textbf{bijective mapping} between a relation type and the ordered types of its arguments; this minimal difference suggests that our models have mostly learned it.
On the contrary on ACE05, this mapping is much more complex (e.g. the relation PART-WHOLE fits 9 pairs of types\footnotemark[4])\footnotetext[4]{see additional details in \autoref{app:sincere:comparison_ace_conll} of the Appendix.}
which explains the larger difference between metrics, whereas the NER F1 scores are comparable.

\subsubsection{Comments on the Ablations}
We must first note that with our full BERT and Span NER baseline, our results do not match those reported by \citet{Eberts2020Span-basedPre-training}.
This can be explained by the slight differences in the models but most likely in the larger ones in tranining procedure and hyperparameters.
Furthermore, we generally observe an important variance over runs, especially for RE.

As expected, the empirical gains mainly come from using BERT, which 
allows the use of simpler decoders for both NER and RE.
Indeed, although our non-contextual IOBES model matches \citep{bekoulis-etal-2018-adversarial} on CoNLL04, the results on ACE05 are overtaken by models using external syntactic information or more sophisticated decoders with a similar BiLSTM encoder.

Comparing the Span-level and sequence tagging approaches for NER is also interesting.
Indeed, although an advantage of Span-level NER is the ability to detect overlapping mentions, its contribution to end-to-end RE on non-overlapping mentions has never been quantified to our knowledge.
Our experiments suggest that it is not beneficial in this case compared to the more classical sequence tagging approach.

\subsection{How to Prevent Future Mistakes?}

The accumulation of mistakes and invalid comparisons should raise questions to both authors and reviewers of end-to-end RE papers.
How was it possible to make them in the first place and not to detect them in the second place?
How can we reduce their chance to occur in the future?

\paragraph{Lack of Reproducibility}

First, it is no secret that the lack of reproducibility is an issue in science in general and Machine Learning in particular, but we think this is a perfect illustration of its symptoms.
Indeed, in the papers we studied, we only found comparisons to reported scores and rarely an attempt to reimplement previous work by different authors.
This is perfectly understandable given the complexity of such a reproduction, in particular in the multitask learning setting of end-to-end RE and often without (documented) source code.

However, this boils down to comparing results obtained in different settings.
We believe that simply evaluating an implementation of the most similar previous work enables to detect differences in metrics or datasets.
But it also allows to properly assess the source of empirical gains  \citep{Lipton2018TroublingScholarship} which could come from different hyperparameter settings \citep{Melis2018OnModels} or 
in-depth changes in the model. 

\paragraph{Need for More Complete Reports}
Although it is often impossible to exactly reproduce previous results even when the source code is provided, we should at least expect that the evaluation setting is always strictly reproduced.
This requires a complete explicit formulation of the evaluation metrics associated with a clear and unambiguous terminology, to which end we advocate for using \citep{bekoulis-etal-2018-adversarial}'s.
Datasets preprocessing and statistics should also be reported
to provide a sanity check.
This should include at least the number of sentences, entity and relation mentions as well as the details of train / test partitions.

\paragraph{Towards a Unified Evaluation Setting}

Finally, in order to reduce confusion, we should aim at unifying our evaluation settings.
We propose to always at least report RE scores with the Strict criterion, which considers both the boundaries and types of arguments.
This view matches the NER metrics and truly assess end-to-end RE performance. 
It also happens to be the most used in previous work.

The Boundaries setting proposes a complementary measure of performance more centered on the relation.
The combination of Strict and Boundaries metrics can thus provide additional insights on the models, as discussed in \autoref{sec:boundvsstrict} where we deduce that models can learn the bijective mapping between argument and relation types in CoNLL04.
However, we believe this discussion on their specificities often lacks in articles where both metrics are reported mostly in order to compare to previous works.
Hence we can only encourage to also report a Boundaries score provided sufficient explanation and exploitation of both metrics.

On the contrary, in our opinion, the Relaxed evaluation, which does not account for argument boundaries, cannot evaluate end-to-end RE since it reduces NER to Entity Classification. Furthermore, some papers report the average of NER and RE metrics \citep{adel-schutze-2017-global, Giorgi2019End-to-endModels}, which we believe is also an incorrect metric since the NER performance is already measured in the RE score.

Using a unified setting would also ease cross-dataset analyses and help to better reflect their often overlooked specificities.

\subsection{Conclusion}

The multiplication of settings in the evaluation of end-to-end Relation Extraction makes the comparison to previous work difficult.
Indeed, in this confusion, numerous articles present unfair comparisons, often overestimating the performance of their proposed model.
Furthermore, besides regular claims of new state-of-the-art results, it complicates the emergence of definitive conclusions on the relevance of each architecture design choice and consequently the proposition of new models.

Hence, our critical literature review epitomizes the need for more rigorous reports of evaluation settings, including detailed datasets statistics, so that we call for a unified end-to-end RE evaluation setting to prevent future mistakes and enable more meaningful cross-domain comparisons.
On that matter, we are glad to notice that the most recent works \citep{wang-lu-2020-two, zhong-chen-2021-frustratingly} adopt this setting and report respectively the first Boundaries scores on CoNLL04 and the first Strict scores on SciERC, that set the base for easier future cross-dataset comparisons. 

Finally, while this section focuses on the necessity to maintain correctness in benchmarks if we want to compare models, we also believe that it is interesting to push model evaluation further than a single global score that cannot reflect key linguistic specificities of individual data samples such as lexical overlap.
This is why we propose to extend the evaluation setting previously used in \autoref{chapter:03:ner} for NER to study the impact of lexical overlap on end-to-end RE in the following section.
\section{Isolating Retention in End-to-end RE}
\label{sec:05:retention}

Indeed, as discussed in \autoref{chapter:03:ner}, lexical overlap with the training set plays an important role in the evaluation of NER.
And because NER is an integral part of end-to-end RE, we can expect that it also plays a role in the final RE performance.
Yet, as for many NLP tasks, benchmarks limit to a single Precision, Recall and F1 report on an held-out test set that overlook important phenomena.

Hence, in this section, we propose to extend our previous study to the case of end-to-end RE, falling with a recent line of works that propose fine-grained NLP evaluation.
We first present these different works in \autoref{subsec:05:shortcoming-benchmarks}.
We then use both test set partition and a contained behavioral testing empirical study to highlight the importance of the retention heuristic in end-to-end RE models.
We finally discuss how future benchmarks could take this phenomenon into account by design.

\subsection{Addressing the Shortcomings of NLP Benchmarks}
\label{subsec:05:shortcoming-benchmarks}

With the recent breakthroughs brought by the use of language model pretraining as a preliminary representation learning step described in \autoref{chapter:02:bert}, NLP models have obtained scores superior to established human baselines on some benchmarks such as SQUAD 1.0 and 2.0 \citep{rajpurkar-etal-2016-squad, rajpurkar-etal-2018-know} for Question Answering (QA) or SuperGLUE \citep{Wang2019SuperGLUE:Systems} composed of several Natural Language Understanding (NLU) tasks, such as QA, Natural Language Inference (NLI), Coreference Resolution or Word Sense Disambiguation.
These impressive ``super-human'' performances, sometimes achieved within a year of their release, provided a motivation to carefully focus on what is exactly measured by these benchmarks because it is simultaneously obvious to practitioners that these models have not achieved a human-level of comprehension.
Hence, it is natural to wonder to what extend these NLU benchmarks really measure ``understanding'' and how they could be fooled by simple heuristics.

Whereas lexical overlap quickly appeared as an essential factor of performance in Named Entity Recognition \citep{palmer-day-1997-statistical} where it is trivial to detect known mentions, only recently can we see a surge in works that propose fine-grained studies of NLP models performance on various tasks.
They aim at finding the linguistic specificities of samples that can explain differences in performance, which can expose simple shallow heuristics adopted by the models.
Exposing a heuristic requires to replace a global test set by several subsets that 
have different characteristics, expected to cause different degrees of difficulty for  
this heuristic.
This can be either achieved by partitioning an existing global test set or designing more or less adversarial examples either manually or automatically.

\subsubsection{Test Set Partition}
As discussed in \autoref{chapter:03:ner}, fine-grained evaluation of models has been developed in Named Entity Recognition by several means.
First, \citet{Augenstein2017GeneralisationAnalysis} propose to separate performance on seen and unseen mentions, setting that we use in our own work \citep{Taille2020ContextualizedGeneralization}.
Without explicitly separating mentions with such an interpretable partition, two concurrent works propose to partition test set mentions into buckets determined by the value of several characteristics, including lexical overlap.

\citet{arora-etal-2020-contextual} consider three properties and split the test set in two halves for each of them: mention length, ambiguity (number of labels a token appears in the training set) and number of occurrences in the training set.
They show that contextual embeddings are more useful on the longer, more ambiguous or less seen mentions.
Their study also tackles Sentiment Analysis with different measures of instance complexity or ambiguity.

In the same spirit, \citet{Fu2020RethinkingStudy, fu-etal-2020-interpretable}, propose to create buckets according to eight properties of individual mentions either local (such as mention, sentence length or proportion of out-of-vocabulary words in the sentence) or global (such as label consistency or frequency).
The \textbf{label consistency} of a test mention is defined as the number of occurrences this mention appears in the training set with the same label over the total number of occurrences.
This takes into account two phenomena: lexical overlap (an unseen mention will have a label consistency of 0) and ambiguity (a mention appearing with only one label will have a label consistency of 1).
They divide the test sets into $m$ buckets and propose to measure the Spearman's rank correlation between these properties and the performance of several NER models on these buckets.
They conclude that label consistency and entity length are the main predictors of performance, respectively positively and negatively correlated to performance.
However, we believe that this correlation approach has a drawback since this measure of correlation depends on the choice of the number of buckets $m$, that is eluded in their study.
And it should aim at two opposite objectives: $m$ should be small to have a more consistent measure of performance inside each bucket and $m$ should be large to have more data points to measure a correlation.

In any case, this work led to the development of \textbf{Explainaboard} \citep{liu-etal-2021-explainaboard}, an interactive leaderboard that enables fine-grained model comparison by separating performance by buckets according to several properties.
At the time of writing this thesis, this leaderboard includes thirteen tasks such as NER, POS, NLI, Summarisation, Machine Translation, Chinese Word Segmentation or Text Classification.

\subsubsection{Out-of-Domain Evaluation}

Another way to test NLP models using preexisting evaluation resources is to perform out-of-domain evaluation.
This can be done by training a model on a given dataset representative of the \textbf{source} domain for a task and testing it on another \textbf{target} dataset.
While this setting is naturally used to study Transfer Learning or Domain Adaptation where target data can be used to improve target performance, it can also be used as a benchmark for out-of-domain generalization capabilities.

Again, this is an evaluation setting that we reuse from \citep{Augenstein2017GeneralisationAnalysis} in \autoref{chapter:03:ner}.
It has also been used by \citep{moosavi-strube-2017-lexical} in Coreference Resolution to study the influence of lexical overlap between coreferent mentions in the train and test sets.
They argue that using lexical features such as word embeddings leads to overfitting to mentions and that the improvement of deep neural networks over previous rule-based baselines fades away when evaluating out-of-domain.
They conclude that contextual features should play a larger role in models to improve their generalization capabilities.

\subsubsection{Adversarial Filtering}

Another set of methods propose to desaturate benchmarks by designing or filtering examples that are particularly difficult for current models: \textbf{adversarial examples}.
For example, \citet{paperno-etal-2016-lambada} propose LAMBADA, a particularly difficult Language Modeling dataset created manually by masking a target word from a final sentence that can only be guessed by considering the previous context passage of at least 50 tokens.
\citet{jia-liang-2017-adversarial} add a distractor sentence at the end of context paragraphs of the SQUAD question answering dataset and show that it heavily disturbs predictions of models, effictively questioning the comprehension capabilities of models.

Other works propose to iteratively collect adversarial data with both state-of-the-art models and humans in the loop.
For example, we can cite Adversarial NLI \citep{nie-etal-2020-adversarial} or Beat the AI in QA \citep{bartolo-etal-2020-beat}.
The creation of such adversarial datasets should be accelerated by the recent introduction of \textbf{Dynabench} \citep{kiela-etal-2021-dynabench}, an open-source platform for the creation of dynamic datasets.

\subsubsection{Behavioral Testing}
Finally, another method is to design controlled modifications or filtering of test examples in order to isolate the effect of given data patterns on performance.
These methods were mainly developed on the Natural Language Inference (NLI) task which aims at predicting if a first sentence (the premise P) and a second one (the hypothesis H) have a logical causation link.
It is traditionally modeled as a classification task with three cases: Entailment if H can be inferred from P, Contradiction if P and H contradict and Neutral otherwise.

\citet{naik-etal-2018-stress} identify several patterns correlated with predictions, e.g. a high lexical overlap between the sentences leads to predict entailment whereas the presence of negation words leads to predict contradiction.
They propose to construct stress tests to study the behaviour of models regarding these patterns.

\citet{mccoy-etal-2019-right} propose to separate shallow lexical overlap heuristics by modifying test instances in a controlled manner.
They first identify these heuristics, for example that about 90\% of training samples with high lexical overlap between the premise and hypothesis are labelled as entailment.
They reframe the problem as binary classification predicting the presence or absence of entailment and create the HANS dataset where there is an equal number of supporting and contradicting examples for each heuristic.
They show that the accuracy of different models is near perfect on supporting examples whereas it is near zero on contradicting ones, except with BERT where it still does not exceeds 20\%.
This supports the fact that even state-of-the-art NLP models adopt shallow heuristics such as lexical overlap in NLI and they are mostly ``right for the wrong reasons'' on standard benchmarks.

This type of experiment can be classified in the broader \textbf{behavioral testing} methods as discussed by \citet{ribeiro-etal-2020-beyond}.
They propose \textbf{CheckList}, a task agnostic framework to push evaluation beyond a single accuracy or F1 score on an held-out dataset by modifying test samples in a controlled manner, inspired by software engineering tests.
They test several commercial models from Google, Amazon and Microsoft and a BERT or RoBERTa based-model in Sentiment Classification, Paraphrase Detection and Question Answering.
For example in Sentiment Classification, they test performance on templates where a positive verb is negated and expect negative prediction but show that this leads to important failure rates (around 75\% at best). Likewise, they propose to replace a Named Entity mention by another one of same type (e.g. replace Chicago by Dallas) expecting that this should not affect model predictions, which fails in around 20\% cases.
They propose more test designs that can help identify pitfalls of current NLP models that are often overlooked by a single evaluation figure.

\subsection{The Case of End-to-end Relation Extraction}
\subsubsection{Shallow heuristics in pipeline Relation Extraction}
Previous work on NER, including our own, showed that lexical overlap plays an important role on final NER performance, that should be taken into account to better evaluate generalization capabilities of models (see \autoref{chapter:03:ner}).
In parallel, \citet{rosenman-etal-2020-exposing} and \citet{peng-etal-2020-learning} expose shallow heuristics in neural models: relying too much on the type of the candidate arguments or on the presence of specific triggers in their contexts.
Indeed, to specifically avoid relying to much on lexical representations of candidate arguments, Relation Classification models often introduce intermediate representations of candidate argument types.
In particular, BERT-based models introduce special tokens corresponding to argument types surrounding the argument mentions \citep{baldini-soares-etal-2019-matching}.

\citet{rosenman-etal-2020-exposing} reveal that RE predictions adopt heuristics based on candidate argument types and trigger words and that their main mistakes come from incorrect linking of an event to its argument.
This can only be detected in sentences with multiple pairs of arguments with types coherent with the same relation type.
They hence propose adversarial filtering of TACRED test samples to create the Challenging RE (CRE) subset where such type or event heuristics will find the correct relations but also predict false positives.
They show that on this test set, the accuracy of SOTA BERT-based models is higher on positive than negative pairs, suggesting that they indeed adopt these heuristics.
Furthermore, their experiments with QA based models suggest that they are less prone to these heuristics.

\citet{peng-etal-2020-learning} propose to study the behaviour of several models when they are given access to the candidate argument mentions, only their type, only their context or context+mention or context+type.
They argue that both context and mentions are crucial for RE and that current RE benchmarks might leak shallow cues in entity mentions.

\subsubsection{How does this apply to end-to-end RE ?}

In the more realistic end-to-end RE setting, we can naturally expect that these NER and RE heuristics are combined.
But we can also expect yet another heuristic: \textbf{the mere retention of training relation triples}.

In this section, we argue that current evaluation benchmarks measure both the desired ability to extract information contained in a text but also the capacity of the model to simply retain labeled (head, predicate, tail) triples during training.
And when the model is evaluated on a sentence expressing a relation seen during training, it is hard to disentangle which of these two behaviours is predominant.
However, we can hypothesize that the model can simply retrieve previously seen information acting like a mere compressed form of knowledge base probed with a relevant query.
Thus, testing on too much examples with seen triples can lead to overestimate the generalizability of a model.

Even without labeled data, LMs are able to learn some relations between words that can be probed with cloze sentences where an argument is masked  \citep{petroni-etal-2019-language}.
This raises the additional question of lexical overlap with the orders of magnitude larger unlabeled LM pretraining corpora that will remain out of scope of this work.

\subsection{Datasets and Models}
\label{sec:models}

We study three recent end-to-end RE models on \textbf{CoNLL04} \citep{roth-yih-2004-linear}, \textbf{ACE05} \citep{Walker2006ACECorpus} and \textbf{SciERC} \citep{luan-etal-2018-multi}.
They rely on various pretrained LMs and for a fairer comparison, we use BERT \citep{devlin-etal-2019-bert} on ACE05 and CoNLL04 and SciBERT \citep{beltagy-etal-2019-scibert} on SciERC\footnotemark[1] .

\footnotetext[1]{More implementation details in \autoref{app:05:retex-implementation} of the Appendix.}

\paragraph{PURE} \citep{zhong-chen-2021-frustratingly}   follows the pipeline approach. 
The NER model is a classical span-based model \citep{sohrab-miwa-2018-deep}. Special tokens corresponding to each predicted entity span are added and used as representation for Relation Classification. For a fairer comparison with other models, we study the approximation model that only requires one pass in each encoder and limits to sentence-level prediction. However, it still requires finetuning and storing two pretrained LMs instead of a single one for the following models.

\begin{figure}[h!]
    \begin{adjustwidth*}{}{-.3\textwidth}
    \caption[Illustration of the PURE model. Figure from \citep{zhong-chen-2021-frustratingly}.]{Illustration of the PURE model. Figure from \citep{zhong-chen-2021-frustratingly}.}
    \label{fig:05:pure}
    
    \centering
    \includegraphics[width=1.3\textwidth]{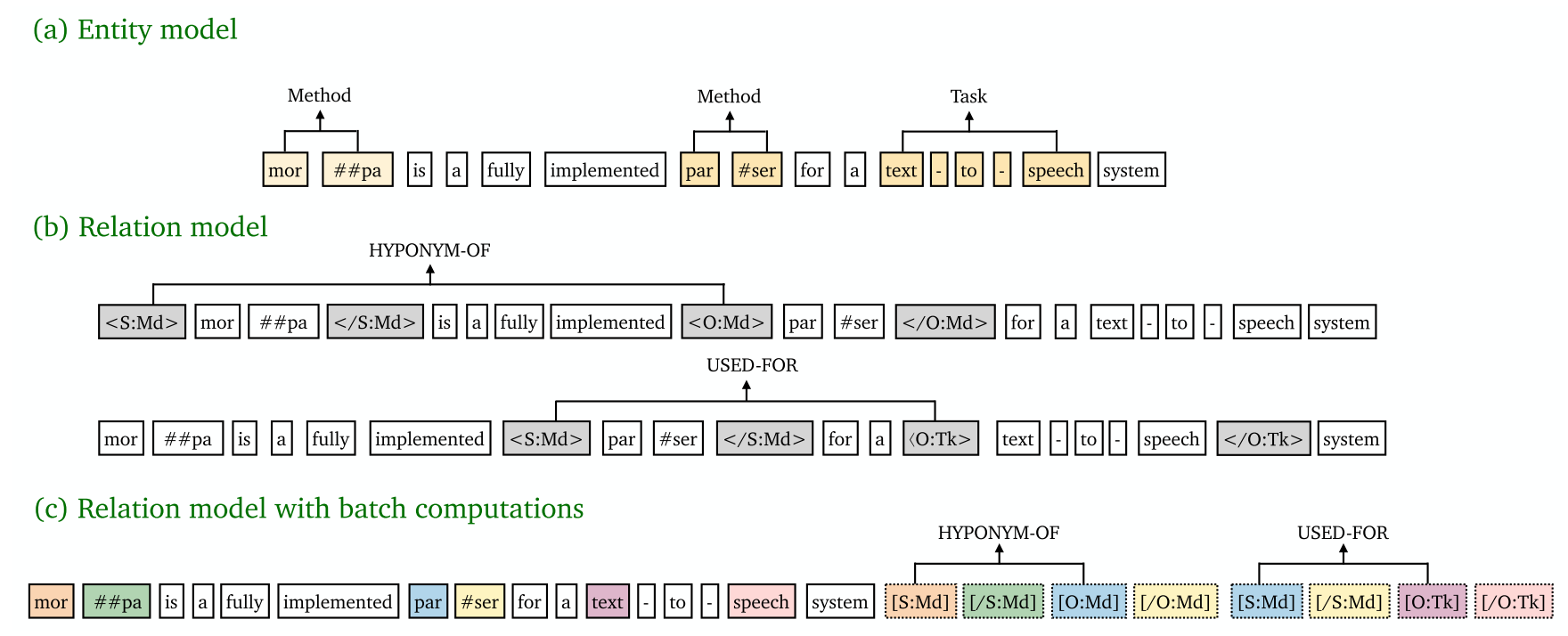}
    
    \end{adjustwidth*}
\end{figure}

\paragraph{SpERT} \citep{Eberts2020Span-basedPre-training}  uses a similar span-based NER module. 
RE is performed based on the filtered representations of candidate arguments as well as a max-pooled representation of their middle context. While Entity Filtering is close to the pipeline approach, the NER and RE modules share a common entity representation and are trained jointly. 
We also study the ablation of the max-pooled context representation that we denote \textbf{Ent-SpERT}.

\begin{figure}[h!]
    \begin{adjustwidth*}{}{-.3\textwidth}
    \caption[Illustration of the SpERT model. Figure from \citep{Eberts2020Span-basedPre-training}.]{Illustration of the SpERT model. Figure from \citep{Eberts2020Span-basedPre-training}.}
    \label{fig:05:spert}
    
    \centering
    \includegraphics[width=1.3\textwidth]{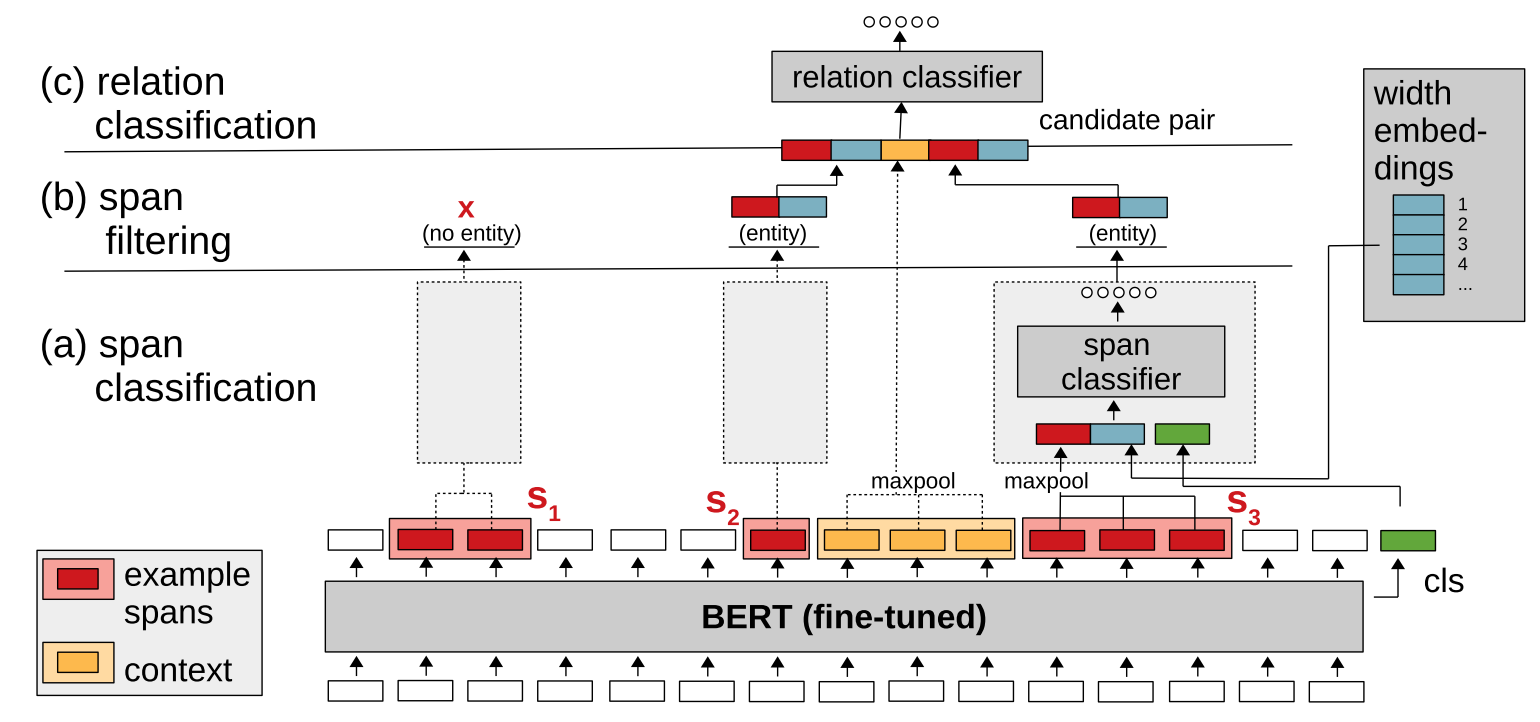}
    
    \end{adjustwidth*}
\end{figure}

\paragraph{Two are better than one (TABTO)} \citep{wang-lu-2020-two} intertwines a sequence encoder and a table encoder in a Table Filling approach \citep{miwa-sasaki-2014-modeling}.
Contrary to previous models the pretrained LM is frozen and both the final hidden states and attention weights are used by the encoders. The prediction is finally performed by a Multi-Dimensional RNN (MD-RNN). Because it is not based on span-level predictions, this model cannot detect nested entities, e.g. on SciERC.

\begin{figure}[h!]
    \begin{adjustwidth*}{}{-.3\textwidth}
    \caption[Illustration of the TABTO model. Figures from \citep{wang-lu-2020-two}.]{Illustration of the TABTO model. Figures from \citep{wang-lu-2020-two}.}
    \label{fig:05:tabto}
    
    \centering
    \includegraphics[width=1.3\textwidth]{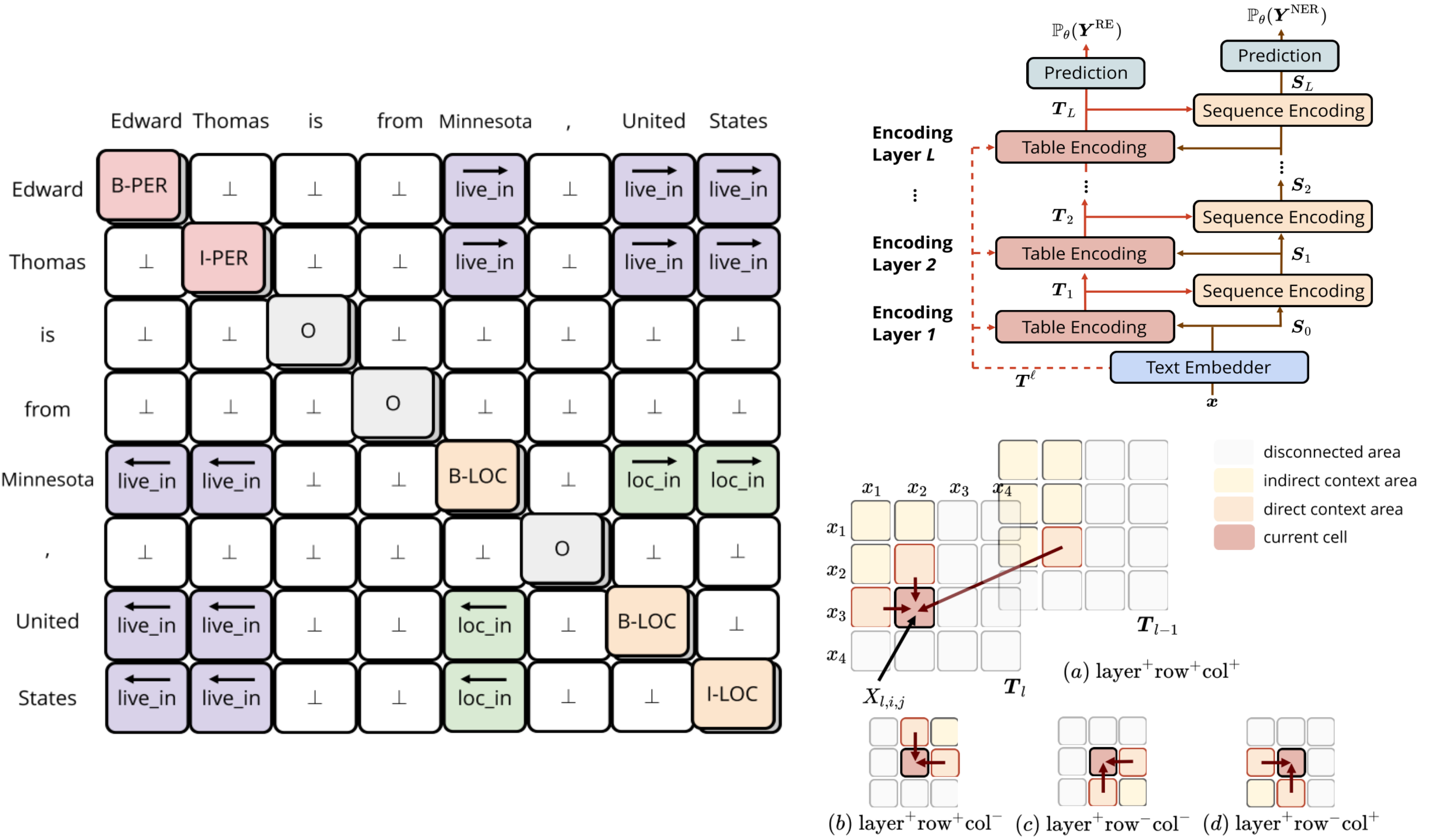}
    
    \end{adjustwidth*}
\end{figure}

    \begin{table}[t]
    \begin{adjustwidth*}{-.1\textwidth}{-.3\textwidth}
    
    \caption[Test NER and RE F1 Scores separated by lexical overlap with the training set.]{Test NER and RE F1 Scores separated by lexical overlap with the training set. Exact Match RE scores are not reported on SciERC where the support is composed of only 5 exactly seen relation instances. Average and standard deviations on five runs.}
    \label{table:05:overlap}
    
    \centering
    \small
    
    \begin{tabularx}{1.4\textwidth}{@{}l*{3}{Y}r*{4}{Y}r*{4}{Y}c@{}}
    
    \toprule
    \multicolumn{1}{c}{\multirow{2}{*}{$\mu$ F1}}
    
     & \multicolumn{3}{c}{NER} &  & \multicolumn{4}{c}{RE Boundaries} & & \multicolumn{4}{c}{RE Strict} \\
    \cline{2-4} \cline{6-9} \cline{11-14} 
    & Seen & Unseen & All & & Exact & Partial & New & All &  &  Exact & Partial & New & All  \\
    
    \midrule
    & \multicolumn{14}{c}{\textbf{ACE05}} \\
    \cline{2-4} \cline{6-9} \cline{11-14} 
    \textit{proportion} &\textit{ 82\%} & \textit{18\%} & & & \textit{23\%} & \textit{63\%} & \textit{14\%} & & & \textit{23\%} & \textit{63\%}& \textit{14\%}\\
    \midrule
    heuristic & 59.2$_{\stddev{\phantom{0.0}}}$ & - & 55.1$_{\stddev{\phantom{0.0}}}$ &  & 37.9$_{\stddev{\phantom{0.0}}}$ & - & - & 23.0$_{\stddev{\phantom{0.0}}}$ &  & 34.3$_{\stddev{\phantom{0.0}}}$ & - & - & 20.8$_{\stddev{\phantom{0.0}}}$ &  \\
    
    Ent-SpERT & 89.0$_{\stddev{0.1}}$ & 74.1$_{\stddev{1.0}}$ & 86.5$_{\stddev{0.2}}$ &  & 77.0$_{\stddev{1.1}}$ & 52.2$_{\stddev{1.1}}$ & 38.9$_{\stddev{1.0}}$ & 57.0$_{\stddev{0.8}}$ &  & 75.1$_{\stddev{1.2}}$ & 48.4$_{\stddev{1.0}}$ & 36.3$_{\stddev{2.0}}$ & 53.9$_{\stddev{0.8}}$ &  \\
    
    SpERT & 89.4$_{\stddev{0.2}}$ & 74.2$_{\stddev{0.8}}$ & 86.8$_{\stddev{0.2}}$ &  & 84.8$_{\stddev{0.8}}$ & 59.6$_{\stddev{0.7}}$ & 42.3$_{\stddev{1.1}}$ & 64.0$_{\stddev{0.6}}$ &  & 82.6$_{\stddev{0.8}}$ & 55.6$_{\stddev{0.7}}$ & 38.4$_{\stddev{1.1}}$ & 60.6$_{\stddev{0.5}}$ &  \\
    
    
    TABTO & 89.7$_{\stddev{0.1}}$ & 77.4$_{\stddev{0.8}}$ & 87.5$_{\stddev{0.2}}$ &  & 85.9$_{\stddev{0.9}}$ & 62.6$_{\stddev{1.8}}$ & 44.6$_{\stddev{2.9}}$ & \textbf{66.4}$_{\stddev{1.3}}$ &  & 81.6$_{\stddev{1.5}}$ & 58.1$_{\stddev{1.6}}$ & 38.5$_{\stddev{3.1}}$ & \textbf{61.7}$_{\stddev{1.1}}$ &  \\
    
    

    PURE & 90.5$_{\stddev{0.2}}$ & 80.0$_{\stddev{0.3}}$ & \textbf{88.7}$_{\stddev{0.1}}$ &  & 86.0$_{\stddev{1.3}}$ & 60.5$_{\stddev{1.0}}$ & 47.1$_{\stddev{1.6}}$ & \textbf{65.1}$_{\stddev{0.7}}$ &  & 84.1$_{\stddev{1.1}}$ & 57.9$_{\stddev{1.3}}$ & 44.0$_{\stddev{2.0}}$ & \textbf{62.6}$_{\stddev{0.9}}$ &  \\
    
    \midrule
     & \multicolumn{14}{c}{\textbf{CoNLL04}} \\
    \cline{2-4} \cline{6-9} \cline{11-14} 
    \textit{proportion} &\textit{ 50\%} & \textit{50\%} & & & \textit{23\%} & \textit{34\%} & \textit{43\%} & & & \textit{23\%} & \textit{34\%}& \textit{43\%} \\
    \midrule
    heuristic & 86.0$_{\stddev{\phantom{0.0}}}$ & - & 59.7$_{\stddev{\phantom{0.0}}}$ &  & 90.9$_{\stddev{\phantom{0.0}}}$ & - & - & 35.5$_{\stddev{\phantom{0.0}}}$ &  & 90.9$_{\stddev{\phantom{0.0}}}$ & - & - & 35.5$_{\stddev{\phantom{0.0}}}$ &  \\
     
    Ent-SpERT & 95.9$_{\stddev{0.3}}$ & 81.9$_{\stddev{0.2}}$ & \textbf{88.9}$_{\stddev{0.2}}$ &  & 92.3$_{\stddev{1.4}}$ & 60.8$_{\stddev{1.4}}$ & 54.6$_{\stddev{1.3}}$ & 64.8$_{\stddev{0.9}}$ &  & 92.3$_{\stddev{1.4}}$ & 60.8$_{\stddev{1.4}}$ & 54.2$_{\stddev{1.2}}$ & 64.7$_{\stddev{0.8}}$ &  \\
    
    SpERT & 95.4$_{\stddev{0.4}}$ & 81.2$_{\stddev{0.4}}$ & 88.3$_{\stddev{0.2}}$ &  & 91.4$_{\stddev{0.6}}$ & 67.0$_{\stddev{1.1}}$ & 59.0$_{\stddev{1.4}}$ & 69.3$_{\stddev{1.2}}$ &  & 91.4$_{\stddev{0.6}}$ & 66.9$_{\stddev{1.1}}$ & 58.5$_{\stddev{1.4}}$ & 69.0$_{\stddev{1.2}}$ &  \\
   
    TABTO & 95.4$_{\stddev{0.4}}$ & 83.1$_{\stddev{0.7}}$ & \textbf{89.2}$_{\stddev{0.5}}$ &  & 92.6$_{\stddev{1.5}}$ & 72.6$_{\stddev{2.1}}$ & 64.8$_{\stddev{1.0}}$ & \textbf{74.0}$_{\stddev{1.4}}$ &  & 92.6$_{\stddev{1.5}}$ & 72.1$_{\stddev{1.8}}$ & 64.7$_{\stddev{1.1}}$ & \textbf{73.8}$_{\stddev{1.2}}$ &  \\
    
    PURE & 95.0$_{\stddev{0.2}}$ & 81.8$_{\stddev{0.2}}$ & 88.4$_{\stddev{0.2}}$ &  & 90.1$_{\stddev{1.3}}$ & 66.6$_{\stddev{1.0}}$ & 58.6$_{\stddev{1.5}}$ & 68.3$_{\stddev{1.0}}$ &  & 89.9$_{\stddev{1.4}}$ & 66.6$_{\stddev{1.0}}$ & 58.5$_{\stddev{1.5}}$ & 68.2$_{\stddev{0.9}}$ &  \\
    
    \midrule
    & \multicolumn{14}{c}{\textbf{SciERC}} \\
    \cline{2-4} \cline{6-9} \cline{11-14} 
    \textit{proportion} &\textit{ 23\%} & \textit{77\%} & & & \textit{<1\%} & \textit{30\%} & \textit{69\%} & & & \textit{<1\%} & \textit{30\%}& \textit{69\%} \\
    \midrule
    
    heuristic & 31.3$_{\stddev{\phantom{0.0}}}$ & - & 20.1$_{\stddev{\phantom{0.0}}}$ & 
    & - & - & - & 0.7$_{\stddev{\phantom{0.0}}}$ &  
    & - & - & - & 0.7$_{\stddev{\phantom{0.0}}}$ \\
    
    Ent-SpERT & 77.6$_{\stddev{1.0}}$ & 64.0$_{\stddev{0.6}}$ & \textbf{67.3}$_{\stddev{0.6}}$ &  
    & - & 48.1$_{\stddev{0.7}}$ & 41.9$_{\stddev{0.6}}$ & 43.8$_{\stddev{0.5}}$ &  
    & - & 38.1$_{\stddev{1.9}}$ & 29.4$_{\stddev{1.1}}$ & 32.1$_{\stddev{1.2}}$ & \\
    
    SpERT & 78.5$_{\stddev{0.5}}$ & 64.2$_{\stddev{0.4}}$ & \textbf{67.6}$_{\stddev{0.3}}$ &  
    & - & 53.1$_{\stddev{1.2}}$ & 46.0$_{\stddev{1.0}}$ & \textbf{48.2}$_{\stddev{1.1}}$ &  
    & - & 43.0$_{\stddev{1.6}}$ & 33.2$_{\stddev{1.1}}$ & \textbf{36.2}$_{\stddev{1.0}}$ & \\
    
    PURE & 78.0$_{\stddev{0.5}}$ & 63.8$_{\stddev{0.6}}$ & \textbf{67.2}$_{\stddev{0.4}}$& & - & 54.0$_{\stddev{0.7}}$ & 44.8$_{\stddev{0.4}}$ & \textbf{47.6}$_{\stddev{0.3}}$ & 
    & - & 42.2$_{\stddev{0.7}}$ & 32.6$_{\stddev{0.7}}$ & \textbf{35.6}$_{\stddev{0.6}}$ & \\
    
    \bottomrule
    
    \end{tabularx}
    
    \end{adjustwidth*}
    \end{table}

\subsection{Partitioning by Lexical Overlap}

Following \citep{Augenstein2017GeneralisationAnalysis}, we partition the entity mentions in the test set based on lexical overlap with the training set.
We distinguish \textit{Seen} and \textit{Unseen} mentions and also extend this partition to relations.
We denote a relation as an \textit{Exact Match} if the same (head, predicate, tail) triple appears in the train set; as a \textit{Partial Match} if one of its arguments appears in the same position in a training relation of same type; and as \textit{New} otherwise.

We implement a naive \textbf{Retention Heuristic} that tags an entity mention or a relation exactly present in the training set with its majority label.
We report micro-averaged Precision, Recall and F1 scores for both NER and RE in \autoref{table:05:overlap}.

An entity mention is considered correct if both its boundaries and type have been correctly predicted.
For RE, we report scores in the \textbf{Boundaries} and \textbf{Strict} settings \citep{bekoulis-etal-2018-adversarial, Taille2020LetsExtraction}.
In the Boundaries setting, a relation is correct if its type is correct and the boundaries of its arguments are correct, without considering the detection of their types.
The Strict setting adds the requirement that the entity type of both argument is correct.

\subsubsection{Dataset Specificities}
We first observe very different statistics of Mention and Relation Lexical Overlap in the three datasets, which can be explained by the singularities of their entities and relations.
In CoNLL04, mentions are mainly Named Entities denoted with proper names while in ACE05 the surface forms are very often common names or even pronouns, which explains the occurrence of training entity mentions such as "it", "which", "people" in test examples.
This also leads to a weaker entity label consistency \citep{fu-etal-2020-interpretable}: "it" is labeled with every possible entity type and appears mostly unlabeled  whereas a mention such as "President Kennedy" is always labeled as a person in CoNLL04.
Similarly, mentions in SciERC are common names which can be tagged with different labels and they can also be nested.
Both the poor label consistency as well as the nested nature of entities hurt the performance of the retention heuristic.

For RE, while SciERC has almost no exact overlap between test and train relations, ACE05 and CoNLL04 have similar levels of exact match.
The larger proportion of partial match in ACE05 is explained by the pronouns that are more likely to co-occur in several instances.
The difference in performance of the heuristic is also explained by a poor relation label consistency.

\subsubsection{Lexical Overlap Bias}
As expected, this first evaluation setting enables to expose an important lexical overlap bias, already discussed in NER, in end-to-end Relation Extraction.
On every dataset and for every model micro F1 scores are the highest for Exact Match relations, then Partial Match and finally totally unseen relations.
This is a first confirmation that retention plays an important role in the measured overall performance of end-to-end RE models.

\subsubsection{Model Comparisons}
While we cannot evaluate TABTO on SciERC because it is unfit for extraction of nested entities, we can notice different hierarchies of models on every dataset suggesting that there is no one-size-fits-all best model, at least in current evaluation settings.

The most obvious comparison is between SpERT and Ent-SpERT where the explicit representation of context is ablated.
This results in a loss of performance on the RE part and especially on partially matching or new relations for which the entity representations pairs have not been seen.
Ent-SpERT is particularly effective on Exact Matches on CoNLL04, suggesting its retention capability.

Other comparisons are more difficult, given the numerous variations between the very structure of each model as well as training procedures.
However, the PURE pipeline setting seems to only be more effective on ACE05 where its NER performance is significantly better, probably because learning a separate NER and RE encoder enables to learn and capture more specific information for each distinctive task.
Even then, TABTO yields better Boundaries performance only penalized on the Strict setting by entity types confusions.
On the contrary, on CoNLL04, TABTO significantly outperforms its counterparts, especially on unseen relations.
This indicates that it proposes a more effective incorporation of contextual information in this case where relation and argument types are mapped bijectively.

On SciERC, performance of all models is already compromised at the NER level before the RE step, which makes further distinction between model performance even more difficult.

\begin{figure}[h!]
    \centering
    \caption[Example of sentence where the relation head and tails are swapped.]{Example of sentence where the relation head and tails are swapped. The Triple (John Wilkes Booth, Kill, President Lincoln) is present in the training set and the retention behaviours lead models to extract this triple when probed with the swapped sentence expressing the reverse relation.}
    \label{fig:05:swapped_ex}
    
    \includegraphics[width=\textwidth]{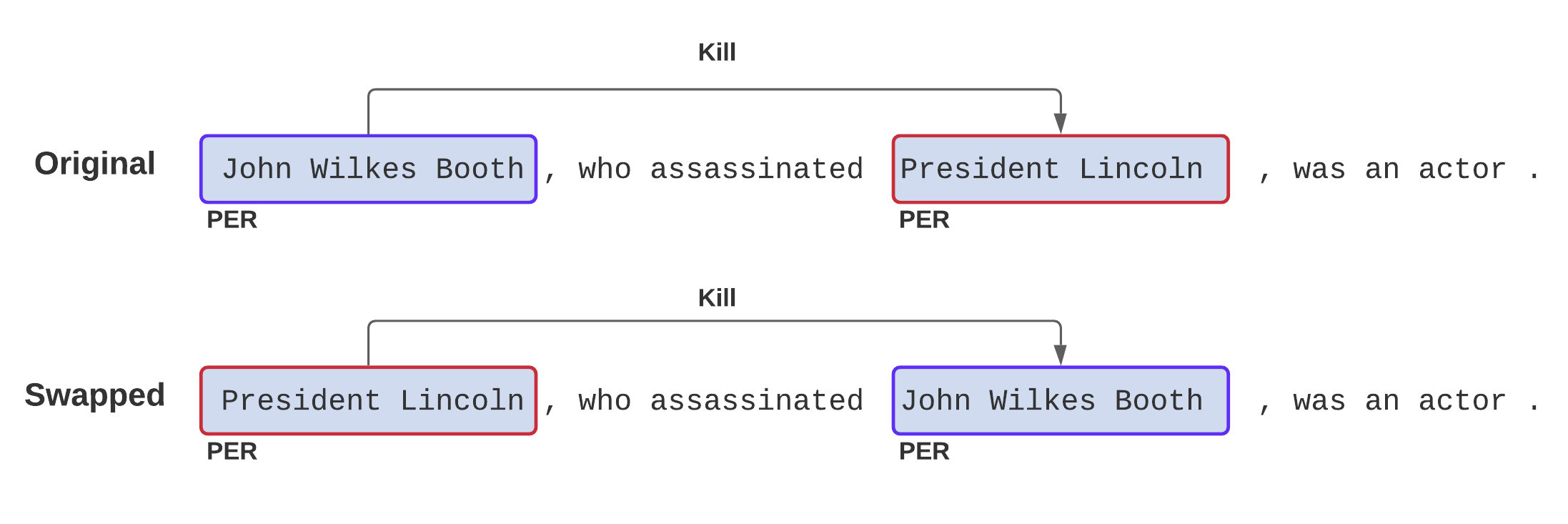}
\end{figure}

    \begin{table}[t]

    \caption[Performance on CoNLL04 test set containing exactly one relation of the corresponding type in its original form (O) and where the relation head and tail are swapped (S).]{Performance on CoNLL04 test set containing exactly one relation of the corresponding type in its original form (O) and where the relation head and tail are swapped (S). NER F1 score is micro-averaged while strict RE score only takes these relations into account. The revRE score corresponds to unwanted extraction of the reverse relation, symptomatic of the retention effect in the swapped setting.}
    \label{table:swap_rev}
    
    \centering
    \small
    
    \begin{tabularx}{\columnwidth}{@{}l*{2}{Y}r*{2}{Y}r*{2}{Y}c@{}}
    
    \toprule
     
    \multicolumn{1}{c}{\multirow{2}{*}{F1}} & \multicolumn{2}{c}{NER $\uparrow$} &  & \multicolumn{2}{c}{RE $\uparrow$} & & \multicolumn{2}{c}{revRE $\downarrow$} \\
      
     \cline{2-3} \cline{5-6} \cline{8-9}
     
     & O & S & & O & S & & O & S &\\
    
    \midrule
    \multicolumn{10}{c}{Kill} \\
    \midrule
     
    Ent-SpERT & 91.6 & 91.7 &  & 85.1 & 35.4 &  & - & 58.5 & \\

    SpERT & 91.4 & \textbf{92.6} &  & 86.2 & 35.0 &  & - & 57.8 & \\
    
    TABTO & \textbf{92.0} & \textbf{92.8} &  & \textbf{89.6} & 27.6 &  & - & 59.5 & \\
    
    PURE & 90.5 & 90.7 &  & 84.1 & \textbf{52.3} &  & - & \textbf{14.3} & \\

    \midrule
    \multicolumn{10}{c}{Located in} \\
    \midrule
    
    Ent-SpERT & \textbf{90.0} & 87.0 &  & 78.3 & 30.3 &  & - & 24.8 & \\
    
    SpERT & 88.6 & 87.7 &  & 75.0 & 24.9 &  & - & 33.5 & \\
    
    TABTO & \textbf{90.1} & \textbf{88.9} &  & \textbf{85.3} & 36.1 &  & - & 34.9 & \\
    
    PURE & 89.0 & 83.7 &  & 81.2 & \textbf{59.3} &  & - & \textbf{5.1} & \\
    \bottomrule
    
    \end{tabularx}
    \end{table}

\subsection{Swapping Relation Heads and Tails}

\newcolumntype{C}[1]{>{\arraybackslash}p{#1}}

    \begin{table}[htp!]
    \begin{adjustwidth*}{}{-.3\textwidth}
    
    \caption[Qualitative examples on swapped sentences for the ``Kill'' relation on CoNLL04]{Some qualitative examples of models' predictions on original (left column) and swapped (right) CoNLL04 sentences for the ``Kill'' relation. Despite a perfect Relation Extraction in the original sentences for all models, swapping head and tails results in several types of errors mainly regarding the direction of the relation. Predictions of incorrect original triples are in red. These examples are obtained from models trained with the same seed (s=0).}
    \label{table:qualitative_kill}
    
    \begin{tabularx}{1.3\textwidth}{@{}llYY@{}}
    
    \toprule
     \textbf{1} & & \multicolumn{2}{C{\textwidth}}{The Warren Commission determined that on Nov. 22 , 1963 , \textbf{A} fired a high-powered rifle at \textbf{B} 's motorcade from the sixth floor of what is now the Dallas County Administration Building , where he worked .} \\
    \midrule
      A, B &  & Lee Harvey Oswald, Kennedy  & Kennedy, Lee Harvey Oswald\\
     \midrule
    Ent-SpERT & & (A,B) & \textcolor{OrangeRed}{(B,A)} \\
    SpERT & & (A,B) & \textcolor{OrangeRed}{(B,A)} \\
    TABTO & & (A,B) & \textcolor{OrangeRed}{(B,A)} \\
    PURE & & (A,B) & \textcolor{OrangeRed}{(B,A)} \\
    \midrule
    
    \midrule
     \textbf{2} & & \multicolumn{2}{C{\textwidth}}{Today 's Highlight in History : Twenty years ago , on June 6 , 1968 , at 1 : 44 a.m. local time , \textbf{B} died at Good Samaritan Hospital in Los Angeles , 25 -LCB- hours after he was shot at the Ambassador Hotel by \textbf{A} .} \\
    \midrule
     A, B &  & Sirhan Bishara Sirhan, Sen. Robert F. Kennedy  &  Sen. Robert F. Kennedy, Sirhan Bishara Sirhan \\
     \midrule
     
    Ent-SpERT & & (A,B) & \textcolor{OrangeRed}{(B,A)} \\
    SpERT & & (A,B) & \textcolor{OrangeRed}{(B,A)} \\
    TABTO & & (A,B) & \textcolor{OrangeRed}{(B,A)} \\
    PURE & & (A,B) & - \\
    \midrule
    
    \midrule
     \textbf{3} & & \multicolumn{2}{C{\textwidth}}{In 1968 , authorities announced the capture in London of \textbf{A} , suspected of the assassination of civil rights leader \textbf{B} .} \\
    \midrule
     A, B &  & James Earl Ray, Dr. Martin Luther King Jr  & Dr. Martin Luther King Jr, James Earl Ray \\
     \midrule
     
    Ent-SpERT & & (A,B) & (A,B) \textcolor{OrangeRed}{(B,A)} \\
    SpERT & & (A,B) & (A,B) \textcolor{OrangeRed}{(B,A)} \\
    TABTO & & (A,B) & (A,B) \\
    PURE & & (A,B) & (A,B) \\
    \midrule
    
    \midrule
     \textbf{4} & & \multicolumn{2}{C{\textwidth}}{The Warren Commission determined that \textbf{A} fired at \textbf{B} from the sixth floor of what is now the Dallas County Administration Building .} \\
    \midrule
     A, B &  & Oswald, Kennedy  & Kennedy, Oswald \\
     \midrule
     
    Ent-SpERT & & (A,B) & - \\
    SpERT & & (A,B) & (A,B) \textcolor{OrangeRed}{(B,A)} \\
    TABTO & & (A,B) & \textcolor{OrangeRed}{(B,A)} \\
    PURE & & (A,B) & (A,B) \\
    
    \bottomrule
    
    \end{tabularx}
    \end{adjustwidth*}
    \end{table}

\newcolumntype{C}[1]{>{\arraybackslash}p{#1}}

    \begin{table}[htp!]
    \begin{adjustwidth*}{}{-.3\textwidth}
    \caption[Qualitative examples on swapped sentences for the ``Located in'' relation on CoNLL04]{Some qualitative examples of models' predictions on original (left column) and swapped (right) CoNLL04 sentences for the ``Located in'' relation. This relation is often simply expressed by an apposition of the head and tail separated by a comma. Predictions of incorrect original triples are in red. These examples are obtained from models trained with the same seed ($s=0$).}
    
    \label{table:qualitative_loc}
    
    
    \begin{tabularx}{1.3\textwidth}{@{}llYY@{}}
    
    \toprule
     \textbf{1} & & \multicolumn{2}{C{\textwidth}}{
Reagan recalled that on the 40th anniversary of the Normandy landings he read a letter from a young woman whose late father had fought at \textbf{A} , a \textbf{B} sector .} \\
    \midrule
      A, B & &  Omaha Beach, Normandy  & Normandy, Omaha Beach\\
    \midrule
    Ent-SpERT & & (A,B) & - \\
    SpERT & & (A,B) & - \\
    TABTO & & (A,B) & - \\
    PURE & & (A,B) & (A,B) \\
    \midrule
    
    \midrule
     \textbf{2} & & \multicolumn{2}{C{\textwidth}}{\textbf{A} , \textbf{B} ( AP )} \\
    \midrule
     A, B &  & MILAN, Italy & Italy, MILAN \\
    \midrule
    
    Ent-SpERT & & (A,B) & (A,B) \\
    SpERT & & (A,B) & (A,B) \\
    TABTO & & (A,B) & \textcolor{red}{(B,A)} \\
    PURE & & (A,B) & - \\
    \midrule
    
    \midrule
    \textbf{3} & & \multicolumn{2}{C{\textwidth}}{In \textbf{A} , downed tree limbs interrupted power in parts of \textbf{B} .} \\
    \midrule
     A, B &  & Indianapolis, Indiana & Indiana, Indianapolis \\
    \midrule
    
    Ent-SpERT & & (A,B) & \textcolor{red}{(B,A)} \\
    SpERT & & (A,B) & \textcolor{red}{(B,A)} \\
    TABTO & & (A,B) & \textcolor{red}{(B,A)} \\
    PURE & & (A,B) & \textcolor{red}{(B,A)} \\
    \midrule
    
    \midrule
    \textbf{4} & & \multicolumn{2}{C{\textwidth}}{The plane , owned by Bradley First Air , of \textbf{A} , \textbf{B} , was carrying cargo to Montreal for Emery Air Freight Corp. , an air freight courier service with a hub at the Dayton airport .} \\
    \midrule
     A, B &  & Ottawa, Canada & Canada, Ottawa \\
    \midrule
    
    Ent-SpERT & & (A,B) (Dayton airport, Canada) & (Dayton airport, Ottawa) \\
    SpERT & & (A,B) (Dayton airport, Canada) & - \\
    TABTO & & (A,B) & (A,B) \\
    PURE & & (A,B) & (A,B) \\
    
    \bottomrule
    
    \end{tabularx}
    \end{adjustwidth*}
    \end{table}

A second experiment to validate that retention is used as a heuristic in models' predictions is to modify their input sentences in a controlled manner similarly to what is proposed in \citep{ribeiro-etal-2020-beyond}.
We propose a very focused experiment that consists in selecting asymmetric relations that occur between entities of same type and swap the head with the tail in the input.
If the model predicts the original triple, then it over relies on the retention heuristic, whereas finding the swapped triple is an evidence of broader context incorporation. We show an example in \autoref{fig:05:swapped_ex}.

Because of the requirements of this experiment, we have to limit to two relations in CoNLL04: ``Kill'' between people and ``Located in'' between locations.
Indeed, CoNLL04 is the only dataset with a bijective mapping between the type of a relation and the types of its arguments and the consistent proper nouns mentions makes the swaps mostly grammatically correct. 
For each relation type, we only consider sentences with exactly one instance of corresponding relation and swap its arguments. 
We only consider this relation in the RE scores reported in \autoref{table:swap_rev}.
We use the strict RE score as well as \textbf{revRE} which measures the extraction of the reverse relation, not expressed in the sentence.

For each relation, the hierarchy of models corresponds to the overall CoNLL04. 
Swapping arguments has a limited effect on NER, mostly for the "Located in" relation.
However, it leads to a drop in RE for every model and the revRE score indicates that SpERT and TABTO predict the reverse relation more often than the newly expressed one.
This is another proof of the retention heuristic of end-to-end models, although it might also be attributed to the language model to the language model. In particular for the ''Located in`` relation, swapped heads and tails are not exactly equivalent since the former are mainly cities and the latter countries.

On the contrary, the PURE model is less prone to information retention, as shown by its revRE scores significantly smaller than the standard RE scores on swapped sentences. Hence, it outperforms SpERT and TABTO on swapped sentences despite being the least effective on the original dataset.The important discrepancy in results can be explained by the different types of representations used by these models. The pipeline approach allows the use of argument type representations in the Relation Classifier whereas most end-to-end models use lexical features in a shared entity representation used for both NER and RE.

These conclusions from quantitative results are validated qualitatively.
We can observe that the four predominant patterns are intuitive behaviours on sentences with swapped relations: retention of the incorrect original triple, prediction of the correct swapped triple and prediction of none or both triples.
We report some examples in \autoref{table:qualitative_kill} and \autoref{table:qualitative_loc}.

\subsection{Conclusion}
In this work, we study three state-of-the-art end-to-end Relation Extraction models in order to highlight their tendency to retain seen relations.
We confirm that retention of seen mentions and relations play an important role in overall RE performance and can explain the relatively higher scores on CoNLL04 and ACE05 compared to SciERC.
Furthermore, our experiment on swapping relation heads and tails tends to show that the intermediate manipulation of type representations instead of lexical features enabled in the pipeline PURE model makes it less prone to over-rely on retention.

While the contained extend of our swapping experiment is an obvious limitation of this work, it shows limitations of both current benchmarks and models. It is an encouragement to propose new benchmarks that might be easily modified by design to probe such lexical overlap heuristics.
Contextual information could for example be contained in templates of that would be filled with different (head, tail) pairs either seen or unseen during training.

Furthermore, pretrained Language Models can already capture relational information between phrases \citep{petroni-etal-2019-language} and further experiments could help distinguish their role in the retention behaviour of RE models. This is especially true regarding the ``Located in'' relation in our experiment that mainly holds between cities and countries.  

\section{Conclusion and Perspectives}
In this chapter, we identify shortcomings in the evaluation of End-to-end Relation Extraction which are obstacles to drawing interesting conclusions useful to develop better models.

First, we point that the several evaluation settings previously used in ERE have led to incorrect comparisons and inconsistent claims in several previous works.
We also claim that it is a major hurdle in the apprehension of the ERE literature and call for a unified evaluation setting reporting RE scores from both Strict and Boundaries settings to enable more meaningful cross-dataset discussions.

Second, we extend our study of the impact of lexical overlap in Named-Entity Recognition to ERE and show that current models tend to simply memorize training triples and that this behaviour can be sufficient to obtain decent performance on common benchmarks such as CoNLL04.
Furthermore, our behavioral testing experiment suggests that a pipeline model that handles intermediary argument type representations is less prone to over-rely on this retention heuristic, although it is still exposed to it.

This study suggests that incorporation of contextual information into relation prediction is a key factor of generalization to new facts that can be overlooked by a single F1 score on standard benchmarks such as CoNLL04 and ACE05.
Hence, it appears that designing benchmarks able to measure this generalization capability is an integral part of the future development of models able to detect new facts.

In the following chapter, we propose a description of a work-in-progress attempt to incorporate such contextual information in a more efficient and interpretable manner, inspired by recent work in BERTology.
\chapter[Towards Explicit Use of Self-Attention Weights in ERE]{Towards Explicit Use of Self-Attention Weights in End-to-end Relation Extraction}
\label{chapter:06:attention}

As discussed in previous chapters, End-to-end Relation Extraction models are exposed to simple lexical overlap heuristics.
It appears that these retention heuristics come from the tendency of models to over-rely on the exact head and tail argument instances, whether they have been seen during training as a labeled relation triple or in the same sentence during Language Model pretraining.
Because the latter lexical overlap with pretraining corpora is arguably very important and only increasing with the ever growing number of web resources used to train ever larger Language Models, it seems important to design models relying less on relation arguments representations and more on their context which is often the key indicator of a relation.

In this chapter, we suggest new architectural ideas that could be of use to better incorporate contextual information in ERE by relying on syntactic knowledge encoded in BERT's attention heads during pretraining.
We limit to the description of these architectures that are still under development.
In particular, we do not report any quantitative or qualitative result in the absence of an exhaustive experimental campaign at the time of writing this thesis, notably regarding the experimental setting we introduced in the previous chapter.

\section{Motivations}
\label{sec:06:motivations}

The recent adoption of the Transformer architecture \citep{Vaswani2017AttentionNeed} as a fundamental tool to obtain state-of-the-art performance in Natural Language Processing suggests that self-attention is an efficient means of incorporating contextual information in individual word representations. 
Another interesting property of attention in general is its interpretability through the weighted selection of parts of the input.  

Before the introduction of BERT, \citet{strubell-etal-2018-linguistically} propose to incorporate linguistic knowledge in the self-attention of a Transformer to improve its performance on Semantic Role Labeling (SRL) in a Multitask Learning setting including Part-of-Speech tagging, Dependency Parsing and Predicate Detection.
They train one attention head to attend to each token's syntactic parent in a Dependency Tree so that its activations can be used by the network as an oracle syntactic structure.

Complementarily to this work which shows that incorporating linguistic knowledge in Transformer self-attention can help improve its performance, some BERTology papers \citep{clark-etal-2019-bert, jawahar-etal-2019-bert, kovaleva-etal-2019-revealing} analyze BERT's self-attention heads patterns and show that after BERT pretraining, some attention heads present linguistically relevant structures close to Dependency Trees.
In particular, \citet{clark-etal-2019-bert} and \citet{jawahar-etal-2019-bert} suggest that after BERT pretraining, some heads correspond to specific syntactic dependency relations between words for example linking a verb to its object or an adjective to the noun it modifies.
These experiments suggest that the Transfer Learning ability of BERT might be explained by these syntactic patterns learned by its attention heads during unsupervised Language Model pretraining. 

Furthermore, several previous works \citep{culotta-sorensen-2004-dependency, miwa-bansal-2016-end} explicitly use Dependency Tree information as input to improve Relation Extraction performance.
In particular, they validated the intuition that the shortest path between the two candidate arguments head words in the Dependency Tree often contains words crucial in the prediction of their relationship.

Motivated by these studies, we propose to explore two architectures to directly incorporate BERT's self-attention patterns into end-to-end Relation Extraction prediction.
They share a similar idea, finetuning a network which takes as input the values of BERT's self-attention heads at every layer.
Indeed, contrary to \citet{strubell-etal-2018-linguistically}, we do not explicitly rely on an additional syntactic parent prediction auxiliary task that would require another source of supervision.
Instead, we hypothesize that we can use the linguistic patterns learned by BERT on a large unlabeled corpus.
Furthermore, because of this pretraining, we cannot supervise a separate attention head for each relation without first selecting the most relevant heads.
Hence we prefer using an approach closer to probing networks: instead of supervising the attention heads so that they directly reflect semantic relations (which is a very hard constraint), we use them as input features for an additional network.
Nevertheless, contrary to probing networks where BERT is frozen in order to study its behaviour, we finetune BERT so that its weights are optimized to handle our End-to-end RE goal.

\section[Supervising Attention Heads as Indicators of Relations]{Supervising Self-attention Heads as Indicators of Relations}
Self-attention can be interpreted as an alignment between each element in its input and every one of them.
Hence, while its input and output are sequence of representations, this alignment scores model relations held between every pair of element in the input sequence.

This view makes it natural to consider attention heads as candidate representations to use for modeling relations between tokens while individual tokens are represented by the hidden representations $h_i^l$. To our knowledge, only syntactic relations have been explored by previous works such as \citep{strubell-etal-2018-linguistically, clark-etal-2019-bert} and our proposal is to explore if and how these attention heads can be used as indicators of semantic relations in an End-to-end Relation Extraction setting.

\subsection{Supervising Attention Heads for Direct Relation Modeling}
We first attempted to follow \citet{strubell-etal-2018-linguistically} who supervise one attention head in a Transformer so that its activation directly predicts a syntactic parent dependencies.
We trained a small randomly initialized Transformer (1 to 3 layers) in which we supervised an attention head for each relation type in a chosen layer in addition to unconstrained heads.

\textbf{Contrary to syntactic parent prediction where each word has exactly one parent, the majority of words are not involved in a semantic relation and a word can be involved in more than one relation.}
To take into account this key difference, we replace the softmax layer, which tends to always select exactly one token aligned for every token, with a\textbf{ sigmoid layer} that enables more flexibility.
Unfortunately, we failed to obtain interesting results on standard benchmarks such as CoNLL04 or ACE05 with such an architecture also relying on a large set of sensitive hyperparameters (number of layers, heads, dimensions, layer in which Relation Extraction heads are supervised).
Note that in this case as in \citep{strubell-etal-2018-linguistically}, the attention weights are directly used for prediction without any projection.

We can hypothesize that training such a network failed because of the hard constraint put on a numerous proportion of attention heads (6 or 7 different relation types over 12 attention heads) but also mainly on a lack of supervision data since CoNLL04 and ACE05 respectively contain 2k and 7k relation instances and this supervision is sparse since the majority of tokens are not involved in a relation.
Hence it seemed useful to take Language Model pretraining as an initialization of attention heads that has been shown to encodes some syntactic information \citep{clark-etal-2019-bert}.

\subsection{Finetuning BERT's Attention Heads}

\begin{figure}[h!]
    \centering
    \caption[Illustration of attention weights in BERT.]{Illustration of attention weights in BERT. Usually, only the last hidden layer's outputs $h_i^L$ are used as input to a classifier and the whole network is finetuned for the final task at hand. We propose to use the attention weights as additional features for Relation Extraction.}
    \label{06:fig:bert_attn}
    
    \includegraphics[width=\textwidth]{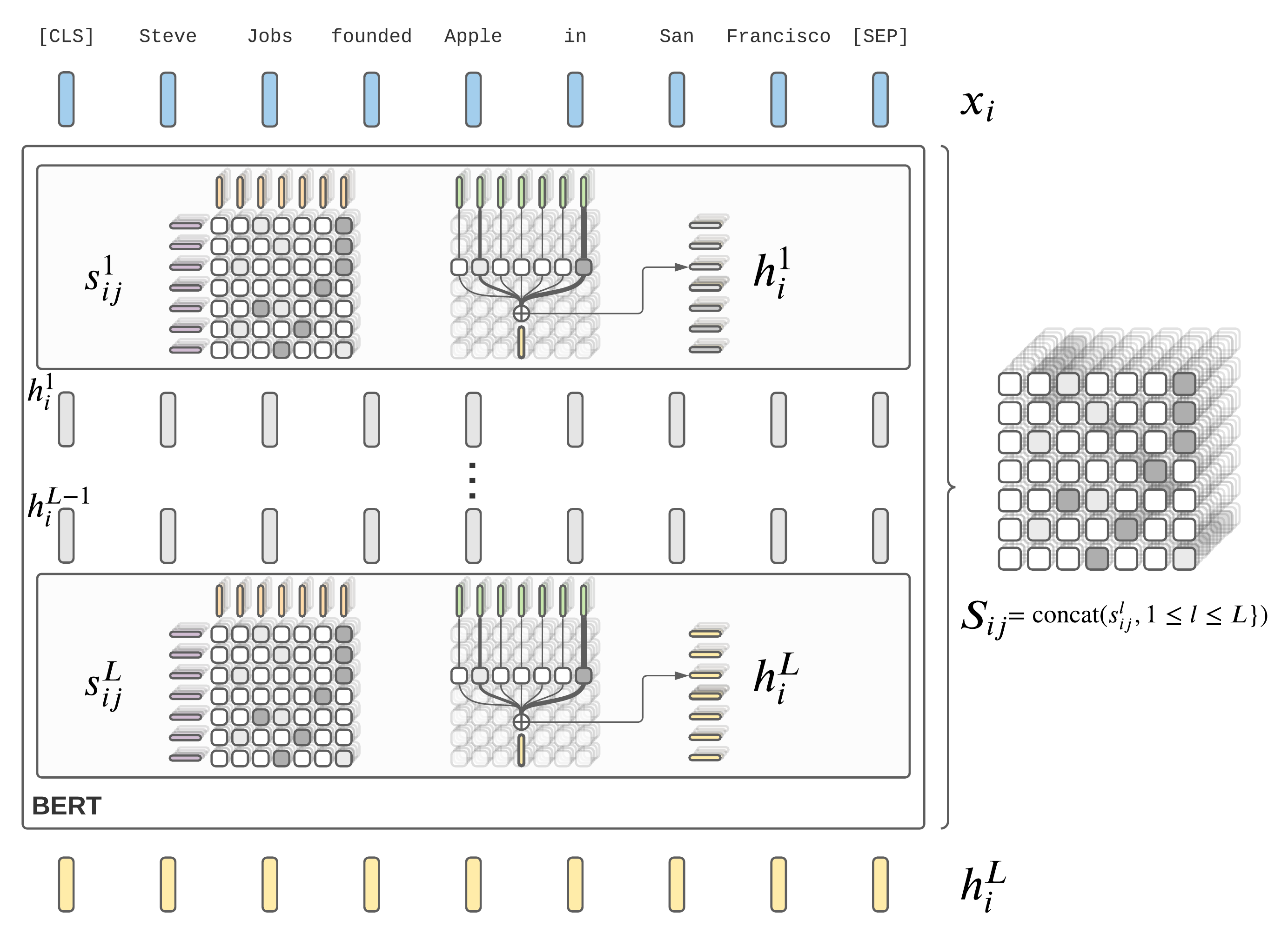}
\end{figure}

To relax the important constraint that attention scores should directly reflect semantic relations between tokens and to leverage unsupervised Language Model pretraining, a natural idea is to use these attention heads as input to a simple network to directly predict relations as modeled by Table Filling.
This is similar to the now standard LM pretraining - finetuning approach except that the RE supervision is made at the level of attention heads: the activations of every attention head at every layer are concatenated and used as a representation of the interaction between every pair of input tokens.
This can be done with a limited time and memory extra cost since these attention scores are already computed in every pass in a Transformer model.
The main additional cost comes from additional dependencies that must be stored during backpropagation in the training phase.

We propose to gather every attention score from the Transformer Language's self-attention layer in a single
input vector representing relations between words.
In order to reduce constraints on the architecture, we concatenate the unnormalized scores $s_{ij}^l$ rather than the softmax normalized attention scores $\alpha_{ij}^l$ that are directly used by the Transformer.

An important point when dealing with BERT's attention weights is how subword tokenization is treated to obtain word-level interpretations.
However, these aggregation details are often omitted from articles and among previously cited papers we only found it discussed in \citep{clark-etal-2019-bert}.
To convert token-to-token weights to word-to-word weights, they distinguish attention \textit{from} a split-up word from attention \textit{to} a split-up word.
Indeed, attention weights from a word must sum to 1 which encourages them to sum the attention weights of the subtokens for attention \textit{to} a split-up word.
On the contrary, they average the weights for attention \textit{from} a split-up word.
Because we do not use normalized scores, we simply obtain word-to-word attention representations by averaging the attention weights of split-up words in both directions.

Another detail is that syntactic patterns in attention heads exposed by \citet{clark-etal-2019-bert} are directed, for example the direct objects of a verb ``attend'' to their verb in given heads.
In order to enable more expressivity, we propose to concatenate the attention scores $s_{ij}^l$ with their transpose.

\begin{equation}
    S_{ij} = \mathrm{concat}(\{s_{ij}^l, s_{ji}^l | 1 \leq l \leq L\}) 
\end{equation}

\begin{figure}[h!]
    \centering
    \caption[Illustration of the First Order Attention architecture.]{Illustration of the First Order Attention architecture. We propose to use attention scores of every layer in BERT as input features for a Table Filling End-to-end Relation Extraction classifier.}
    \label{fig:06:foa}
    
    \includegraphics[width=\textwidth]{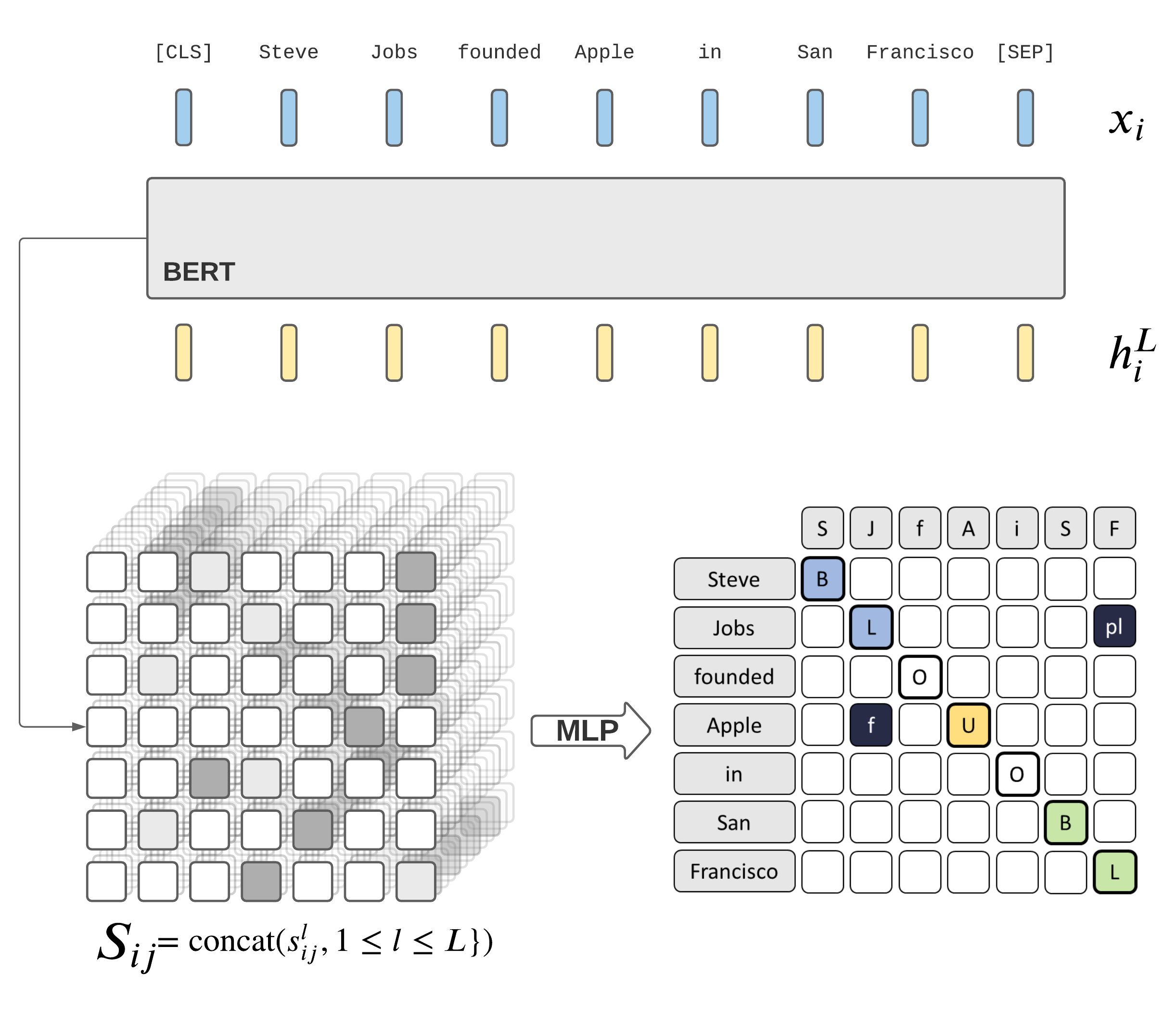}
\end{figure}

Our first architectural proposal simply consists in using the obtained $S_{ij}$ vectors as input representations of pairs of words ($i,j$) to use in a Table Filling classifier.
Following the general trend, this classifier could be as simple as a 1-hidden layer Multilayer Perceptron.
Several strategies can be employed for the NER input representations in sequence labeling : using the traditional $h_i^L$, the newly obtained $S_{ii}$ or a concatenation of both.
We call this approach First Order Attention and propose an illustration in \autoref{fig:06:foa}.

\section{Second Order Attention}

\begin{figure}[htp!]
\begin{adjustwidth*}{}{-.1\textwidth}
    \caption[Illustration of the Second Order Attention architecture.]{Illustration of the Second Order Attention architecture. BERT's attention scores are used to compute an alignment between every pair of candidate arguments $(i,j)$ and every word $k$. A candidate argument pair is then represented by a weighted sum over the $h_k$ representations of every word in the sentence and not with the individual argument representations $h_i$ and $h_j$.
    This could help focus more on the expression of the predicate that is a more general indicator of a relation than its specific arguments.}
    \label{fig:06:soa}
    
    \centering
    \includegraphics[width=1.1\textwidth]{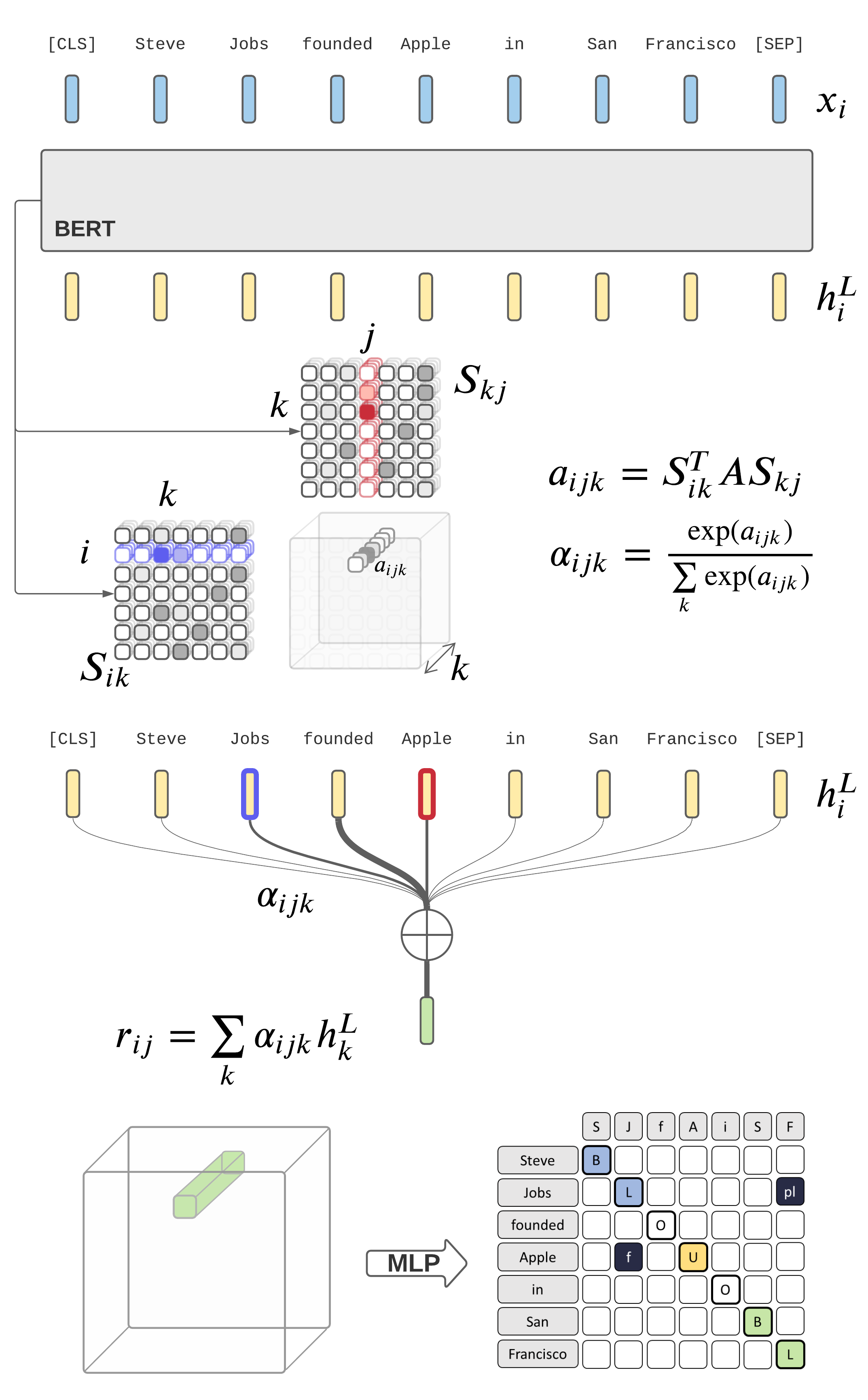}
\end{adjustwidth*}
\end{figure}

While new in its spirit, the previously described finetuning of attention heads is still exposed to one of the main drawbacks of neural end-to-end RE systems: retention of seen facts.
Indeed, \textbf{we can view attention scores as modeling a bilinear interaction between tokens}.
Hence, while this term is complementary to the linear combination classicaly used when concatenating candidate arguments representations, in both cases, context is only implicitly incorporated by contextual embeddings in the representations of candidate arguments.
Yet, better relying on context seems the only way to reduce the effect of simple memorization of argument pairs and better generalize to the detection of unseen facts.

A classical method to explicitly incorporate contextual information is to use a piecewise approach, splitting the document at the candidate arguments and pooling the representations of the words inside the three obtained subsentences.
The middle context in particular is believed to more often contain key indicators of the predicate in English which leads a model such as SpERT \citep{Eberts2020Span-basedPre-training} to only use this piece. 
However, while pooling word representation is known as a strong baseline to obtain effective sentence representations \citep{Arora2017AEmbeddings}, piecewise pooling seems suboptimal.
First because the predicate is not always expressed between the arguments of the relations but mostly because it boils down to lose the structure of the sentence in a way similar to a bag of words representation.
In particular when the context between arguments is long, and possibly containing other entity mentions, such type of representations intuitively rely on the presence of trigger words indicative of a relation and are not adapted to associate a relation head with the correct tail as described by \citet{rosenman-etal-2020-exposing}.

On the contrary, several works anterior to Language Model pretraining propose to use syntactic information from Dependency Parsing for Relation Extraction with ground truth features \citep{miwa-bansal-2016-end} or the predictions of a pretrained model \citep{zhang-etal-2017-end}.
The hypothesis is that the predicate of a relation is often expressed in the shortest path between the arguments, more precisely by a verb whose subject and object are the arguments of a relation.
Having access to this syntactic structure thus seems relevant for better association between arguments and to reduce the dependency on shallow heuristics such as relying on trigger words or memorization of seen triples.
Yet, as described in \autoref{sec:02:bertology}, \citet{clark-etal-2019-bert} suggest that some of BERT's attention heads learn such types of syntactic relations. 

\subsection{Model}

The core idea of our model is similar to attention : weighting representations of context words in order to focus on the most relevant ones.
However, contrary to self-attention where this weighting is computed by the interaction of every pair of words, here we would like an interaction between every triple of words in order to model the (head, predicate, tail) structure.
Our intuition is that in order to prevent the model to overfit on the exact head and tail representations it should be explicitly given access to a representation of the words expressing the predicate. 
Hence, an attention mechanism could learn to focus on this most relevant part of the context.
However, we also want to use the syntactic knowledge learnt during LM pretraining and not simply learn pairs of keys and queries from scratch on small labeled datasets.

The idea of this model that we name \textbf{Second Order Attention} (SOA) is thus to use the self-attention weights already computed in a BERT pass and use them as indicator of compatible paths containing one word linking a candidate head to a candidate tail.
This architecture proposal is illustrated in \autoref{fig:06:soa}.

More formally, given a candidate head token with index $i$ and a candidate tail token $j$, we compute an attention score $\alpha_{ijk}$ that aims at reflecting the compatibility of every token $k$ as being a relevant indicator of a relation between $i$ and $j$.
This weight is used to obtain a representation vector for each pair $ij$ by weighting BERT's last layer hidden states $h_k^L$ as traditional token representations.
\begin{equation}
    r_{ij} = \sum \limits_k \alpha_{ijk} h_k^L
\end{equation}

Similarly to our First Order Attention model, we concatenate BERT attention scores at every layer before softmax activation $s_{ij}^l$ along with their transpose to allow syntactic dependencies to be modeled in both directions.

\begin{equation}
    S_{ij} = \mathrm{concat}(\{s_{ij}^l, s_{ji}^l | 1 \leq l \leq L\}) 
\end{equation}

The $\alpha_{ijk}$ are obtained with a bilinear function of the attention scores $S_{ik}$, between the candidate head $i$ and a context token $k$, and $S_{kj}$, between the same context token and the candidate tail $j$. These scores are normalized with the softmax function.

\begin{equation}
    a_{ijk} = \frac{\mathrm{exp}(S_{ik}^T A S_{kj})}{\sum\limits_k \mathrm{exp}(S_{ik}^T A S_{kj})}
\end{equation}

The bilinear mapping is preferred to a simple dot product in order to enable the model to learn to associate different heads. Intuitively, this enables to associate the head modeling the syntactic relation Subject-Verb with the one modeling the Verb-Object dependency.

In order to enable this mechanism to attend to several dependencies, perhaps relevant for different relation types we can envision an extension with multiple SOA heads that could learn to focus on complementary syntactic patterns.

A major limitation of this model is its memory complexity in $O(n^3)$ where $n$ is the number of tokens during the computation of the $\alpha_{ijk}$ and $r_{ijk}$.
A first strategy to deal with this is to project the attention scores heads, e.g. of dimension 144 in BERT$_{BASE}$, to a lower dimension to reduce the complexity of the bilinear product.
We can also switch the Table Filling modeling of End-to-end Relation Extraction to Entity Filtering or even the pipeline setting to reduce the number of candidate pairs $(i,j)$ to consider.

\subsection{Future Experiments}

We led preliminary experiments on CoNLL04 and ACE05 with traditional global metrics that served for the development of the SOA model.
Experiments with several combinations of traditional argument representations $[h_i, h_j]$, FOA attention scores $S_{ij}$ and SOA contextual representations $r_{ij}$ suggest that FOA and SOA only bring marginal quantitative gains if any on the dev scores of these benchmarks with traditional metrics.
This encouraged us to develop the empirical study presented in \autoref{sec:05:retention} and introduce metrics better able to focus on the extraction of unseen facts.
Unfortunately, this new evaluation setting is yet to be used with our two architecture proposal and could help propose refinements on these models still under development.

In particular, comparing performance of the currently implemented Table Filling approach with Entity Filtering or pipeline structure will be insightful both in terms of quantitative performance and time and memory efficiency.
Furthermore, we believe that SciERC is also a more interesting and challenging benchmark on which the SOA architecture might be even more relevant due to its very low lexical overlap.
Finally, beyond mere quantitative performance, the SOA architecture can offer one or several interpretable weightings of context words for every candidate arguments that could be compared with ground truth syntactic relations.

On a final note, if we managed to find a solution to better tackle the cubic complexity of Second Order Attention, we might then envision to extend it to n-th order attention for which longer paths between arguments could be considered in order to model longer dependency paths, for example including words indicative of a negation.
\chapter{Conclusion}
\label{chapter:07:conclusion}

Throughout this thesis, we have tackled the issue of generalization beyond facts seen during training in deep Named-Entity Recognition and Relation Extraction models.
In this final chapter, we first recapitulate our work (\autoref{sec:07:synopsis}) and summarize our main findings and their value and contribution to the literature (\autoref{sec:07:findings}).
We then discuss the limitations of this work (\autoref{sec:07:limitations}) and provide our perspectives on possible extensions of this research (\autoref{sec:07:future}) and the future of Information Extraction in the Language Models era (\autoref{sec:07:era}). 

\section{Synopsis}
\label{sec:07:synopsis}
In this dissertation, we proposed to analyze how state-of-the-art models based on the recently introduced Language Model pretraining strategy generalize to the detection of unseen entities and relations in End-to-end Relation Extraction.

In \autoref{chapter:02:bert}, we proposed an overview of the different shifts in paradigm involved in the evolution from handcrafted features to learned word representations with static then contextual embeddings. We provided an introduction to the main Deep Learning architectures and to the nowadays inescapable BERT model.

\autoref{chapter:03:ner} presented a first study focused on Named Entity Recognition and the performance of different contextual embeddings models with a focus on generalization to unseen mentions as well as out of domain.

In \autoref{chapter:04:re-taxonomy}, we suggested a taxonomy of the numerous previously proposed End-to-end Relation Extraction models.
We highlighted a triple evolution of models in terms of NER strategy, joint learning modeling and word representations.

In \autoref{chapter:05:rethinking}, we first identified several incorrect comparisons in previous works and proposed a double ablation study of span-level NER and BERT pretraining that we viewed as missing from the literature.
We then proposed to quantify the impact of lexical overlap in End-to-end Relation Extraction and measure the capacity of state-of-the-art models to generalize to unseen mentions and relations.

Finally, \autoref{chapter:06:attention} presented preliminary reflections towards using BERT's attention weights as syntactic information in order to better incorporate contextual information and improve generalization. 

\section{Findings and Contributions}
\label{sec:07:findings}

Our work mainly aimed to provide an evaluation of Named Entity Recognition and End-to-end Relation Extraction models that better reflects generalization to facts unseen during training.
Because of the concurrent introduction of contextual embeddings and because contextualization seemed essential for generalization in Information Extraction, analyzing the impact of Language Model pretraining in Entity and Relation Extraction is also an important part of our study.
In this context, our main findings are the following:

\paragraph{}
\textbf{The effectiveness of contextual embeddings in Named Entity Recognition
can be mainly explained by their improved performance on unseen entities and out-of-domain.} 
This result from \autoref{sec:03:contener} is important in an industrial application context where we can expect that lexical overlap between inference and training is much less important than on academic benchmarks that have a training set often an order of magnitude larger than the test set.
It indicates that the benefits of using contextual embeddings in this context have been underestimated and are worth the additional computational cost.

\paragraph{}
\textbf{Language Model Pretraining and more specifically BERT, is also the main explanation for better End-to-end Relation Extraction performance.}
We can draw this conclusion from our implementation of a double ablation study comparing Span-level to sequence tagging NER and BERT pretraining to a BiLSTM with GloVe embeddings in \autoref{sec:05:sincere}.
Indeed, the ablation of BERT led to important drops in performance while the ablation of Span-level NER had a more limited impact, rather positive.
Hence, despite the multiplication of End-to-end Relation Extraction settings, it is difficult to conclude on the real effectiveness of recently proposed models over previous models for example relying on dependency tree structures \citep{miwa-bansal-2016-end, zhang-etal-2017-end} and which could be combined with contextual word representations.

\paragraph{}
\textbf{Despite the relative effectiveness of contextual embeddings over previous static representations, state-of-the-art Named Entity Recognition and End-to-end Relation Extraction models are still biased towards the detection of seen mentions and relations.}
Indeed, our studies on NER (\autoref{sec:03:contener}) and ERE (\autoref{sec:05:retention}) reveal a lexical overlap bias that makes models more proficient on mentions and relations seen during training. This bias is only encouraged by the high overlap between testing and training entity and relation mentions in traditional benchmarks such as Ontonotes, CoNLL03, CoNLL04 and ACE05.
This implies a reduced proficiency for the key useful application to extract previously unknown facts from textual corpora, for example in Knowledge Base Construction.
Furthermore, in an application perspective, this leads to a representation bias: locations or people with names absent or rare in the training data are more likely to be misdetected during inference.

\paragraph{}
\textbf{The intermediate type representation used in some pipeline models makes them less prone to this retention heuristic.}
Indeed, the comparison of the PURE pipeline model with two end-to-end counterparts in our behavioral study in \autoref{sec:05:retention} indicates that it less often predicts the original triple from swapped sentences.
This intuitive result that should be confirmed with additional experiments provides an interesting indication in the design of architectures better able to generalize beyond the simple memorization of training triples.  

\paragraph{}
Additionally to these findings, we believe that \textbf{our work contributes to making the apprehension of the End-to-end Relation Extraction literature easier}.
First, by proposing a rich taxonomy of previously proposed models in \autoref{chapter:04:re-taxonomy} but mostly by identifying and correcting several erroneous comparisons in previous literature and calling for a cleaner unified evaluation setting in \autoref{sec:05:sincere}.

\section{Limitations}
\label{sec:07:limitations}

Despite the previously enumerated findings and contributions, we can discuss several limitations of our work.

\paragraph{}
\textbf{First, our work is limited in its scope to English corpora mainly representative of the news domain.}
Indeed, whether in NER or ERE, our studies limit to the mainly used datasets which are in one hand CoNLL03 and OntoNotes, in the other hand CoNLL04 and ACE05. Although we also experimented with more recent benchmarks such as WNUT for NER in the Twitter domain and SciERC for ERE on the scientific domain, more languages, domains and benchmarks could be explored to strengthen the scope of our findings.
In particular, due to constraint on argument types, our behavioral study in \autoref{sec:05:retention} is limited to two relations in CoNLL04 which adds up to a few hundreds test samples and it deserves to be extended.

\paragraph{}
\textbf{Second, lexical overlap is not the only linguistic phenomenon that impacts the performance of NER or ERE models.}
We chose to focus on the difference of performance on seen and unseen test facts which is a key issue in the development of real-world applications but other linguistic phenomena have an impact on performance.
For example in NER, concurrent complementary works by  \citet{arora-etal-2020-contextual} and \citet{fu-etal-2020-interpretable} also identify entity mention length and number of different training labels as factors of performance of NER algorithms. As a matter of fact, these properties as well as the distance between the arguments of a relation should also impact the performance of ERE models.

\paragraph{}
\textbf{Finally, lexical overlap is not only present between the test set and the training set but also with the orders of magnitude larger corpora used to pretrain Language Models.}
This overlap which is only increasing with recent Language Models trained on ever growing corpora with ever more parameters is also an important part of retention of facts in pretrained LMs, as indicated by the work of \citet{petroni-etal-2019-language}.
This implies an important limitation in every NLP study that uses word representations obtained by a Language Model trained on a corpus more recent than the NLP benchmark at hand.
In our case, using a BERT model pretrained on Wikipedia articles from 2018 as an initialization to a NER model evaluated on Reuters articles from 1996 to 1997 in CoNLL03 \textbf{boils down to predicting the past using data from the future, which has limited real-world applications.}
With Language Models trained on larger and larger corpora sometimes obtained by automatically crawling the internet, we cannot even exclude the fact that some NLP benchmarks might be included in the pretraining corpora of some Language Models.

\textbf{This leads to ignore the diachronicity of language and the generalization of models through time.}
In particular new named entities such as people or organizations regularly appear and disappear from the headlines of news articles. Entire new lexical fields can even suddenly become prominent like with the recent COVID-19 crisis.
For now the mainly considered solution to this problem is to regularly retrain entire Language Models which is not particularly cheap nor environment friendly.

\section{Future Research}
\label{sec:07:future}

Based on our findings and to address some limitations, our work opens perspectives for future research in three interconnected areas: the fine-grained analysis of End-to-end Relation Extraction models performance, the creation of new datasets and the design of models better apt to extract unseen facts.

\paragraph{}
\textbf{First, an immediate follow-up of this work is to finish the experiments on our proposal of First and Second Order Attention models, described in \autoref{chapter:06:attention}.}
This requires not only experimenting on current standard datasets and metrics but also using our metrics partitioned by lexical overlap and our behavioral study to see if this proposal can at least reduce the retention tendency of models, improve extraction of unseen facts, or even perhaps improve global performance.
A key drawback of this method is its memory complexity, storing a weight for every triple of head $i$, context word $k$ and tail $j$. A first method to address it is to switch from the Table Filling approach to an Entity Filtering or even Pipeline setting that would filter out candidate heads and tails not detected by the NER submodule.
Another lead is to get inspiration from works that propose linear attention models to reduce the complexity of this approach.

\paragraph{}
\textbf{Another follow-up is to broaden the scope of our behavioral study on retention in End-to-end Relation Extraction .}
However, as discussed in \autoref{sec:05:retention}, because swapping entity mentions on datasets such as ACE05 or SciERC leads to ungrammatical sentences, this would likely demand creating a new dataset.
Furthermore, with this retention aspect in mind, we believe that this new dataset should enable to separate the context which expresses a relation with the particular relation instance arguments in order to better measure the impact of both aspects on extraction.
It could also be the occasion to propose a multilingual dataset for End-to-end Relation Extraction for which to our knowledge only English, Chinese and Arabic datasets have been proposed with the ACE initiative.
Building such a dataset is a research project per se and would demand additional reflection on what other linguistic phenomena could be isolated to better understand the real capacities and limitations of Information Extraction models.
We do believe that enhancing our understanding of the true capacities of models is a necessary step towards improving them and that currently used datasets and metrics are not totally aligned with the true objective of Relation Extraction which is extracting new facts from raw text.

\paragraph{}
\textbf{Furthermore, because context is key in the extraction of unseen facts, we believe reversing the pipeline Entity and Relation Extraction structure is worth exploring.}
Indeed, the recent work by \citet{zhong-chen-2021-frustratingly} suggests that a BERT-based pipeline model is enough to obtain state-of-the-art results, although they omit comparison on CoNLL04 and do not report the previous best performance on SciERC.
Our experiments indicate that the comparison is much more nuanced although the intermediate type representations used in their pipeline model enable to reduce the dependency on mere retention.
We believe that the End-to-end setting has the potential to outperform the pipeline for more difficult cases than those proposed by current datasets, such as with more entity and relation types and on longer sentences containing more than a pair of entities.
And one under explored approach of End-to-end Relation Extraction is a \textbf{predicate first approach}.

Indeed, joint training was mainly introduced to better model the interdependency between the two tasks whereas the pipeline only enables the dependency of relation prediction on previous entity detection.
Yet, the structure of the mainly used end-to-end Entity Filtering models keeps this hierarchy with NER performed before RE.
This is also true for Question Answering based models that first identify entities then relations and to a lesser extend for the newly proposed Table Filling model by \citet{wang-lu-2020-two}.
Yet, the presence of two candidate entities does not guarantee the presence of a relation whereas the opposite is true.
Hence, we think that using the opposite approach to first identify a predicate then its arguments could bring complementary capabilities to the models and that we should aim to use information from relation to enhance NER capabilities.
This approach is close to what is performed in \textbf{Semantic Role Labeling} and it could be interesting to unify Relation Extraction with Semantic Role Labeling to better take context into account and truly model the interdependency between relation and argument predictions.

\paragraph{}
\textbf{Finally, the retention phenomenon we have evidenced in our study should motivate us to rethink the way Relation Extraction datasets are collected.}
Indeed, the important overlap between test and train relations in datasets such as CoNLL04 or ACE05 is symptomatic of a bias in the data collection process.
In particular, in CoNLL04, the important number of occurrences of some relations throughout the dataset such as (Oswald, Kill, Kennedy) or (James Earl Ray, Kill Martin Luther King) is an evidence that these sentences were selected based on the presence of arguments of previously known relations.

In essence, this is the process of \textbf{distant supervision} \citep{mintz-etal-2009-distant} which has been widely used in pipeline Relation Extraction as an automatic data annotation strategy and in the creation of large datasets such as NYT-Freebase \citep{Riedel2010ModelingText} for training and evaluating models.
It has even been used as an effective pretraining strategy for a model such as Matching the Blank \citep{baldini-soares-etal-2019-matching} for which one of the arguments is masked and must be retrieved.

However, it consists in only selecting sentences which contain both arguments of previously known relations and prevents extending datasets to more rarely expressed relations. This only reinforce the lexical overlap between relations and comforts the retention heuristic.
Furthermore, there is no guarantee that a sentence containing the two arguments of a relation actually expresses this relation.
That is why \textbf{we believe we could shift from distant supervision to an opposite weak supervision strategy: annotating the presence of a relation at a sentence-level.}

Indeed, even though this requires human annotation, annotating the presence of a relation at a sentence-level is much quicker than also identifying the arguments of the relation and could be assisted automatically by detecting trigger words such as predicate verbs when applicable.
The risk of lexical overlap of these trigger words should in turn be accounted for but seems less problematic than the overlap of arguments since relations are actually communicated through a limited set of expressions.

\section{Information Extraction in Language Models Era}
\label{sec:07:era}

As a final thought, we can more widely wonder about the future of Information Extraction following the rapid development of Transformer Language Models as the backbone of every NLP model.

\paragraph{}
\textbf{First, this unification of architectures shifts the focus from model design to choices regarding the quantity and quality of data, different strategies for supervision or even new framings of NLP tasks.}
Thus, some recent works propose to explore cross-task Transfer Learning by casting numerous NLP tasks as Question Answering \citep{McCann2018TheAnswering} or Language Modeling with specific prompts \citep{Raffel2020ExploringTransformer} so that they can be tackled by a single model.
Recent findings suggest that such models exhibit impressive zero-shot cross-task transfer capabilities when trained for multitask learning \citep{Wei2021FinetunedLearners, Sanh2021MultitaskGeneralization}.

For now, to our knowledge, Information Extraction tasks such as Named Entity Recognition or Relation Extraction have mainly been explored in the Question Answering setting.
This includes zero-shot Relation Extraction \citep{levy-etal-2017-zero} or ERE in the Multiturn QA model by \citet{li-etal-2019-entity}.
These approaches are based on extractive QA models more than on a text generation approach to ensure the presence of extracted entity mentions in the original context, which is not guaranteed in a text-to-text model.

In ERE, the MT-QA model can be an interesting lead to better associate a relation to its arguments as suggested by \citet{rosenman-etal-2020-exposing} for QA models in pipeline Relation Classification. 
This could help reduce retention heuristics, however in the absence of a public implementation and sufficient ablations the work by \citet{li-etal-2019-entity} fails to clearly assert the superiority of this method over the traditional classification method.

\paragraph{}
\textbf{Second, some works wonder if Language Models can be used as Knowledge Bases, making Language Model pretraining itself an Information Extraction algorithm over gigantic corpora.}
This line of works initiated by \citet{petroni-etal-2019-language} proposes to explore world knowledge encoded in Language Models with cloze-style prompts that must be completed by the Language Model, such as "Apple was founded by [MASK]".
They show that some factual knowledge can indeed be retrieved from Language Models with such prompts which open the way for using them as Knowledge Bases.
A recent work by \citet{de-cao-etal-2021-editing} even propose to edit some erroneous facts retrieved with such Language Models while minimizing the impact on other facts.

However, while this line of works is interesting to analyze the information encoded by Language Models during pretraining and expose the induced biases, such use of Language Models seems highly unreliable.
Indeed, \textbf{any text generation model can be subject to hallucinations: outputting factually false expressions.}
This is the case when generation is more closely conditioned on an input such as in Neural Machine Translation \citep{raunak-etal-2021-curious} and we can only expect it to be worse when the only conditioning comes from a beginning prompt.

Indeed, we must keep in mind that Language Models are only trained to predict the most probable sequence of following words based on a pretraining corpus, only having access to words and their distribution as a proxy for meaning.
While recent NLP successes seem to validate the distributional hypothesis and suggest that form can help to encode some semantic information, we should still refrain our tendency to assign anthropomorphic capabilities such as reasoning or understanding to such statistical models. 

Although in some cases a LM can indeed retrieve a stereotypical sentence expressing a true fact about a real-world entity, such as "Apple was founded by Steve Jobs", the lack of guarantee on when the model is correct or hallucinates is a serious hurdle for adoption.
On the contrary, we have guarantees about the information contained or not in a Knowledge Base and its structure can be used to implement logical rules to model reasoning or infer new facts.

\paragraph{}
\textbf{Hence, we believe that Information Extraction and its key Knowledge Base Construction application have a crucial role in the future of NLP as the foundation for Knowledge Aggregation and Reasoning Models that complement Language Models.}

We can take Fact Checking as an example, in a context where every internet user has the potential to generate content, either sourced and factual or not, and where Language Models can generate fluent text sometimes difficult to distinguish from human-generated text.
We could use Information Extraction to create knowledge graphs on various text sources and find where they agree and contradict with simple rules in a more scalable manner than using a Natural Language Inference model on every pair of sentences.

Language Modeling would still be the core of Information Extraction models but the conversion to a symbolic representation of facts, unique to Information Extraction, currently seems necessary to explicitly access, control and curate the intermediate interpretation of text by NLP models.

\cleardoublepage\part*{Résumé de la Thèse}

\chapter[Contextualisation et Généralisation en EER]{Contextualisation et Généralisation en Extraction d'Entités et de Relations}
\section{Introduction}
\label{sec:08:intro}
Au cours de la dernière décennie, les réseaux de neurones sont devenus incontournables
dans le Traitement Automatique du Langage (TAL), notamment pour leur capacité à
apprendre des représentations de mots à partir de grands corpus non étiquetés.
Ces plongements de mots peuvent ensuite être transférés et raffinés pour des applications diverses au cours d'une phase d'entrainement supervisé.
Plus récemment, en 2018, le transfert de modèles de langue pré-entraînés et la préservation de leurs capacités de contextualisation ont permis 
d'atteindre des performances sans précédent sur pratiquement tous les benchmarks de TAL,
surpassant parfois même des performances humaines de référence.
Cependant, alors que ces modèles atteignent des scores impressionnants, leurs capacités de compréhension apparaissent toujours assez
peu développées, révèlant les limites des jeux de données de référence pour identifier leurs facteurs de performance et pour mesurer précisément leur capacité de compréhension.

Dans cette thèse, nous nous focalisons sur une application du TAL, l'Extraction d'Entités et de Relations.
C'est une tâche cruciale de l'Extraction d'Information (EI) qui vise à convertir l'information exprimée dans un texte en une base de données structurée. 
Dans notre cas, nous souhaitons identifier les mentions d'entités - comme des personnes, des organisations ou des lieux - ainsi que les relations exprimées entre elles.
Cela permet par exemple de construire un Graphe de Connaisances utile pour formaliser un raisonnement logique sur des faits et inférer de nouvelles connaissances.

Plus précisément, nous étudions le comportement des modèles état de l'art en ce qui concerne la généralisation à des faits inconnus en Reconnaissance d'Entités Nommées (REN) et en Extraction d'Entités et de Relations (EER) sur des corpus en langue anglaise.
En effet, les benchmarks traditionnels présentent un recoupement lexical important entre les mentions et les relations utilisées pour l'entraînement et l'évaluation des modèles.
Au contraire, l'intérêt principal de l'Extraction d'Information est d'extraire des informations inconnues jusqu'alors.

Nous commençons par introduire les concepts ayant mené à l'introduction des modèles de langue et en particulier à BERT qui est aujourd'hui le nouveau standard des modèles de TAL  (\autoref{sec:08:bert}). 

Nous proposons ensuite une première étude empirique centrée sur l'Extraction d'Entités et l'impact des représentations contextuelles récentes induites par préentraitement de modèles de langue (\autoref{sec:08:contener}).

Puis, nous abordons la tâche d'EER dans la \autoref{sec:08:sincere}. 
Nous identifions d'abord des comparaisons incorrectes dans plusieurs précédents articles afin de remettre en perspective leurs différentes propositions avant d'étendre notre étude à l'impact du recoupement des mentions et des relations avec le jeu d'entrainement.

Finalement, nous partageons une idée d'architecture, en phase de développement préliminaire, dont le but est de permettre une meilleure incorporation du contexte en Extraction de Relations grâce à l'utilisation explicite des poids d'attention d'un modèle de type BERT (\autoref{sec:08:soa}).

\section{BERT et les modèles de langue préentrainés}
\label{sec:08:bert}

\subsection{Introduction à l'Apprentissage Profond}
Alors que les premiers algorithmes de TAL étaient basées sur des règles, par exemple pour créer un système conversationnel comme ELIZA \citep{Weizenbaum1966ELIZA:Machine} mais aussi dans les premières tentatives d'Extraction d'Entités \citep{Rau1991ExtractingText}, maintenir de telles règles présente des limitations apparentes.
En effet, bien qu'interprétables leur rigidité demande un important travail de réflexion et d'essai erreur par un expert du domaine afin de les concevoir et les maintenir.
De plus leur spécificité demande un travail d'adaptation pour pouvoir les transférer d'un type d'application ou de document à un autre.  

Afin de pallier ces problèmes, il est apparu utile de pouvoir apprendre ces règles automatiquement à partir de données annotées en utilisant des algorithmes d'\textbf{Apprentissage Automatique} ou \textit{Machine Learning} (ML).
Cependant, la première étape dans l'utilisation de tels algorithmes est la représentation de données qui à son tour était traditionnellement conçue par des experts.

Pour les mêmes raisons, il a semblé intéressant d'apprendre ces représentations directement à partir de données, ce qui est l'objet de l'Apprentissage de Représentations, branche du ML.
En particulier, \textbf{l'Apprentissage Profond} ou \textit{Deep Learning} (DL)\citep{LeCun2015DeepLearning} propose d'utiliser des \textbf{réseaux de neurones artificiels} pour apprendre ces représentations.
Ces modèles ont pris de l'ampleur au début des années 2010, avec l'apparition conjointe d'implémentations distribuées et accélérées sur des \textit{Graphical Processing Units} (GPU) et de jeux de données massifs tels que ImageNet \citep{Russakovsky2015ImageNetChallenge} qui ont permis d'asseoir la supériorité du réseau de neurones AlexNet \citep{Krizhevsky2012ImageNetNetworks} sur l'état de l'art précédent en classification d'images.

Les neurones artificiels sont des modèles simples de \textbf{combinaison linéaire} d'un vecteur d'entrée suivi d'une fonction d'activation non linéaire.
Leurs \textbf{paramètres} sont les \textbf{poids} de cette combinaison linéaire, initialisés aléatoirement et itérativement modifiés pour miniser une \textbf{fonction de coût différentiable} par \textbf{rétropropagation du gradient} \citep{LeCun1989BackpropagationRecognition}.

La capacité de modélisation des réseaux de neurones provient de l'architecture hiérarchique par couches des ces neurones, spécifiquement conçu pour chaque problème.
On peut citer les \textbf{réseaux de neurones convolutifs} (CNN pour \textit{Convolutional Neural Networks}) particulièrement adaptés au traitement d'images ou les \textbf{réseaux de neurones récurrents} (RNN pour \textit{Recurrent Neural Networks}) adaptés au traitement de séquences dont le texte vu comme séquence de mots.

\subsection{L'architecture Transformer}

Bien qu'adaptés au TAL, les RNN ont des difficultés à modéliser correctement les interactions à longues distances dans la séquence.
Des variantes telles que les \textbf{LSTM} \citep{Hochreiter1997LongMemory} proposant de pallier ce problème se sont imposées vers 2014 mais elles souffraient toujours d'un temps de calcul linéaire dans la longueur de la séquence.

En 2017, \citet{Vaswani2017AttentionNeed} proposent l'architecture \textbf{Transformer} pour traiter ces deux problèmes et atteignent des performances état de l'art en Traduction Automatique.
Contrairement aux RNNs ou la séquence est traitée un mot après l'autre, ici chaque mot peut être traité simultanément au prix d'une consommation de mémoire plus importante.

Cette architecture se base sur le \textbf{mécanisme d'attention} qui propose de pondérer les représentations de plusieurs éléments d'entrée selon un score de pertinence, introduit notamment pour la Traduction Automatique \citep{Bahdanau2015NeuralTranslate}.
Ici, le mécanisme employé est \textbf{l'auto-attention}, l'attention entre tous les éléments de la séquence d'entrée et eux mêmes.

L'implémentation de \citet{Vaswani2017AttentionNeed} peut s'écrire:

\begin{enumerate}
    \item Soit une séquence de vecteurs d'entrée $x_{i}$, les trois vecteurs requête $q_i$, clé $k_i$, valeur $v_i$ de même dimension $d_k$ sont obtenus par projection linéaire des $x_{i}$
    \item Pour toute paire $(i,j)$, un score d'alignement est obtenu suivant $\alpha_{i,j} = \frac{exp(q_i^T k_j / \sqrt{d_k})}{\sum_{j} exp(q_i^T k_j / \sqrt{d_k})}$
    \item La représentation de sortie est obtenue suivant $h_i=\sum_{j} \alpha_{i,j} v_{j}$
\end{enumerate}

Ceci est souvent résumé en une seule équation : 
\begin{equation*}
    \text { Attention }(Q, K, V)=\operatorname{softmax}\left(\frac{Q K^{T}}{\sqrt{d_{k}}}\right) V
\end{equation*}

L'auto-attention permet au réseau d'apprendre des motifs d'interaction entre mots d'une séquence, aussi éloignés soient-ils.
Afin de permettre l'apprentissage simultané de différents type d'interactions, \citet{Vaswani2017AttentionNeed} propose un \textbf{mécanisme multi-têtes}, en dupliquant simplement ce mécanisme typiquement avec 12 ou 16 têtes d'attention.

\begin{figure}[!h]
\begin{adjustwidth*}{}{-.3\textwidth}
\caption[Schéma de l'Auto-Attention Multitêtes.]{Schéma de l'Auto-Attention Multitêtes. 
Pour chaque élément d'entrée et dans chaque tête d'attention, les vecteurs requête, clé et valeur
sont obtenus par projection linéaire.
Chaque élément $i$ est représenté par la somme de toutes les valeurs d'entrée $v_j$ pondérée par les scores d'alignement entre son vecteur requête $q_i$ et toutes les clés $k_j$.}
\label{fig:08:mhsa}
\centering\includegraphics[width=1.3\textwidth]{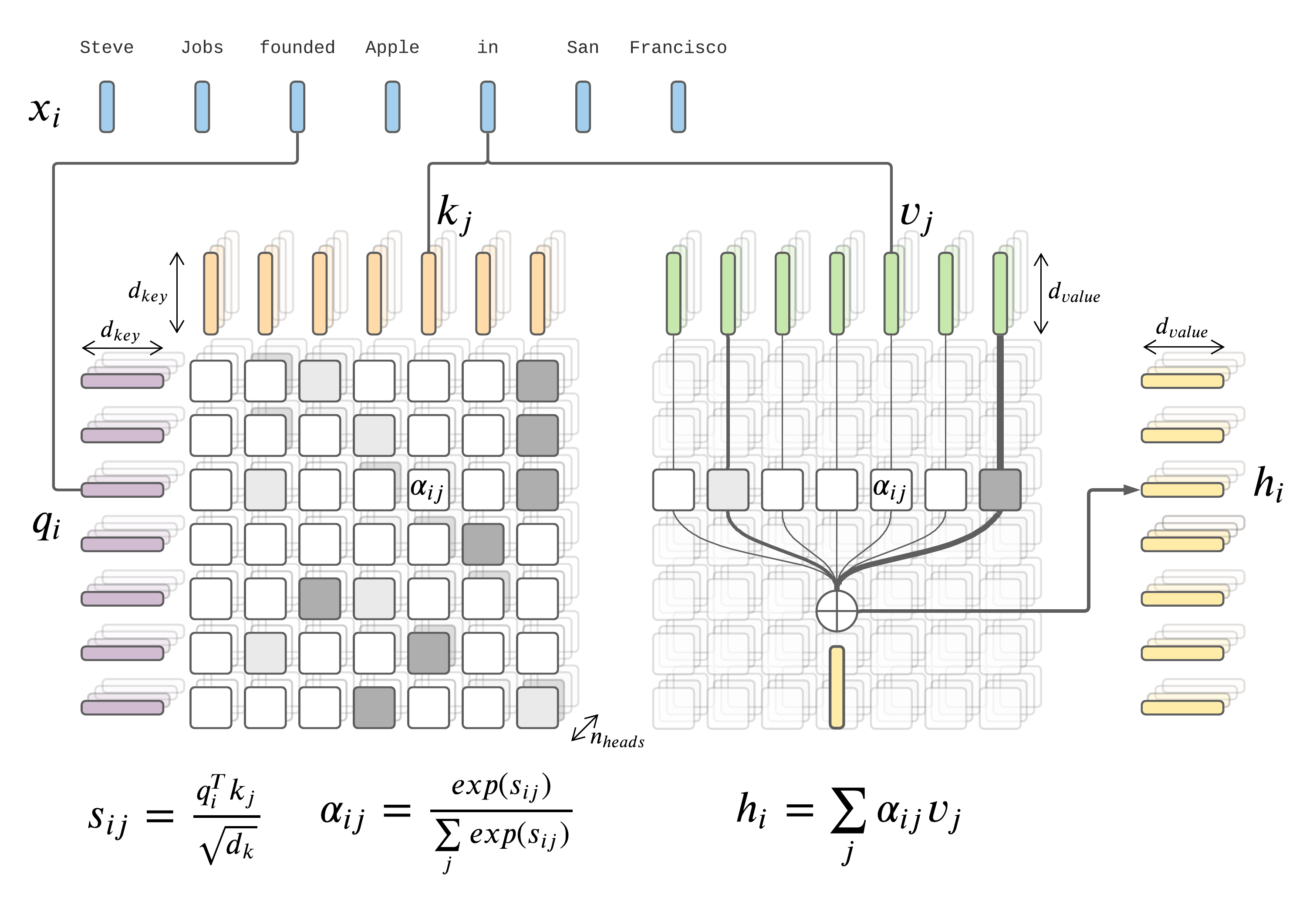}
\end{adjustwidth*}
\end{figure}

Un réseau Transformer est constitué essentiellement d'une succession de couches d'auto-attention multitêtes ou \textit{MultiHead Self-Attention} (MHSA) dont une illustration est proposée dans la \autoref{fig:08:mhsa}.

\subsection{Représentations de Mots et Modèles de Langue}

La représentation vectorielle naïve d'un mot $w$ dans un vocabulaire $V$ est la représentation \textit{one-hot} pour laquelle le vecteur a la taille du vocabulaire $|V|$ et où toutes ses composantes sont nulles exceptée celle correspondant à son index dans $V$.
Cependant cela présente plusieurs difficultés: les vecteurs obtenus sont ainsi des \textbf{vecteurs creux de grande dimension} et ils sont \textbf{orthogonaux deux à deux}, empêchant de modéliser la notion de similarité sémantique.
Ainsi les vecteurs de ``chien'' et ``chat'' sont aussi différents que ``chien'' et ``chaise''. 

Parallèlement, les \textbf{Modèles de Langue} ou \textit{Language Modèles} (LM) sont des modèles statistiques dont le but est de prédire la probabilité d'apparition d'un mot étant donné la séquence de mots qui précède  $P(w_{k} | w_{1}, ..., w_{k-1})$.
Ils sont historiquement utilisés pour des tâches de génération de texte telles que la complétion automatique de phrase ou la traduction automatique.
Traditionnellement basés sur des approches de comptage des occurrences de sous-séquences, \citet{Bengio2003AModel} proposent de les modéliser par un réseau de neurones et notamment d'apprendre des représentations de mots de petite dimension comparée à la taille du vocabulaire et reflétant une similarité sémantique.
Ils appellent ces représentations les \textit{Word Embeddings} ou \textbf{Plongements de Mots} en français.

Progressivement, il est apparu que ces LM neuronaux pouvaient servir à obtenir des représentations de mots qui pouvaient être transférées à diverses tâches de TAL, la modélisation du langage servant comme une tâche de pré-entrainement efficace permettant aux réseaux de neurones d'établir de nouveaux résultats état de l'art en apprenant une notion de similarité basée sur \textbf{l'hypothèse distributionnelle} selon laquelle un mot est caractérisé par son contexte \citep{Firth1957A1930-1955}.
Ainsi différents modèles ont été proposés comme \textbf{SENNA} \citep{Collobert2008ALearning}, \textbf{Word2Vec} \citep{Mikolov2013DistributedCompositionality} ou \textbf{GloVe} \citep{pennington-etal-2014-glove}.

Cependant, deux limites demeurent avec de tels plongements de mots statiques: un mot non rencontré pendant l'entraînement n'a pas de représentation apprise et un mot possédant polysémique n'a qu'une représentation.

Pour représenter des mots inconnus, des sous-réseaux prenant en entrée des représentations de caractères ont été proposés et sont notés charCNN ou charLSTM en fonction de l'architecture employée \citep{ling-etal-2015-finding}.
Ils peuvent apprendre des caractéristiques morphologiques telles que la présence de radicaux ou d'affixes. 
D'autres solutions telles que WordPiece \citep{Wu2016GooglesTranslation} proposent de décomposer un mot en sous-mots fréquents et d'apprendre leur représentation avec des LM.

Pour traiter la polysémie et désambiguïser différents sens le contexte semble encore une fois essentiel.
Alors qu'avec les plongements de mots traditionnels un mot ne possède qu'une représentation quel que soit son contexte, il paraît utile d'utiliser ce dernier dans la représentation.
Pour ce faire, une méthode retenue est simplement de transférer le modèle de langue complet au lieu de sa première couche comprenant les représentations statiques.
En effet, le LM est entrainé à prédire un mot en fonction de son contexte.
Par exemple dans la phrase ``Georges Washington lived in Washington D.C.'' il peut apprendre que Georges apparait dans le contexte de personnes alors que ``in'' ou ``D.C'' dans celui de lieux pour modifier les deux représentations du même mot ``Washington''.

C'est ainsi qu'ont été proposé les \textbf{plongements de mots contextuels} dérivés de modèles de langue.
D'abord basés sur une architecture récurrente avec \textbf{ELMo} \citep{peters-etal-2018-deep} puis Transformer avec \textbf{BERT} \citep{devlin-etal-2019-bert}.
Ces modèles et notamment BERT pour lequel la capacité de parallélisation du Transformer permet d'être entrainé sur des corpus plus massifs de textes ont entrainé un bond des performances en TAL telles que mesurées sur les benchmarks traditionnels.
BERT est désormais une architecture de base massivement adoptée dans les modèles état de l'art de TAL.
Cela inclut notamment les modèles de Reconnaissance d'Entités Nommées et d'Extraction de Relations pour lesquels nous proposons une étude plus approfondie des performances, notamment au regard de leur capacité de généralisation.

\section[Plongements de Mots Contextuels et Généralisation en REN]{Plongements de Mots Contextuels et Généralisation en Reconnaissance d'Entités Nommées}
\label{sec:08:contener}

\subsection{Reconnaissance d'Entités Nommées et BiLSTM-CRF}

La \textbf{Reconnaissance d'Entités Nommées} (REN) consiste à détecter les mentions textuelles d'entités telles que des personnes, organisations ou lieux et à les classifier selon leur type.
Cette tâche est traditionnellement modélisée comme de l'étiquetage de séquence : on cherche à prédire pour chaque mot un label qui désigne à la fois le type de l'entité et la position du mot dans la mention, i.e. B (\textit{beginning}) pour le premier mot, E (\textit{end}) pour le dernier, I (\textit{inside}) pour les mots intermédiaires, S (\textit{single}) pour les mentions en un mot et O (\textit{outside}) pour tous les mots ne désignant pas d'entité.

L'architecture neuronale classique pour aborder ce problème était le BiLSTM-CRF \citep{Huang2015BidirectionalTagging} avant l'apparition de BERT.
Cette dernière combine un réseau de neurones récurrent bidirectionnel, le BiLSTM avec un \textit{Conditional Random Field}, un modèle graphique probabiliste visant à apprendre les transitions probables entre labels successifs.
Cette architecture dominait les benchmarks anglais traditionnels tels que CoNLL03 d'abord avec des plongements de mots statiques \citep{Huang2015BidirectionalTagging}, puis des plongements de caractères \citep{lample-etal-2016-neural} et enfin avec les plongements de mots contextuels \citep{peters-etal-2018-deep}.
Toutefois, l'évaluation utilisée se limite à un score F1 global, moyenne harmonique entre la précision et le rappel, qui ne reflète pas les caractéristiques individuelles des exemples de test.
En particulier, pour mesurer les performances des algorithmes, ont les évalue sur des phrases jamais rencontrées pendant l'entraînement, toutefois les mentions recherchées peuvent l'être, ce \textbf{recoupement lexical} entre données de test et d'entrainement a un impact sur les performances déjà mis en avant par \citet{Augenstein2017GeneralisationAnalysis} sur des modèles datant d'avant 2011.
On propose d'étendre leur étude aux plongements de mots contextuels.

\begin{figure}[!h]
\begin{adjustwidth*}{}{-.3\textwidth}
\caption[Schéma de l'architecture BiLSTM-CRF.]{Schéma de l'architecture BiLSTM-CRF. 
Pour calculer la représentation du mot ``Apple'', un LSTM dans le sens direct prend en compte le contexte gauche dans la représentations $\textbf{l}_{Apple}$ qui est concaténée à $\textbf{r}_{Apple}$ calculée par un LSTM dans le sens contraire qui prend en compte le contexte droit.}
\label{fig:08:iobes}
\centering\includegraphics[width=1.3\textwidth]{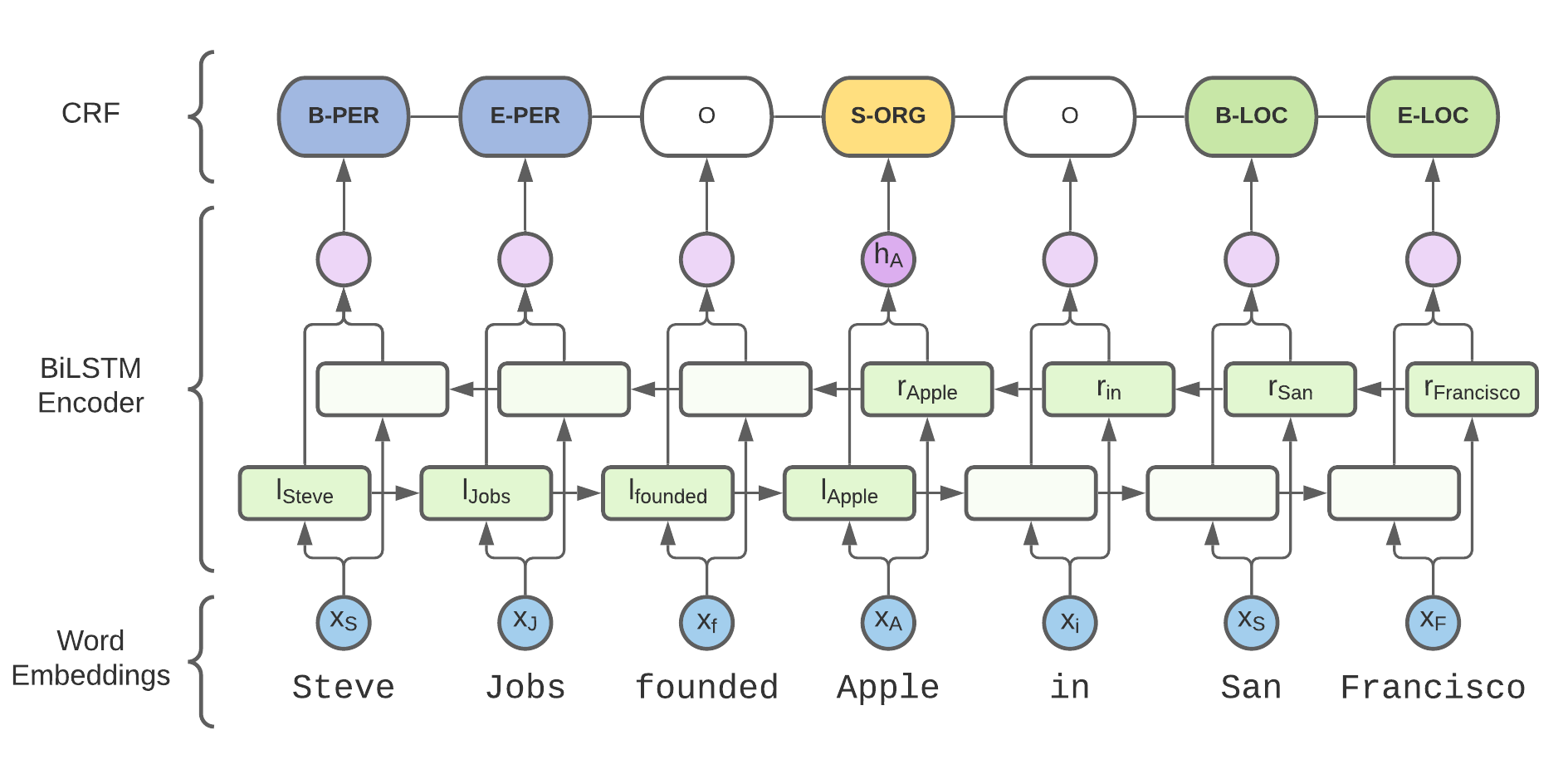}
\end{adjustwidth*}
\end{figure}

\subsection{Recoupement Lexical}
Afin de mesurer l'impact du recoupement lexical, nous proposons une partition des mentions de test selon leur recoupement un peu plus fine que celle utilisée par \citet{Augenstein2017GeneralisationAnalysis}.
Une mention est un \textbf{Exact Match (EM)} si elle apparaît sous l'exacte même forme sensible à la capitalisation dans le jeu d'entraînement et annotée avec le même type.
C'est un \textbf{Partial Match (PM)} si au moins un des mots non vides de la mention apparaît dans une mention de même type.
Toutes les autres mentions sont désignées comme \textbf{nouvelles (New)}.

Nous choisissons d'utiliser trois jeux de données en anglais pour notre étude empirique.
\textbf{CoNLL03} \citep{tjong-kim-sang-de-meulder-2003-introduction}, le benchmark standard de REN composé d'articles Reuters datés de 1996 et annotés pour quatre types : Organisation (ORG), Personne (PER), Localité (LOC) et Divers (MISC).
\textbf{OntoNotes 5.0 } \citep{Weischedel2013OntoNotesLDC2013T19} qui est composé de documents de six domaines dont des articles de presse, des conversations téléphonique et des forums web annotés pour la REN et la Résolution de Coréférence. 
Il est annoté manuellement pour onze types d'entités et sept types de valeurs qui sont généralement traités sans distinction.
\textbf{WNUT17} \citep{derczynski-etal-2017-results} qui est plus petit, spécifique aux contenus générés par les utilisateurs comme des tweets ou des commentaires Youtube ou Reddit et qui est conçu sans recoupement lexical.

\textbf{CoNLL03 et OntoNotes, les principaux benchmarks anglais présentent un recoupement lexical irréaliste} dans des cas d'utilisation concrète avec plus de la moitié des mentions de test rencontrées pendant l'entrainement.
Cela conduit a surpondérer la mémorisation des types pour les mots rencontrés au détriment de la capacité de généralisation aux nouveaux mots, pour laquelle nous proposons de mesurer l'impact des plongements de mots contextuels.

Nous évaluons également leur capacité d'adaption de domaine en entrainant nos modèles sur CoNLL03 et en les testant sur des versions avec labels alignés de OntoNotes et WNUT dénotées par $\ast$.

\subsection{Représentations de Mots étudiées}

\paragraph{Plongements de mots classiques}
Nous prenons \textbf{GloVe} \citep{pennington-etal-2014-glove} comme base de référence des plongements traditionnels.
Bien que les plongements GloVe soient calculés sur un corpus important pour capturer une similarité sémantique basée sur la co-occurrence, cette représentation est purement lexicale puisque chaque mot est aligné à une unique représentation.
Les plongements sont initialisés avec GloVe 840B et leurs valeurs sont affinées pendant l'entraînement. 

\paragraph{Plongements de mots à l'échelle des caractères}
Nous reproduisons le \textbf{Char-BiLSTM} de \citet{lample-etal-2016-neural}, un BiLSTM au niveau de chaque mot qui apprend sa représentation à partir des plongements de ses caractères pour tenir compte de caractéristiques orthographiques et morphologiques. Le charBiLSTM est entraîné conjointement au réseau de REN et ses sorties sont concaténées aux plongements GloVe.
Nous introduisons également \textbf{ELMo[0]}, le charCNN utilisé comme première couche de ELMo.

\paragraph{Plongements de mots contextuels}
Contrairement aux représentations précédentes, les plongements de mots contextuels prennent en compte le contexte d'un mot dans sa représentation.
Pour ce faire, un modèle de langue est préentrainé sur un corpus non annoté et on prend sa représentation interne de la prédiction d'un mot sachant son contexte.
\textbf{ELMo} \citep{peters-etal-2018-deep} utilise un réseau convolutif à l'échelle des caractères (Char-CNN) pour obtenir un plongement de mot indépendant du contexte et la concaténation de modèles de langue LSTM à deux couches en sens avant et inverse pour la contextualisation.
\textbf{BERT} \citep{devlin-etal-2019-bert} adopte des plongements de sous-mots 
et apprend une représentation dépendant des contextes droits et gauches en entraînant l'encodeur d'un Transformer \citep{Vaswani2017AttentionNeed} pour un modèle de langue masqué et la prédiction de la phrase suivante.
Nous utilisons le modèle  ``\textit{BERT\textsubscript{LARGE} feature-based}" pour une comparaison plus juste : les poids du modèle de langue sont gelés et nous concaténons les états cachés de ses quatre dernières couches.
\textbf{Flair} \citep{akbik-etal-2018-contextual} emploie directement un modèle de langue à l'échelle du caractère. Comme pour ELMo, deux modèles de langue LSTM de sens opposés sont entraînés et leurs sorties concaténées.

\subsection{Expériences et Résultats}

Nous proposons d'utiliser un modèle \textbf{BiLSTM-CRF} dont on fait varier uniquement les représentations des mots en entrée afin d'évaluer les capacités de généralisation de ces dernières.
De plus, nous introduisons aussi un modèle \textbf{Map-CRF} ou le BiLSTM est remplacé par une projection linéaire.

Cela permet de séparer précisément l'impact de la contextualisation supervisée par la tâche $\mathcal{C}_{REN}$ (passage de Map à BiLSTM) de la contextualisation non supervisée due au Modèle de Langue dans ELMo $\mathcal{C}_{LM}$ (passage de ELMo[0] à ELMo).

 \begin{table}[h]
 
 \begin{adjustwidth*}{-.1\textwidth}{-.3\textwidth}
\caption{Scores micro-F1 des modèles entrainés sur CoNLL03 et testés en intra-domaine et extra-domaine sur OntoNotes$^\ast$ and WNUT$^\ast$. 
Moyenne de 5 entrainement, écart-types en indice.}
\label{table:08:ood}
\centering

 \begin{tabularx}{1.3\textwidth}{@{}l@{\hspace{0em}}lr*{4}{Y}r*{4}{Y}r*{4}{Y}r@{}}
 \toprule


 &  & & \multicolumn{4}{c}{CoNLL03} & \hspace{.5em} & \multicolumn{4}{c}{OntoNotes$^\ast$} & \hspace{.5em} & \multicolumn{4}{c}{WNUT$^\ast$}\\
  \cmidrule{4-7} \cmidrule{9-12} \cmidrule{14-17} 
 & Emb. & & EM & PM & New & All & & EM & PM & New & All & & EM & PM & New & All \\
\midrule

 \parbox[t]{4mm}{\multirow{6}{*}{\rotatebox[origin=c]{90}{\textbf{BiLSTM-CRF}}}} 
                      & BERT                   &    & 95.7$_{\stddev{.1}}$ & 88.8$_{\stddev{.3}}$ & 82.2$_{\stddev{.3}}$ & 90.5$_{\stddev{.1}}$ &
                                                    & 95.1$_{\stddev{.1}}$ & 82.9$_{\stddev{.5}}$ & 73.5$_{\stddev{.4}}$ & \textbf{85.0}$_{\stddev{.3}}$ & 
                                                    & 57.4$_{\stddev{1.0}}$ & 56.3$_{\stddev{1.2}}$ & 32.4$_{\stddev{.8}}$ & 37.6$_{\stddev{.8}}$\\
                                                    
                      & ELMo                   &    & 95.9$_{\stddev{.1}}$ & 89.2$_{\stddev{.5}}$ & 85.8$_{\stddev{.7}}$ & \textbf{91.8}$_{\stddev{.3}}$ &
                                                    & 94.3$_{\stddev{.1}}$ & 79.2$_{\stddev{.2}}$ & 72.4$_{\stddev{.4}}$ & 83.4$_{\stddev{.2}}$ &
                                                    & 55.8$_{\stddev{1.2}}$ & 52.7$_{\stddev{1.1}}$ & 36.5$_{\stddev{1.5}}$ & \textbf{41.0}$_{\stddev{1.2}}$\\
                                                    
                      & Flair                  &    & 95.4$_{\stddev{.1}}$ & 88.1$_{\stddev{.6}}$ & 83.5$_{\stddev{.5}}$ & 90.6$_{\stddev{.2}}$ &
                                                    & 94.0$_{\stddev{.3}}$ & 76.1$_{\stddev{1.1}}$ & 62.1$_{\stddev{.5}}$ & 79.0$_{\stddev{.5}}$ &
                                                    & 56.2$_{\stddev{2.2}}$ & 49.4$_{\stddev{3.4}}$ & 29.1$_{\stddev{3.3}}$ & 34.9$_{\stddev{2.9}}$\\
\cdashline{2-17}
                      & ELMo[0] &    & 95.8$_{\stddev{.1}}$ & 87.2$_{\stddev{.2}}$ & 83.5$_{\stddev{.4}}$ & 90.7$_{\stddev{.1}}$ &
                                                    & 93.6$_{\stddev{.1}}$ & 76.8$_{\stddev{.6}}$ & 66.1$_{\stddev{.3}}$ & 80.5$_{\stddev{.2}}$ &
                                                    & 52.3$_{\stddev{1.2}}$ & 50.8$_{\stddev{1.5}}$ & 32.6$_{\stddev{2.2}}$ & 37.6$_{\stddev{1.8}}$\\
                                                    
                      & G + char              &     & 95.3$_{\stddev{.3}}$ & 85.5$_{\stddev{.7}}$ & 83.1$_{\stddev{.7}}$ & 89.9$_{\stddev{.5}}$ &
                                                    & 93.9$_{\stddev{.2}}$ & 73.9$_{\stddev{1.1}}$ & 60.4$_{\stddev{.7}}$ & 77.9$_{\stddev{.5}}$ &
                                                    & 55.9$_{\stddev{.8}}$ & 46.8$_{\stddev{1.8}}$ & 19.6$_{\stddev{1.6}}$ & 27.2$_{\stddev{1.3}}$\\
                                                    
                      & GloVe                  &    & 95.1$_{\stddev{.4}}$ & 85.3$_{\stddev{.5}}$ & 81.1$_{\stddev{.5}}$ & 89.3$_{\stddev{.4}}$ &
                                                    & 93.7$_{\stddev{.2}}$ & 73.0$_{\stddev{1.2}}$ & 57.4$_{\stddev{1.8}}$ & 76.9$_{\stddev{.9}}$ &
                                                    & 53.9$_{\stddev{1.2}}$ & 46.3$_{\stddev{1.5}}$ & 13.3$_{\stddev{1.4}}$ & 27.1$_{\stddev{1.0}}$\\
\cmidrule{1-17}
\parbox[t]{4mm}{\multirow{6}{*}{\rotatebox[origin=c]{90}{\textbf{Map-CRF}}}}
                      & BERT                   &    & 93.2$_{\stddev{.3}}$ & 85.8$_{\stddev{.4}}$ & 73.7$_{\stddev{.8}}$ & 86.2$_{\stddev{.4}}$ &
                                                    & 93.5$_{\stddev{.2}}$ & 77.8$_{\stddev{.5}}$ & 67.8$_{\stddev{.9}}$ & 80.9$_{\stddev{.4}}$ &
                                                    & 57.4$_{\stddev{.3}}$ & 53.5$_{\stddev{2.6}}$ & 33.9$_{\stddev{.6}}$ & 38.4$_{\stddev{.4}}$\\
                                                    
                      & ELMo                   &    & 93.7$_{\stddev{.2}}$ & 87.2$_{\stddev{.6}}$ & 80.1$_{\stddev{.3}}$ & \textbf{88.7}$_{\stddev{.2}}$ &
                                                    & 93.6$_{\stddev{.1}}$ & 79.1$_{\stddev{.5}}$ & 69.5$_{\stddev{.4}}$ & \textbf{82.2}$_{\stddev{.3}}$ &
                                                    & 61.1$_{\stddev{.7}}$ & 53.0$_{\stddev{.9}}$ & 37.5$_{\stddev{.7}}$ & \textbf{42.4}$_{\stddev{.6}}$\\

                      & Flair                   &   & 94.3$_{\stddev{.1}}$ & 85.1$_{\stddev{.3}}$ & 78.6$_{\stddev{.3}}$ & 88.1$_{\stddev{.03}}$ &
                                                    & 93.2$_{\stddev{.1}}$ & 74.0$_{\stddev{.3}}$ & 59.6$_{\stddev{.2}}$ & 77.5$_{\stddev{.2}}$ &
                                                    & 52.5$_{\stddev{1.2}}$ & 50.6$_{\stddev{.4}}$ & 28.8$_{\stddev{.5}}$ & 33.7$_{\stddev{.5}}$\\
\cdashline{2-17}
                      & ELMo[0]                 &   & 92.2$_{\stddev{.3}}$ & 80.5$_{\stddev{1.0}}$ & 68.6$_{\stddev{.4}}$ & 83.4$_{\stddev{.4}}$ &
                                                    & 91.6$_{\stddev{.4}}$ & 69.6$_{\stddev{1.0}}$ & 56.8$_{\stddev{1.5}}$ & 75.0$_{\stddev{1.0}}$ &
                                                    & 51.9$_{\stddev{1.1}}$ & 42.6$_{\stddev{.9}}$ & 32.4$_{\stddev{.3}}$ & 35.8$_{\stddev{.4}}$\\

                      & G + char                &   & 93.1$_{\stddev{.3}}$ & 80.7$_{\stddev{.9}}$ & 69.8$_{\stddev{.7}}$ & 84.4$_{\stddev{.4}}$ &
                                                    & 91.8$_{\stddev{.3}}$ & 69.3$_{\stddev{.3}}$ & 55.6$_{\stddev{1.1}}$ & 74.8$_{\stddev{.5}}$ & 
                                                    & 50.6$_{\stddev{.9}}$ & 42.5$_{\stddev{1.4}}$ & 20.6$_{\stddev{2.8}}$ & 28.7$_{\stddev{2.5}}$\\
                                                    
                      & GloVe                   &   & 92.2$_{\stddev{.1}}$ & 77.0$_{\stddev{.4}}$ & 61.7$_{\stddev{.3}}$ & 81.5$_{\stddev{.05}}$ &
                                                    & 89.6$_{\stddev{.3}}$ & 62.8$_{\stddev{.6}}$ & 38.5$_{\stddev{.4}}$ & 68.1$_{\stddev{.4}}$ & 
                                                    & 46.8$_{\stddev{.8}}$ & 41.3$_{\stddev{.5}}$ & 3.2$_{\stddev{.2}}$ & 18.9$_{\stddev{.7}}$\\
 \bottomrule
 \end{tabularx}

\end{adjustwidth*}
\end{table}

Nous observons d'abord comme attendu que dans toutes les configurations $F1_{EM}~>~F1_{PM}~>~F1_{new}$ avec plus de 10 points d'écart entre EM et New.
Cela confirme un biais de recoupement lexical qui est néanmoins comblé par les deux contextualisation évoquées: $\mathcal{C}_{REN}$ et $\mathcal{C}_{LM}$.

La comparaison des deux montre que $\mathcal{C}_{REN}$ est plus bénéfique que $\mathcal{C}_{LM}$ sur CoNLL03, son domaine de supervision, alors que $\mathcal{C}_{LM}$ est particulièrement utile en extra-domaine, d'autant plus que le domaine cible paraît éloigné du domaine source.
Ces deux formes de contextualisations peuvent par ailleurs être complémentaires, leur combinaison permettant d'obtenir les meilleurs résultats en intra-domaine ou en extra-domaine sur OntoNotes$^\ast$.

\subsection{Conclusion}

Les benchmarks actuels de REN sont donc biaisés en faveur des mentions déjà rencontrées, à l'exact opposé des applications
concrètes. 
D'où la nécessité de séparer les performances par degré de recoupement des mentions pour mieux évaluer les capacités de généralisation.
Dans ce cadre, les plongements contextuels bénéficient plus significativement aux mentions non rencontrées, d'autant plus en extra domaine.

\section{Repenser l'Évaluation en Extraction d'Entités et de \mbox{Relations}}
\label{sec:08:sincere}

Une fois ce travail effectué pour la REN, nous pouvons l'étendre à l'Extraction jointe d'Entités et de Relations.
Dans cette tâche, il faut comme en REN extraire et classifier les mentions d'entités mais aussi les relations exprimées entre elles.
On se limite généralement aux relations binaires dirigées qui peuvent être représentées par un \textbf{triplet (sujet, prédicat, objet)} tel que (Apple, fondé par, Steve Jobs).
Une première approche appelée \textbf{pipeline} consiste à considérer l'extraction d'entité et de relations comme séparées: deux modèles sont entrainés séparément et appliqués séquentiellement.
Cependant, une approche d'apprentissage d'\textbf{entrainement conjoint} a également été proposé dans le but de mieux modéliser l'interdépendance apparente entre les deux tâches.

Toutefois, de multiples architectures et cadres d'évaluation ont été proposés, ce qui a entrainé une confusion et des \textbf{comparaisons incorrectes} dans la litérature.
Nous identifions donc d'abord ces erreurs avant d'étendre notre étude à l'impact du recoupement lexical en Extraction jointe d'Entités et de Relations (EER).

\subsection{Différentes architectures en EER}

L'approche naive pour extraire les entités et les relations est l'approche \textbf{pipeline} ou l'Extraction de Relations est traitée comme classification étant donnée une phrase et deux arguments candidats.
Lors de l'entrainement, ces candidats sont les mentions annotées comme vérité terrain alors qu'en inférence ce sont les prédictions d'un modèle de REN indépendant.
L'architecture classique en Classification de Relations est une architecture par morceaux dite \textbf{Piecewise} \citep{zeng-etal-2015-distant} dans laquelle la phrase est découpée en trois morceaux après avoir été encodée. 
Le morceau à gauche du premier argument, celui à droite du second argument et celui entre les deux arguments les représentations de chaque mots d'un morceau sont aggrégées par \textbf{pooling} en une représentation unique du morceau.
L'intuition derrière cette architecture est le fait qu'au moins en anglais, une relation est plus souvent exprimée dans le morceau entre ses arguments.

Avec l'émergence de BERT, des architectures plus simples basées sur des Modèles de Langue Transformer ont été proposées \citep{Alt2019ImprovingRepresentations, Soares2019MatchingLearning}.
Par exemple, des tokens spéciaux représentatifs du type des arguments candidats peuvent être insérés dans la phrase et utilisés pour la classification de relations.
Au delà de cette architecture pipeline, de nombreuses propositions ont été faites pour modéliser l'interaction entre extraction d'entités et de relations.

\paragraph{}
Une première famille de modèles sont les \textbf{modèles incrémentaux} \citep{li-ji-2014-incremental, katiyar-cardie-2017-going} qui parcourent une phrase mot à mot et font deux prédictions à chaque étape.
D'abord la prédiction classique des labels de REN et si elle correspond au dernier mot d'une entité une prédiction de relation entre cette dernière entité dectectée et toutes celles précédemment detectées.

Dans un même esprit mais dans une volonté simplificatrice, \citet{miwa-sasaki-2014-modeling} proposent de modéliser l'EER comme le remplissage séquentiel d'un tableau appelé \textbf{Table Filling}.
Ce tableau contient une case pour chaque paire de mots de la phrase et ses éléments diagonaux contiennent l'information sur les entités tandis que les autres représentent les relations.
Cette approche a été reprises par \citet{gupta-etal-2016-table} et \citet{zhang-etal-2017-end} avec des RNN et des plongements de mots statiques et plus récemment par \citet{wang-lu-2020-two} qui utilisent un modèle basé sur BERT ainsi qu'un RNN Multidimensionnel.

Une manière plus simple de modéliser l'interaction entre les deux tâches est de partager un réseau encodeur dont les sorties sont utilisées par deux décodeurs séparés dans l'approche dite \textbf{Shared Encoder}.
C'est cette approche qui a été la plus majoritairement utilisée, d'abord avec un encodeur BiLSTM \citep{miwa-bansal-2016-end, adel-schutze-2017-global,zhang-etal-2017-end, sun-etal-2018-extracting, bekoulis-etal-2018-adversarial, Nguyen2019End-to-endAttention, Sanh2019ATasks} puis BERT \citep{wadden-etal-2019-entity, Eberts2020Span-basedPre-training}. Il est a noté que ces derniers modèles modifient également la REN qui n'est plus modélisée comme étiquetage de séquence mais comme classification de chaque n-grammes d'une phrase dans une approche dite \textbf{span-based} \citep{sohrab-miwa-2018-deep}. Elle a l'avantage de pouvoir détecter des mentions imbriquées l'une dans l'autre ou qui se chevauchent, ce qui est notamment utile dans le domaine des textes biomédicaux.

Enfin, une approche plus originale propose de modéliser l'EER comme des réponses successives à un modèle prédéfini de questions \citep{li-etal-2019-entity}. Par exemple pour la relation ``fondé par'' on peut d'abord demander ``Quelle entreprise est mentionnée dans le texte ?'' puis en fonction de la réponse X du modèle ``Qui a fondé X ?''.

\subsection{Attention aux Comparaisons Incorrectes en EER}
Dans cette multiplication d'architectures, il est difficile de dégager avec certitude la meilleure approche, malgré l'utilisation de benchmarks communs.
D'abord parce qu'on observe une triple évolution dans ces architectures: la modélisation des interactions a évolué d'une approche incrémentale à l'utilisation d'encodeurs partagés, la REN traditionnellement traitée comme étiquetage de séquence est désormais abordée à l'échelle des n-grammes et évidemment les représentations de mots statiques ont été remplacées par BERT.
Dans cette triple évolution, il est difficile de comparer deux modèles lorsqu'il y a plus d'un facteur de variation et malheureusement, les études ablatives correspondantes sont très rarement menées.

De plus, on observe aussi une multiplication de cadre de mesure de performance qui a malheureusement mené à plusieurs comparaisons incorrectes dans des travaux précédents.
En effet, bien que comme pour la REN on utilise les scores de Précision, Rappel et F1 comme métriques, encore faut-il s'accorder sur les critères utilisés pour considérer une relation comme correctement détectée.

Si le type et la direction de la relation doivent être corrects, différents critères ont été utilisés concernant la détection de leurs arguments, comme décrit par \citet{bekoulis-etal-2018-adversarial}:

\textit{\textbf{Strict}}: les limites des arguments ainsi que le type d'entité doivent être corrects.

\textit{\textbf{Boundaries}}: seules les limites des arguments doivent être correctes.

\textit{\textbf{Relaxed}}: une entité à plusieurs mots est correcte si un de ces mots est classifié avec le bon type.

\begin{figure}[htb!]
\begin{adjustwidth*}{}{-.3\textwidth}
    \caption{Illustration des différents critères d'évaluation en EER.}
    \label{08:fig:ere-eval}
    \centering
    \includegraphics[width=1.3\textwidth]{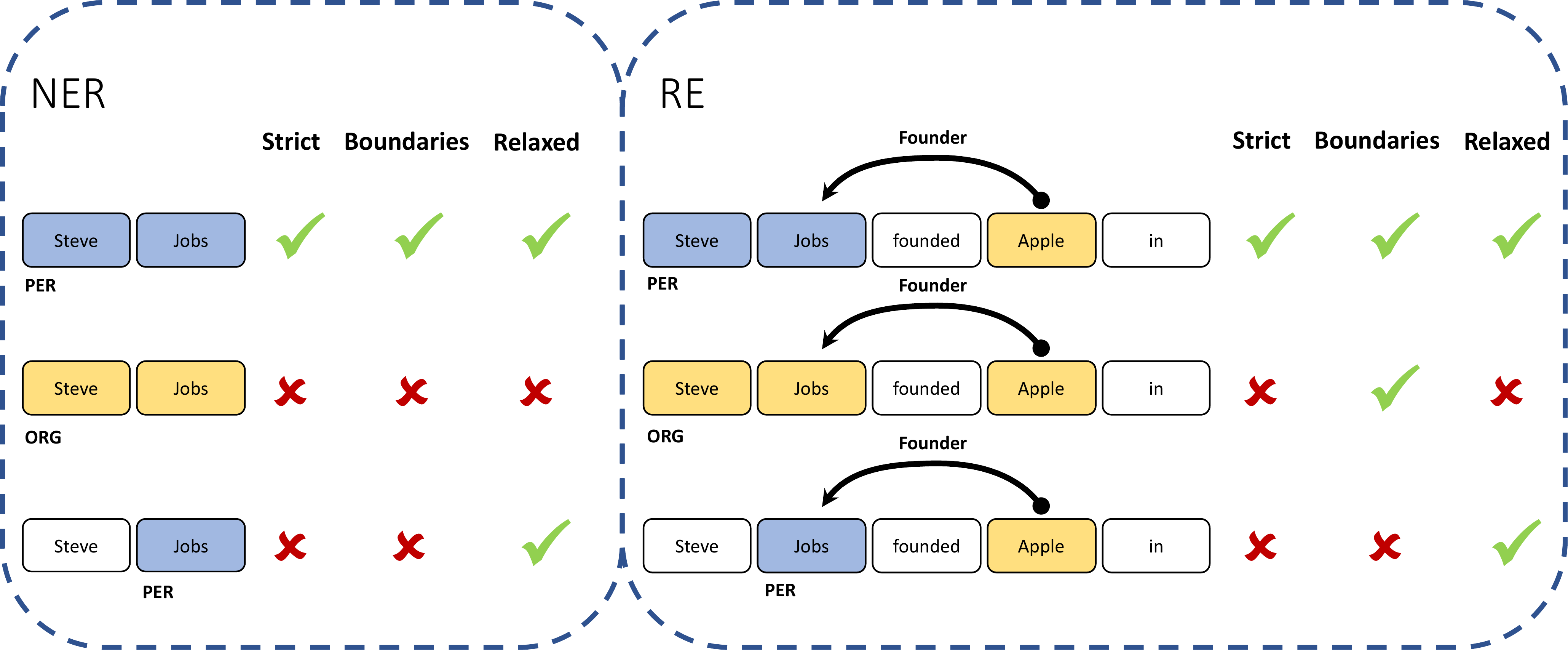}
\end{adjustwidth*}
\end{figure}

\paragraph{}
Nous identifions donc des erreurs dans plusieurs précédents articles.
La plus fréquente est la confusion des critères Strict et Boundaries,  plusieurs articles \citep{Zheng2017JointNetwork, luan-etal-2019-general, wadden-etal-2019-entity, Sanh2019ATasks} comparant des résultats plus lâches à ceux précédemment rapportées en Strict.
\textbf{Nous évaluons que cette erreur peut conduire à une surestimation de l'ordre de 5\% des performances F1 sur un dataset standard comme ACE05}, alors que cela a peu d'impact sur CoNLL04 qui a des caractéristiques différentes.
Afin d'éviter de nouvelles erreurs, \textbf{nous invitons à une plus grande rigueur dans le rapport des cadres d'évaluation}, ce qui comprend les statistiques détaillées des datasets, parfois indument modifiés entre deux articles.
\textbf{Nous appelons aussi à rapporter toujours un score Strict et Boundaries pour unifier les cadres expérimentaux} et faciliter les comparaisons de modèles quel que soit le jeu de données utilisé.

\paragraph{}
Nous profitons également de cette étude empirique pour effectuer l'ablation de deux facteurs de variations sur le modèle récent SpERT \citep{Eberts2020Span-basedPre-training}: l'utilisation de BERT et la modélisation de la REN en span-based.
Nous confirmons que \textbf{les gains de performances tels que mesurés par les benchmarks ACE05 et CoNLL04 sont très largement dûs à BERT} alors que l'utilisation du span-based n'a que peu d'impact, voire un impact négatif.

\subsection{Séparer la Rétention de l'Extraction en EER}

Bien qu'il soit d'abord nécessaire de maintenir l'intégrité des résultats pour permettre des comparaisons, nous restons convaincus qu'un unique score global ne peut refléter les caractéristiques linguistiques de chaque exemple, en particulier concernant le recoupement lexical.
Dans le cadre de l'Extraction de Relations comme en Extraction d'Entités, un modèle peut ainsi simplement mémoriser une relation vue en entrainement et mesurer ses performances sur cette relation présente un biais important.
C'est pourquoi nous proposons d'étendre l'étude empirique menée sur la REN à l'EER avec deux expériences sur trois modèles état de l'art basés sur BERT et représentatifs de trois approches différentes.
\textbf{PURE} \citep{zhong-chen-2021-frustratingly} pour la pipeline, \textbf{SpERT} \citep{Eberts2020Span-basedPre-training} pour le Shared Encoder en Entity Filtering et \textbf{TABTO} \citep{wang-lu-2020-two} pour le Table Filling.

\paragraph{Partition des Performances par Recoupement Lexical}
D'abord, comme dans la \autoref{sec:08:contener}, nous proposons de séparer les performances selon le recoupement lexical des mentions et des relations avec le jeu d'entrainement.
Nous utilisons les deux benchmarks standard \textbf{CoNLL04} \citep{roth-yih-2004-linear} et \textbf{ACE05} \citep{Walker2006ACECorpus} composés d'articles de presse ainsi que SciERC \citep{luan-etal-2018-multi} composé de résumés d'articles scientifiques et où les relations sont entre des concepts scientifiques comme des tâches, méthodes ou métriques.

Nous séparons les mentions d'évaluation en deux groupes: \textit{seen} si la mention apparaît sous la même forme avec le même label en entrainement et \textit{unseen} sinon.

Les relations sont séparées en trois: \textit{Exact Match (EM)} si la relation apparaît en entrainement sous l'exact même forme (sujet, prédicat, objet), \textit{Partial Match (PM)} si un de ses arguments apparaît au même endroit dans une relation de même type et \textit{New} sinon. 

    \begin{table}[t]
    \begin{adjustwidth*}{-.1\textwidth}{-.3\textwidth}
    
    
    \caption[Scores F1 de test en Extraction d'Entités et de Relations séparés par recoupement lexical avec le jeu d'entrainement.]{Scores F1 de test en Extraction d'Entités et de Relations séparés par recoupement lexical avec le jeu d'entrainement. Les scores Exact Match ne sont pas rapportés sur SciERC ou leur support n'est composé que de 5 instances. Moyennes et écarts types sur cinq entrainements.}
    \label{table:08:overlap}
    
    \centering
    \small
    
    \begin{tabularx}{1.4\textwidth}{@{}l*{3}{Y}r*{4}{Y}r*{4}{Y}c@{}}
    
    \toprule
    \multicolumn{1}{c}{\multirow{2}{*}{$\mu$ F1}}
    
     & \multicolumn{3}{c}{Entités} &  & \multicolumn{4}{c}{Relations Boundaries} & & \multicolumn{4}{c}{Relations Strict} \\
    \cline{2-4} \cline{6-9} \cline{11-14} 
    & Seen & Unseen & All & & Exact & Partial & New & All &  &  Exact & Partial & New & All  \\
    
    \midrule
    & \multicolumn{14}{c}{\textbf{ACE05}} \\
    \cline{2-4} \cline{6-9} \cline{11-14} 
    \textit{proportion} &\textit{ 82\%} & \textit{18\%} & & & \textit{23\%} & \textit{63\%} & \textit{14\%} & & & \textit{23\%} & \textit{63\%}& \textit{14\%}\\
    \midrule
    heuristic & 59.2$_{\stddev{\phantom{0.0}}}$ & - & 55.1$_{\stddev{\phantom{0.0}}}$ &  & 37.9$_{\stddev{\phantom{0.0}}}$ & - & - & 23.0$_{\stddev{\phantom{0.0}}}$ &  & 34.3$_{\stddev{\phantom{0.0}}}$ & - & - & 20.8$_{\stddev{\phantom{0.0}}}$ &  \\
    
    
    SpERT & 89.4$_{\stddev{0.2}}$ & 74.2$_{\stddev{0.8}}$ & 86.8$_{\stddev{0.2}}$ &  & 84.8$_{\stddev{0.8}}$ & 59.6$_{\stddev{0.7}}$ & 42.3$_{\stddev{1.1}}$ & 64.0$_{\stddev{0.6}}$ &  & 82.6$_{\stddev{0.8}}$ & 55.6$_{\stddev{0.7}}$ & 38.4$_{\stddev{1.1}}$ & 60.6$_{\stddev{0.5}}$ &  \\
    
    
    TABTO & 89.7$_{\stddev{0.1}}$ & 77.4$_{\stddev{0.8}}$ & 87.5$_{\stddev{0.2}}$ &  & 85.9$_{\stddev{0.9}}$ & 62.6$_{\stddev{1.8}}$ & 44.6$_{\stddev{2.9}}$ & \textbf{66.4}$_{\stddev{1.3}}$ &  & 81.6$_{\stddev{1.5}}$ & 58.1$_{\stddev{1.6}}$ & 38.5$_{\stddev{3.1}}$ & \textbf{61.7}$_{\stddev{1.1}}$ &  \\
    
    

    PURE & 90.5$_{\stddev{0.2}}$ & 80.0$_{\stddev{0.3}}$ & \textbf{88.7}$_{\stddev{0.1}}$ &  & 86.0$_{\stddev{1.3}}$ & 60.5$_{\stddev{1.0}}$ & 47.1$_{\stddev{1.6}}$ & \textbf{65.1}$_{\stddev{0.7}}$ &  & 84.1$_{\stddev{1.1}}$ & 57.9$_{\stddev{1.3}}$ & 44.0$_{\stddev{2.0}}$ & \textbf{62.6}$_{\stddev{0.9}}$ &  \\
    
    \midrule
     & \multicolumn{14}{c}{\textbf{CoNLL04}} \\
    \cline{2-4} \cline{6-9} \cline{11-14} 
    \textit{proportion} &\textit{ 50\%} & \textit{50\%} & & & \textit{23\%} & \textit{34\%} & \textit{43\%} & & & \textit{23\%} & \textit{34\%}& \textit{43\%} \\
    \midrule
    heuristic & 86.0$_{\stddev{\phantom{0.0}}}$ & - & 59.7$_{\stddev{\phantom{0.0}}}$ &  & 90.9$_{\stddev{\phantom{0.0}}}$ & - & - & 35.5$_{\stddev{\phantom{0.0}}}$ &  & 90.9$_{\stddev{\phantom{0.0}}}$ & - & - & 35.5$_{\stddev{\phantom{0.0}}}$ &  \\
     
    
    SpERT & 95.4$_{\stddev{0.4}}$ & 81.2$_{\stddev{0.4}}$ & 88.3$_{\stddev{0.2}}$ &  & 91.4$_{\stddev{0.6}}$ & 67.0$_{\stddev{1.1}}$ & 59.0$_{\stddev{1.4}}$ & 69.3$_{\stddev{1.2}}$ &  & 91.4$_{\stddev{0.6}}$ & 66.9$_{\stddev{1.1}}$ & 58.5$_{\stddev{1.4}}$ & 69.0$_{\stddev{1.2}}$ &  \\
   
    TABTO & 95.4$_{\stddev{0.4}}$ & 83.1$_{\stddev{0.7}}$ & \textbf{89.2}$_{\stddev{0.5}}$ &  & 92.6$_{\stddev{1.5}}$ & 72.6$_{\stddev{2.1}}$ & 64.8$_{\stddev{1.0}}$ & \textbf{74.0}$_{\stddev{1.4}}$ &  & 92.6$_{\stddev{1.5}}$ & 72.1$_{\stddev{1.8}}$ & 64.7$_{\stddev{1.1}}$ & \textbf{73.8}$_{\stddev{1.2}}$ &  \\
    
    PURE & 95.0$_{\stddev{0.2}}$ & 81.8$_{\stddev{0.2}}$ & 88.4$_{\stddev{0.2}}$ &  & 90.1$_{\stddev{1.3}}$ & 66.6$_{\stddev{1.0}}$ & 58.6$_{\stddev{1.5}}$ & 68.3$_{\stddev{1.0}}$ &  & 89.9$_{\stddev{1.4}}$ & 66.6$_{\stddev{1.0}}$ & 58.5$_{\stddev{1.5}}$ & 68.2$_{\stddev{0.9}}$ &  \\
    
    \midrule
    & \multicolumn{14}{c}{\textbf{SciERC}} \\
    \cline{2-4} \cline{6-9} \cline{11-14} 
    \textit{proportion} &\textit{ 23\%} & \textit{77\%} & & & \textit{<1\%} & \textit{30\%} & \textit{69\%} & & & \textit{<1\%} & \textit{30\%}& \textit{69\%} \\
    \midrule
    
    heuristic & 31.3$_{\stddev{\phantom{0.0}}}$ & - & 20.1$_{\stddev{\phantom{0.0}}}$ & 
    & - & - & - & 0.7$_{\stddev{\phantom{0.0}}}$ &  
    & - & - & - & 0.7$_{\stddev{\phantom{0.0}}}$ \\
    
    
    SpERT & 78.5$_{\stddev{0.5}}$ & 64.2$_{\stddev{0.4}}$ & \textbf{67.6}$_{\stddev{0.3}}$ &  
    & - & 53.1$_{\stddev{1.2}}$ & 46.0$_{\stddev{1.0}}$ & \textbf{48.2}$_{\stddev{1.1}}$ &  
    & - & 43.0$_{\stddev{1.6}}$ & 33.2$_{\stddev{1.1}}$ & \textbf{36.2}$_{\stddev{1.0}}$ & \\
    
    PURE & 78.0$_{\stddev{0.5}}$ & 63.8$_{\stddev{0.6}}$ & \textbf{67.2}$_{\stddev{0.4}}$& & - & 54.0$_{\stddev{0.7}}$ & 44.8$_{\stddev{0.4}}$ & \textbf{47.6}$_{\stddev{0.3}}$ & 
    & - & 42.2$_{\stddev{0.7}}$ & 32.6$_{\stddev{0.7}}$ & \textbf{35.6}$_{\stddev{0.6}}$ & \\
    
    \bottomrule
    
    \end{tabularx}
    
    \end{adjustwidth*}
    \end{table}

Nous pouvons d'abord observer que ACE05 et CoNLL04 présentent des proportions de recoupement lexical importantes, que ce soit au niveau des mentions d'entités ou de relations dont au moins un des arguments
a déjà été rencontré dans une position similaire.
Au contraire, SciERC a un recoupement réduit, que ce soit pour les entités ou les relations, pour lesquelles la proportion d'Exact Match est quasi nulle.
Cette différence explique des performances largement inférieures sur ce dernier dataset offrant des conditions plus difficiles.

Ensuite, cette expérience confirme un biais de rétention important avec des performances meilleures sur les relations vues, exactement ou partiellement.
Néanmoins, il est toujours difficile de trancher sur une meilleure approche en EER puisque la hiérarchie des modèles est différente d'un jeu de données à l'autre.
Néanmoins ces résultats mettent en perspective les affirmations de \citet{zhong-chen-2021-frustratingly} sur la supériorité de l'approche pipeline qu'ils n'ont pas évaluée sur CoNLL04.

\paragraph{Étude Comportementale par Inversion du Sujet et de l'Objet}
Nous proposons une seconde expérience, limitée à deux types de relations dans le dataset CoNLL04 qui ont la particularité d'avoir des arguments de même types. Les relations ``Kill'' entre deux personnes et ``Located in'' entre deux lieux.
En effet, nous proposons d'inverser le sujet et l'objet de la relation dans une étude comportementale inspirée par les travaux de \citet{mccoy-etal-2019-right} et \citet{ribeiro-etal-2020-beyond}.

\begin{figure}[h!]
    \centering
    \caption[Exemple de phrase après inversion sujet-objet.]{Exemple de phrase après inversion sujet-objet. Le triplet (John Wilkes Booth, Kill, President Lincoln) est présent dans le jeu d'entrainement et l'heuristique de rétention conduit les modèles à l'extraire même dans la phrase qui exprime la relation inverse.}
    \label{fig:08:swapped_ex}
    
    \includegraphics[width=\textwidth]{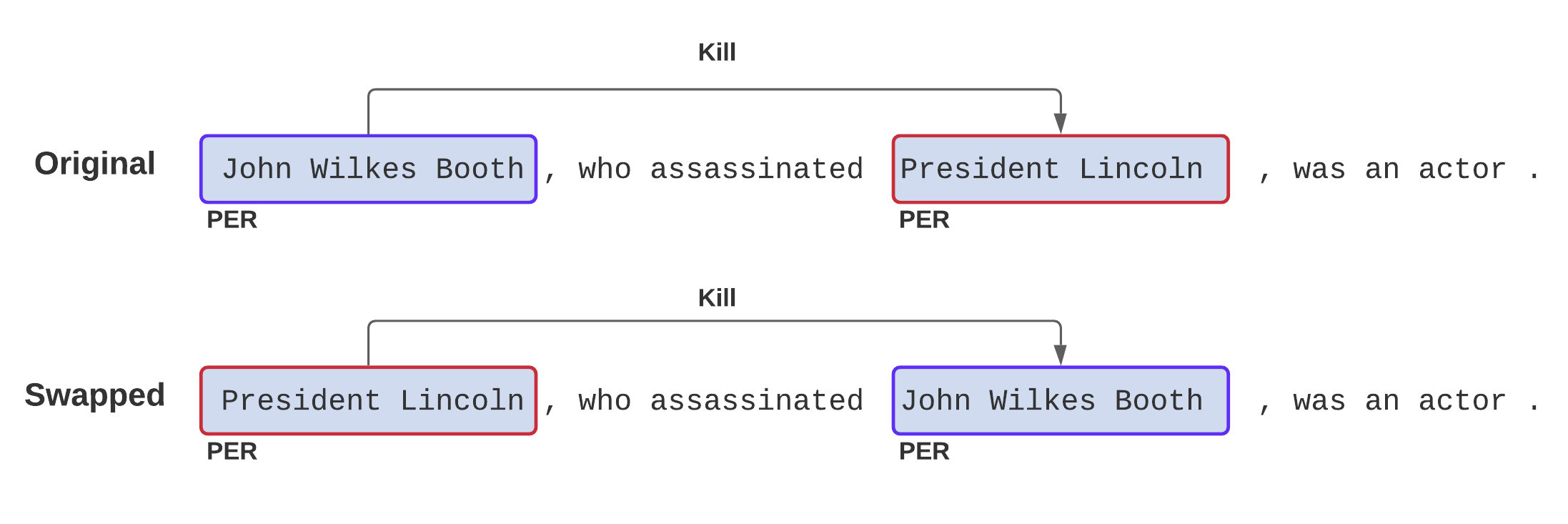}
\end{figure}

    \begin{table}[t]
    
    \caption[Résultats de l'étude comportementale d'inversion sujet-objet sur le test de CoNLL04.]{Résultats de l'étude comportementale d'inversion sujet-objet sur le test de CoNLL04. 
    Performance sur la phrase originale (O) et inversée (I).
    Le score F1 de REN est obtenu par micro-moyenne tandis que le score de relation RE Strict ne prend en compte que le type de relation spécifique.
    Le score revRE correspond à l'extraction non désirée de la relation originale dans une phrase inversée, symptomatique d'un effet de rétention.}
    \label{table:08:swap_rev}
    
    \centering
    \small
    
    \begin{tabularx}{.8\textwidth}{@{}l*{2}{Y}r*{2}{Y}r*{2}{Y}c@{}}
    
    \toprule
     
    \multicolumn{1}{c}{\multirow{2}{*}{F1}} & \multicolumn{2}{c}{Entité $\uparrow$} &  & \multicolumn{2}{c}{RE $\uparrow$} & & \multicolumn{2}{c}{revRE $\downarrow$} \\
      
     \cline{2-3} \cline{5-6} \cline{8-9}
     
     & O & I & & O & I & & O & I &\\
    
    \midrule
    \multicolumn{10}{c}{Kill} \\
    \midrule
     

    SpERT & 91.4 & \textbf{92.6} &  & 86.2 & 35.0 &  & - & 57.8 & \\
    
    TABTO & \textbf{92.0} & \textbf{92.8} &  & \textbf{89.6} & 27.6 &  & - & 59.5 & \\
    
    PURE & 90.5 & 90.7 &  & 84.1 & \textbf{52.3} &  & - & \textbf{14.3} & \\

    \midrule
    \multicolumn{10}{c}{Located in} \\
    \midrule
    
    
    SpERT & 88.6 & 87.7 &  & 75.0 & 24.9 &  & - & 33.5 & \\
    
    TABTO & \textbf{90.1} & \textbf{88.9} &  & \textbf{85.3} & 36.1 &  & - & 34.9 & \\
    
    PURE & 89.0 & 83.7 &  & 81.2 & \textbf{59.3} &  & - & \textbf{5.1} & \\
    \bottomrule
    
    \end{tabularx}
    \end{table}

La dégradation des performances lors de l'inversion est symptomatique d'un effet de rétention de l'information vue en entrainement au détriment d'une capacité de généralisation aux nouvelles relations.
Cela suggère que les modèles actuels basent excessivement leurs prédictions sur les arguments exacts des relations et pas assez sur leur contexte.

Ce comportement est validé par l'introduction de la métrique revRE qui mesure l'extraction de la relation originale dans les phrases inversées.
Nous observons que les modèles SpERT et TABTO prédisent plus souvent cette relation originale que la relation inversée pourtant exprimée dans la phrase inversée.

Au contraire, le modèle pipeline PURE semble plus robuste à la rétention.
Nous pouvons l'expliquer par l'utilisation de mots spéciaux correspondant au type des arguments qui sont utilisés pour la prédiction de relation.
Ainsi, il ne s'appuie moins directement sur les mentions exactes des arguments pour la prédiction.

\section[Utilisation Explicite des Poids d'Attention en EER]{Vers une Utilisation Explicite des Poids d'Attention en Extraction d'Entités et de Relations}
\label{sec:08:soa}

Après avoir mis en évidence le phénomène de rétention en EER, nous décrivons une architecture conçue pour incorporer plus explicitement le contexte dans la prédiction de relations afin d'améliorer la généralisation aux relations non rencontrées pendant l'entrainement.
Cette architecture est encore en cours de développement et nous ne rapportons donc pas de résultats quantitatifs ou  qualitatifs.

\subsection{Motivations}
La récente adoption de l'architecture Transformer \citep{Vaswani2017AttentionNeed} en TAL, suggère que le mécanisme d'auto-attention est un moyen efficace d'incorporer le contexte dans les représentations individuelles de mots.

Avant même l'appartition de BERT, \citet{strubell-etal-2018-linguistically} ont proposé d'incorporer de l'information linguistique dans l'attention d'un Transformer en entrainant une tete à prédire le parent syntactique de chaque mot d'une phrase.
Ils ont démontré que cela permettait d'améliorer les performances pour l'étiquetage de rôles sémantiques consistant notamment à trouver le sujet ou l'objet d'un verbe.

En complément de ce travail, de nombreux articles ont cherché à comprendre les raisons de la performance de BERT et certains ont mis en évidence \citep{clark-etal-2019-bert, jawahar-etal-2019-bert, kovaleva-etal-2019-revealing} que le préentrainement de Modèle de Language conduisait certaines têtes d'attention à refléter des relations syntactiques telles que la relation sujet-verbe ou verbe-objet.

 Forts de ces travaux, nous proposons d'utiliser explicitement ces têtes d'attention de BERT comme une représentation des relations syntactiques entre tous les mots d'une phrase.
 Nous pensons que la conservation de cette structure, perdue dans l'approche piecewise, peut permettre une meilleure incorporation du contexte en extraction de relations.

\begin{figure}[htp!]
\begin{adjustwidth*}{}{-.3\textwidth}
    \centering
    \caption[Illustration du modèle d'Attention du Second Ordre.]{Illustration du modèle d'Attention du Second Ordre.
    Les scores d'attention de BERT sont utilisés pour obtenir un alignement entre chaque paire d'arguments candidats $(i,j)$ et chaque mot $k$.
    Une paire d'arguments candidats est alors représentée par une somme pondérée des représentations des mots du contexte $h_k$ et non les représentations individuelles des arguments $h_i$ et $h_j$.
    Cela peut permettre de se focaliser davantage sur l'expression du prédicat qui est un meilleur indicateur de la présence d'une relation que ses arguments.} 
    
    \label{fig:08:soa}
    
    \includegraphics[width=1.1\textwidth]{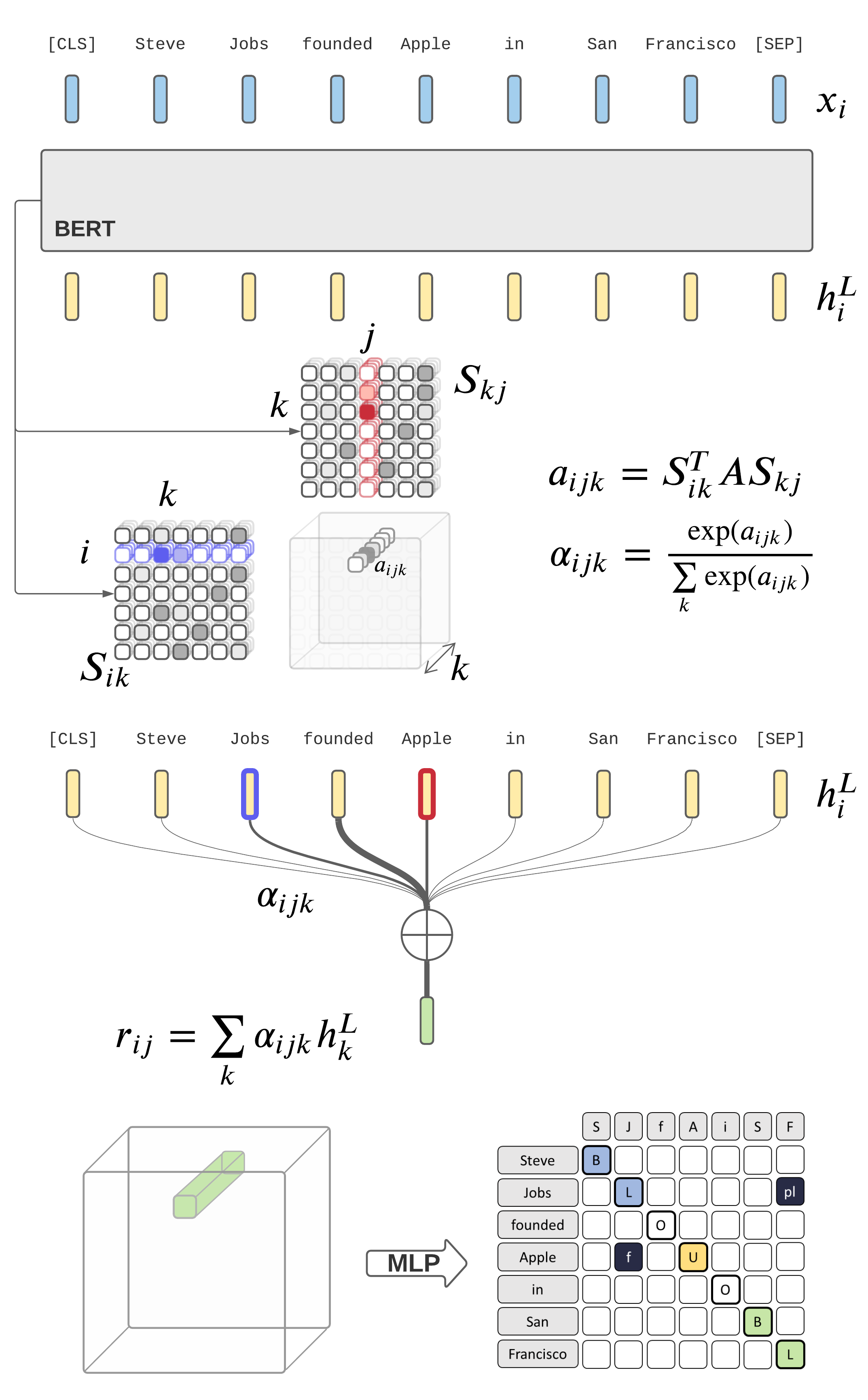}
\end{adjustwidth*}
\end{figure}

\subsection{Attention du Second Ordre}
L'idée majeure de notre modèle est similaire à l'attention: pondérer les représentations de mots du contexte pour se focaliser sur les plus pertinentes.
Cependant, contrairement à l'auto-attention où cette pondération est effectuée par l'interaction de chaque paire de mots, ici nous voulons une interaction dans chaque triplet de mots pour modéliser la structure (sujet, prédicat, objet).
Notre intuition est que pour éviter le surapprentissage des arguments exacts d'une relation, le modèle doit avoir plus directement accès à une représentation du prédicat, souvent exprimé par un mot clé du contexte.

Pour ce faire, nous proposons le modèle d'\textbf{Attention du Second Ordre (ASO)} qui utilise les poids d'auto-attention déjà calculés lors de toute utilisation de BERT comme des indications de cohérence des chemins reliant le sujet à l'objet par exactement un mot du contexte.
Cette architecture est illustrée dans la \autoref{fig:08:soa}.

\paragraph{}
Plus formellement, soit un mot candidat pour le sujet d'index $i$ et un mot candidat pour l'objet $j$, nous calculons un score d'attention  $\alpha_{ijk}$ qui a pour but de refléter la pertinence de chaque mot $k$ du contexte comme indicateur de la relation entre $i$ et $j$.
Ces poids sont utilisés pour pondérer pour chaque paire $ij$ les états cachés de la dernière couche de BERT $h_k^L$ traditionnellement utilisés comme représentations de mots.

\begin{equation}
    r_{ij} = \sum \limits_k \alpha_{ijk} h_k^L
\end{equation}

Nous concaténons les scores d'attention avant normalisation par Softmax $s_{ij}^l$ de chaque couche de BERT ainsi que leur transposés, pour permettre la modélisation des dépendances syntactiques dans les deux directions.

\begin{equation}
    S_{ij} = \mathrm{concat}(\{s_{ij}^l, s_{ji}^l | 1 \leq l \leq L\}) 
\end{equation}

Les $\alpha_{ijk}$ sont obtenus par une application bilinéaire des scores d'attention $S_{ik}$, entre le sujet candidat $i$ et un mot du contexte $k$, et les $S_{kj}$, entre ce même mot et l'objet candidat $j$.
Ces scores sont normalisés par une fonction Softmax.

\begin{equation}
    a_{ijk} = \frac{\mathrm{exp}(S_{ik}^T A S_{kj})}{\sum\limits_k \mathrm{exp}(S_{ik}^T A S_{kj})}
\end{equation}

L'application bilinéaire apprise est préférée à un simple produit scalaire pour permettre au modèle d'associer différentes des têtes d'attention de BERT.
Cela permet intuitivement d'apprendre à associer la tête la plus proche de refléter la relation Sujet-Verbe avec celle reflétant plutôt la relation Verbe-Objet.

Afin de permettre à ce mécanisme ASO d'apprendre plusieurs formes d'associations, peut-être pertinentes pour différents types de relations, nous pouvons envisager une extension multi-têtes.

Les résultats préliminaires n'indiquent pas de gain de performance par rapport à une représentation traditionnelle basée sur les arguments candidats tels que mesurés par les métriques globales sur CoNLL04 et ACE05.
Cela nous a poussé à développer la métrique séparée par recoupement lexical précédemment présentée et qu'il reste à utiliser sur un tel modèle.
De plus, des expériences sur SciERC qui présente un recoupement lexical plus limité pourraient révéler le potentiel de ce modèle.

\section{Conclusion}
\label{sec:08:ccl}

\subsection{Résultats et Contributions}
Dans cette thèse, nous analysons comment les modèles état de l'art basés sur les Modèles de Langue préentrainés généralisent à la détection de nouvelles entités et relations en Extraction d'Entités et de Relations.
Pour ce faire, nous menons plusieurs études empiriques dans lesquelles nous séparons les performances selon le recoupement des mentions et des relations avec le jeu d'entraînement.

Nous constatons que les modèles de langage pré-entraînés sont principalement bénéfiques pour détecter les mentions non connues, en particulier dans des genres de textes nouveaux.
Bien que cela les rende adaptés à des cas d'utilisation concrets, il existe toujours un écart de performance important entre les mentions connues et inconnues, ce qui nuit à la généralisation à de nouveaux faits.

En particulier, même les modèles d'Extraction d'Entités et de Relations les plus récents reposent sur une heuristique de rétention superficielle, basant plus leur
prédiction sur les arguments des relations que sur leur contexte.

Dans ce travail, nous consolidons également les bases de l'évaluation de l'Extraction dd'Entités et de Relations qui ont été sapées par des comparaisons incorrectes et nous proposons une base pour une évaluation et une compréhension plus fines des modèles concernant leur généralisation à de nouvelles relations.
Enfin, nous proposons le modèle d'Attention du Second Ordre qui a le potentiel d'améliorer l'incorporation du contexte et la généralisation en Extraction de Relations.

\subsection{Limitations}
Malgré ces résultats et contributions, nous pouvons identifier plusieurs limitations de nos travaux.
D'abord, nos travaux sont limités dans leur étendue à quelques jeux de données en anglais, principalement composés d'articles de presse.
Il pourrait ainsi être intéressant de les étendre à des genres de textes plus divers et dans plusieurs langues.

De plus, le recoupement lexical n'est pas le seul phénomène linguistique qui impacte les performances des modèles de REN et d'EER. En REN en particulier, les travaux exécutés simultanément aux notres par  \citet{arora-etal-2020-contextual} et \citet{fu-etal-2020-interpretable} identifient également la longueur des mentions ou le nombre de labels différents dans le jeu d'entrainement comme facteurs de performance.

Enfin, le recoupement lexical n'est pas seulement valable entre jeux d'évaluation et d'entrainement mais aussi avec le corpus de préentrainement des Modèles de Langue, beaucoup plus vaste.
Et qui doit avoir un impact sur la rétention de relations, comme suggéré par \citet{petroni-etal-2019-language}.

\subsection{Travaux Futurs}
Nous pouvons imaginer plusieurs pistes de travaux futurs.
Dans un premier temps, une extension immédiate de cette thèse est l'achèvement des expériences sur le modèle d'Attention du Second Ordre, afin de statuer sur ses capacités de répondre ou non à la problématique de généralisation à la détection de nouveaux faits.

Deuxièmement, malgré nos efforts, il est toujours difficile de dégager les avantages et inconvénients de chaque méthode d'EER et en particulier de trancher entre une approche pipeline ou une approche jointe.

Enfin, les problématiques de rétention nous invite à repenser la manière dont les jeux de données sont collectés.
En effet, pour faciliter le travail d'annotation les phrases contenant certaines paires d'arguments de relations connues ont été filtrées ce qui a augmenté de fait le recoupement lexical dans un dataset comme CoNLL04.
Ce principe aussi utilisé en supervision distante ne peut qu'encourager la mémorisation des arguments qui nuit à la généralisation.
Ainsi, nous proposons un filtrage inverse, basé sur le prédicat pour lequel la problématique de recoupement lexical nous semble moins problématique.
En effet, seule un faible nombre d'expressions différentes sont utilisées pour exprimer un type de relations donné.


\appendix
\cleardoublepage
\part*{Appendix}

\chapter[Contextual Embeddings in NER]{Contextual Embeddings in Named Entity Recognition}
This chapter proposes additional results and implementations details regarding our empirical study on the impact of contextual embeddings on generalization in NER presented in \autoref{sec:03:contener}. 

\section{Influence of Domain}

We first report out-of-domain evaluation of the BiLSTM-CRF model separated by OntoNotes genres in \autoref{table:supplementary-genre}.
We observe that the performance is always the best on broadcast news and the worst on magazine and web text domains. 
This is natural since CoNLL03 is composed of news stories.
Nevertheless, contextualized embeddings particularly benefit to the genres with the worst non contextual performances and lead to more homogeneous results.
We can also notice that the non contextual ELMo[0] representation improves over GloVe+char, mainly for the most difficult genres and for unseen mentions.

\begin{table}[h!]
\caption{Per genre out-of-domain micro-F1 scores of the BiLSTM-CRF model trained on CoNLL03. Results are averaged over 5 runs.}
\label{table:supplementary-genre}
\begin{center}
 \resizebox{\textwidth}{!}{
 \begin{tabular}{@{}lrrrrrrrrrrrrrrrrrrrrr@{}}
 \toprule

  & \multicolumn{4}{c}{broadcast conversation} & & \multicolumn{4}{c}{broadcast news} & & \multicolumn{4}{c}{news wire}\\
  \cmidrule{2-5} \cmidrule{7-10} \cmidrule{12-15}
  & Exact & Partial & New & All & & Exact & Partial & New & All & & Exact & Partial & New & All\\
  \midrule
BERT            & 96.7 & 81.5 & 77.8 & 87.2 &
                & 95.2 & 81.2 & 76.5 & 88.4 &
                & 95.0 & 83.9 & 74.0 & 84.7 \\
 
ELMo            & 95.6 & 77.5 & 76.1 & 85.0 &
                & 95.5 & 79.0 & 77.8 & 88.6 &
                & 92.4 & 81.1 & 73.2 & 82.9 \\
                
Flair           & 95.1 & 71.9 & 57.3 & 78.0 &
                & 94.8 & 77.4 & 71.4 & 86.5 &
                & 93.3 & 79.4 & 66.6 & 80.4 \\

\hdashline
ELMo[0]         & 95.5 & 76.2 & 68.3 & 82.6 &
                & 95.1 & 79.2 & 75.9 & 88.0 &
                & 91.4 & 78.3 & 66.4 & 80.0 \\

GloVe + char    & 94.8 & 75.8 & 61.5 & 80.4 &
                & 95.1 & 75.6 & 71.6 & 86.3 &
                & 92.5 & 75.2 & 61.2 & 78.0 \\
                
 \midrule
 & \multicolumn{4}{c}{magazine} & & \multicolumn{4}{c}{telephone conversation} & & \multicolumn{4}{c}{web text}\\
 \cmidrule{2-5} \cmidrule{7-10} \cmidrule{12-15}
  & Exact & Partial & New & All & & Exact & Partial & New & All & & Exact & Partial & New & All\\
 \midrule
  BERT          & 96.3 & 82.5 & 72.8 & 82.4 &
                & 93.1 & 88.9 & 78.1 & 84.5 &
                & 92.1 & 81.6 & 66.1 & 79.5 \\
  
 ELMo           & 97.7 & 71.3 & 68.7 & 78.1 &
                & 91.1 & 86.5 & 79.3 & 84.0 &
                & 92.5 & 82.4 & 66.2 & 79.9 \\
                
 Flair          & 96.8 & 67.4 & 55.7 & 71.1 &
                & 89.9 & 74.7 & 66.4 & 73.5 &
                & 91.2 & 77.2 & 51.4 & 72.1 \\

\hdashline            

ELMo[0]         & 97.3 & 67.1 & 60.6 & 73.4 &
                & 90.0 & 85.9 & 71.2 & 79.2 & 
                & 91.2 & 75.1 & 59.7 & 75.1 \\
                
 GloVe + char   & 97.1 & 64.6 & 56.0 & 70.7 &
                & 91.3 & 84.4 & 72.2 & 79.7 &
                & 91.1 & 69.1 & 48.9 & 69.2 \\
                
 \bottomrule
 \end{tabular}
 }
\end{center}
\end{table}

\section{Implementation Details}
\paragraph{Metric}
We compute separate Precision, Recall and F1 for \textit{exact}, \textit{partial} and \textit{new} mentions. 
For Precision we split predictions by novelty a posteriori.
All experiments are validated on the development set global micro-F1 score.

\paragraph{Pretrained Models}
The models are implemented in PyTorch.
We use the cased Glove 840B\footnotemark[1] embeddings trained on Common Crawl; the original implementations of ELMo\footnotemark[2] and Flair\footnotemark[3] and the PyTorch reimplementation of the pretrained BERT models from Huggingface\footnotemark[4].
An implementation of our experiments is released at \href{https://github.com/btaille/contener}{github.com/btaille/contener}.

\paragraph{Hyperparameters}
We use the IOBES tagging scheme for NER supervision and no data preprocessing.
Preliminary parameters search leads to fix a batch size of 64 and a learning rate of 0.001 across experiments.
We use a 0.5 dropout rate for regularization at the embedding layer and after the BiLSTM or linear projection.
In the char-BiLSTM, the character embeddings have dimension 100 and the BiLSTM has a 25 hidden size in each direction, as in \citep{lample-etal-2016-neural}.
The maximum number of epochs is set to 100 and we use early stopping with patience 5 validated on development global micro-F1.
For each configuration, we use the best performing optimization method on development set between SGD and Adam with $\beta_1=0.9$ and $\beta_2=0.999$.
In practice, SGD leads to better results for GloVe baselines which have lower dimensions.

All experiments are run with five different random seeds and we report the average of these runs.

\footnotetext[1]{\href{https://nlp.stanford.edu/projects/glove}{nlp.stanford.edu/projects/glove}}
\footnotetext[2]{\href{https://github.com/allenai/allennlp}{github.com/allenai/allennlp}}
\footnotetext[3]{\href{https://github.com/zalandoresearch/flair}{github.com/zalandoresearch/flair}}
\footnotetext[4]{\href{https://github.com/huggingface/transformers}{github.com/huggingface/transformers}}

\subsection{Mapping OntoNotes to CoNLL03}
For out-of-domain evaluation, we remap OntoNotes annotations to match CoNLL03 types.
We leave ORG and PER types as is and map LOC + GPE in OntoNotes to LOC in CoNLL03 and NORP + LANGUAGE to MISC.
The obtained dataset is referred to as Ontonotes$^\ast$.

Contrary to \citet{Augenstein2017GeneralisationAnalysis}, we choose to keep the MISC tag from CoNLL03 for our mapping and find that it corresponds to NORP and LANGUAGE in OntoNotes. Additionally, our mapping differs for LOC since they add FACILITY to GPE and LOC in OntoNotes. We find that some Facilities in OntoNotes indeed fit LOC but some would rather be classified as ORG in CoNLL03. Likewise, some Events fit MISC but they are exceptions.  Nevertheless, these two types only represent 3\% of entity types mentions in OntoNotes and are thus negligible.

For normalization, we remove ``the" at the beginning and ``'s" at the end of OntoNotes mentions.

\subsection{Additional Dataset Statistics}
We report general statistics of CoNLL03 and OntoNotes in \autoref{table:supplementary-datasets}. We can see that OntoNotes is much larger than CoNLL03 with around four times the number of mention occurrences and five times the number of tokens. 

Development set lexical overlaps for CoNLL03, OntoNotes and OntoNotes$^{\ast}$ are shown in \autoref{table:03:overlap-dev}. The numbers are close to test sets overlaps except for CoNLL03 for which the overlap is worse in the dev set. This raises an additional issue since models are validated with a different distribution of mention novelty which is even more biased towards exact match mentions. 

Finally, we report the lexical overlap of the original OntoNotes dataset for every entity and value type. We expected that value types mostly contribute to partial overlap because of matching units but it seems that even the numbers are overlapping leading to 72\% exact overlap against 64\% for entity types.

\begin{table}[t]
\caption{Dataset statistics. We report both the number of mention occurrences and unique mentions. We take type into account to compute the latter.}
\label{table:supplementary-datasets}

\begin{center}
\resizebox{\textwidth}{!}{
\begin{tabular}{@{}lrrrrrrrrrrr@{}}
\toprule
                    &   \multicolumn{3}{c}{CoNLL03}     &   &  \multicolumn{3}{c}{OntoNotes} &   &  \multicolumn{3}{c}{OntoNotes$^\ast$} \\
\cmidrule{2-4} \cmidrule{6-8} \cmidrule{10-12}
                    & Train     & Dev       & Test      &   & Train     & Dev       & Test &   & Train     & Dev       & Test\\
\midrule
Sentences           & 14,041    & 3,250     & 3,453     &   & 59,924    & 8,528     & 8,262 &   & 59,924    & 8,528     & 8,262\\
Tokens              & 203,621   & 51,362    & 46,435    &   & 1,088,503 & 147,724   & 152,728  &   & 1,088,503 & 147,724   & 152,728 \\
Mentions            & 23,499    & 5,942     & 5,648     &   & 81,828    & 11,066     & 11,257  &   & 52,342 & 7,112 & 7,065 \\
Unique              & 8,220     & 2,854     & 2,701     &   & 25,707     & 4,935     & 4,907  &   & 14,661 & 2,768 & 2,663 \\

\midrule
                    &        &    & &             &  \multicolumn{3}{c}{WNUT} &   &  \multicolumn{3}{c}{WNUT$^\ast$} \\
 \cmidrule{6-8} \cmidrule{10-12}
                    &   &        &       &          & Train     & Dev       & Test &        & Train         & Dev       & Test\\
\midrule
Sentences           &    &      &      &            & 3,394     & 1,009     & 1,287 &       & 3,394         & 1,009     & 1,287\\
Tokens              &  &     &     &                & 62,730    & 15,733    & 23,394  &     & 62,730        & 15,733    & 23,394 \\
Mentions            &    &      &       &           & 1,975     & 836       & 1,079  &      &  1,429       &    578  & 645 \\
Unique              &    &    &      &              & 1,604    & 747        & 955    &      & 1,120          &  499    & 561 \\

\bottomrule
\end{tabular}
}
\end{center}
\end{table}

\begin{table}[h]
\begin{adjustwidth*}{}{-.3\textwidth}
\caption{Per type lexical overlap of dev mention occurrences with respective train set in-domain and with CoNLL03 train set in the out-of-domain scenario. (EM / PM = \textit{exact / partial match})
}
\label{table:03:overlap-dev}
\scriptsize
\begin{tabularx}{1.3\textwidth}{@{}ll*{5}{Y}rYr*{5}{Y}rYr*{4}{Y}r@{}}
\toprule
& &\multicolumn{5}{c}{CoNLL03} & & ON &  & \multicolumn{5}{c}{OntoNotes$^{\ast}$} & & WNUT & & \multicolumn{4}{c}{WNUT$^{\ast}$}\\
\cmidrule{3-7} \cmidrule{9-9} \cmidrule{11-15} \cmidrule{17-17} \cmidrule{19-22}
 & & LOC & MISC & ORG & PER & ALL & & ALL& & LOC & MISC & ORG & PER & ALL & &  ALL &  & LOC & ORG & PER & ALL\\ 
\midrule
\parbox[t]{3mm}{\multirow{3}{*}{\rotatebox[origin=c]{90}{\textbf{Self}}}} 
& EM & 86\% & 79\% & 67\% & 43\% & 67\% &
        & 67\% & 
       
        & 85\% & 93\% & 58\% & 48\% & 69\% &
         & - & 
         & - & - &- &- \\
        
& PM & 2\% & 7\% & 19\% & 33\% & 16\% & 
        & 23\% &
        
        & 6\% & 1\% & 31\% & 35\% & 20\% &
        & 15\% & 
        & 15\% & 3\% & 15\% & 14\% \\
        
& New  & 14\% & 22\% & 29\% & 43\% & 28\% &
        & 9\% & 
        
        & 7\% & 5\% & 14\% & 15\% & 11\% &
        & 85\% &
        & 85\% & 97\% & 85\% & 86\% \\
        
\midrule
\parbox[t]{3mm}{\multirow{3}{*}{\rotatebox[origin=c]{90}{\textbf{CoNLL03}}}}  
        &  EM   &-  & - &  -& - &-  &
        &- & 
        
        & 71\% & 85\% & 22\% & 12\% & 44\%  &
        & - & 
        & 24\% & 12\% & 6\% & 9\%\\
        
& PM &-  & - &-  &  -& - & 
        & - &
        & 7\% & 4\% & 47\% & 43\% & 27\% &
        & - & 
        & 9\% & 15\% & 19\% & 17\% \\
        
& New  & -  & - & -  & - & - &
        & - & 
        & 22\% & 11\% & 31\% & 44\% & 29\%&
        & - &
        & 69\% & 74\% & 74\% & 73\%\\
        
\bottomrule
\end{tabularx}
\end{adjustwidth*}

\end{table}

\begin{table}[h]

\begin{adjustwidth*}{}{-.3\textwidth}
 \caption{Lexical overlap with train set in English OntoNotes for the eleven entity types and seven value types.}
\label{table:OntoNotes-overlap}
 \begin{center}
 \resizebox{1.3\textwidth}{!}{
 
    \begin{tabular}{@{}llccccccccccc@{}}
    \toprule
                    &  & \multicolumn{11}{c}{Entities} \\
    \cmidrule{3-13}
                    &  & EVENT & FAC & GPE & LANG  & LAW & LOC & NORP & ORG & PER & PROD & WOA \\  
    \midrule
    \textbf{Dev}    & Exact     & 22\%  & 20\%  & 87\%  & 79\%  & 15\%   & 48\%  & 93\%  & 55\%  & 47\%  & 42\%  & 15\%  \\
                    & Partial   & 39\%  & 49\%  & 4\%   & 0\%   & 70\%   & 36\%  & 1\%   & 33\%  & 35\%  & 25\%  & 54\%  \\ 
                    & New       & 38\%  & 31\%  & 9\%   & 21\%  & 15\%   & 16\%  & 6\%   & 11\%  & 17\%  & 33\%  & 30\%  \\ 
    \noalign{\vskip 3mm}
    \textbf{Test}   & Exact     & 52\%  & 17\%  & 89\%  & 77\%  & 20\%  & 60\%  & 87\%  & 51\%  & 48\%  & 47\%  & 22\%  \\ 
                    & Partial   & 37\%  & 65\%  & 4\%   & 0\%   & 55\%  & 31\%  & 8\%   & 35\%  & 37\%  & 16\%  & 45\%  \\ 
                    & New       & 11\%  & 18\%  & 7\%   & 23\%  & 25\%  & 9\%   & 5\%   & 14\%  & 15\%  & 37\%  & 33\%  \\ 
    \midrule
                    & &  \multicolumn{7}{c}{Values}  &   & \multicolumn{3}{c}{All} \\
    \cmidrule{3-9}  \cmidrule{11-13}
                    & & CARD  & DATE  & MONEY & ORD   & PERC  & QUANT & TIME    &   & ENT   & VAL   & ALL \\
    \midrule
    \textbf{Dev}    & Exact   & 82\%  & 74\%  & 39\%  & 93\%  & 70\%  & 13\%   & 67\%   &   &  64\%   & 72\% & 67\% \\  
                    & Partial & 11\%  & 24\%  & 55\%  & 0\%   & 20\%  & 86\%   & 30\%   &   &  23\%   & 23\% & 23\% \\ 
                    & New     & 7\%   & 3\%   & 6\%   & 7\%   & 10\%  & 1\%    & 2\%    &   &  13\%   & 5\%  & 10\% \\ 
    \noalign{\vskip 3mm}
    \textbf{Test}   & Exact   & 83\%  & 77\%  & 32\%  & 97\%  & 62\%  & 24\%  & 58\%    &   &  64\%   & 72\% & 67\% \\  
                    & Partial & 11\%  & 22\%  & 63\%  & 1\%   & 24\%  & 74\%  & 38\%    &   &  24\%   & 24\% & 24\% \\  
                    & New     & 6\%   & 1\%   & 4\%   & 3\%   & 13\%  & 2\%   & 4\%     &   &  12\%   &  4\% & 9\% \\  
    \bottomrule
    \end{tabular}}

\end{center}
\end{adjustwidth*}
\end{table}

\chapter[Let's Stop Incorrect Comparisons in ERE !]{Let's Stop Incorrect Comparisons in End-to-end Relation Extraction !}
This chapter proposes additional results and implementations details regarding our work presented in  \autoref{sec:05:sincere} on the identification of incorrect comparisons in the ERE literature and the double ablation study of the Span-level NER modeling and BERT representations from the SpERT model. 

\renewcommand{\arraystretch}{1.2}

\section{Additional Implementation Details}
We used an Nvidia V100 server with 16BG VRAM for our experiments. 
They can be run with a single Nvidia GTX 1080 with 8GB VRAM with the same hyperparameters as experimented during prototyping.
We report the average number of epochs and time for every configuration in \autoref{table:app_time}.
We report the number of parameters in our models in \autoref{table:app_params}.

\begin{table}[h]

\caption{Average number of epochs before early stopping and corresponding runtime in minutes for a training with early stopping on the dev RE Strict $\mu$ F1 score.}
\label{table:app_time}

\centering
\small
\begin{tabularx}{.7\textwidth}{@{}lYRlYR@{}}

\toprule
            & \multicolumn{2}{c}{CoNLL04} & & \multicolumn{2}{c}{ACE05}  \\
            \cline{2-3} \cline{5-6}
Model       & Ep. &  Time & &  Ep. &  Time \\

\midrule

BERT + Span & 52 & 166 & & 25 & 160\\
BERT + BILOU& 16 & 20 & & 22 & 50\\
BiLSTM + Span & 20 & 52 & & 17 & 100\\
BiLSTM + BILOU & 14 & 7 & & 14 & 18\\
\bottomrule
\end{tabularx}
\end{table}

\begin{table}[h]

\caption{Number of parameters in the different modules of our models.}
\label{table:app_params}
\small
\centering
\begin{tabularx}{.7\textwidth}{@{}lRR@{}}

\toprule
           
Module      &   CoNLL04 &  ACE05  \\
\midrule

BERT Embedder & 108 M & 108 M\\
GloVe Embedder & 2.6 M & 5.6 M\\
charBiLSTM  & 34 k & 35 k\\
\midrule
BiLSTM Encoder & 2.3 M & 2.3 M\\
\midrule
Span NER & 4 k & 7 k \\
BILOU NER & 13 k & 22 k\\
\midrule
RE Decoder & 12 k & 14 k \\

\bottomrule
BERT + Span & 108 M & 108 M \\
BERT + BILOU & 108 M & 108 M \\
BiLSTM + Span & 5 M & 8 M \\
BiLSTM + BILOU & 5 M & 8 M \\
\bottomrule
\end{tabularx}

\end{table}

\section{Additional Datasets Statistics}
We provide more detailed statistics on the two datasets we used for our experimental study in \autoref{table:app_data_stats_conll04} and \autoref{table:app_data_stats_ace05}.
We believe that reporting the number of sentences, entity mentions and relation mentions per training partition is a minimum to enable sanity checks ensuring data integrity.

\begin{table}[h]

\caption[Detailed statistics of our CoNLL04 dataset compared with previous works]{Detailed statistics of our CoNLL04 dataset, as preprocessed by \citet{Eberts2020Span-basedPre-training} \protect\footnotemark[1].
We compare to previously reported statistics \citep{roth-yih-2004-linear,gupta-etal-2016-table,adel-schutze-2017-global}.
The test sets from \citep{gupta-etal-2016-table}, \citep{adel-schutze-2017-global} and \citep{Eberts2020Span-basedPre-training} are supposedly the same but we observe differences.
Only \citet{Eberts2020Span-basedPre-training} released their complete training partition.}
\label{table:app_data_stats_conll04}

\small
\centering
\begin{tabularx}{.8\textwidth}{@{}ll*{4}{R}@{}}

\toprule
            & Reference & Train & Dev & Test & Total  \\
\midrule
Sentences & (R\&Y, 04)  &   -   & -   &  -   &  1437 \\
            & (G, 16) & 922 & 231 & 288 & 1441 \\
            & Ours &  922 & 231 & 288 & 1441 \\
\midrule
Tokens    & (A\&S, 17) & 23,711 & 6,119 & 7,384 & 37,274\\
          &Ours & 26,525 & 6,993 & 8,336 & 41,854 \\
\midrule
Entities &  (R\&Y, 04)  &   -   & -   &  -   & 5,336 \\
        &  (A\&S, 17) & 3,373 & 858 & 1,071 & 5,302\\
        & Ours & 3,377 & 893 & 1,079 & 5,349 \\
\midrule
Relations &  (R\&Y, 04)  & - & - & - & 2,040  \\
          & (A\&S, 17) & 1,270 & 351 & 422 & 2,043 \\    
          & Ours & 1,283 & 343 & 422 & 2,048\\
\bottomrule

\end{tabularx}
\end{table}

\begin{table}[h]

\caption[Detailed statistics of our ACE05 dataset compared with previous works]{Detailed statistics of our ACE05 dataset, following \citet{miwa-bansal-2016-end}'s preprocessing scripts\protect\footnotemark[2]. 
We compare to previously reported statistics by \citep{li-ji-2014-incremental}.
The large difference in the number of sentences is likely due to a different sentence tokenizer.
}
\label{table:app_data_stats_ace05}

\small
\centering
\begin{tabularx}{.8\textwidth}{@{}ll*{4}{R}@{}}

\toprule
            & Reference & Train & Dev & Test & Total  \\
\midrule
Documents   & (L\&J, 14) & 351 & 80 & 80 & 511\\
            & Ours & 351 & 80 & 80 & 511\\
\midrule
Sentences   & (L\&J, 14)  &  7,273 & 1,765 & 1,535  & 10,573 \\
            
            & Ours & 10,051 & 2,420 & 2,050 & 14,521 \\
\midrule
Tokens    & Ours & 144,783 & 35,548 & 30,595 & 210,926 \\
\midrule
Entities & (L\&J, 14) &  26,470   & 6,421   &  5,476  & 38,367 \\
         & Ours & 26,473 & 6,421 & 5,476 & 38,370 \\
\midrule
Relations & (L\&J, 14)  & 4,779 & 1,179 & 1,147 & 7,105 \\
          & Ours & 4,785 & 1,181 & 1,151 & 7,117\\
\bottomrule

\end{tabularx}
\end{table}

\footnotetext[1]{\href{https://github.com/markus-eberts/spert}{github.com/markus-eberts/spert}}
\footnotetext[2]{\href{https://github.com/tticoin/LSTM-ER}{github.com/tticoin/LSTM-ER}}

\section{Additional Comparison of ACE05 and CoNLL04}
\label{app:sincere:comparison_ace_conll}
\begin{figure}[h]
    \caption{Distribution of the number of entity and relation mentions per sentence in ACE05 and CoNLL04.}
    \label{fig:add_ent_rel}
    \centering
    \includegraphics[width=0.93\textwidth]{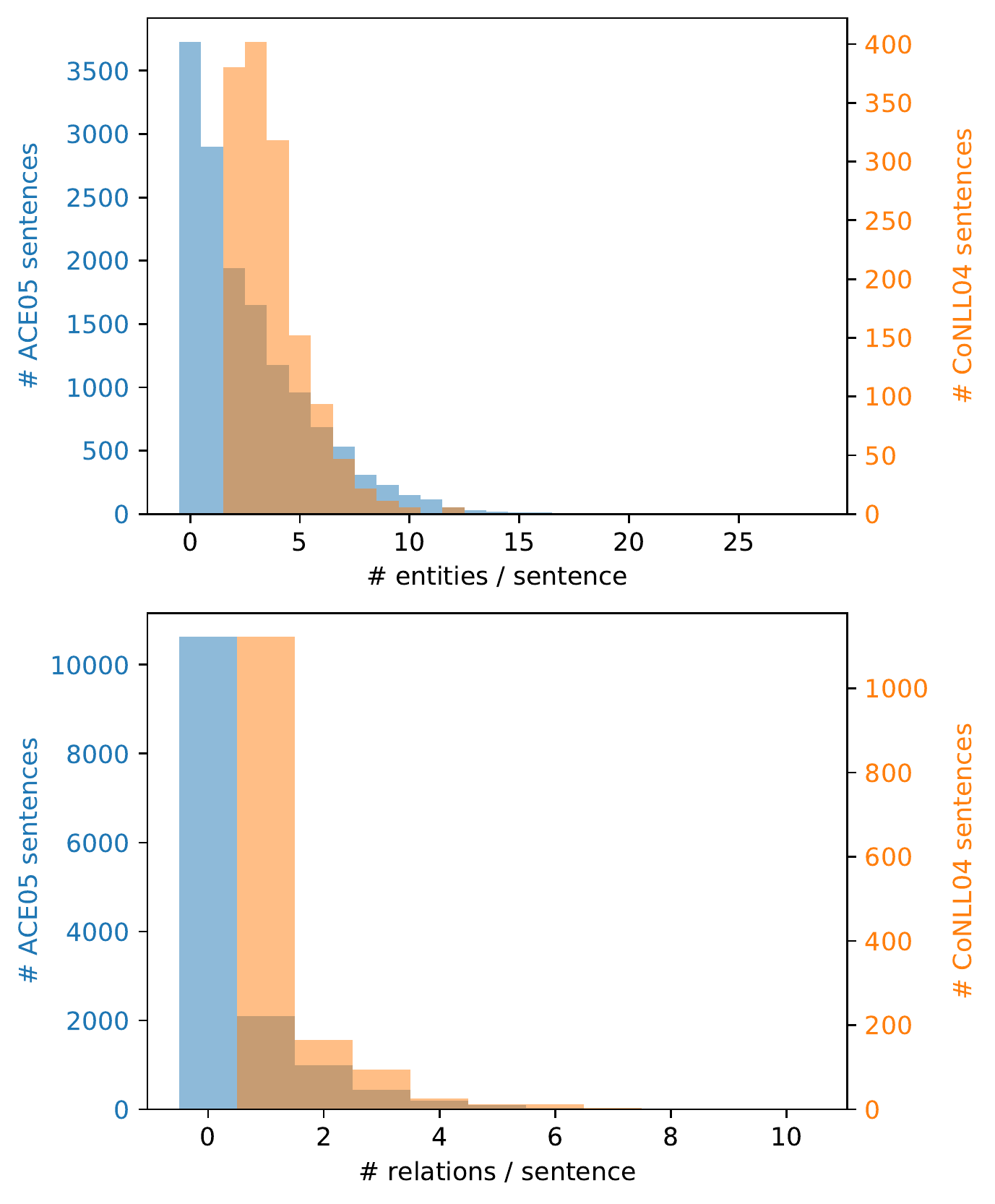}
\end{figure}

\begin{figure}[h]
    \begin{adjustwidth*}{}{-.3\textwidth}
    \centering
    \begin{minipage}{0.5\textwidth}
        \centering

        \caption{Occurrences of each relation / argument types combination in CoNLL04.}
        \label{fig:add_conll04}
    
    \includegraphics[width=\textwidth]{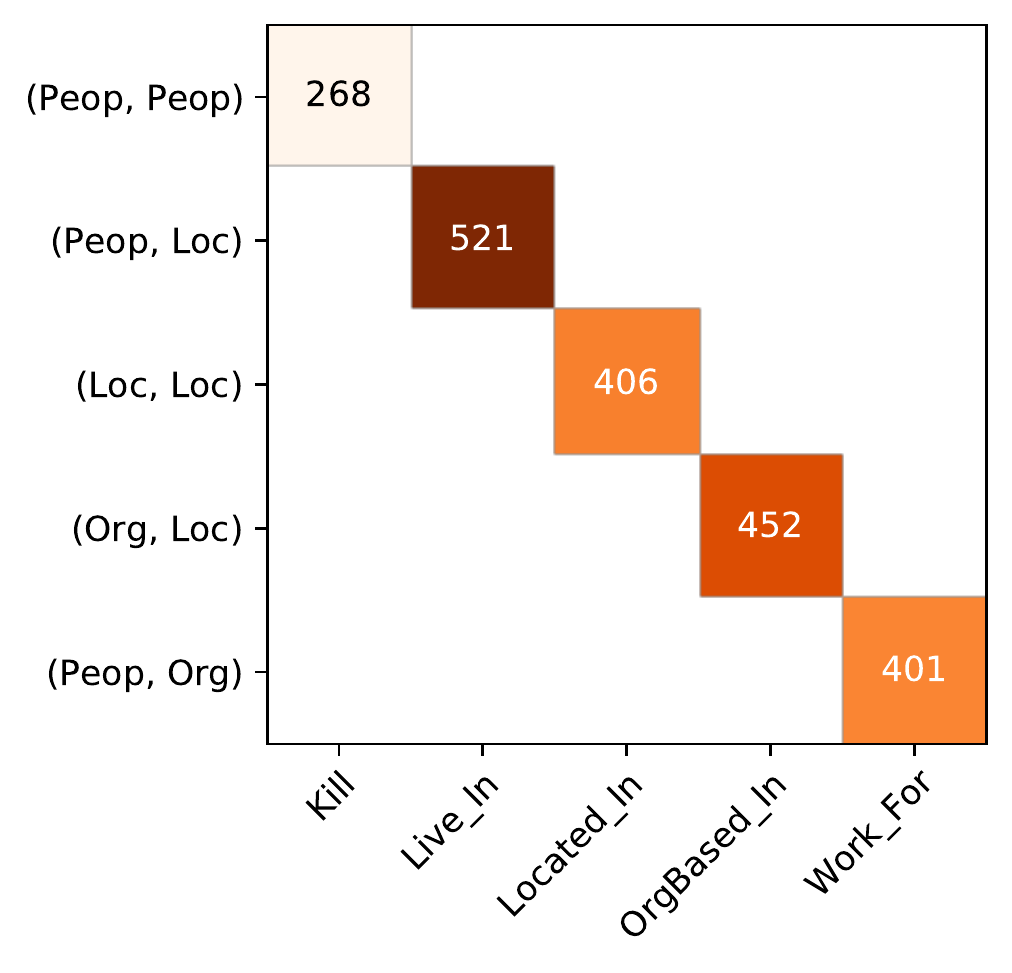}
    
    \end{minipage}
    \hfill
    \begin{minipage}{0.7\textwidth}
        \centering
        \caption{Occurrences of each relation / argument types combination in ACE05.}
        \label{fig:add_ace05}
        
        \includegraphics[height=\textheight]{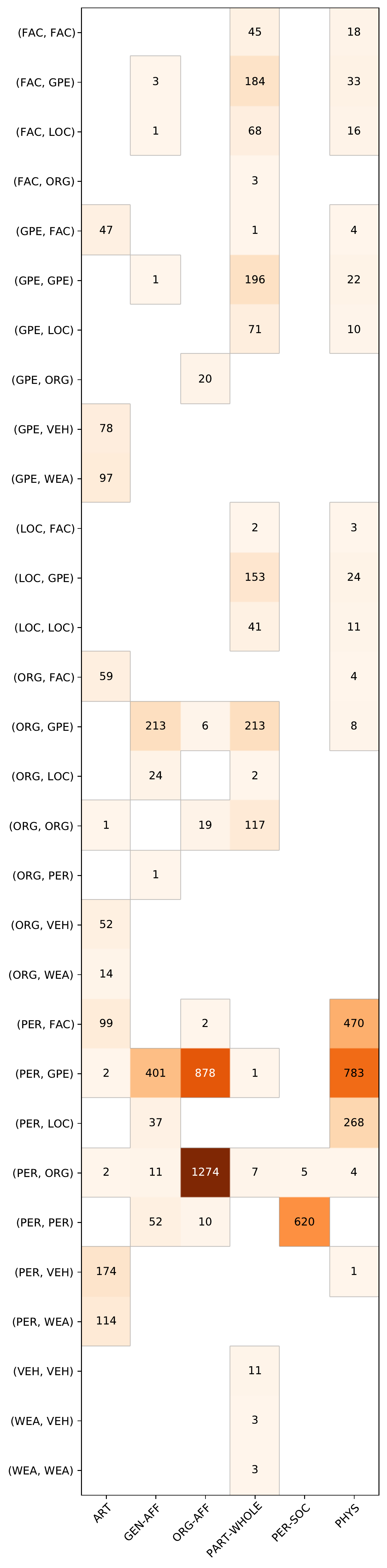}
    \end{minipage}
    \end{adjustwidth*}
\end{figure}

ACE05 and CoNLL04 have key differences we propose to visualize with global statistics.
First, in CoNLL04 every sentence contains at least two entity mentions and one relation while the majority of ACE05 contains no entities nor relations as depicted in Fig.~\autoref{fig:add_ent_rel}.We can also notice that among sentences containing relations, a higher proportion of ACE05 contain several of them.
Second, the variety of combinations between relation types and argument types makes RE on ACE05 much more difficult than on CoNLL04 (Fig. \autoref{fig:add_conll04} and \autoref{fig:add_ace05}).

\chapter[Separating Retention from Extraction in ERE]{Separating Retention from Extraction in End-to-end Relation Extraction}
This chapter proposes additional results and implementations details regarding our empirical study on isolating the impact of lexical overlap on the performance of state-of-the-art End-to-end Relation Extraction presented in \autoref{sec:05:retention}.

\section{Implementation Details}
\label{app:05:retex-implementation}

For every model, we use the original code associated with the papers with the default best performing hyperparameters unless stated otherwise.
We run 5 runs on a single NVIDIA 2080Ti GPU for each of them on each dataset. 
For CoNLL04 and ACE05, we train each model with both the cased and uncased versions of BERT$_{BASE}$ and only keep the best performing setting. 

\paragraph{PURE} \cite{zhong-chen-2021-frustratingly} \footnotemark[1] 
We use the approximation model and limit use a \textit{context window} of 0 to only use the current sentence for prediction and be able to compare with other models.
For ACE05, we use the standard\textit{ bert-base-uncased} LM but use the \textit{bert-base-cased} version on CoNLL04 which results in a significant +2.4 absolute improvement in RE Strict micro F1 score.

\paragraph{SpERT} \cite{Eberts2020Span-basedPre-training} \footnotemark[2] We use the original implementation as is with \textit{bert-base-cased} for both ACE05 and CoNLL04 since the uncased version is not beneficial, even on ACE05 where there are fewer proper nouns.
For the Ent-SpERT ablation, we simply remove the max-pooled context representation from the final concatenation in the RE module. 
This modifies the RE classifier's input dimension from the original 2354 to 1586.

\paragraph{Two are better than one (TABTO)} \cite{wang-lu-2020-two} \footnotemark[3] We use the original implementation with \textit{bert-base-uncased} for both ACE05 and CoNLL04 since the cased version is not beneficial on CoNLL04.

\footnotetext[1]{\href{https://github.com/princeton-nlp/PURE}{github.com/princeton-nlp/PURE}}
\footnotetext[2]{\href{https://github.com/lavis-nlp/spert}{github.com/lavis-nlp/spert}}
\footnotetext[3]{\href{https://github.com/LorrinWWW/two-are-better-than-one}{github.com/LorrinWWW/two-are-better-than-one}}

\section{Datasets Statistics}

We present general datasets statistics in \autoref{table:data_stats}.

    \begin{table}[h]
    
    \caption{ERE Datasets Statistics}
    \label{table:data_stats}
    
    \centering
    \begin{tabularx}{.7\textwidth}{@{}lRRR@{}}
    
    \toprule
   
    \textbf{ACE05} & Train & Dev & Test \\
  
    \midrule
    
    Sentences & 10,051 & 2,424 & 2,050 \\
    Mentions & 26,473 & 6,338 & 5,476 \\
    Relations & 4,788 & 1,131 & 1,151 \\
    
    \midrule
    \textbf{CoNLL04} & Train & Dev & Test \\
    \midrule
    
    Sentences & 922 & 231 & 288 \\
    Mentions & 3,377 & 893 & 1,079 \\
    Relations & 1,283 & 343 & 422 \\
    
    \midrule
    \textbf{SciERC} & Train & Dev & Test \\
    \midrule
    
    Sentences & 1,861 & 275 & 551 \\
    Mentions &  5,598 & 811 & 1,685 \\
    Relations & 3,219 & 455 & 974 \\
    
    \bottomrule
    \end{tabularx}
    \end{table}

We also compute average values of some entity and relation attributes inspired by \cite{fu-etal-2020-interpretable} and reported in \autoref{table:data_consistency}.

We report two of their entity attributes: \textbf{entity length} in number of tokens (\textbf{eLen}) and \textbf{entity label consistency (eCon)}.
Given a test entity mention, its label consistency is the number of occurrences in the training set with the same type divided by its total number of occurrences.
It is zero for unseen mentions.
Because eCon reflects both the ambiguity of labels for seen entities and the proportion of unseen entities, we propose to introduce the \textbf{eCon*} score that only averages label consistency of seen mentions and \textbf{eLex}, the proportion of entities with lexical overlap with the train set.

We introduce similar scores for relations. \textbf{Relation label consistency (rCon)} extends label consistency for triples. \textbf{Argument types label constitency  (aCon)} considers the labels of every pair of mentions of corresponding types in the training set. Because pairs of types are all seen during training we do not decompose aCon into aCon* and aLex. \textbf{Argument length  (aLen)} is the sum of the lengths of the head and tail mentions. \textbf{Argument distance (aDist)} is the number of tokens between the head and the tail of a relation.

We present a more complete report of overall Precision, Recall and F1 scores that can be interpreted in light of these statistics in \autoref{table:overlap_prf}.

    \begin{table}[t]
    \begin{adjustwidth*}{}{-.3\textwidth}
    \caption{Average of some entity and relation attributes in the test set.}
    \label{table:data_consistency}
    
    \centering
    \begin{tabularx}{\textwidth}{@{}l*{4}{Y}r*{8}{Y}@{}}
    
    \toprule
   
           & \multicolumn{4}{c}{Entities} & & \multicolumn{8}{c}{Relations} \\
            
            \cline{2-5} \cline{7-12}
            & eCon & eCon* & eLex & eLen & 
            & rCon & rCon* & rLex & aCon & aLen & aDist \\
           
    \midrule
    
    \textbf{ACE05}   & 65\% & 78\% & 82\% & 1.1 &
            & 15\% & 62\% & 23\%  & 7.1\% & 2.3 & 2.8 \\
    
    \textbf{CoNLL04} & 49\% & 98\% & 50\% & 1.5 &
            & 21\% & 91\% & 23\% & 29\% & 3.8 & 5.8 \\
            
    \textbf{SciERC}  & 17\% & 74\% & 23\% & 1.6 &
            & 0.4\% & 74\% & 0.5\% & 13\% & 4.7 & 5.3\\
            
    \bottomrule
    
    \end{tabularx}
    \end{adjustwidth*}
    \end{table}

    \begin{table}[h]
    \begin{adjustwidth*}{}{-.3\textwidth}
    \caption[Overall micro-averaged Test NER and Strict RE Precision, Recall and F1 scores.  Average and standard deviations on five runs.]{Overall micro-averaged Test NER and Strict RE Precision, Recall and F1 scores.  Average and standard deviations on five runs.
    We can observe that the recall of the heuristic is correlated with the proportions of seen entities or triples (eLex or rLex).
    Its particularly high precision on CoNLL04 seems rather linked to the important label consistency of seen entities and relation (eCon* and rCon*).}
    \label{table:overlap_prf}
    
    \small
    \centering
    \begin{tabularx}{1.3\textwidth}{@{}l*{3}{Y}r*{3}{Y}r*{3}{Y}c@{}}
    
    \toprule
    \multicolumn{1}{c}{\multirow{2}{*}{$\mu$ F1}}
    
     & \multicolumn{3}{c}{NER} &  & \multicolumn{3}{c}{RE Boundaries} & & \multicolumn{3}{c}{RE Strict} \\
    \cline{2-4} \cline{6-8} \cline{10-12} 
    & P & R & F1 & & P & R & F1 & & P & R & F1 &  \\
    
    \midrule
    & \multicolumn{12}{c}{\textbf{ACE05}} \\
    \midrule
    
    heuristic & 44.7$_{\stddev{\phantom{0.0}}}$ & 71.9$_{\stddev{\phantom{0.0}}}$ & 55.1$_{\stddev{\phantom{0.0}}}$ &  & 23.6$_{\stddev{\phantom{0.0}}}$ & 22.3$_{\stddev{\phantom{0.0}}}$ & 23.0$_{\stddev{\phantom{0.0}}}$ &  & 21.4$_{\stddev{\phantom{0.0}}}$ & 20.2$_{\stddev{\phantom{0.0}}}$ & 20.8$_{\stddev{\phantom{0.0}}}$ & \\

    Ent-SpERT & 86.7$_{\stddev{0.3}}$ & 86.3$_{\stddev{0.3}}$ & 86.5$_{\stddev{0.2}}$ &  & 56.7$_{\stddev{1.0}}$ & 57.4$_{\stddev{0.7}}$ & 57.0$_{\stddev{0.8}}$ &  & 53.5$_{\stddev{1.0}}$ & 54.2$_{\stddev{0.8}}$ & 53.9$_{\stddev{0.8}}$ & \\
    
    SpERT & 87.2$_{\stddev{0.2}}$ & 86.5$_{\stddev{0.3}}$ & 86.8$_{\stddev{0.2}}$ &  & 68.1$_{\stddev{1.1}}$ & 60.5$_{\stddev{0.5}}$ & 64.0$_{\stddev{0.6}}$ &  & 64.4$_{\stddev{1.1}}$ & 57.2$_{\stddev{0.4}}$ & 60.6$_{\stddev{0.5}}$ & \\
    
    TABTO & 86.7$_{\stddev{0.3}}$ & 88.3$_{\stddev{0.6}}$ & 87.5$_{\stddev{0.2}}$ &  & 71.0$_{\stddev{2.7}}$ & 62.5$_{\stddev{2.5}}$ & 66.4$_{\stddev{1.3}}$ &  & 66.1$_{\stddev{2.6}}$ & 58.1$_{\stddev{2.1}}$ & 61.8$_{\stddev{1.1}}$ & \\
    
    PURE & 88.8$_{\stddev{0.3}}$ & 88.6$_{\stddev{0.1}}$ & 88.7$_{\stddev{0.1}}$ &  & 67.4$_{\stddev{0.8}}$ & 63.0$_{\stddev{0.8}}$ & 65.1$_{\stddev{0.7}}$ &  & 64.8$_{\stddev{1.0}}$ & 60.5$_{\stddev{1.0}}$ & 62.6$_{\stddev{0.9}}$ & \\
 
    \midrule
     & \multicolumn{12}{c}{\textbf{CoNLL04}} \\
    \midrule
    
    heuristic & 75.9$_{\stddev{\phantom{0.0}}}$ & 49.2$_{\stddev{\phantom{0.0}}}$ & 59.7$_{\stddev{\phantom{0.0}}}$ &  & 84.1$_{\stddev{\phantom{0.0}}}$ & 22.5$_{\stddev{\phantom{0.0}}}$ & 35.5$_{\stddev{\phantom{0.0}}}$ &  & 84.1$_{\stddev{\phantom{0.0}}}$ & 22.5$_{\stddev{\phantom{0.0}}}$ & 35.5$_{\stddev{\phantom{0.0}}}$ & \\

Ent-SpERT & 88.4$_{\stddev{0.6}}$ & 89.3$_{\stddev{0.7}}$ & 88.9$_{\stddev{0.2}}$ &  & 59.3$_{\stddev{0.5}}$ & 71.3$_{\stddev{1.5}}$ & 64.8$_{\stddev{0.9}}$ &  & 59.2$_{\stddev{0.5}}$ & 71.2$_{\stddev{1.5}}$ & 64.7$_{\stddev{0.8}}$ & \\

SpERT & 87.9$_{\stddev{0.6}}$ & 88.7$_{\stddev{0.3}}$ & 88.3$_{\stddev{0.2}}$ &  & 69.7$_{\stddev{2.3}}$ & 69.0$_{\stddev{0.5}}$ & 69.3$_{\stddev{1.2}}$ &  & 69.4$_{\stddev{2.3}}$ & 68.7$_{\stddev{0.6}}$ & 69.0$_{\stddev{1.2}}$ & \\

TABTO & 89.0$_{\stddev{0.7}}$ & 89.3$_{\stddev{0.3}}$ & 89.2$_{\stddev{0.5}}$ &  & 75.6$_{\stddev{3.2}}$ & 72.6$_{\stddev{1.9}}$ & 74.0$_{\stddev{1.4}}$ &  & 75.4$_{\stddev{3.1}}$ & 72.4$_{\stddev{1.8}}$ & 73.8$_{\stddev{1.2}}$ & \\

PURE & 88.3$_{\stddev{0.4}}$ & 88.5$_{\stddev{0.5}}$ & 88.4$_{\stddev{0.2}}$ &  & 68.6$_{\stddev{2.0}}$ & 68.2$_{\stddev{1.6}}$ & 68.3$_{\stddev{1.0}}$ &  & 68.5$_{\stddev{2.0}}$ & 68.1$_{\stddev{1.5}}$ & 68.2$_{\stddev{0.9}}$ & \\

    \midrule
    & \multicolumn{12}{c}{\textbf{SciERC}} \\
   \midrule

    heuristic & 18.8$_{\stddev{\phantom{0.0}}}$ & 21.5$_{\stddev{\phantom{0.0}}}$ & 20.1$_{\stddev{\phantom{0.0}}}$ &  & 3.5$_{\stddev{\phantom{0.0}}}$ & 0.4$_{\stddev{\phantom{0.0}}}$ & 0.7$_{\stddev{\phantom{0.0}}}$ &  & 3.5$_{\stddev{\phantom{0.0}}}$ & 0.4$_{\stddev{\phantom{0.0}}}$ & 0.7$_{\stddev{\phantom{0.0}}}$ & \\

    Ent-SpERT & 68.0$_{\stddev{0.3}}$ & 66.6$_{\stddev{0.9}}$ & 67.3$_{\stddev{0.6}}$ &  & 44.8$_{\stddev{0.7}}$ & 42.9$_{\stddev{1.0}}$ & 43.8$_{\stddev{0.5}}$ &  & 32.9$_{\stddev{0.9}}$ & 31.5$_{\stddev{1.5}}$ & 32.1$_{\stddev{1.2}}$ & \\
    
    SpERT & 67.6$_{\stddev{0.5}}$ & 67.6$_{\stddev{0.2}}$ & 67.6$_{\stddev{0.3}}$ &  & 49.3$_{\stddev{1.4}}$ & 47.2$_{\stddev{1.3}}$ & 48.2$_{\stddev{1.1}}$ &  & 37.0$_{\stddev{1.3}}$ & 35.4$_{\stddev{1.0}}$ & 36.2$_{\stddev{1.0}}$ & \\
    
    PURE & 68.2$_{\stddev{0.6}}$ & 66.2$_{\stddev{0.9}}$ & 67.2$_{\stddev{0.4}}$ &  & 50.2$_{\stddev{0.9}}$ & 45.2$_{\stddev{1.0}}$ & 47.6$_{\stddev{0.3}}$ &  & 37.6$_{\stddev{1.2}}$ & 33.8$_{\stddev{0.7}}$ & 35.6$_{\stddev{0.6}}$ & \\

    \bottomrule
    
    \end{tabularx}
    \end{adjustwidth*}
    \end{table}

    \begin{table}[h]
    \begin{adjustwidth*}{}{-.3\textwidth}
    
    \caption{Entity and Relation Types of end-to-end RE datasets. SciERC presents two types of symmetric relations denoted with a *.}
    \label{table:data_types}
    
    \centering
    \begin{tabularx}{\textwidth}{@{}lXX@{}}
    
    \toprule
   
    Dataset  & Entity Types & Relation Types \\
  
    \midrule
    ACE05   & Facility, Geo-political Entity, Location, Person, Vehicle, Weapon 
            & Artifact, Gen-affiliation, Org-affiliation, Part-whole, Person-social, Physical \\
    
    \midrule
    CoNLL04 & Location, Organization, Other, Person 
            & Kill, Live in, Located in, Organization based in, Work for \\
            
    \midrule
    SciERC & Generic, Material, Method, Metric, Other Scientific Term, Task  
            & Compare*, Conjunction*, Evaluate for, Feature of, Hyponym of, Part of, Used for \\

    \bottomrule
    
    \end{tabularx}
    \end{adjustwidth*}
    \end{table}


    \begin{table}[h]
    \begin{adjustwidth*}{}{-.3\textwidth}
    \caption{Detailed results of the Swap Relation Experiment with Precision, Recall and F1 scores.}
    \label{table:swap_prf_rev}
   
    \small
    \centering
    \begin{tabularx}{1.3\textwidth}{@{}lll*{3}{Y}r*{3}{Y}r*{3}{Y}c@{}}
    
    \toprule
     
     & & & \multicolumn{3}{c}{NER $\uparrow$} &  & \multicolumn{3}{c}{RE Strict $\uparrow$} &  & \multicolumn{3}{c}{Reverse RE Strict $\downarrow$} \\
      
     \cline{4-6} \cline{8-10} \cline{12-14}
     & & & P & R & F1 & & P & R & F1 & & P & R & F \\
     \midrule
     
    \parbox[t]{2mm}{\multirow{8}{*}{\rotatebox[origin=c]{90}{Kill}}} 
    & \parbox[t]{2mm}{\multirow{4}{*}{\rotatebox[origin=c]{90}{Original}}}

    & Ent-SpERT & 91.7$_{\stddev{0.4}}$ & 91.5$_{\stddev{0.7}}$ & 91.6$_{\stddev{0.4}}$ &  & 82.9$_{\stddev{2.7}}$ & 87.6$_{\stddev{1.8}}$ & 85.1$_{\stddev{0.9}}$ &  & - & - & - & \\
    
    & & SpERT & 91.7$_{\stddev{2.1}}$ & 91.0$_{\stddev{1.0}}$ & 91.4$_{\stddev{1.2}}$ &  & 88.1$_{\stddev{3.1}}$ & 84.4$_{\stddev{1.4}}$ & 86.2$_{\stddev{1.4}}$ &  & - & - & - & \\

    & & TABTO & 91.8$_{\stddev{0.6}}$ & 92.2$_{\stddev{0.5}}$ & \textbf{92.0}$_{\stddev{0.4}}$ &  & 88.8$_{\stddev{1.6}}$ & 90.7$_{\stddev{3.3}}$ & \textbf{89.6}$_{\stddev{1.3}}$ &  & - & - & - & \\
    
    & & PURE & 91.5$_{\stddev{0.9}}$ & 89.6$_{\stddev{0.6}}$ & 90.5$_{\stddev{0.6}}$ &  & 87.2$_{\stddev{2.1}}$ & 81.3$_{\stddev{1.1}}$ & 84.1$_{\stddev{1.2}}$ &  & - & - & - & \\
    
    \cline{2-14}
    & \parbox[t]{2mm}{\multirow{4}{*}{\rotatebox[origin=c]{90}{Swap}}} 
    
    & Ent-SpERT & 91.3$_{\stddev{0.9}}$ & 92.1$_{\stddev{0.7}}$ & 91.7$_{\stddev{0.7}}$ &  & 31.8$_{\stddev{5.3}}$ & 40.0$_{\stddev{8.3}}$ & 35.4$_{\stddev{6.5}}$ &  & 52.8$_{\stddev{5.6}}$ & 65.8$_{\stddev{7.2}}$ & 58.5$_{\stddev{5.7}}$ & \\
    
    & & SpERT & 92.6$_{\stddev{1.8}}$ & 92.6$_{\stddev{0.8}}$ & \textbf{92.6}$_{\stddev{1.2}}$ &  & 33.0$_{\stddev{4.4}}$ & 37.3$_{\stddev{7.4}}$ & 35.0$_{\stddev{5.6}}$ &  & 54.8$_{\stddev{5.1}}$ & 61.3$_{\stddev{4.1}}$ & 57.8$_{\stddev{4.0}}$ & \\

    & & TABTO & 92.8$_{\stddev{0.8}}$ & 92.7$_{\stddev{0.9}}$ & \textbf{92.8}$_{\stddev{0.7}}$ &  & 26.8$_{\stddev{3.6}}$ & 28.4$_{\stddev{4.1}}$ & 27.6$_{\stddev{3.8}}$ &  & 57.8$_{\stddev{3.1}}$ & 61.3$_{\stddev{3.0}}$ & 59.5$_{\stddev{2.8}}$ & \\
    
    & & PURE & 92.0$_{\stddev{0.5}}$ & 89.5$_{\stddev{1.0}}$ & 90.7$_{\stddev{0.5}}$ &  & 65.2$_{\stddev{6.0}}$ & 44.0$_{\stddev{7.4}}$ & \textbf{52.3}$_{\stddev{6.5}}$ &  & 17.8$_{\stddev{2.3}}$ & 12.0$_{\stddev{2.3}}$ & \textbf{14.3}$_{\stddev{2.2}}$ & \\
   
   \midrule
    
    \parbox[t]{2mm}{\multirow{8}{*}{\rotatebox[origin=c]{90}{Located in}}} 
    & \parbox[t]{2mm}{\multirow{4}{*}{\rotatebox[origin=c]{90}{Original}}} 
    
    & Ent-SpERT & 90.1$_{\stddev{0.8}}$ & 89.8$_{\stddev{1.5}}$ & \textbf{90.0}$_{\stddev{0.7}}$ &  
                & 80.8$_{\stddev{3.7}}$ & 76.2$_{\stddev{3.2}}$ & 78.3$_{\stddev{2.4}}$ &  
                & - & - & - & \\
    
    & & SpERT   & 89.8$_{\stddev{1.2}}$ & 87.5$_{\stddev{1.5}}$ & 88.6$_{\stddev{1.1}}$ &  
                & 77.2$_{\stddev{2.8}}$ & 73.0$_{\stddev{3.0}}$ & 75.0$_{\stddev{2.0}}$ & 
                & - & - & - & \\

    & & TABTO   & 90.1$_{\stddev{1.3}}$ & 90.0$_{\stddev{1.8}}$ & \textbf{90.1}$_{\stddev{1.5}}$ &  
                & 93.0$_{\stddev{3.3}}$ & 78.9$_{\stddev{4.6}}$ & \textbf{85.3}$_{\stddev{3.9}}$ & 
                & - & - & - & \\

    & & PURE    & 88.6$_{\stddev{1.1}}$ & 89.4$_{\stddev{1.8}}$ & 89.0$_{\stddev{1.0}}$ &  
                & 89.3$_{\stddev{4.0}}$ & 74.6$_{\stddev{3.7}}$ & 81.2$_{\stddev{2.6}}$ & 
                & - & - & - & \\
    
    \cline{2-14}
    & \parbox[t]{2mm}{\multirow{4}{*}{\rotatebox[origin=c]{90}{Swap}}} 
    
    &  Ent-SpERT    & 86.7$_{\stddev{1.9}}$ & 87.4$_{\stddev{2.7}}$ & 87.0$_{\stddev{2.1}}$ &  
                    & 38.0$_{\stddev{8.5}}$ & 25.4$_{\stddev{2.8}}$ & 30.3$_{\stddev{4.6}}$ &  
                    & 30.2$_{\stddev{5.2}}$ & 21.1$_{\stddev{5.8}}$ & 24.8$_{\stddev{5.7}}$ & \\
    
    & & SpERT       & 87.3$_{\stddev{1.4}}$ & 88.0$_{\stddev{0.9}}$ & 87.7$_{\stddev{1.1}}$ & 
                    & 34.8$_{\stddev{14.8}}$ & 19.5$_{\stddev{6.7}}$ & 24.9$_{\stddev{9.2}}$ &  
                    & 45.6$_{\stddev{17.0}}$ & 26.5$_{\stddev{10.5}}$ & 33.5$_{\stddev{13.0}}$ & \\

    & & TABTO       & 89.0$_{\stddev{0.6}}$ & 88.8$_{\stddev{0.9}}$ & \textbf{88.9}$_{\stddev{0.8}}$ &
                    & 46.5$_{\stddev{6.6}}$ & 29.7$_{\stddev{5.7}}$ & 36.1$_{\stddev{5.8}}$ &  
                    & 45.2$_{\stddev{5.2}}$ & 28.6$_{\stddev{3.7}}$ & 34.9$_{\stddev{3.6}}$ & \\

    & & PURE        & 82.7$_{\stddev{0.8}}$ & 84.6$_{\stddev{0.8}}$ & 83.7$_{\stddev{0.5}}$ &  
                    & 74.9$_{\stddev{7.6}}$ & 49.7$_{\stddev{4.7}}$ & \textbf{59.3}$_{\stddev{3.0}}$ &  
                    & 6.5$_{\stddev{1.8}}$ & 4.3$_{\stddev{1.3}}$ & \textbf{5.1}$_{\stddev{1.5}}$ & \\
    
    \bottomrule
    
    \end{tabularx}
    \end{adjustwidth*}
    \end{table}

\cleardoublepage

\manualmark
\markboth{\spacedlowsmallcaps{\bibname}}{\spacedlowsmallcaps{\bibname}} 
\refstepcounter{dummy}
%

\part*{Bibliography}
\addcontentsline{toc}{chapter}{\tocEntry{\bibname}}
\label{app:bibliography}

\printbibliography
\cleardoublepage\pagestyle{empty}

\hfill

\vfill

\pdfbookmark[0]{Colophon}{colophon}
\section*{Colophon}
This document was typeset using the typographical look-and-feel \texttt{classicthesis} developed by Andr\'e Miede. 
The style was inspired by Robert Bringhurst's seminal book on typography ``\emph{The Elements of Typographic Style}''. 
\texttt{classicthesis} is available for both \LaTeX\ and \mLyX: 
\begin{center}
\url{https://bitbucket.org/amiede/classicthesis/}
\end{center}
Happy users of \texttt{classicthesis} usually send a real postcard to the author, a collection of postcards received so far is featured here: 
\begin{center}
\url{http://postcards.miede.de/}
\end{center}
 
\bigskip
\clearpage

\begingroup 
    \let\clearpage\relax
    \let\cleardoublepage\relax
    \let\cleardoublepage\relax
    \voffset = -1cm
    
\begin{titlepage}
    \begin{addmargin}[-4cm]{-2cm}
    
    
        
        
        
    

	
        
    \vspace{-4cm}
    \begin{center}
        \large  
        \begingroup
            \color{Maroon}\spacedallcaps{\myTitle} \\ \bigskip
        \endgroup
        
        \large
        \spacedlowsmallcaps{\myName}
        

    \end{center} 
    
    \paragraph{Abstract}
    During the past decade, neural networks have become prominent in Natural Language Processing (NLP), notably for their capacity to learn relevant word representations from large unlabeled corpora.
These word embeddings can then be transferred and finetuned for diverse end applications during a supervised training phase. 
In 2018, the transfer of entire pretrained Language Models and the preservation of their contextualization capacities enabled to reach unprecedented performance on virtually every NLP 
benchmark.
However, as models reach such impressive scores, their comprehension abilities still appear as shallow, which reveal limitations of benchmarks to provide useful insights on their factors of performance and to accurately measure understanding capabilities. 

In this thesis, we study the behaviour of state-of-the-art models regarding generalization to facts unseen during training in two important Information Extraction tasks: Named Entity Recognition (NER) and Relation Extraction (RE).
Indeed, traditional benchmarks present important lexical overlap between mentions and relations used for training and evaluating models, whereas the main interest of Information Extraction is to extract previously unknown information.
We propose empirical studies to separate performance based on mention and relation overlap with the training set and find that pretrained Language Models are mainly beneficial to detect unseen mentions, in particular out-of-domain.
While this makes them suited for real use cases, there is still a gap in performance between seen and unseen that hurts generalization to new facts.
In particular, even state-of-the-art ERE models rely on a shallow retention heuristic, basing their prediction more on arguments surface forms than context.

    
    \paragraph{Résumé}
    Au cours de la dernière décennie, les réseaux de neurones sont devenus incontournables
dans le Traitement Automatique du Langage (TAL), notamment pour leur capacité à
apprendre des représentations de mots à partir de grands corpus non étiquetés.
Ces plongements de mots peuvent ensuite être transférés et raffinés pour des applications diverses au cours d'une phase d'entraînement supervisé.
En 2018, le transfert de modèles de langue pré-entraînés et la préservation de leurs capacités de contextualisation ont permis 
d'atteindre des performances sans précédent sur pratiquement tous les benchmarks de TAL.
Cependant, alors que ces modèles atteignent des scores impressionnants, leurs capacités de compréhension apparaissent toujours assez
peu développées, révélant les limites des jeux de données de référence pour identifier leurs facteurs de performance et pour mesurer précisément leur capacité de compréhension.

Dans cette thèse, nous étudions le comportement des modèles état de l'art en ce qui concerne la généralisation à des faits inconnus dans deux tâches importantes en Extraction d'Information : la Reconnaissance d'Entités Nommées et l'Extraction de Relations.
En effet, les benchmarks traditionnels présentent un recoupement lexical important entre les mentions et les relations utilisées pour
l'entraînement et l'évaluation des modèles.
Au contraire, l'intérêt principal de l'Extraction d'Information est d'extraire des informations inconnues jusqu'alors.
Nous proposons plusieurs études empiriques pour séparer les performances selon le recoupement des mentions et des relations avec le jeu d'entraînement.
Nous constatons que les modèles de langage pré-entraînés sont principalement bénéfiques pour détecter les mentions non connues, en particulier dans des genres de textes nouveaux.
Bien que cela les rende adaptés à des cas d'utilisation concrets, il existe toujours un écart de performance important entre les mentions connues et inconnues, ce qui nuit à la généralisation à de nouveaux faits.
En particulier, même les modèles d'Extraction d'Entités et de Relations les plus récents reposent sur une heuristique de rétention superficielle, basant plus leur
prédiction sur les arguments des relations que sur leur contexte.
  \end{addmargin}       
\end{titlepage} 
\endgroup

\end{document}